\RequirePackage{fix-cm}
\RequirePackage{amsmath}	
\documentclass[twocolumn]{svjour3}           
\smartqed  
\newcommand{\vect}[1]{\mathbf{#1}}
\newcommand{\mat}[1]{\mathbf{#1}}

\newcommand{\diffs}[3]{\frac{\partial^2 #1}{
\ifx#2#3 
\partial #2^2
\else
\partial #2 \partial #3
\fi
}}

\newcommand{\ev}{\vect{e}}

\newcommand{\fv}{\vect{f}}

\newcommand{\pv}{\vect{p}}

\newcommand{\ddpv}{\ddot{\vect{p}}}

\newcommand{\uv}{\vect{u}}
\newcommand{\vv}{\vect{v}}

\newcommand{\xv}{\vect{x}}

\newcommand{\yv}{\vect{y}}

\newcommand{\zv}{\vect{z}}

\newcommand{\gammav}{\bm{\gamma}}

\newcommand{\sigmav}{\bm{\sigma}}

\newcommand{\tauv}{\bm{\tau}}

\newcommand{\omegav}{\bm{\omega}}

\newcommand{\etav}{\bm{\eta}}

\newcommand{\Gv}{\vect{G}}

\newcommand{\Gm}{\mat{G}}
\newcommand{\Jm}{\mat{J}}

\newcommand{\Mm}{\mat{M}}

\newcommand{\Qm}{\mat{Q}}
\newcommand{\Rm}{\mat{R}}

\newcommand{\Tm}{\mat{T}}

\newcommand{\wFrame}{\mathcal{F}_W}		
\newcommand{\xW}{\vect{x}_W}			
\newcommand{\yW}{\vect{y}_W}			
\newcommand{\zW}{\vect{z}_W}			
\newcommand{\originW}{O_W}				
\newcommand{\bFrame}{\mathcal{F}_B}		
\newcommand{\xB}{\vect{x}_B}			
\newcommand{\yB}{\vect{y}_B}			
\newcommand{\zB}{\vect{z}_B}			
\newcommand{\originB}{O_B}             
\newcommand{\aFrame}{\mathcal{F}_{A_i}}		
\newcommand{\xA}{\vect{x}_{A_i}}			
\newcommand{\yA}{\vect{y}_{A_i}}			
\newcommand{\zA}{\vect{z}_{A_i}}			
\newcommand{\originA}{O_{A_i}}				

\newcommand{\doubleu}{\text{w}}	

\usepackage{graphicx}
\usepackage{array}
\usepackage{url}
\usepackage{amsmath}
\usepackage{amssymb}
\usepackage{amsthm}
\usepackage{amsfonts}
\usepackage{mathtools}
\usepackage{bm}
\usepackage{dsfont}
\usepackage[usenames,dvipsnames,table]{xcolor}
\usepackage{hyperref} 
\usepackage{etoolbox}
\makeatletter
\patchcmd\@combinedblfloats{\box\@outputbox}{\unvbox\@outputbox}{}{%
   \errmessage{\noexpand\@combinedblfloats could not be patched}%
}%
 \makeatother

\usepackage{pifont}
\newcommand{\cmark}{\ding{51}}%
\newcommand{\xmark}{\ding{55}}%

\usepackage{times}

\usepackage[normalem]{ulem} 

\usepackage[ruled,vlined,linesnumbered]{algorithm2e}
\usepackage[noend]{algorithmic}

\usepackage{subfiles}
\makeatletter
\def  \input@path{{./figures/tikz/}}
\makeatother

\usepackage[per-mode=symbol,binary-units]{siunitx}
\usepackage{cite} 

\graphicspath{{figures/}{figures/ipe/}{figures/matlab/}}

\hypersetup{
    colorlinks,
    citecolor=black,
    filecolor=black,
    linkcolor=black,
    urlcolor=black,
    pdfauthor={},
    pdfsubject={},
    pdftitle={}
}

\usepackage{verbatim}
\usepackage{latexsym}
\usepackage{booktabs}
\usepackage{multirow}	
\usepackage{epsfig,epstopdf}
\usepackage[inline]{enumitem} 

\newif\ifreview
\reviewfalse
\ifreview
	\usepackage[textsize=footnotesize,textwidth=-0.5em]{todonotes}
	\paperwidth=\dimexpr \paperwidth + 6cm\relax
	\oddsidemargin=\dimexpr\oddsidemargin + 3cm\relax
	\evensidemargin=\dimexpr\evensidemargin + 3cm\relax
	\marginparwidth=\dimexpr \marginparwidth + 3cm\relax
\else
\usepackage[textsize=footnotesize,disable]{todonotes}
\fi

\usepackage{tikz}
\usetikzlibrary{shapes,fit,arrows,calc,intersections,
	positioning,backgrounds,decorations.markings, patterns}

\newcommand{\norm}[1]{\lVert#1\rVert}

\definecolor{lightcyan}{rgb}{0.88, 1.0, 1.0} 
\newcolumntype{g}{>{\columncolor{lightcyan}}c}  
\newcolumntype{C}{>{$}c<{$}} 
\newcommand{\GG}[1]{} 
\newcommand\citedb[1]{\centering\scriptsize\cite{#1}}

\newcommand{\rank}{\mathrm{rank}}

\newcommand\figWidth{0.75}

\newcommand\figWidthLim{0.87}
\newcommand\hspaceLim{\hspace{4em}}

\newif\iftableFlag	
\tableFlagtrue			
\begin{document}

\title{Nonlinear Model Predictive Control with Enhanced Actuator Model\\ for Multi-Rotor Aerial Vehicles with Generic Designs
\thanks{This research was partially supported by 
the cooperation program “INTERREG Deutschland-Nederland” as part of the SPECTORS project number 143081 
and by the European Union’s Horizon 2020 research and innovation program  grant agreement ID: 871479 AERIAL-CORE.}
}

\author{Davide Bicego \and Jacopo Mazzetto \and Ruggero Carli \and Marcello Farina \and Antonio Franchi
}

\institute{
D.B. and A.F. are with the Robotics and Mechatronics group, University of Twente, Enschede, The Netherlands and with LAAS-CNRS, Universit\'e de Toulouse, CNRS, Toulouse, France. J.M. and R.C. are with Department of Information Engineering, University of Padova, Padova, Italy. M.F. is with Dipartimento di Elettronica, Informazione e Bioingegneria, Politecnico of Milano, Milan, Italy. Corresponding author: D.B., e-mail: \href{mailto:d.bicego@utwente.nl}{d.bicego@utwente.nl}.
}

\date{Received: date / Accepted: date}

\maketitle

\begin{abstract}
In this paper, we propose, discuss, and validate an online Nonlinear Model Predictive Control (NMPC) method
for multi-rotor aerial systems with arbitrarily positioned and oriented rotors which simultaneously addresses the local reference trajectory planning and tracking problems.
This work brings into question some common modeling and control design choices that are
typically adopted to guarantee robustness and reliability but which may severely limit the attainable performance.  
Unlike most of state of the art works, the proposed method 
takes advantages of a unified nonlinear model which aims to describe the whole robot dynamics by explicitly including a realistic physical description of the actuator dynamics and limitations.  
As a matter of fact, our solution
does not resort to common simplifications such as: 1) linear model approximation, 2) cascaded control paradigm used to decouple the translational and the rotational dynamics of the rigid body, 3) use of low-level reactive trackers for the stabilization of the internal loop, and 4) unconstrained optimization resolution or use of fictitious constraints. 
More in detail, we consider as control inputs the derivatives of the propeller forces and propose a novel method to suitably identify the actuator limitations by leveraging experimental data. Differently from previous
approaches, the constraints of the optimization problem are defined only by the
real physics of the actuators, avoiding conservative -- and often not physical -- input/state saturations which are present, e.g., in cascaded approaches.
The control algorithm is implemented using a state-of-the-art Real Time Iteration (RTI) scheme with partial sensitivity update method. The performances of the control system are finally validated by means of real-time simulations and in real experiments, with a large spectrum of heterogeneous multi-rotor systems: an \emph{under-actuated} quadrotor, a \emph{fully-actuated} hexarotor, a multi-rotor with \emph{orientable} propellers, and a multi-rotor with an unexpected \emph{rotor failure}.
To the best of our knowledge, this is the first time that a predictive controller framework with all the valuable aforementioned features is presented and extensively validated in real-time experiments and simulations.
\end{abstract}

\vspace*{-1.5em}
\section{Introduction}
In the last decade, thanks to the development of both new hardware technologies and software algorithms, the employment of Multi-Rotor Aerial Vehicles (MRAVs) has significantly spread across a wide set of challenging real-life applications, thanks to their vertical take-off and landing (VTOL) and hovering capabilities, their agility, relatively compact structure, good robustness, and low cost.
Classical multi-rotor platforms with \emph{under-actuated} dynamics (e.g., the popular quadrotors), have been extensively studied by the scientific community and widely employed in \emph{contact-less} civil applications such as aerial photography, visual inspection of infrastructures, area patrolling, crop monitoring, and urban search and rescue (USAR) missions~\cite{merino2012unmanned}. The total thrust direction in the body frame of these platforms is fixed and a re-orienta\-tion of the robot chassis is needed to continuously steer the exerted force towards the desired direction. We refer to the vehicles in this class as unidirectional-thrust (UDT) aerial vehicles.

On the other hand, recent platforms characterized by particular actuator arrangements can exploit the mul\-ti\-di\-rec\-tion\-al-thrust (MDT) capability, i.e., the possibility to exert forces in more than one direction without the need to re-orient their body frame, allowing to partially decouple the robot rotational dynamics from the translational one. A subset of this class is represented by the so-called \emph{fully-actuated} systems, for which the control force can be varied in all directions, disregarding the actuator constraints.
Vehicles of this kind have been demonstrated to be particularly suitable for the accomplishment of aerial physical interactions tasks~\cite{2019h-RylMusPieCatAntCacFra,2018g-StaBicSabAreMisFra}, i.e., operations which require an \emph{active contact} and a consequent exchange of energy between the robots and the surrounding environment. Examples of such operations are grasping, transportation, and manipulation of loads, contact-based inspection tasks, and building/decommissioning of structures.

\iftableFlag{
\begin{table*}[t!]
\def\vDim{1cm}
	\caption{Overview of the paper contributions w.r.t. relevant works in the state of the art. A:~capability to drive platforms that can independently control (at least partially) their position and orientation, B:~full nonlinear model and control (non-cascaded) of the system dynamics, C:~extended/enhanced model of the actuators dynamics including low level constraints, D:~controller validated through real experiments with online computation, E:~framework suitable to control arbitrarily-designed MRAVs. \cmark: implemented, \xmark: not implemented.}
	\label{tab:contributions}
	\centering
	\renewcommand\arraystretch{1.15}
	\resizebox{\textwidth}{!}{%
	\begin{tabular}{*{21}{c} | c}
	\hline
	&
	\rotatebox{90}{\parbox{\vDim}{\citedb{alexis2014robust}}} &
	\rotatebox{90}{\parbox{\vDim}{\citedb{alexis2016robust}}} &
	\rotatebox{90}{\parbox{\vDim}{\citedb{baca2016embedded}}} &
	\rotatebox{90}{\parbox{\vDim}{\citedb{BANGURA201411773}}} & 
	\rotatebox{90}{\parbox{\vDim}{\citedb{bouffard2012learning}}} &
	\rotatebox{90}{\parbox{\vDim}{\citedb{darivianakis2014hybrid}}} &
	\rotatebox{90}{\parbox{\vDim}{\citedb{foehnonboard}}} &
	\rotatebox{90}{\parbox{\vDim}{\citedb{geisert2016trajectory}}} &
	\rotatebox{90}{\parbox{\vDim}{\citedb{hofer2016application}}} &
	\rotatebox{90}{\parbox{\vDim}{\citedb{kamel2015fast}}} &
	\rotatebox{90}{\parbox{\vDim}{\citedb{kamel2017linear}}} &
	\rotatebox{90}{\parbox{\vDim}{\citedb{ligthart2017experimentally}}} &
	\rotatebox{90}{\parbox{\vDim}{\citedb{lin2016model}}} &
	\rotatebox{90}{\parbox{\vDim}{\citedb{liu2012explicit}}} &
	\rotatebox{90}{\parbox{\vDim}{\citedb{liu2015robust}}} &
	\rotatebox{90}{\parbox{\vDim}{\citedb{mueller2013model}}} &
	\rotatebox{90}{\parbox{\vDim}{\citedb{mueller2015computationally}}} &
	\rotatebox{90}{\parbox{\vDim}{\citedb{neunert2016fast}}} &
	\rotatebox{90}{\parbox{\vDim}{\citedb{papachristos2013model}}} &
	\rotatebox{90}{\parbox{\vDim}{\citedb{papachristos2016dual}}} &
	\rotatebox{90}{\parbox{\vDim}{\centering THIS PAPER}}
	\\		\hline
	A & 
	\xmark &	
	\xmark & 
	\xmark & 
	\xmark & 
	\xmark & 
	\xmark & 
	\xmark & 
	\cmark & 
	\cmark & 
	\xmark & 
	\xmark & 
	\xmark & 
	\xmark & 
	\cmark & 
	\xmark & 
	\xmark & 
	\xmark & 
	\cmark & 
	\xmark & 
	\xmark & 
	\cmark  	
	\\		\hline	
	B & 
	\xmark &	
	\xmark & 
	\xmark & 
	\xmark & 
	\xmark & 
	\xmark & 
	\xmark & 
	\cmark & 
	\xmark & 
	\cmark & 
	\cmark & 
	\xmark & 
	\xmark & 
	\cmark & 
	\xmark & 
	\xmark & 
	\xmark & 
	\cmark & 
	\xmark & 
	\xmark & 
	\cmark  	
	\\		\hline	
	C & 
	\xmark &	
	\xmark & 
	\xmark & 
	\xmark & 
	\xmark & 
	\xmark & 
	\xmark & 
	\xmark & 
	\xmark & 
	\xmark & 
	\xmark & 
	\xmark & 
	\xmark & 
	\xmark & 
	\xmark & 
	\xmark & 
	\xmark & 
	\xmark & 
	\xmark & 
	\xmark & 
	\cmark  	
	\\		\hline	
	D & 
	\cmark &	
	\cmark & 
	\cmark & 
	\cmark & 
	\cmark & 
	\cmark & 
	\cmark & 
	\xmark & 
	\cmark & 
	\cmark & 
	\cmark & 
	\cmark & 
	\xmark & 
	\cmark & 
	\cmark & 
	\cmark & 
	\cmark & 
	\cmark & 
	\cmark & 
	\cmark & 
	\cmark  	
	\\		\hline	
	E & 
	\xmark &	
	\xmark & 
	\xmark & 
	\xmark & 
	\xmark & 
	\xmark & 
	\xmark & 
	\xmark & 
	\xmark & 
	\xmark & 
	\xmark & 
	\xmark & 
	\xmark & 
	\xmark & 
	\xmark & 
	\xmark & 
	\xmark & 
	\xmark & 
	\xmark & 
	\xmark & 
	\cmark  	
	\\		\hline	
	\end{tabular}
	}	
\end{table*}
} 
\fi
{}

Many different control strategies for MRAVs have been designed for trajectory tracking. The most common controllers implemented on these systems are PIDs (i.e., Proportional, Integrative and Derivative) designed based on models, either linearized around the hovering condition as in~\cite{michael2010grasp}, or obtained with feedback linearization as in~\cite{2010-LeeLeoMcc,2011-MelKum,goodarzi2013geometric}. Other control methods applied to MRAV include, but are not limited to, adaptive control~\cite{dai2014adaptive}, back-stepping and sliding-mo\-de~\cite{2005-BouSie}.
The interested reader is addressed to~\cite{2013-HuaHamMorSam} for a detailed overview about available control strategies for under-actuated MRAVs, while an extension of~\cite{2010-LeeLeoMcc} to the fully-actuated case has been proposed in our previous work~\cite{2018d-FraCarBicRyl}.
The main limitations of the mentioned algorithms are: (i) they are \emph{not predictive}, in the sense that the control input at any time instant is not computed with the objective of optimizing the system performance on a future time horizon, possibly based on a reference motion trajectory; (ii) they are not able to enforce the fulfillment of limitations on input and state variables, which might be crucial for safety reasons.

In the last decades, intense research has been devoted to the development, testing, and implementation of Model Predictive Control (MPC), a model-based optimization-based predictive control method which has gained large popularity especially in the process and chemical industries.
More recently, thanks to the growing availability of increasingly efficient embedded computers, the popularity of MPC is broadening to safety and time-critical applications with fast dynamics, e.g., in the automotive and robotic fields.
MPC is nowadays theoretically well founded and its popularity is mainly related to the following facts. First, it is able to optimize, in a predictive fashion, the system behavior on a given future time horizon based on the system model. Also, in view of the fact that (at least in its most common implementation) it is based on the iterative solution 
of a constrained optimal control problem (OCP), it allows to enforce dynamic constraints on the state and the inputs of a physical system. Furthermore, since the related OCP is solved at each sampling instant as new state measurements get available, it is able to mitigate for possible model perturbations.

Regarding the application of MPC to MRAVs, several notable works have been done in the past few years. On the one hand, some papers tackle the problems of \emph{offline} generating (by solving a suitable OCP) a reference trajectory, feasible with respect to (w.r.t.) the state limits of the system while avoiding possible fixed obstacles, e.g.,~\cite{
mueller2013model,mueller2015computationally,liu2015robust,lin2016model}. Other references to MPC in this perspective can be found in a recent review of motion planning methods for swarms of aerial robots~\cite{chung2018survey}.
Other works, instead, are devoted to closed-loop schemes, that allow for stabilization of the vehicle dynamics and possibly for local trajectory planning.

Here we will focus on the latter class. In this framework, cascaded control schemes are very common, that rely on the decoupling between the translational and the rotational dynamics of the rigid body. In the majority of the works that rely on this approach, e.g.,~\cite{BANGURA201411773,
darivianakis2014hybrid,kamel2017linear,foehnonboard,alexis2016robust},
MPC is used for position control, while the inner-loop attitude control task is obtained using unconstrained regulators (e.g., Lyapunov-based, PIDs, etc.). On the contrary, in~\cite{kamel2015fast,ligthart2017experimentally}, the authors employ MPC for control of the inner rotational loop. 
In either ways, the common strategy is to stabilize the rotational dynamics in an inner loop and to use the rotation configuration (or the angular velocity) as a virtual input commanded by the outer position-control loop.
However, cascaded control methods do not allow to exploit the potentialities of the vehicles at their best, in our opinion. Indeed, the problem with this decoupled approach is the introduction of \textit{fictitious} (non-real from a physical point of view) constraints in the virtual inputs, i.e., in the state variables that represent the interface between the two nested controlled systems. 
As a matter of fact, any constraint imposed on state variables such as the linear velocity, acceleration, jerk, snap, or on the orientation (e.g., Euler angles) and the angular velocity, constitutes a heuristic limitation which does not model accurately the real physical constraints of the real system. 
On the other hand, in our
work, the complete system dynamics is modeled in a non-cascaded way and only the limitations on the individual motor thrust forces and their rates are enforced.

As a matter of fact, the only constraints that play the major role in the platform dynamics are the maximum and minimum torques that can be attained by the motors which drive the propellers. Such limits cause a maximum  speed (mainly due to air drag), a minimum speed (mainly due to electronic reactions), a maximum acceleration (mainly due to motor/propeller inertia), and a maximum deceleration (main\-ly due to nonlinear active breaking).  Any simplification which replaces such real constraints with fictitious constraints in the configuration/state of the platform results,  unavoidably, in a reduced control performance w.r.t. the real dynamic potential of the robot.
In support for the need for a ``whole system'' control, a few recent works, e.g.,~\cite{liu2012explicit,neunert2016fast,
hofer2016application,geisert2016trajectory,foehnonboard},  
avoid cascaded configuration as well.
However, such works either do not include any realistic model of the actuator dynamics and constraints in the control design, or they do not demonstrate the capability of the proposed methods to perform online control of the robot in challenging real experiments. Furthermore, they do not offer a generic framework for the seamless control of both under-actuated and fully-actuated MRAVs with generic designs. On the other hand, the simultaneous accomplishment of all these three objectives constitute the main contribution of this paper.

Another common source of performance limitation is the use of linear/linearized models, see e.g., in~\cite{alexis2016robust,baca2016embedded,BANGURA201411773,bouffard2012learning,darivianakis2014hybrid,foehnonboard,
hofer2016application,papachristos2016dual}. Such models have the advantage of typically requiring less computation, in relation to the online resolution of the OCP, but at the detriment of maximum attainable performance. On the other hand, the MPC scheme for local planning and tracking presented in this paper uses a full-order nonlinear model which includes an innovative data-driven description of the actuator dynamics.
To effectively limit the increased computational burden, the control algorithm is implemented using a state-of-the-art RTI scheme with partial sensitivity update method, as further explained in the paper.

Therefore, despite the field of MPC-based control for MRAVs is already deeply studied, we believe there is still a considerable margin for interesting research investigation, in particular in relation to the employment of more precise models which take into account more representative constraints for the actuators, can be applied to arbitrarily-de\-signed MRAVs, and are 
validated through real experiments with online computation, as demonstrated by the novel and unique results in this regard presented in this paper.

To summarize, the contributions of this paper are threefold. First, the take advantage of a novel actuator model that allows to consider as control inputs the derivatives of the forces generated by the multi-rotor vehicle and to leverage the vehicle dynamic capabilities in a better way.
Second, the development of a control framework suitable to seamlessly deal with UDT and MDT MRAVs. Third, an extensive and comprehensive validation of the controller by means of real-time simulations and experiments performed with heterogeneous MRAVs, i.e., both with \emph{under-ac\-tu\-a\-ted} and \emph{fully-actuated} aerial robots, and both with \emph{fixed} and \emph{orientable} propellers.
To the best of our knowledge, this is the first time that a framework with all such relevant characteristics is successfully tested online to control non-specific aerial vehicles with arbitrary propeller arrangements.
Following the discussion above, Table~\ref{tab:contributions} provides a summary of the contribution of this paper compared to the main works in the literature.

This paper is structured as follows. First, the mathematical model of a MDT MRAV is described in details, with focus on the novel actuator model development and identification. Then, we describe the MPC implementation details. Finally, we present an extensive and thorough validation campaign, conducted with four heterogeneous robot platforms. The results of both realistic simulations and real experiments for the control of an under-actuated, a fully-actuated, and a convertible MRAV are presented, compared, and discussed. Furthermore, the stabilization of a fully-actuated platform subject to a rotor failure is also targeted.
A few summarizing considerations and hints on future work conclude the article.

\textbf{Notation}. In this paper, we denote (column) vectors and matrices in bold font, with lower and upper cases, respectively. The transpose operator is indicated with the superscript $\bullet^{\top}$. Letter superscripts of vectors represent the reference frame w.r.t which these vectors are expressed. $\bf{1}_{i,j}$ denotes the matrix with $i$ rows and $j$ columns with all the elements equal to $1$. $\mathbf{A} \otimes \mathbf{B}$ denotes the Kronecker product between the matrices $\mathbf{A}$ and $\mathbf{B}$.
For the reader's ease, we collected in Tab.~\ref{tab:modeling_symbols} the main symbols related to the modeling used in the paper.
\begin{table}[t]
	\caption{Overview of the main symbols used in this paper.}
	\label{tab:modeling_symbols}
	\centering
	\renewcommand\arraystretch{1.1}
	\resizebox{0.9\columnwidth}{!}{
	\begin{tabular}{lc}
		\hline
		\bf Definition& \bf Symbol \\ \hline
		World Inertial Frame	& 	$\wFrame$ \\
		Multi-rotor Body Frame	& 	$\bFrame$ \\
		Actuator frame ($i$-th) & $\aFrame$ \\
		Position, velocity, acceleration of $\originB$ in $\wFrame$	& 	$\pv$, $\dot{\pv}$, $\ddot{\pv}$ \\
		Rotation matrix representing $\bFrame$ w.r.t. $\wFrame$	&	$\Rm$ \\
		Angular velocity of $\bFrame$ w.r.t. $\wFrame$, expressed in $\bFrame$	&	$\omegav$ \\
		Angular acceleration of $\bFrame$ w.r.t. $\wFrame$, expressed in $\bFrame$	&	$\dot{\omegav}$ \\
		Position of $\originA$ in $\bFrame$	& 	$\pv_{A_i}^B$ \\
		Rotation matrix representing $\aFrame$ w.r.t. $\bFrame$	&	$\Rm_{A_i}^B$ \\
		Mass of the vehicle	&	$m$ \\
		Vehicle's inertia matrix w.r.t. to $\originB$, expressed in $\bFrame$	&	$\Jm$ \\
		Gravity acceleration & $g$ \\
		Total force acting on the CoM & $\fv_B$ \\
		Total moment acting on the CoM & $\tauv_B$ \\
		\hline	
	\end{tabular}
	}
\end{table}

\section{Modeling of MRAVs with generic design}

\subsection{Model of a multi-rotor platform}

\begin{figure}[t]
	\centering
	\includegraphics[width=0.9\columnwidth]{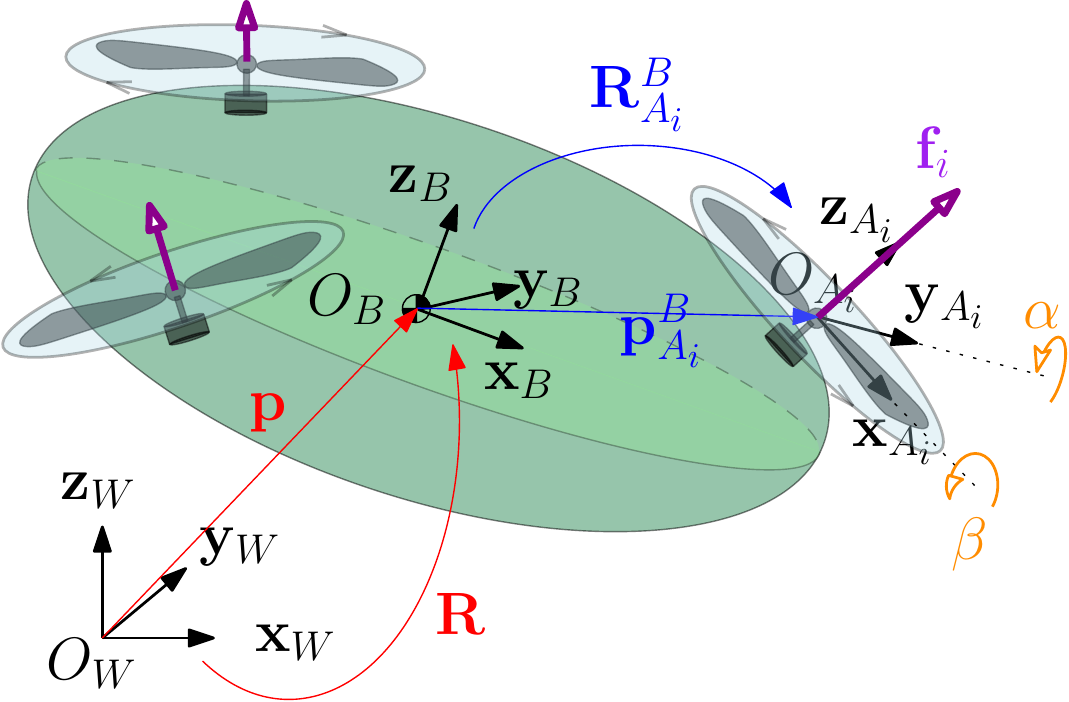}
	\caption{Schematic representation of a MDT MRAV with its reference frames.}
	\label{fig:modeling}
\end{figure}

Multi-rotor platforms are modeled as rigid bodies having mass $m$, actuated by $n \in \mathbb{N} \setminus \{0\} $ spinning motors coupled with propellers, i.e., $n=4$ and $n=6$ in the particular quadrotor and hexarotor models, respectively. Keeping $n$ generic allows to express the model in a non-specific form.
With reference to Fig.~\ref{fig:modeling}, we denote with $\wFrame = \originW,\,\{\xW ,\allowbreak\yW ,\zW \}$ and $\bFrame = \originB,\,\{\xB,\yB,\zB\}$ the world inertial frame and the body frame attached to the MRAV, respectively. The origin of $\bFrame$, i.e., $\originB$, is chosen coincident with the Center of Mass (CoM) of the aerial platform and its position w.r.t. $\originW$, in $\wFrame$, is denoted with $\pv_B^W \in \mathbb{R}^3$, shortly indicated with $\pv$ in the following. The orientation of $\bFrame$ w.r.t. $\wFrame$ is represented by the rotation matrix $\Rm_B^W \in \mathbb{R}^{3 \times 3}$, denoted with $\Rm$ for ease of notation.
We also define with $\aFrame = \originA,\,\{\xA ,\yA ,\zA \}$ the reference frame related to the $i$-th actuator, $i\in \{1,\dots,n \}$, with $\originA$ attached to the thrust generation point and $\zA$ aligned with the thrust direction.
Thanks to this convention, the actuator force expressed in its frame is $\fv_i^{A_i}=f_i \ev_3$, where $\ev_i, \; \scriptstyle{i = 1,\,2,\,3}$ represents the $i$-th vector of the canonical basis of $\mathbb{R}^3$.
The position of $\originA$ w.r.t $\originB$, in $\bFrame$, is indicated with $\pv_{A_i}^B$, while the orientation of $\aFrame$ w.r.t. $\bFrame$ is represented with $\Rm_{A_i}^B$.
The positive definite matrix $\Jm \in \mathbb{R}^{3\times3}$ denotes the vehicle inertia matrix w.r.t. $\originB$, expressed in $\bFrame$.
The angular velocity of $\bFrame$ w.r.t. $\wFrame$, expressed in $\bFrame$, is indicated with $\omegav_{B}^B \in \mathbb{R}^3$ and compactly denoted as $\omegav$ in the following.
The vehicle orientation kinematics, accounting for the evolution of the rotation matrix $\Rm$, is described by the well-known equation
\begin{align}
\label{eq:rotation_kinematics}
	\dot{\Rm}=\Rm\left[\omegav\right]_{\times}
\end{align}
where $\left[\bullet\right]_{\times}\in so(3)$ represents, in general, the skew symmetric matrix associated to any vector $\bullet \in \mathbb{R}^3$.

Using the Newton-Euler formalism, we can derive the dynamics of the aerial platform in order to relate the motion of its CoM, in particular its linear and angular accelerations ($\ddot{\pv}$ and $\dot{\omegav}$, respectively), to the sum of the forces $\fv_B$ and the torques $\tauv_B$ acting on this particular point of the rigid body.
As traditionally done, we express the translational dynamics in world frame, while keeping the rotational one in body frame. This allows to slightly simplify the form of the equations.
Combining them in a compact form, we obtain
\begin{align}
\renewcommand\arraystretch{1.3}
\label{eq:Newton_Euler_Law}
\begin{bmatrix}
m\mat I_3 & \mathbf{0}_3\\
\mathbf{0}_3 & \Jm
\end{bmatrix}
\begin{bmatrix}
\ddpv \\
\dot{\omegav}
\end{bmatrix}
=
\begin{bmatrix}
-mg\ev_3 \\
-\omegav\times\Jm\omegav
\end{bmatrix}
+
\begin{bmatrix}
\Rm & \mathbf{0}_3\\
\mathbf{0}_3 & \mat I_3
\end{bmatrix}
\begin{bmatrix}
\fv_B^B \\
\tauv_B^B
\end{bmatrix}
\end{align}
where $g$ is the gravitational acceleration and $\mathbf{I}_3 \in \mathbb{R}^{3 \times 3}$ is the identity matrix of order $3$. In order to expand~\eqref{eq:Newton_Euler_Law}, one must explicit the dependence of the body wrench on the forces generated by actuators. The vector $\fv_B^B$ is the sum of the actuator forces, properly rotated in body frame, i.e.,
\begin{align}
\label{eq:body_force}
\fv_B^B
= \sum\limits_{i=1}^{n} \fv_i^B
= \sum\limits_{i=1}^{n} \Rm_{A_i}^B \fv_i^{A_i}
= \sum\limits_{i=1}^{n} \Rm_{A_i}^B \ev_{3}f_i.
\end{align}
On the other hand, the body torque is the result of the moments $\tauv_{f_i}$ created by the actuator forces due to their leverage arms and the drag torques $\tauv_{d_i}$ which are a byproduct of the counteracting reaction of the air to the propeller rotation.
\begin{align}
\label{eq:body_torque}
\tauv_B^B
&= \sum\limits_{i=1}^{n} \tauv_{f_i}^B + \tauv_{d_i}^B
= \sum\limits_{i=1}^{n} \pv_{A_i}^B \times \fv_i^B + c_i c_f^{\tau} \fv_i^B \nonumber \\
&= \sum\limits_{i=1}^{n} \big( [\pv_{A_i}^B]_{\times} + c_ic_f^{\tau}\mathbf{I}_3 \big) \Rm_{A_i}^B \ev_3 f_i.
\end{align}
The constant parameter $c_f^\tau>0$ is characteristic of the type of propeller and is defined as the intensity ratio between the thrust produced by the propeller rotation and the generated drag torque. Furthermore, $c_i$ is a variable whose value is equal to $-1$ (respectively, $+1$) in the case the direction of the induced drag torque is opposite (respectively, the same) w.r.t. the generated thrust force, that is the case for a propeller spinning counter-clockwise (respectively, clockwise) w.r.t. its thrust direction. Such coefficient models the fact that the drag torque is always opposed w.r.t. the rotor velocity.
In particular, the model used in this paper assumes that the sense of rotation of each rotor is fixed and cannot be reversed.
Furthermore, the collective pitch of the propeller blades is modeled as constant. As a consequence, the generated thrust cannot be flipped. Thus, swash-plate designs are out of the scope of this work.
Finally, $f_i$ is the intensity of the produced force, which is related to the controllable spinning rate w$_i$ of motor $i$ by means of the quadratic relation
\begin{align}
\label{eq:f_i}
f_i=c_f\doubleu_i^2\quad
\end{align}
where $c_f > 0$ is another propeller-dependent constant parameter to be experimentally identified.
Note that~\eqref{eq:f_i} is a well-established model in the literature, that has been validated experimentally, e.g., in~\cite{2015-RylBueRob}.

We underline that one goal of this paper is to define and guarantee the compliance of the system with meaningful bounds for the actuators, and not to accurately model the physics of the thrust generation. To this purpose, the interest reader is addressed to~\cite{khan2013toward}. Leaving the dependence of the model equations on $f_i$, see~\eqref{eq:body_force}-\eqref{eq:body_torque}, allows the proposed MPC framework to be seamlessly adaptable to the particular thrust generation model specified by the user. Therefore, also different and more accurate thrust models, such as, e.g.,~\cite{2018v-AreMerFra} can be easily integrated in our framework.

From~\eqref{eq:body_force} and~\eqref{eq:body_torque}, the body wrench can be expressed as a linear combination of the forces produced by the $n$ actuators.
Once defined
$
\gammav
=
	\begin{bmatrix}
	f_1 \cdots f_n
	\end{bmatrix}
	^{\top}
$,
we can write
\begin{align}
\renewcommand\arraystretch{1.3}
\label{eq:alloc_mat}
\begin{bmatrix}
\fv_B^B \\
\tauv_B^B \\
\end{bmatrix}
=
\begin{bmatrix}
	\Gm_1 \\
	\Gm_2 \\
\end{bmatrix}
\gammav
=
\Gm
\gammav
\end{align}
where $\Gm \in \mathbb{R}^{6 \times n}$ is the \emph{allocation matrix}.
In particular, its sub-blocks $\Gm_1$ and $\Gm_2$ map the actuator forces to the body forces and moments, respectively.
Moreover, the $j$-th column of $\Gm$, $j \in \{ 1,\dots,n \}$, refers to the contribution of the $j$-th actuator force to the total body wrench, being
\begin{align}
\renewcommand\arraystretch{1.3}
\label{eq:alloc_mat_detail}
\Gm(:,j)=
\begin{bmatrix}
\Rm_{A_j}^B\ev_3 \\
\Big( [\pv_{A_j}^B]_{\times} + c_jc_f^{\tau}\mathbf{I}_3 \Big) \Rm_{A_j} \ev_3
\end{bmatrix}.
\end{align}
The matrix $\Gm$ maps the vector of actuator force intensities, that belongs to a
subset\footnote{Such subset is the Cartesian product of the scalar subsets $\mathbb{F}_i \subset \mathbb{R}^+$ which contain the feasible force values that each actuator can exert.}
of an $n$-dimensional space, to body wrenches laying in a subset of a $6$-dimensional space. Remark the fact that in the case of a \emph{fully-actuated} MRAV, the allocation matrix has full-rank, while for an \emph{under-actuated} vehicle it has a number of rank deficiencies equal to its under-actuation degree. In the particular case of a UDT platform, we have that $\rank({\Gm})=4$, with $\rank({\Gm_2})=3$ and $\rank(\allowbreak{\Gm_1})=1$. This reflects the vehicle capability to exert a body torque in all the directions, disregarding the actuator limits, but a body force along only one direction, i.e., the one of the $\zB$ axis.
A detailed analysis of the allocation matrix rank has been presented
in~\cite{2018r-MorBicRylFra}, in the particular configuration of a hexarotor with synchronized dual-tilting propellers. The theoretical problem of designing an omni-directional (OD) aerial vehicle, that is a fully-actuated MRAV that can produce any body force inside a spherical shell independently from the body torque, has instead been investigated in~\cite{2016-BreDan,2016-ParHerJonKimLee,2018e-TogFra}.

The model defined by the equations~\eqref{eq:Newton_Euler_Law}-\eqref{eq:body_torque} describes the dynamics of a MRAV with arbitrarily positioned and rotated actuators. Nevertheless, it contains, like all models, a certain degree of simplification w.r.t. the real system. In the particular case, it neglects the contributions of the gyroscopic effect induced by the conservation of the angular momentum of the propellers, the blade flapping and the rotor induced drag reactions.
As far as the gyroscopic effect is concerned, its contribution could be taken into account by adding to the right-side part of the rotational dynamics in~\eqref{eq:Newton_Euler_Law} a modified version of~(3) in~\cite{BANGURA201411773} that takes into account the fact that the actuators may have different orientations w.r.t. $\mathcal{F}_B$. As one can easily figure out, each term in such equation is scaled with $\mathbf{J}_{A_i}$, that is the inertia tensor of the rotating part of the $i$-th actuator (composed of the propeller and the rotor).
For MRAVs with actuators of small-medium size, that are the ones on which this work focuses its attention, the entries of this matrix are typically 2-3 orders of magnitude smaller than the ones of $\Jm$. Therefore, the contribution of the gyroscopic effect can be safely neglected in~\eqref{eq:Newton_Euler_Law}.
Regarding the blade flapping and the rotor induced drag effects, they are mainly associated with the flexibility and the rigidity of the rotors, respectively~\cite{mahony2012multirotor}, and are generated by the interaction of the air with the translating propellers.
The results of these aerodynamic effects can be typically observed in UDT platforms as exogenous lateral forces in the $x$-$y$ plane of the rotors. 
In the scope of MDT MRAVs, this analysis would be complex to be precisely evaluated and would require to measure the relative speed of the vehicle w.r.t. the wind and to model the possible interactions between the air-flows of different propellers, which is outside the scope of this paper. Moreover, it should be remarked that the behavior of small-medium size rotor-crafts is much more dominated by their thruster characteristics than to aerodynamic forces, cf.~\cite{khan2013toward}.

For these reasons, in the line of~\cite{mueller2013model,kamel2015fast} and many other relevant works, further motivated by the results presented in~\cite{2015-RylBueRob}, we decided to neglect the first-order contribution of these two reactions and all other second-order effects arising at very high speed and highly dynamic MRAV maneuvers.

The model developed so far is known in the literature and has been presented for completeness and self-consisten\-cy. 
The true contribution brought by this work regarding the modeling of a MRAV is described in the following paragraph, where we detail a methodology aimed to take into account the dynamics and the limitations of the actuators in a simple and effective way.

\subsection{State-dependent actuator bounds}

In our previous work~\cite{2018d-FraCarBicRyl}, we already showed the importance of keeping into account the rotor velocity constraints in the MRAV control strategy in order to preserve the system stability.
As also claimed in~\cite{bemporad2009hierarchical}, further improvements in the control of MRAVs could be attained by extending the nonlinear model in order to include the motor/blade dynamics and treating the motor voltages as the commanded inputs. However, this would require to accurately model the significant nonlinearities introduced by the \emph{active braking}, to control the system a high rate ($\ge 1$ KHz) and at low latency ($\le 1$ ms), and the availability of further measurements (e.g., the motor currents and spinning velocities). 

In~\cite{geisert2016trajectory} a trade-off solution is proposed, considering the rotor accelerations as control input.
This strategy allows to put constraints on both the motor velocities and their derivatives. Doing so, the simplistic hypothesis that the spinning velocities of the rotors (and the generated forces, by consequence) can be changed instantaneously, implicitly done by other works in the literature, is abandoned. Constraints on the rotor accelerations are enforced in the OCP resolution, assuming the lower and upper bounds as constant.

However, as corroborated by experimental data, the capability of the rotors to accelerate depends on the motor currents, the blade dynamics, and on other nonlinear effects hidden in the electrical level that could additionally induce an asymmetry between the acceleration and the deceleration constraints, which will in turn indirectly depend on the rotor velocity.
For these reasons, 
we believe that the extended MRAV model should rely on a methodology that can assess the actuators dynamics and constraints in a more accurate way.
Since the presence of strong nonlinearities in the closed-loop dynamics of the actuators prevents the use of Bode plots analysis or other linear methods, we propose to derive the model from available data in an alternative way.
More specifically, first we experimentally assess how the spinning rate $\doubleu_i$ and the acceleration
$\dot{\doubleu}_i$ of the rotors, each of which is regulated by an independent embedded {Electronic Speed Controller} (ESC), should be properly constrained in order to prevent the risk of damaging the motors and to guarantee an accurate force tracking. 
Secondly, we derive proper constraints for the actuator forces and their derivatives, used by the MPC, in relation to the particular model used to describe the thrust generation. This confers generality to our approach, making it compatible with any other thrust generation model that one wants to adopt.

\subsubsection{Experimental assessment of the limitations on the spinning rate \texorpdfstring{$\doubleu$}{} and the acceleration \texorpdfstring{$\dot{\doubleu}$}{} of the rotors}

In this paragraph, we present a procedure to experimentally identify appropriate rotor acceleration limits as function of the velocity set-points to the ESCs, allowing to account for the aforementioned nonlinearities in a simple yet effective way. To do this, we use a simple testbed composed of a single Brush-Less Direct-Current BL-DC electric motor that is fixed on a mechanical structure, endowed with a propeller and controlled by a dedicated ESC. The latter is connected to a computer via a serial cable. Using a suitable application, the user should be able to specify the desired rotor velocity $\doubleu_d$, read the measurement $\doubleu$, and measure or estimate the current in input to the motor.

As far as the lower and upper bounds for the rotor velocities are concerned, they can be experimentally identified by producing velocity commands that cause the currents to be at the safety limits, with a certain security margin. This information is available from the motors data-sheets. Such velocity limits should be combined with the ones imposed by the low-level speed controllers, if any.

\begin{figure}[t]
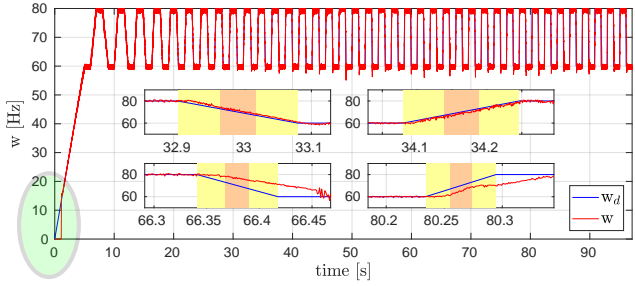

	\centering
	\subfile{velocity_ramps.tex}
	\caption{Trajectory for the identification of the input limits at $\doubleu^*=70$ Hz (that is the average spinning rotor velocities while the platform hovers) for the hexarotor setup. A series of ramps with increasing slope, which corresponds to growing acceleration commands, is sent to one actuator. The top and bottom sub-plots outline intervals where the tracking of the velocity command is good and bad, respectively. In particular, remark on the green ellipse that once the motor is activated, it has a minimum spinning velocity $\underline{\doubleu} = 16$ Hz below which it can't physically rotate. This has to be kept into account by the MPC.}
	\label{fig:velocity_ramps}
\end{figure}

In order to identify the constraints on the rotor accelerations, i.e., $\underline{\dot{\doubleu}}$ and $\overline{\dot{\doubleu}}$, the actuator should be provided with increasing acceleration commands, centered at different velocity set-points in order to appreciate the dependence of the constraints on the rotor velocity.
The profile of the desired rotor velocity trajectory, depicted in Fig.~\ref{fig:velocity_ramps}, is a sequence of ramps (highlighted with yellow rectangles) centered at given set-points $\doubleu_h^*,\, h\in \mathbb{N} \setminus \{0\}$, that are chosen in order to equally span the feasible set $[\underline{\doubleu},\overline{\doubleu}]$. The ramp segments are designed with increasing slopes (both positive and negative) over time and separated by rest-intervals where $\dot{\doubleu}_{d}=0$, needed to avoid overheating the motor.
At this point, the tracking error $e_f$ of the generated force $f$, mapped from the measured rotor velocity via the thrust generation model, w.r.t. a given desired value $f_d$ can be used as the metric to define the acceleration bounds. Using~\eqref{eq:f_i}, we have
\begin{align}
\label{eq:e_f}
	e_f(e_\doubleu,\doubleu_d) &= f_d - f = c_f\ (2\doubleu_d e_\doubleu - {e_\doubleu}^2)
\end{align}
where $e_{\doubleu}= \doubleu_d - \doubleu$ is the velocity error.
After a standard post processing of the data, mostly consisting in a low-pass filtering of the measured velocity in order to reduce high-frequency noise, by visual inspection of the force error associated with the acceleration intervals centered at each $\doubleu_h^*$, the user can determine the velocity-dependent acceleration limits in relation to the force tracking accuracy (s)he is willing to achieve, i.e., those that guarantee an average force inaccuracy below a chosen threshold $\epsilon_f$. Connecting these values using a linear interpolation, it is possible to have an approximation of $\underline{\dot{\doubleu}}$ and $\overline{\dot{\doubleu}}$ as a function of $\doubleu$.

\subsubsection{Definition of the constraints on \texorpdfstring{$f$}{f} and \texorpdfstring{$\dot{f}$}{fdot}}

First of all, the values $\underline{\doubleu}$ and $\overline{\doubleu}$ can be translated into the force constraints $\underline{f}$ and $\overline{f}$ by using the force generation mo\-del~\eqref{eq:f_i}.
Secondly, once the functions $\underline{\dot{\doubleu}}(\doubleu)$ and $\overline{\dot{\doubleu}}(\doubleu)$ are available, in order to convert them into constraints on force derivatives, one can easily compute the time-derivative of~\eqref{eq:f_i}, obtaining
\begin{align}
\label{eq:dot_f_i}
	\dot{f} = \frac{\partial f}{\partial \doubleu} \frac{\partial \doubleu}{\partial t} = 2c_f \doubleu 		\dot{\doubleu}.
\end{align}
The expression of the state dependent input constraints $\overline{\dot{f}}(f)$ and $\underline{\dot{f}}(f)$ are finally obtained from~\eqref{eq:f_i} and~\eqref{eq:dot_f_i}.
We stress the fact that another model for the thrust generation might be used. In that case~\eqref{eq:f_i}, and consequently~\eqref{eq:e_f} and~\eqref{eq:dot_f_i}, should be changed according to the new thrust model. 

As a final remark, it should be noted that the proposed procedure does not require a force/torque sensor.

\subsubsection{Application of the identification procedure to hardware setup}

In the following, we describe how we concretely apply the previously-described procedure to two different hardware setups.
The first one, shortly named \textit{setup I}, is composed of a Mikro\-Kopter\footnote{\url{http://www.mikrokopter.de/en/home}} electric motor MK3638 coupled with a 12X4.5' propeller and controlled by a BL-Ctrl V2.0 ESC. The low-level control of the rotor velocity is performed in closed-loop employing the Adaptive Bias Adaptive Gain (AB\-AG) algorithm, whose details can be found in~\cite{2017c-FraMal}.

In this setup, being the one of our (custom-made) fully-actuated hexarotors, the constraints on the minimum and maximum velocities are related to the properties of the \emph{closed-loop} rotor velocity controller. Specifically, the actual rotor velocity is estimated by the low-level controller without any additional sensor and the quality of such estimation is proportional to the rotor speed. This causes the velocity to have a lower bound, in order to be properly estimated by the controller with a certain precision. On the other hand, the limited arithmetic capabilities of the ESC micro-controller (which allows only 8-bit additions and has no floating point unit) translates into a velocity upper bound, cf.~\cite{2017c-FraMal}.
In this case, we identified $\underline{\doubleu}=16$\,\si{Hz} and $\overline{\doubleu}=102$\,\si{Hz}. In particular, the upper limit satisfies the maximum current limitation of $20$\,\si{A} reported in the motor data-sheet. Finally, using~\eqref{eq:f_i} we obtained the limits $\underline{f}\approx 0.25$\,\si{N} and $\overline{f}\approx 10.3$\,\si{N} used to constrain the OCP resolution in the MPC algorithm.

As far as the identification of the acceleration limits is concerned, we generated a set of increasing $\dot{\doubleu}$ spanning the range $\pm[20,300]$\,\si{Hz/s} with a step of $10$\,\si{Hz/s}, centered at a given average velocity level $\doubleu_h^*$. Each ramp fragment takes values in the set $[\doubleu_h^*-\delta_h,\doubleu_h^*+\delta_h]$, with $\delta_h=10 \si{Hz}$.
With reference to Fig.~\ref{fig:velocity_ramps}, for each ramp we select the $30\%$ of the total samples which are centered in the middle of the interval (highlighted with orange rectangles in Fig.~\ref{fig:velocity_ramps}) and compute the correspondent force error using~\eqref{eq:e_f}.
The operation is repeated at different set-points $\doubleu_h^*$ in the set $[30,90]$\,\si{Hz} with a step of $10$ \,\si{Hz}, in order to span the set of admissible velocities previously estimated.
The plots of the force error trends related to \textit{setup I} are shown in Fig.~\ref{fig:plots_acceleration_limits}. 
In each subplot, notice that the number of samples related to increasing values of $|\dot{\doubleu}|$ is gradually decreasing. This happens because 
an increase in the ramp slopes is associated with a decrease in the time duration associated with the segments.
Remark three facts:
\begin{enumerate*}
\item[(i)] At the same velocity set-points, increasing force errors are associated with increasing acceleration values, on average. This suggests that high acceleration references (of both signs) are difficult to be tracked and fosters the idea to constrain them with lower and upper bounds.
\item[(ii)] For different set-point velocities, the profile of the force error at corresponding acceleration intervals is different. This confirms the claim that the limits are velocity-dependent. In particular, we observe that while increasing values of set-points seem to cause increasing force error for positive accelerations, such trend is not pursued by negative accelerations. A reasonable explanation for such effect could be the fact that the active braking, which intervenes only for negative accelerations, is not behaving in the same way for different velocity levels.
\item[(iii)] At the same velocity set-points, the force error $e_f$ associated with negative accelerations is larger, on average, w.r.t. the one associated with positive accelerations. This reveals that, despite the use of the active-braking, the deceleration of a rotor produces a worse force tracking than the corresponding acceleration.
\end{enumerate*}
\begin{table}[b]
	\renewcommand\arraystretch{2}
	\caption{Identified acceleration limits for \textit{setup I}.}
	\label{tab:limits}
	\centering
	\resizebox{0.85\columnwidth}{!}{%
	\begin{tabular}{cccccccc}
		\hline
		$\doubleu\ [\si{Hz}]$ & $30$ & $40$ & $50$ & $60$ & $70$ & $80$ & $90$ \\ 
		\hline
		$\underline{\dot{\doubleu}}\ [\si{Hz/s}]$ & $-120$ & $-160$ & $-200$ & $-140$ & $-160$ & $-160$ & $-140$\\
		$\overline{\dot{\doubleu}}\ [\si{Hz/s}]$ & $200$ & $200$ & $200$ & $160$ & $180$ & $180$ & $180$ \\
		\hline
	\end{tabular}
	}
\end{table}

\begin{figure}[t]
	\centering
	\includegraphics[width=\columnwidth]{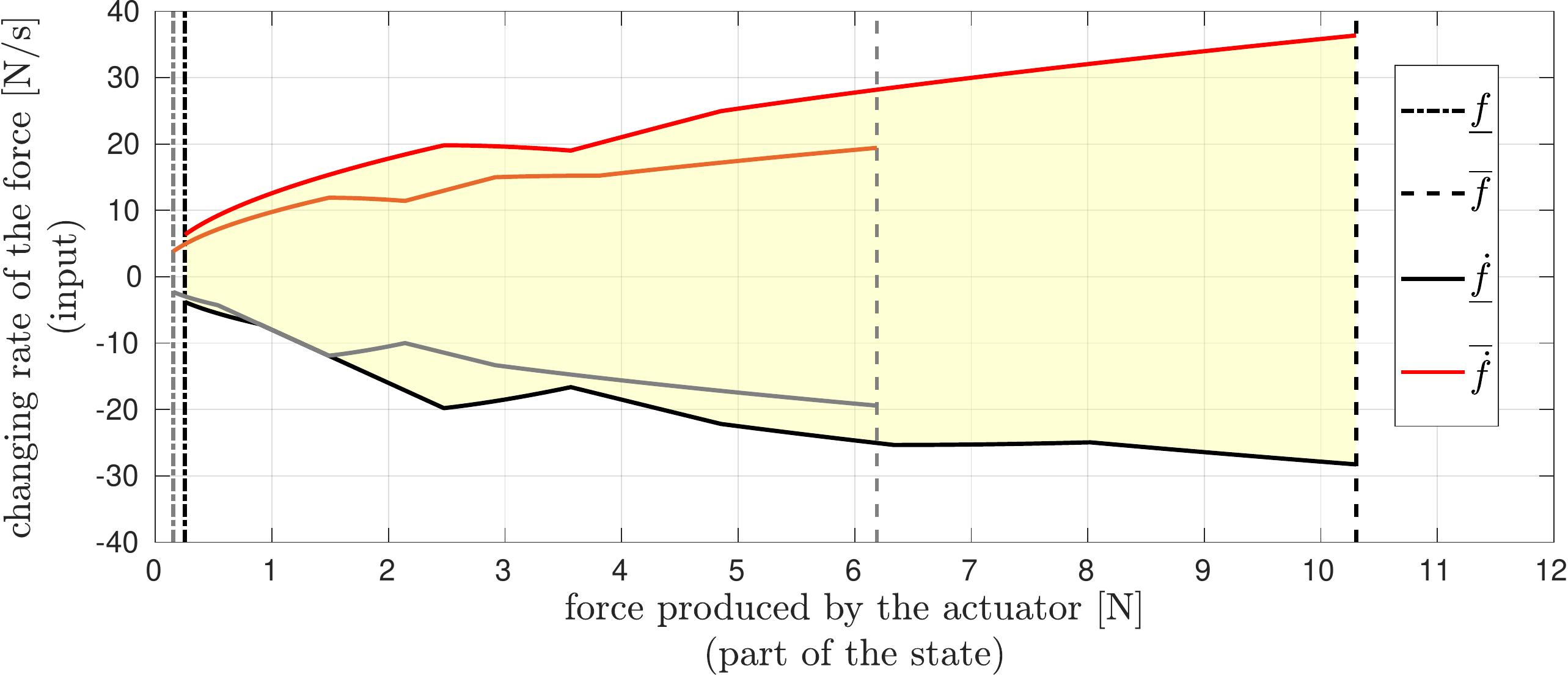}
	\caption{State and input constraints given to the NMPC in relation to the two hardware setups used in the experiments. Darker and lighter colored lines are referred to \textit{setup I} and \textit{setup II}, respectively.}
	\label{fig:state_input_limits}
\end{figure}

\begin{figure*}[ht!]
	\centering
	\includegraphics[width=\figWidthLim\columnwidth]{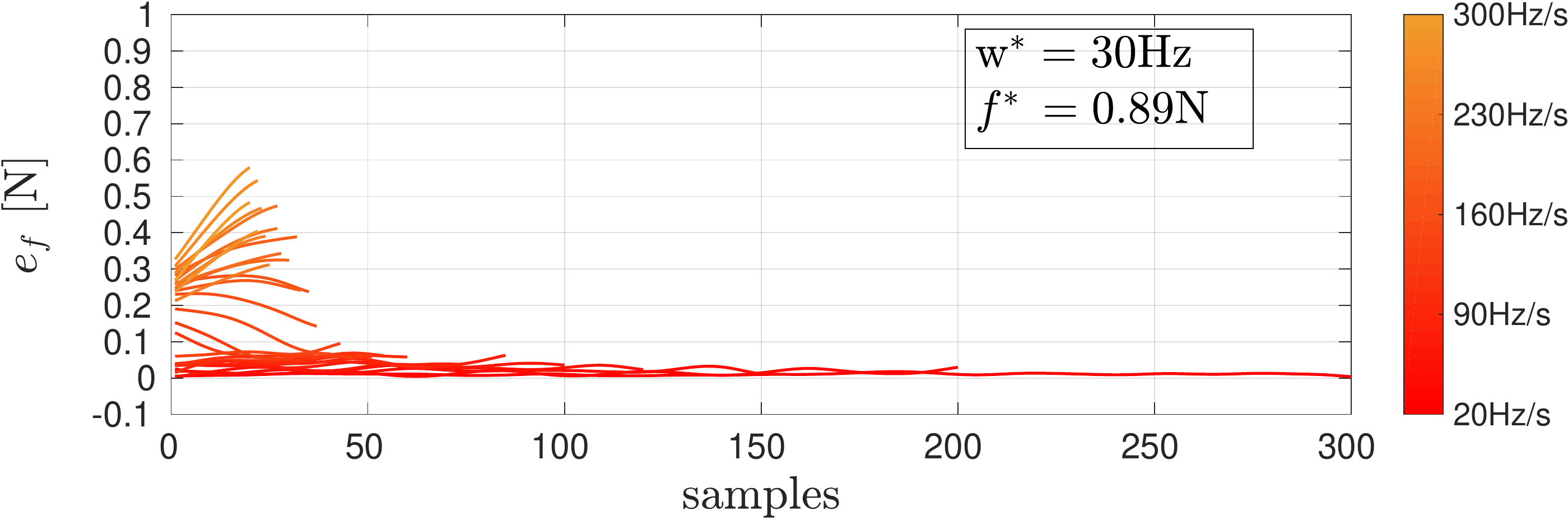} \hspaceLim
	\includegraphics[width=\figWidthLim\columnwidth]{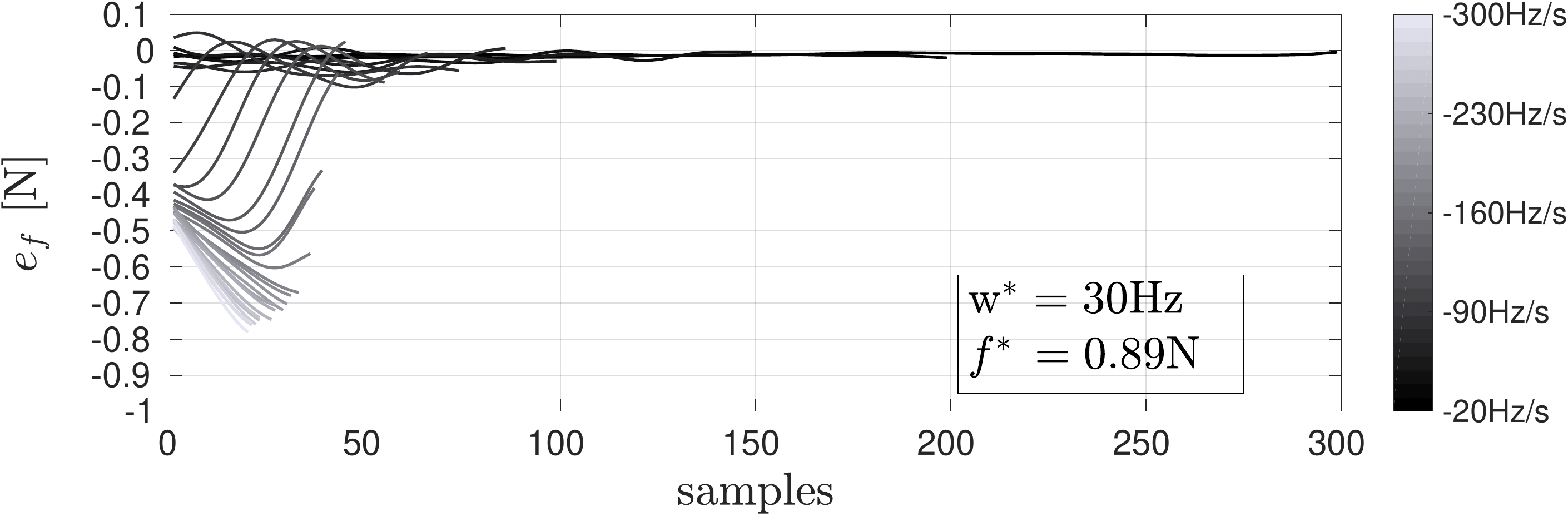}	\\
	\includegraphics[width=\figWidthLim\columnwidth]{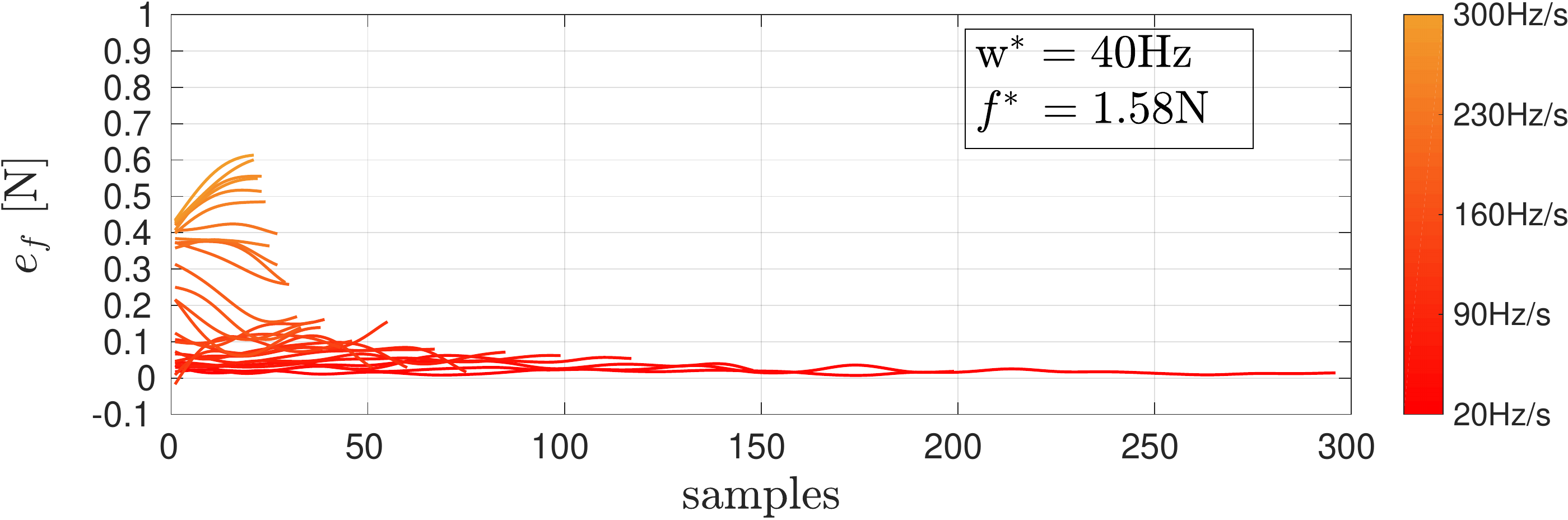} \hspaceLim
	\includegraphics[width=\figWidthLim\columnwidth]{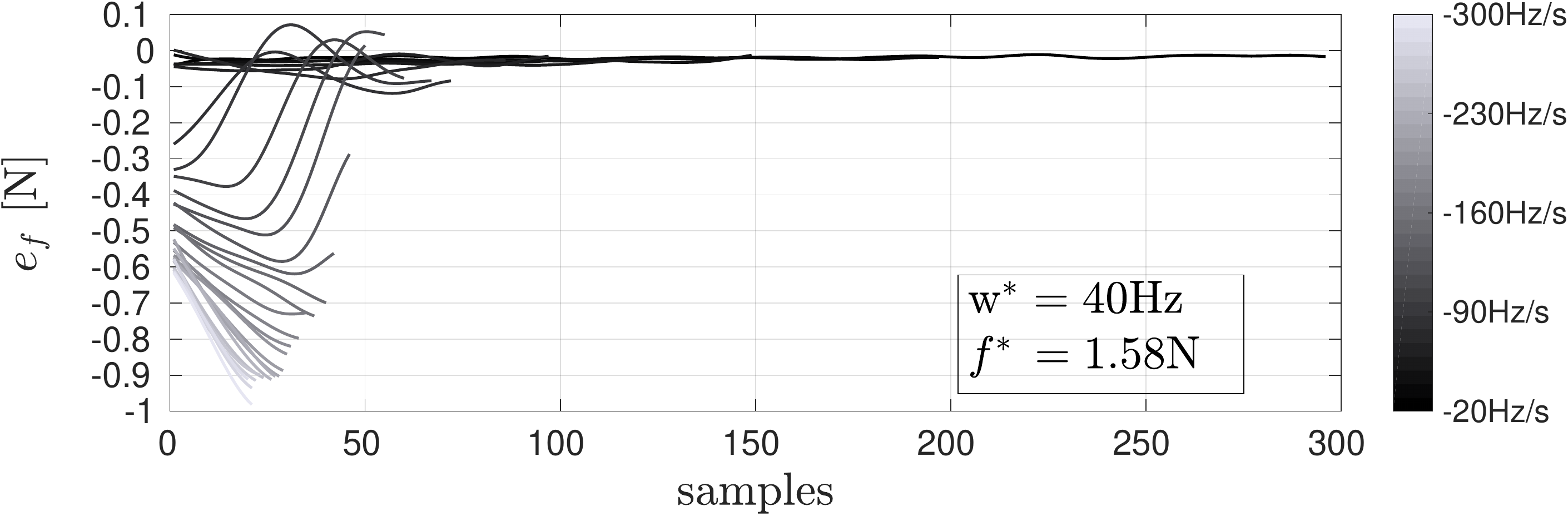}	\\
	\includegraphics[width=\figWidthLim\columnwidth]{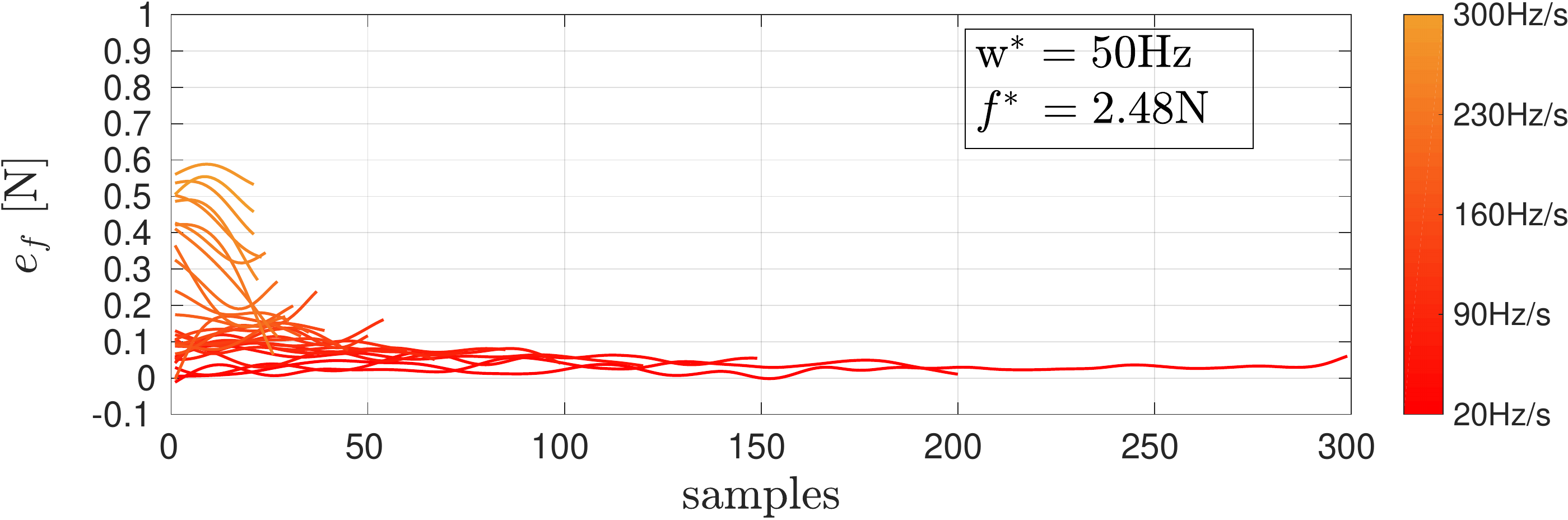} \hspaceLim
	\includegraphics[width=\figWidthLim\columnwidth]{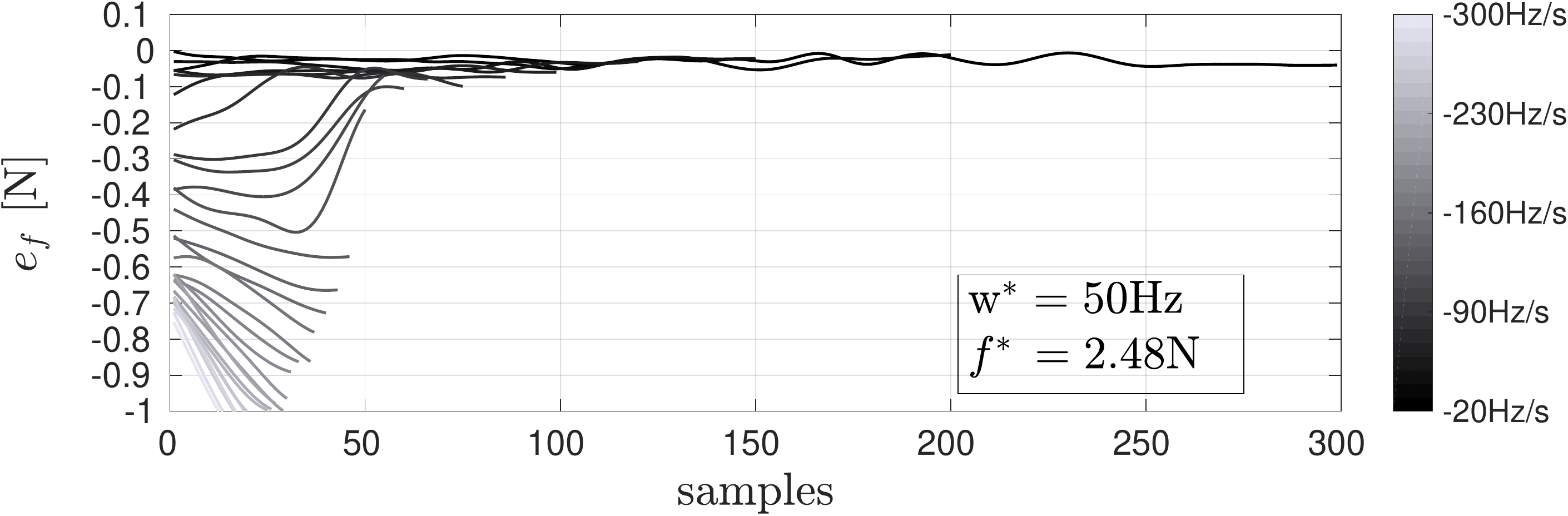}	\\
	\includegraphics[width=\figWidthLim\columnwidth]{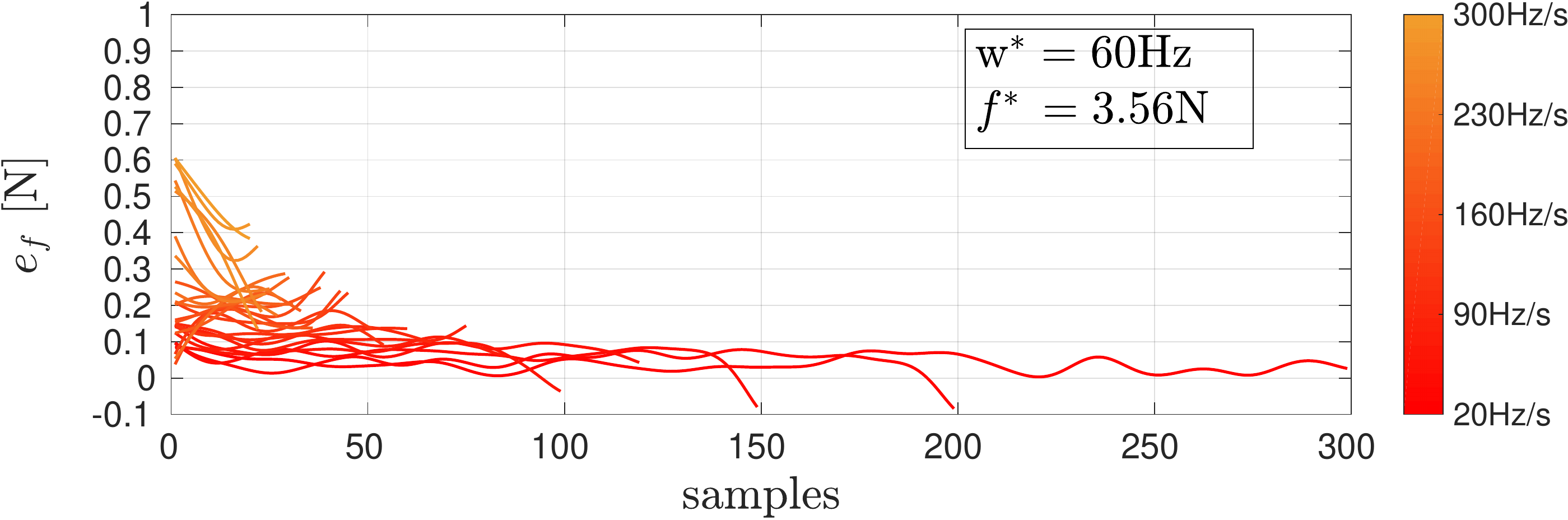} \hspaceLim
	\includegraphics[width=\figWidthLim\columnwidth]{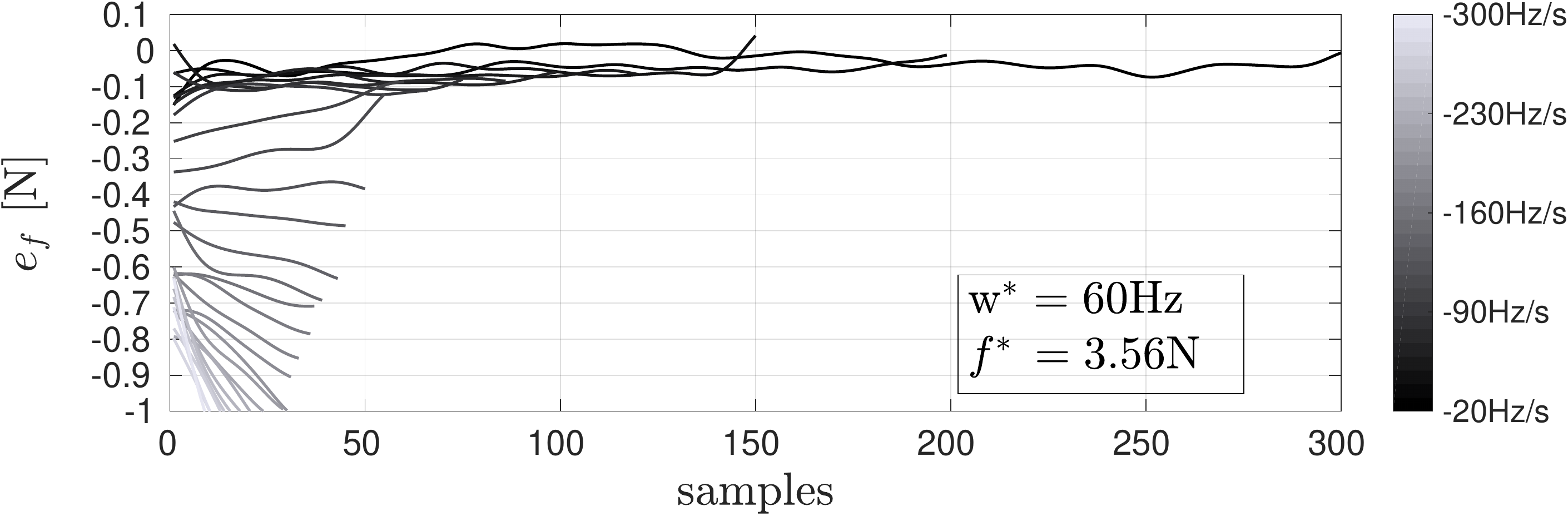}	\\
	\includegraphics[width=\figWidthLim\columnwidth]{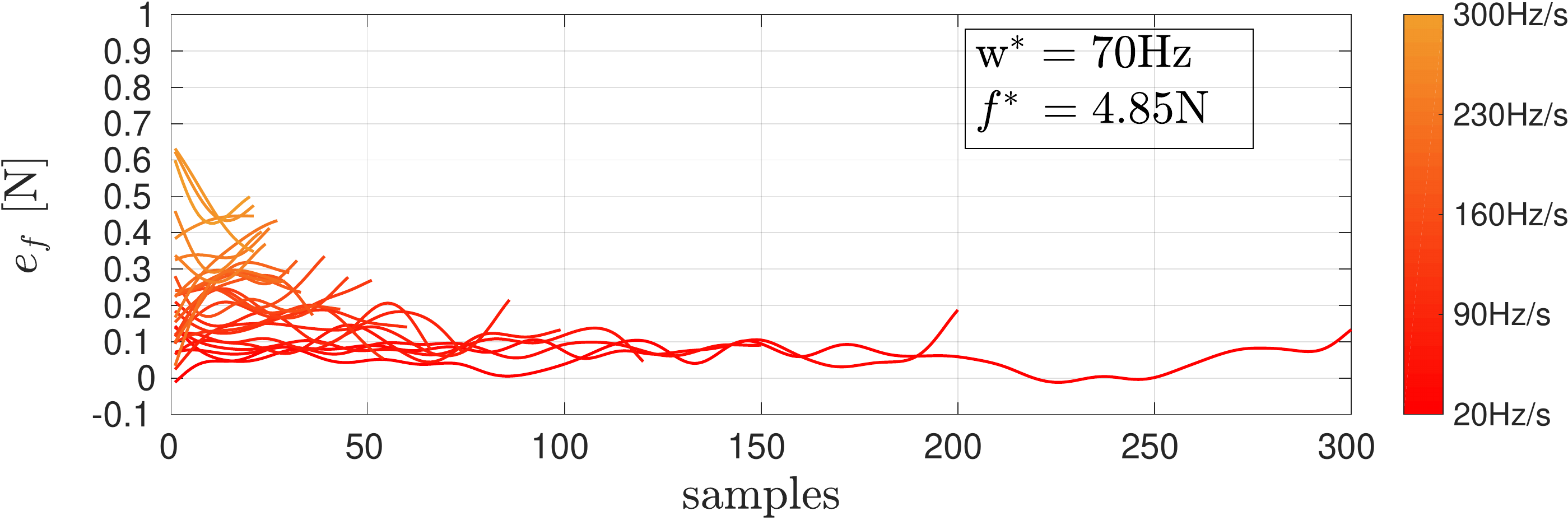} \hspaceLim
	\includegraphics[width=\figWidthLim\columnwidth]{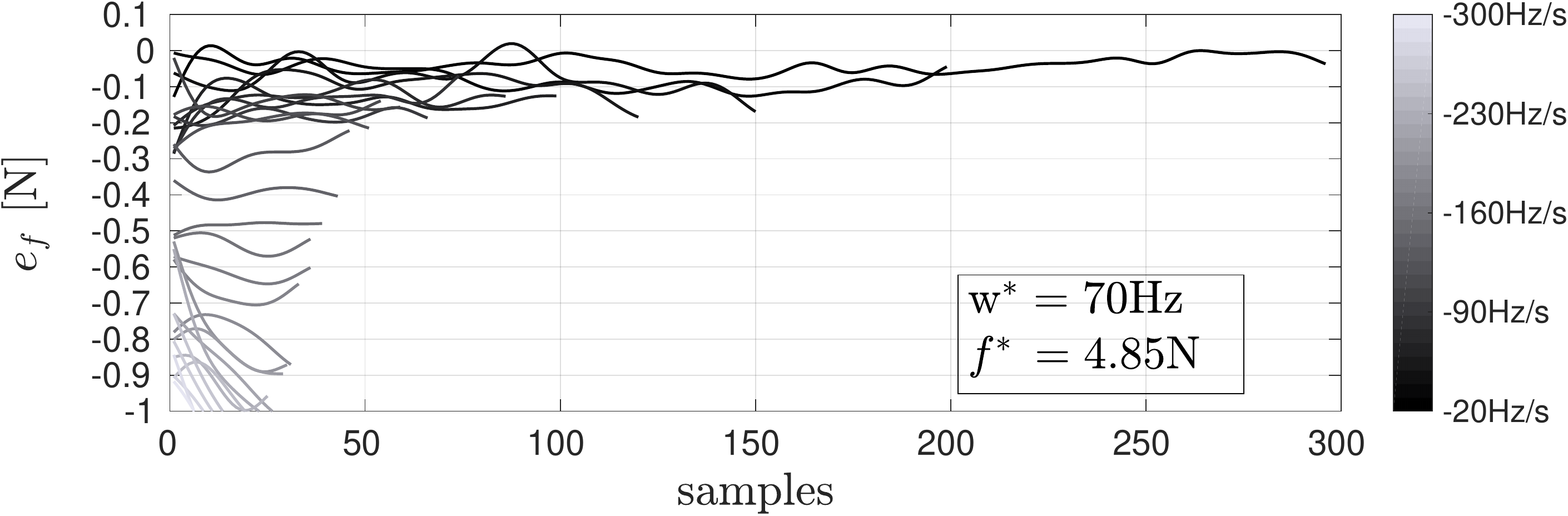}	\\
	\includegraphics[width=\figWidthLim\columnwidth]{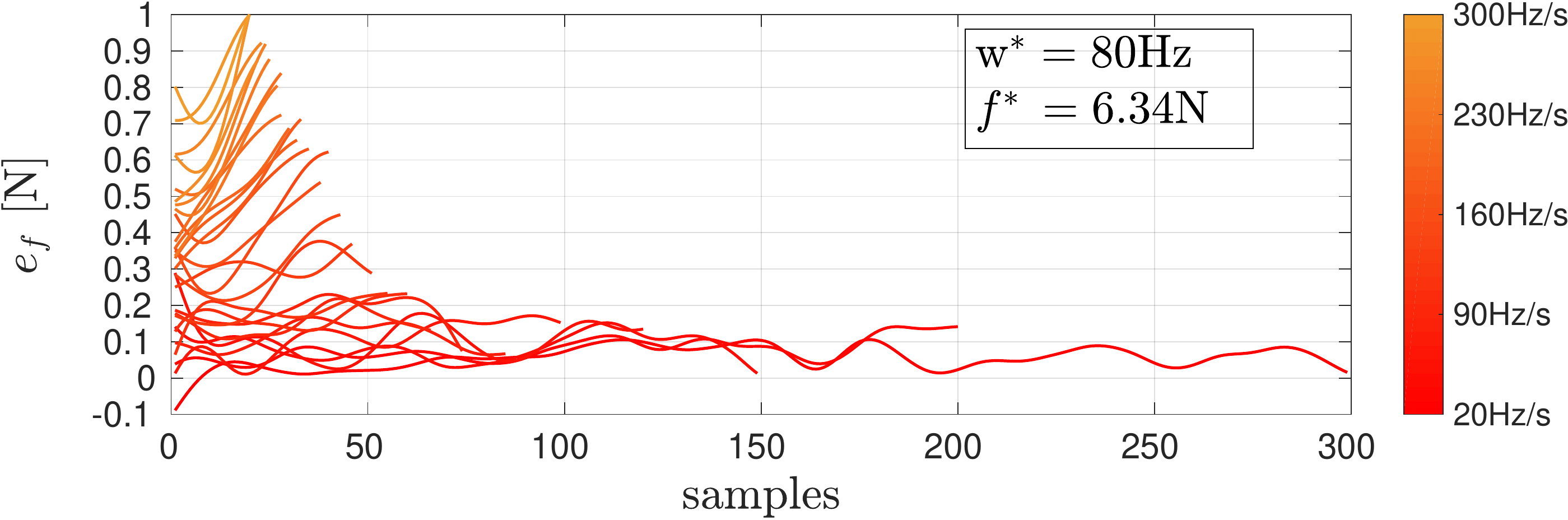} \hspaceLim
	\includegraphics[width=\figWidthLim\columnwidth]{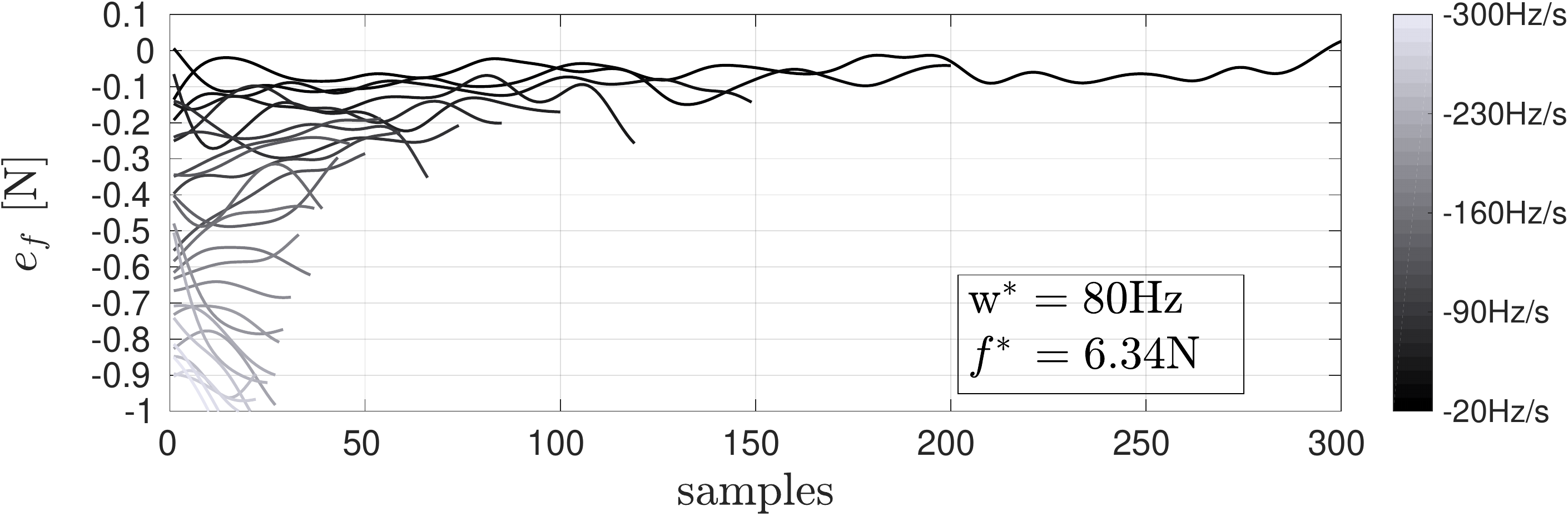}	\\
	\includegraphics[width=\figWidthLim\columnwidth]{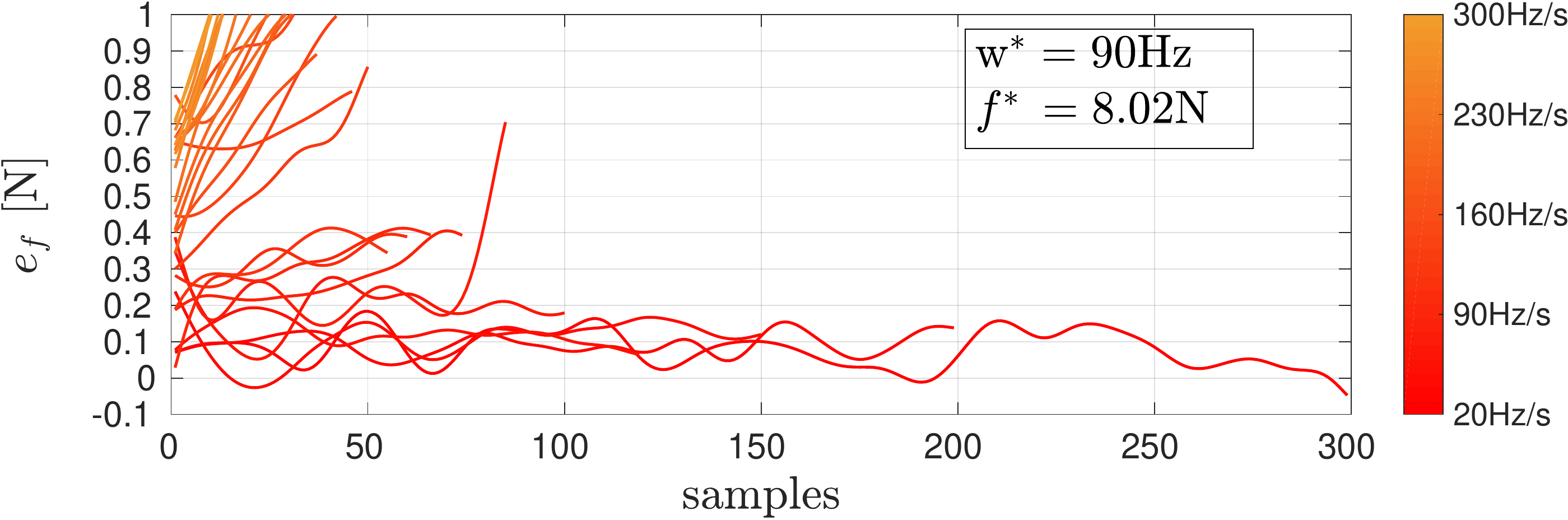} \hspaceLim
	\includegraphics[width=\figWidthLim\columnwidth]{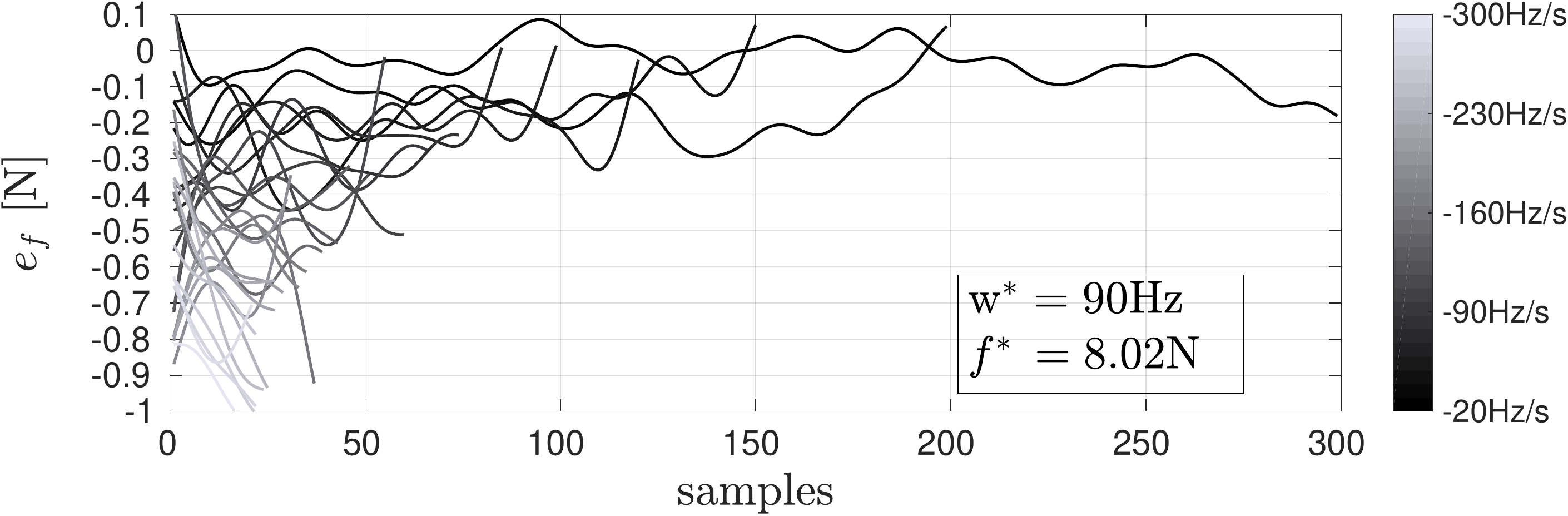}
	\caption{Plots of the force error trends, associated to the acceleration intervals at different set-point velocities $\doubleu_h^*$, related to \textit{setup I}. Positive and negative acc. are depicted on the left and right column, respectively. The acceleration-dependent errors are represented with different color shades.} \vspace*{-2em}
	\label{fig:plots_acceleration_limits}
\end{figure*}

In order to identify the acceleration limits $\underline{\dot{\doubleu}}$ and $\overline{\dot{\doubleu}}$, we defined $\epsilon_f\approx0.2$\,\si{N} as the force error threshold, admitting slightly bigger values ($\approx0.3$\,\si{N}) at high velocity set-points. As we will see in the experimental validation plots, such value generates conservative limits that preserve the platform stability also during agile trajectory tracking. As a general rule, such threshold value shall depend on the particular robot task.
The identified acceleration limits related to \textit{setup I} are collected in Tab.~\ref{tab:limits}, where velocity data are expressed in $\si{Hz}$, while acceleration ones in $\si{Hz/s}$. 
Interpolating these values with linear functions and using~\eqref{eq:f_i} and~\eqref{eq:dot_f_i}, allowed us to obtain the force derivative constraints as function of the instantaneous thrust forces.

A second hardware setup (\textit{setup II}) is analyzed, i.e., that one of the available under-actuated quadrotor, which combines a MK2832/35 motor with a 10X4.5' propeller from MikroKop\-ter, controlled by the same ESC and closed-loop algorithm of \textit{setup I}. 
The profile of the constraints for the actuator forces and their derivatives, related to the two setups, are depicted with different colors in the plot of Fig.~\ref{fig:state_input_limits}, where the admissible set of values for both cases are represented with the yellow area.
Consistently with the previous results, the limits on positive and negative thrust derivatives are not perfectly symmetric.

\subsection{State-space model for discrete-time control}

Let us define the state vector $\mathbf{x}$ and the input vector $\mathbf{u}$ as
 \begin{align}
	\label{eq:state_vector}
	\mathbf{x} &:=
	\begin{bmatrix}
		\pv^\top\,\dot{\pv}^{\top}\,\etav^{\top}\,\omegav^{\top}\,\gammav^{\top}
	\end{bmatrix}
	^{\top}
	\\
	\label{eq:input_vector}
	\mathbf{u} &:= \dot{\gammav}
\end{align}
with $\etav \in \mathbb{R}^{n_{\eta}}$ being the vector used for concisely representing the platform orientation. Specifically, for the experimental validation we chose a minimum representation with three angles (see the section related to the experimental validation for a detailed discussion about pros and cons of minimal representations).
With reference to~\eqref{eq:state_vector}, it is worth to remark the fact that the actuator forces $\gammav$, which in other works were either discarded from the model or assumed as the input, are considered here as part of the state.
In particular, all the quantities which compose the state are assumed to be measurable (cf. Sec.~\ref{subsec:setup} for a discussion of the employed sensors). In view of this, the control scheme will be implemented without resorting to a dedicated state-observer.

The expression of the map $\mathbf{f}(\bullet)$ relating $\dot{\mathbf{x}}$ to $\mathbf{x}$ and $\mathbf{u}$, i.e.,
\begin{align}\label{eq:NMPC_model}
\dot{\mathbf{x}}(t) = \mathbf{f}(\mathbf{x}(t),\mathbf{u}(t)),
\end{align}
can be obtained from~\eqref{eq:rotation_kinematics}-\eqref{eq:body_torque}, according to the previous definition of $\mathbf{x}$ and $\mathbf{u}$.
For digital control purposes, the continuous-time model in~\eqref{eq:NMPC_model} is discretized (in the particualr case using a fixed step $4^{th}$ order explicit Runge-Kutta integrator) yielding the following discrete-time model
\begin{align}
\label{eq:discretized_NMPC_model}
\xv_{k+1} = \boldsymbol{\phi}(\xv_k,\uv_k),\quad k=0,1,\ldots,N-1
\end{align}
where, for ease of notation, $\xv_k=\xv(kT)$ being $T$ the MPC sampling time and $\uv(t)=\uv_k$ for $t \in \left[kT, (k+1)T \right)$.

\section{NMPC for MRAVs with generic design}

The goal of this section is to devise an MPC controller able to simultaneously address the problem of local reference trajectory planning and that of stabilizing the vehicle dynamics. Specifically, we aim to track a reference trajectory denoted $\left(\pv_r(t), \etav_r(t)\right)$ given by a generic global planner. We assume $\left(\pv_r(t), \etav_r(t)\right)$ to be twice continuously differentiable.  
In order to guarantee smoothness properties of the generated trajectory, we force our algorithm to be able to drive also the derivatives of the state variables toward the corresponding ones of the reference trajectory. Therefore, we introduce the following enlarged reference signal 
\begin{align}
\hspace{-0.5em}
 {\bf y}_r(t) =\left[\pv^\top_r(t) \,\, \dot{\pv}^\top_r(t)  \,\, \ddot{\pv}^\top_r(t) \,\, \etav^\top_r(t) \,\, {\omegav}^\top_r(t) \,\, \dot{\omegav}^\top_r(t)  \right]^\top
\end{align}
 and, accordingly, we define the output map as
\begin{align}
    {\bf y}(t) = {\bf h} \left(\xv(t), \uv(t) \right) = 
 \left[
 \begin{array}{c}
 \pv(t)\\
 \dot{\pv}(t)\\
   \ddot{\pv}\left(\xv(t), \uv(t) \right) \\
   \etav(t) \\
    {\omegav}(t)\\
     \dot{\omegav}\left(\xv(t), \uv(t) \right)
 \end{array}
 \right] .
\end{align}
For clarity observe that $\pv(t), \dot{\pv}(t),    \etav(t), {\omegav}(t)$ are measured sub-vectors of the state, while $\ddot{\pv},\dot{\omegav}$ are functions of $\xv(t), \uv(t)$, and, in particular, sub-components of the map $\fv$ in \eqref{eq:NMPC_model}.

Finally, we define $ {\bf y}_{r,k}$,  ${\bf y}_{k}$ as the discretized version of $ {\bf y}_r(t)$ and ${\bf y}(t)$, respectively, i.e., $ {\bf y}_{r,k} = {\bf y}_r(kT)$,  ${\bf y}_{k}={\bf y}(kT)$.
The OCP to be solved at time $kT$, given the current state $\xv_k$, is formulated as

\begin{align}
	\label{eq:cost_function}
	\min_{\begin{matrix}
	\hat{\xv}_0, \ldots, \hat{\xv}_N \\
	\hat{\uv}_0, \ldots, \hat{\uv}_{N-1}
	\end{matrix}} \,
	&\sum_{h=0}^{N-1}\left\{ \norm{\hat{{\bf y}}_{h} - {\bf y}_{r,k+h}}_{\mathbf{Q}_h}^2  +\norm{\hat{\uv}_h}_{\mathbf{R}_h}^2  \right\} + \nonumber \\
	&\qquad \qquad + \norm{\hat{{\bf y}}_{N} - {\bf y}_{r,k+N}}_{\mathbf{Q}_N}^2	\\
	\label{eq:meas_const}
	\text{s.t.}\,\,\,\, \,\,\,\,\,&\mathbf{\hat{x}}_0=\xv_k\\
	\label{eq:dyn_const}
	&\hat{\xv}_{h+1}=\boldsymbol{\phi}(\hat{\xv}_h,\hat{\uv}_h),\,\,\, \, {\scriptstyle h=0,1,\ldots,N-1} ,\\
   \label{eq:output}
	&\hat{\bf y}_{h}={\bf{h}}(\hat{\xv}_h,\hat{\uv}_h),\,\,\,\, {\scriptstyle h=0,1,\ldots,N} ,\\
	\label{eq:state_const}
	&\underline{\gammav} \leq \Mm \hat{\xv}_h \leq \overline{\gammav},\,\,\,\, {\scriptstyle h=0,1,\ldots,N} ,\\
	\label{eq:input_const}
	&\underline{\dot{\gammav}}_{k+h}\leq \hat{\uv}_h \leq \overline{\dot{\gammav}}_{k+h},\,\,\,\, {\scriptstyle h=0,1,\ldots,N-1},
\end{align}

\vspace*{1em}
\noindent
where $\mathbf{Q}_h$, $\mathbf{R}_h$ are semidefinite positive matrices and matrix ${\Mm}$ is defined in order to select only the $n$ elements of the state $\xv$ corresponding to the actuator forces, that is,
\begin{align}
\label{eq:state_input_const}
{\Mm} =
\begin{bmatrix}
\mathbf{0}_{n \times (9+n_{\eta})} & \mathbf{I}_{n}
\end{bmatrix}.
\end{align}
The bounds $\overline{\gammav}, \underline{\gammav}$, depend on the quantities $\overline{f}$, $\underline{f}$ characterized in the previous section and, compactly, they are defined as
\begin{align}
\overline{\gammav} &= \mathbf{1}_{6 \times 1} \otimes \overline{f} \\
\underline{\gammav} &= \mathbf{1}_{6 \times 1} \otimes \underline{f}.
\end{align}

\noindent
Furthermore, the bounds
$\underline{\dot{\gammav}}_{k+h}, \overline{\dot{\gammav}}_{k+h}$, $h=0,1,\ldots,N-1$, depend on the time-varying and state dependent quantities $\overline{\dot{f}}(f)$, $\underline{\dot{f}}(f)$ and, precisely, they should be defined as
\begin{align}
\overline{\dot{\gammav}}_{k+h} &= \left[ \overline{\dot{f}}(f_{1,k+h}), \ldots,  \overline{\dot{f}}(f_{n,k+h}) \right]^\top \\
\underline{\dot{\gammav}}_{k+h} &= \left[ \underline{\dot{f}}(f_{1,k+h}), \ldots,  \underline{\dot{f}}(f_{n,k+h}) \right]^\top
\end{align}
Observe that constraints in \eqref{eq:input_const}, with $\overline{\dot{\gammav}}_{k+h}$ and $\underline{\dot{\gammav}}_{k+h}$ defined as above are highly non-linear. In order to retain linearity of these constraints in the OCP, we consider the following alternative definition for $\overline{\dot{\gammav}}_{k+h}$ and $\underline{\dot{\gammav}}_{k+h}$
\begin{align}
\overline{\dot{\gammav}}_{k+h} &= \left[ \overline{\dot{f}}(\tilde{f}_{1,k+h}), \ldots,  \overline{\dot{f}}(\tilde{f}_{n,k+h}) \right]^\top \\
\underline{\dot{\gammav}}_{k+h} &= \left[ \underline{\dot{f}}(\tilde{f}_{1,k+h}), \ldots,  \underline{\dot{f}}(\tilde{f}_{n,k+h}) \right]^\top
\end{align}
where $\tilde{f}_{i,k+h}$ are independent of the decision variables of the OCP at time $t=kT$. Different choices can be taken, for example keeping the constraints constant along the horizon
$$
\tilde{f}_{i,k+h}= f_{i,k}, \,\,\, h=0, \ldots, N-1.
$$
Alternatively, $\tilde{f}_{i,k+h}$ can be selected in a time-varying fashion based on the solution of the previous OCP obtained at instant $t=(k-1)T$.

\begin{figure*}[t!]
	\centering
	\includegraphics[width=0.8\textwidth]{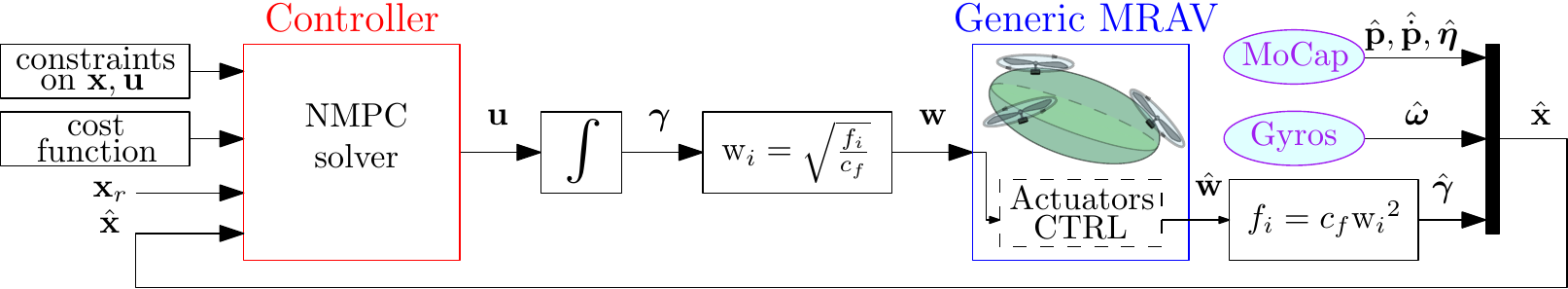}
	\caption{Block diagram of the experimental setup architecture. The main components are highlighted with different colors.}
	\label{fig:block_scheme_exp}
\end{figure*}

The solution to the OCP, at a given time step $k$ consists of the optimal values $\hat{\xv}_{0|k}$, $\ldots$, $\hat{\xv}_{N|k}$, $\hat{\uv}_{0|k}$, $\ldots$, $\hat{\uv}_{N-1|k}$. According to the receding horizon principle~\cite{MAYNE2000}, the input value ${\uv}_{k}=\hat{\uv}_{0|k}$ is applied, and the procedure is repeated at the subsequent time step $k+1$.

Some remarks are due at this point, concerning the problem formulation. 
Regarding the stability-related properties of our scheme, first of all note that the problem addressed here consists of tracking a trajectory generated, possibly without any regard of the vehicle model, by a generic global planner.
In particular, we avoid on purpose any feasibility assumptions of the reference trajectory w.r.t. the robot, in order to test the NMPC framework capability to locally re-generate and track a trajectory which is compatible with the system dynamics and with the actuator constraints. This motivates the claim that the proposed algorithm can seamlessly deal with arbitrarily-designed MRAVs, without the need for a preliminary analysis on the system dynamics. For example, if the system is under-actuated and differentially flat, our framework will automatically recognize such property and exploit it, thus considerably simplifying the reference trajectory generation problem.

Under this general assumption, stability (in a strict sense) of the reference trajectory cannot be guaranteed, since the guarantee to track the given set-point is ensured only provided that the trajectory is generated compatibly with the system dynamics and the actuator constraints. 
For particular implementations of nonlinear MPC, in case of feasible set-points or reference trajectories, the stability of the closed-loop system can be conferred, for example, by designing particular terminal constraints and appropriate terminal penalties on the state. A compelling work reviewing the main essential principles that ensures stability has been presented in the survey~\cite{MAYNE2000}. However, the difficulty in explicitly computing the terminal set and the terminal cost function for general nonlinear systems remains quite dissuasive in real-life applications~\cite{alamir2018stability}.

On the other hand, it has been shown that under the assumption that the reference trajectory is consistent with the vehicle dynamics (e.g., as in~\cite{Alessandretti13}), stability guarantees could be provided by
selecting a sufficiently large prediction horizon length $N$ relying on~\cite{Gruene10,ALAMIR19951353}. Such approach has been applied for path following in a robotic scenario, e.g., in~\cite{MEHREZ20179852}. 
In line with many other works dealing with NMPC applied to aerial vehicles, we preferred to heuristically follow this methodology. To determine the minimum length of the horizon, which directly affects the dimension of the optimization problem, we have performed preliminary simulations with different trajectories, robot models, and prediction horizon lengths, carrying out a trade-off between stability performance and computational burden. Considering that a formal proof of the controller stability is out of the scope of this paper, we leave the analytic study of the optimal prediction horizon length as well as a formal discussion of the closed-loop system stability for future work.
Practically, throughout all the experimental tests, the N\-MPC algorithm was always able to stabilize the robot along a re-computed trajectory, even in case the reference one is (on purpose) not feasible with respect to the robot dynamics and actuator constraints.

In relation to the stability problem, it is worth to briefly discuss the controllability and observability of the systems to be controlled. Regarding the former property, which is related to the capability of the input to affect the evolution of the system state, we leveraged previous theoretical results to assess the existence of feasible input trajectories capable to steer the state of the system towards the tested reference trajectories. Whenever the specific motion was not feasible for the particular dynamic constraints of the platform at hand, we allowed the local re-generation of a feasible trajectory. In this way, we ensure the system controllability to ``the closest'' feasible trajectory in relation to the original reference.
In particular, while fully-actuated systems can track a full-pose decoupled trajectory provided that the actuator constraints are not violated~\cite{2018d-FraCarBicRyl}, we know from previous theoretical results that the trajectory of under-actuated vehicles has to satisfy the flatness-property~\cite{2018c-FaeFraSca}. Moreover, results on the reduced controllability of particular fully-actuated MRAVs after a propeller failure are available from~\cite{2017f-MicRylFra,2018a-MicRylFra}. 
Ultimately, we also tested the controllability of all the presented MRAV systems in relation to the target trajectories in a preliminary phase of extensive simulations. 

As far as the observability is concerned, which describes the possibility of inferring the internal state of a system from knowledge of its external outputs, we do not have any concern in this sense as the full state of the aerial vehicle, cf.~\eqref{eq:state_vector}, is assumed measurable from the available sensors. 
Considering that we do not make use of any state estimator in this work, a formal analysis of the system observability is not needed, in this case. 
Details on the design of a fault-tolerance NMPC scheme for systems with sensor faults, which falls outside the scope of this paper, can be found in~\cite{knudsen2016sensor}.

Another important feature of MPC-based algorithms is recursive feasibility, i.e., the guarantee that the OCP always admits a solution. In our practical implementation we have adopted the widespread solution of guaranteeing it by enforcing the (slightly tightened) constraints in a soft way using slack variables.  
Practically, alongside our extensive experimental campaign, the algorithm was always able to find a solution.

To conclude the extensive presentation of the proposed NMC scheme, the implementation details related to the resolution of the OCP are discussed in the following. 

\subsection{Implementation details for the OCP resolution}

The control algorithm is implemented using the state-of-the-art Real Time Iteration (RTI) scheme, see~\cite{diehl2002real}, embedding the multiple shooting method, cf.~\cite{bock1984multiple}. 
The RTI scheme performs a single sequential quadratic Programming (SQP) iteration to solve the OCP. To do this, a linearization of the system constraints~\eqref{eq:dyn_const} and~\eqref{eq:output} is performed to obtain a quadratic programming (QP) problem, to be solved at each sampling time. To reduce the computational time, in~\cite{chen2017fast} a procedure called partial sensitivity update is proposed, where the constraint linearization is updated only if the dynamics around the generated trajectory exhibits a certain degree of nonlinearity. To reduce the computational complexity, the QP problem is condensed using the algorithms discussed in~\cite{andersson2013general}. The required linear algebra routines are implemented using OpenBLAS~\footnote{\url{https://github.com/xianyi/OpenBLAS}}. The resulting dense QP is solved by qpOASES, see~\cite{Ferreau2014}, which employs on-line active-set method with warm-start strategy.

According to the common practice, the sampling time must be selected to be as small as possible, to make the control system sufficiently reactive. On the other hand, it must also be sufficiently larger than the average computational time. However note that, despite this, there is no guarantee that a solution to the OCP is always available in due time at each time step. If, at a given instant (say at step $k$), the time required to compute the solution is occasionally larger than a given threshold, a back-up solution must be taken to guarantee reliability of the control system. In this paper, this solution consists of taking the possibly sub-optimal but admissible value $\uv_k=\hat{\uv}_{1|k-1}$, computed as part of the solution to the OCP at time $k-1$.

\section{Experimental validation}

In this section we show and thoroughly discuss the experimental results obtained from the application of the proposed NMPC algorithm to the aerial robot prototypes built, and in some cases conceived, in the laboratory facility of LAAS-CNRS - the interested reader is as well referred to the attached multimedia file.
First of all, we present the experimental setup, with a focus on the description of both the hardware and the software components, and on the implementation details of the NMPC strategy. Then, we analyze the outcomes of the experiments achieved with two different kinds of MRAVs, i.e., an under-actuated UDT quadrotor and a MDT (in particular, also fully-actuated) hexarotor with tilted propellers.
The goal of this investigation is to demonstrate the precision of our approach and, above all, its potential applicability to any arbitrarily-designed MRAV. The robots are required to perform tracking experiments: the reference trajectories are designed both to test the re-generation capability of the algorithm, to highlight the different behaviors of UDT and MDT platforms, and to assess the solution compliance with the constraints previously identified in the case of fast maneuvers.

In order to better appreciate the results achieved in the experimental validation with the proposed NMPC framework, we point the reader to the attached video.

\subsection{Experimental setup}\label{subsec:setup}

The experimental setup architecture, whose block diagram is portrayed in Fig.~\ref{fig:block_scheme_exp}, can be conceptually divided into three main components: the NMPC controller, which periodically computes the input of the actuator controllers (the ESCs), the physical aerial robot to be controlled, and the sensors, used to retrieve the information about the MRAV state that is employed as feedback in the closed-loop control strategy. Each block exchanges information with the others thanks to a properly-designed software architecture.

The predictive controller is implemented using MAT\-MPC, a recently-developed MATLAB-based nonlinear MPC toolbox, see~\cite{chen2019matmpc}. Its algorithmic routines are written using MATLAB C API and available as MEX functions. The tool supports fixed step Runge-Kutta (RK) integrator for multiple shooting and obtains the derivatives that are needed to perform the optimization from the toolbox CasADi\footnote{\url{https://github.com/casadi/casadi/wiki}}.
The OCP described by~\eqref{eq:cost_function}-\eqref{eq:input_const} is solved by the external solver qp\-OASES\footnote{\url{https://projects.coin-or.org/qpOASES/wiki}} by~\cite{Ferreau2014}, integrating the RTI algorithm.
Such an implementation has been chosen mainly due to the particular ease of test and development of MATLAB/Simulink\textsuperscript{\textregistered} compared to pure C/C++.
The presented NMPC algorithm is executed on a ground-station PC equipped with an Intel\textsuperscript{\textregistered} 2.60GHz Core\textsuperscript{TM} i7-6700HQ CPU (x8) and 32 GB RAM which runs the Linux Ubuntu 16.04 LTS operating system.
As it can be observed from Fig.~\ref{fig:block_scheme_exp}, the control input $\uv$, which provides the actuators' force derivatives references, is integrated and then converted into a rotor velocity command \textbf{w}, thanks to the inversion of the force generation model. The resulting velocity set-points are finally transferred to the module of the low-level controllers on-board the aerial platform, by means of a serial cable.

\begin{figure}[t]
	\begin{center}
		\includegraphics[width=0.45\columnwidth]{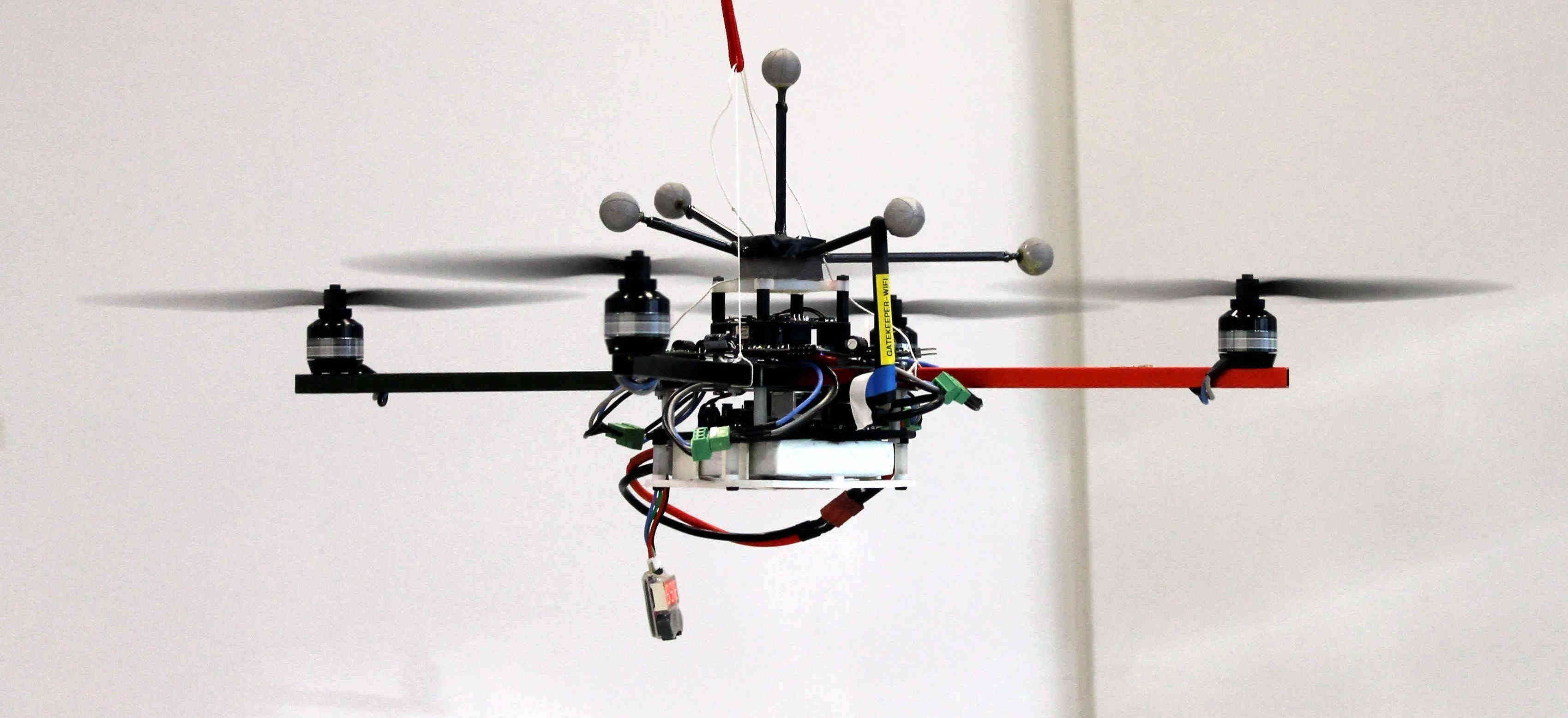}
		\includegraphics[width=0.45\columnwidth]{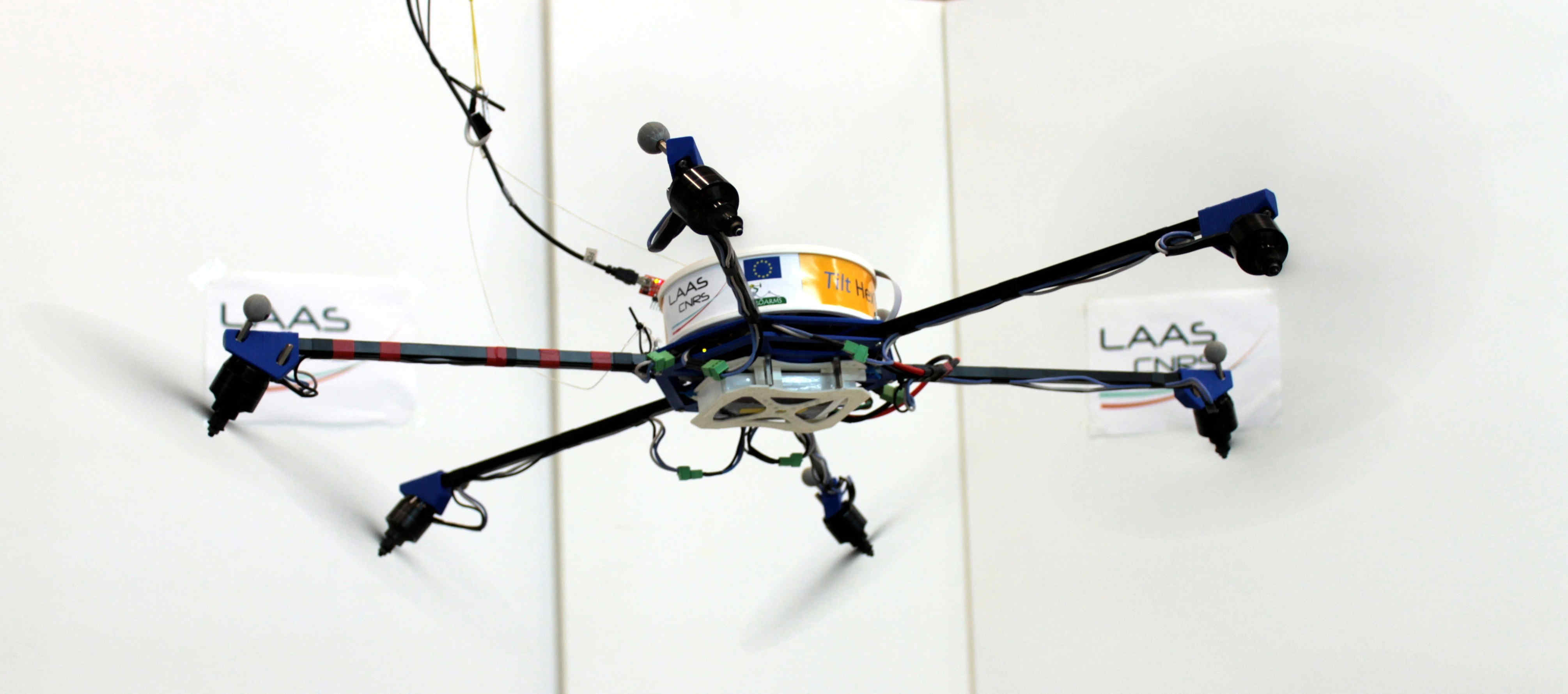}
	\end{center}
	\caption{Photos of the quadrotor (left) and the Tilt-Hex (right).}
	\label{fig:robots}
\end{figure}

As far as the aerial robots are concerned, we tested our control algorithm with the two heterogeneous MRAVs depicted in Fig.~\ref{fig:robots}. The first one, shown on the left, is an under-actuated UDT platform, a quadrotor, with four collinear rotors. Apart from some custom-made features realized in-house with 3D printed components, like the battery support, most of the platform is built by assembling off-the-shelf parts from MikroKopter. The second robot, illustrated on the right of the Fig.~\ref{fig:robots}, is a fully-actuated MDT platform with six non-collinear tilted rotors, from which it inherits the name `Tilt-Hex'. In this case, the prototype has been completely designed and manufactured in our laboratory, and has already been presented in some of our previous contributions~\cite{2018d-FraCarBicRyl,2019h-RylMusPieCatAntCacFra}.
The values of the physical parameters used in the MRAVs models are summarized in Tabs.~\ref{tab:robots_parameters_quad} and~\ref{tab:robots_parameters_tilthex}. In particular, $\alpha$ and $\beta$ are defined as the actuator rotation angles around $\xv_{A_i}$ and $\yv_{A_i}$, respectively, as shown in Fig.~\ref{fig:modeling}.

\begin{table}[t]
	\renewcommand\arraystretch{1.3}
	\caption{Physical parameters of the quadrotor.}
	\label{tab:robots_parameters_quad}
	\centering
	\resizebox{\figWidth\columnwidth}{!}{%
	\begin{tabular}{CCC}
		\hline
		\multicolumn{3}{c}{\textbf{Quadrotor}} \\
		\hline
		\text{Parameter}  & \text{Value} & \text{Unit} \\
		\hline
		m & 1.042 & \si{Kg} \\
		\Jm(:,1) & [0.015\ 0 \ 0]^{\top} & \si{Kg\,m^2} \\
		\Jm(:,2) & [0 \ 0.015 \ 0]^{\top} & \si{Kg\,m^2} \\
		\Jm(:,3) & [0 \ 0 \ 0.015]^{\top} & \si{Kg\,m^2} \\
		c_i & (-1)^{i-1} & [\ ] \\
		c_f^{\tau} & \num{1.69e-2} & \si{m} \\
		c_f & \num{5.95e-4} & \si{N/Hz^2} \\
		\hline
		\Rm_{A_i}^B & \Rm_z \big ((i-1)\frac{\pi}{2})\big)\Rm_x(\alpha)\Rm_y(\beta) & [\ ] \\
		\pv_{A_i}^B & \Rm_z \big((i-1)\frac{\pi}{2})\big) [l\ 0\ 0]^{\top} & [\ ] \\
		\alpha & 0 & \si{deg} \\
		\beta & 0 & \si{deg} \\
		l & 0.23 & \si{m} \\
		\hline
	\end{tabular}	} 
\end{table}

The main sensors integrated in our experimental framework are the onboard gyroscope, the Motion Capture system, and speedometers of each propeller rotational speed:
\begin{itemize}
\item the Gyroscope measures the rotational velocity of the vehicle around each of the body frame axis;
\item the Motion Capture (MoCap) system provides the information regarding the robot position and orientation w.r.t. the inertial reference frame, whose origin is fixed in a particular point of the robots workspace.
The platform linear velocity is numerically computed online from the position measurements, using multi-sample least squares model fitting;
\item the rotor spinning velocities are measured by the low-level ESC controller by computing the time elapsed between two phase switches (which depends on the motor number of poles) and reducing the measurement noise with an exponential moving average filter.  Ultimately, the rotor velocities are converted into the actuator forces, thanks to the force generation model, and used to complete the information of the measured full-state $\hat{\xv}$ of the MRAV.
\end{itemize}
Note that the accelerometers have been disregarded from the sensor fusion since we assessed that the noise in their measurements was causing an offset in the estimation of the linear velocity, which motivated the numerical computation of the latter. In general, the effect of such velocity offset on the tracking performance is quite more evident on predictive controllers w.r.t. reactive ones, given the fact that a wrong state estimation generates an erroneous evolution of the model internally simulated and, in turn, a misleading control input that finally produces an inaccurate trajectory tracking.

\begin{table}[t]
	\renewcommand\arraystretch{1.3}
	\caption{Physical parameters of the Tilt-Hex.}
	\label{tab:robots_parameters_tilthex}
	\centering	
	\resizebox{\figWidth\columnwidth}{!}{%
	\begin{tabular}{CCC}
		\hline
		\multicolumn{3}{c}{\textbf{Tilt-Hex}} \\
		\hline
		\text{Parameter}  & \text{Value} & \text{Unit} \\
		\hline
		m & 1.86 & \si{Kg} \\
		\Jm(:,1) & [0.11 \ 0 \ 0]^{\top} & \si{Kg\,m^2} \\
		\Jm(:,2) & [0 \ 0.11 \ 0]^{\top} & \si{Kg\,m^2} \\
		\Jm(:,3) & [0 \ 0 \ 0.19]^{\top} & \si{Kg\,m^2} \\
		c_i & (-1)^{i-1} & [\ ] \\
		c_f^{\tau} & \num{1.9e-2} & \si{m} \\
		c_f & \num{9.9e-4} & \si{N/Hz^2} \\
		\hline
		\Rm_{A_i}^B & \Rm_z \big ((i-1)\frac{\pi}{3})\big)\Rm_x(\alpha_i)\Rm_y(\beta) & [\ ] \\
		\pv_{A_i}^B & \Rm_z \big((i-1)\frac{\pi}{3})\big) [\ell\ 0\ 0]^{\top} & [\ ] \\
		\alpha_i & (-1)^i\ 35 & \si{deg} \\
		\beta & -25 & \si{deg} \\
		\ell & 0.368 & \si{m} \\
		\hline
	\end{tabular}	} 
\end{table}

In order to design the software architecture, we rely on the GenoM3\footnote{\url{https://git.openrobots.org/projects/genom3/wiki}} abstraction level, which allows to encapsulate software functions inside independent components. More in detail, it is used as a wrapper for the robot low-level controller and the sensors.
This allows to obtain high flexibility in the development and in the use of the components. With reference to our architecture, the software in MATLAB /Simulink\textsuperscript{\textregistered} communicates with the GenoM3 modules using the Robot Operating System (ROS) middleware, which is compliant with the soft real-time constraints required for our experiments, i.e., a control bandwidth larger than $200\si{Hz}$ and a latency smaller than $10\si{ms}$. Since MATLAB/Simulink\textsuperscript{\textregistered} is not meant for a hard real-time execution, the hardware is commanded via the GenoM3 components, which essentially behave like drivers.

\subsubsection{Implementation details}

For all the experiments presented in this paper, we chose a prediction horizon of $t_H=1\si{s}$, sampled at $N+1=11$ shooting points. Therefore, the discretization time of the nonlinear MPC algorithm, being the length of one of the $N$ intervals, results $T=0.1\si{s}$. Even though the internal MPC prediction is performed at $10\si{Hz}$, the controller runs at a frequency  always larger than $200\si{Hz}$. Such technique, employed by many state-of-the-art contributions, e.g.,~\cite{kamel2017linear}, allows the predictive algorithm to simulate the model along a wider prediction horizon with less computational effort. Indeed, as observed by~\cite{falanga2018pampc}, the number of discretization nodes roughly increases the computational time $t_{\text{solv}}$ by $O(\allowbreak N^2)$. Basically, one should guarantees a control sample time $T_{\text{ctrl}}$ at least equal to time $t_{\text{solv}}$ needed for the algorithm to solve the OCP. On the other hand, the prediction horizon should be long enough to cover at least the time of one controller iteration. In mathematical terms, this translates in the following chain of inequalities
\begin{align}
\label{eq:timing}
t_{\text{solv}} \le T_{\text{ctrl}} \le T \le t_H
\end{align}

As far as the representation of the robot orientation is concerned, for the particular experiments presented in this paper we decided to use a minimal parametrization with three angles, in particular the $3-2-1$ one (yaw-pitch-roll), i.e.,
\begin{align}
\label{eq:YPR}
	\etav =
	\begin{bmatrix}
	\phi&\theta&\psi
	\end{bmatrix}^{\top}
\end{align}
With reference to this ordered sequence, we have that
\begin{align}
\label{eq:R_convention}
\begin{aligned}
	\Rm &= \Rm_{\zv}(\psi) \Rm_{\yv}(\theta) \Rm_{\xv}(\phi) \\
	&=
	\begin{bmatrix}
	c_{\theta}c_{\psi} & s_{\phi}s_{\theta}c_{\psi}-c_{\phi}s_{\psi} & s_{\phi}s_{\psi}+c_{\phi}s_{\theta}c_{\psi} \\
	c_{\theta}s_{\psi} & c_{\phi}c_{\psi}+s_{\phi}s_{\theta}s_{\psi} & c_{\phi}s_{\theta}s_{\psi}-s_{\phi}c_{\psi} \\
	-s_{\theta} & s_{\phi}c_{\theta} & c_{\phi}c_{\theta}
	\end{bmatrix}
\end{aligned}
\end{align}
where $\Rm_{\bullet}(\alpha)$ denotes a rotation around one of the main body frame axes $\{ \xv, \, \yv, \, \zv \}_B$ of an angle $\alpha$, while $s_{\alpha}$, $c_{\alpha}$ indicate $\sin(\alpha)$ and $\cos(\alpha)$, respectively. Using this convention, we can express the body frame angular velocity as a function of the vector $\dot{\etav}$, that contains the so-called \emph{Euler rates}
\begin{align}
\label{eq:omega_B_rates}
\omegav = \Tm \dot{\etav}
\end{align}
In particular, with reference to the specific parametrization of~\eqref{eq:R_convention}, we have
\begin{align}
\label{eq:T_B}
	\Tm =
	\begin{bmatrix}
	1 & 0 & -s_{\theta} \\
	0 & c_{\phi} & s_{\phi}c_{\theta} \\
	0 & -s_{\phi} & c_{\phi}c_{\theta}
	\end{bmatrix}.
\end{align}
Inverting~\eqref{eq:omega_B_rates} allows to write explicitly the Euler rates as a function of the body angular velocity (expressed in body frame) in the NMPC model dynamics.
This representation, like all the minimal parametrizations given by three angles, has a singularity, which in the specific case occurs when $\theta=\pi/2$. In general, all these conventions should be avoided if the robot orientation is supposed to evolve in the complete $SO(3)$ manifold. However, in the particular case of the trajectories that we have tested, we safely used this representation by explicitly avoiding singular configurations for the platform pose.
We chose to not use the re-arranged elements of $\Rm$ or a unit quaternion for a simple matter of convenience. Indeed, in such cases a larger state vector would have been needed.
Furthermore, additional constraints, e.g., the orthogonality of the rotation matrix or the unitary-norm for the quaternion, should have been added in the resolution of the OCP, thus increasing the solver computational time and, by consequence, slowing down the available bandwidth of the controller. This can easily be dealt with, of course, by using a more powerful computation unit.
It should be underlined that the proposed framework does not depend on the particular orientation representation and easily adapts to the others without the need to deal with additional theoretical issues.

The cost function weights in~\eqref{eq:cost_function} are specified at the beginning of the description of each experiment and simulation. In general, they have been chosen on a case-dependent basis taking into account heuristic considerations and often following a \emph{trial-and-error} procedure. The automatic tuning of such weights is an important topic which is left for future work. Throughout all experiments and simulations presented in the paper, the input terms in the cost function have not been considered, i.e., the entries of the weights $\Rm_h$ related to the input are equal to zero. 
This has been done with the goal to exploit the MRAVs potentialities until their limits by taking advantage of the actuator dynamics up to their bounds. Therefore, we decided to test our NMPC algorithm by discarding these regularization terms.
In all the performed tests, including the most agile ones, we never encountered problems in the regularity of the input evolution.
Furthermore, despite the strong accelerations of some of the reference state trajectories to the NMPC algorithm, we never triggered the activation of the slack variables.

Finally, regarding the choice of the input bounds along the prediction horizon, we selected
\begin{align}
\label{eq:input_bounds_horizon}
\tilde{f}_{i,k+h}= f_{i,k}, \,\,\, h=0, \ldots, N-1
\end{align}
i.e., the limits are kept constant along the future window. This choice has been motivated by a matter of simplicity of implementation. 
A more rigorous choice could be to select the time-varying $\tilde{f}_{i,k+h}$ in relation to the predicted state evolution at the previous control step for $t=(k-1)T$. The comparison within the results produced by these two configurations is also left as future investigation.

\subsection{Experiments with the quadrotor}

According to the choices we made for the state and input vectors, defined by~\eqref{eq:state_vector}-\eqref{eq:input_vector}, and for the orientation description associated to~\eqref{eq:YPR}, in the case of the quadrotor model we have $\xv \in \mathbb{R}^{16}$ and $\uv \in \mathbb{R}^{4}$. With this configuration, the average NMPC solver time is $t_{\text{solv}}=3.5\si{ms}$. In the following, we present the tracking results obtained by the quadrotor with two different trajectories. The first one combines a sinusoidal chirp motion along one component of the position with a steadily horizontal and constant-heading desired orientation. Such dynamic and decoupled motion, which was designed on purpose to be dynamically unfeasible for this vehicle (see previous discussion), is also given as reference to the Tilt-Hex in order to compare the performances of the two platforms and, in particular, to highlight the different behaviors of UDT and MDT aerial robots. For a numerical comparison of the results, we refer the reader to Tab.~\ref{tab:exp_comp}. On the other hand, we designed the second reference motion in order to test the controller compliance with the actuator bounds, when dealing with a discontinuous trajectory.
In particular, these tests highlight the importance of the control compliance with the input constraints for the preservation of the system stability. Also in this case, comparative numerical results are outlined in Tab.~\ref{tab:exp_comp}.

\begin{table}[t]
	\renewcommand\arraystretch{1.3}
	\caption{Parameters used in the quadrotor experiments.}
	\label{tab:qr_exp}
	\centering
	\resizebox{\figWidth\columnwidth}{!}{%
	\begin{tabular}{CCC}
		\hline
		\text{Parameter}  & \text{Value} & \text{Unit} \\
		\hline
		\nu & 1.2 & \si{m} \\
		\xi & 0.025 & \frac{\si{rad}}{\si{s^2}} \\
		\bar{t} & 44.84 & \si{s} \\
		\boldsymbol{\epsilon}_{\pv} & [-0.5\, 0.2\, 0.2]^{\top} & \si{m} \\
		\rho & 0.125:0.125:1.75 & [\ ] \\
		\hline
		\Qm_{\pv}(j,j)|_{j=1,2,3} & 500,300,300 & [\ ] \\
		\Qm_{\dot{\pv}}(j,j)|_{j=1,2,3} & 1.9,1.9,1.9 & [\ ] \\
		\Qm_{\etav}(j,j)|_{j=1,2,3} & 0.1,0.1,40 & [\ ] \\
		\Qm_{\omegav}(j,j)|_{j=1,2,3} & 0.15,0.15,0.15 & [\ ] \\
		\Qm_{\ddot{\pv}}(j,j)|_{j=1,2,3} & 0,0,0 & [\ ] \\
		\Qm_{\dot{\omegav}}(j,j)|_{j=1,2,3} & 0,0,0 & [\ ] \\
		\Rm_h(j,j)|_{j=1,\dots{},4} & 0,0,0,0 & [\ ] \\
		\hline
	\end{tabular}	
} 
\end{table}

\subsubsection{Position chirp trajectory}

In the first experiment, the quadrotor is required to track a position reference $\pv_r=[c(t) \, 0 \, 0]^{\top}$, where the chirp signal $c(t)$ is a sine with varying frequency, with amplitude $\nu=1.2$ m, and a triangular frequency that linearly increases from $\xi_0=0$ rad/s to $\xi_{\bar{t}}=1.12$ rad/s with a slope $\xi=0.025$ $\si{rad/s^2}$ in the interval $[0,\bar{t}[$, $\bar{t}=44.84$ $\si{s}$, and then decreases with a slope $-\xi$ in the interval $[\bar{t},2\bar{t}[$. In mathematical terms, this translates into
\begin{align}
	\label{eq:chirp}
	c(t) \!=\! \nu \sin{\bigl(\xi(t)\ t\bigr)},\
	\xi(t)
	\!=\!
	\begin{cases}
		\xi t &\!\text{if $t\in[0,\bar{t}[$}\\
	   \xi (\bar{t}-t) &\!\text{if $t\in[\bar{t},2\bar{t}[$}
	\end{cases}
\end{align}
On the other hand, the attitude reference is constantly $\etav_r=[0 \, 0 \, 0]^{\top}$. Moreover, the desired position derivatives $\dot{\pv}_r$, $\ddot{\pv}_r$ and the rotational derivatives $\omegav_r$, $\dot{\omegav}_r$ are consistent with the definitions of $\pv_r$ and $\etav_r$, respectively.
Regarding the form of the diagonal matrices employed to weight the different error terms inside the cost function, we used $\Qm_k = diag(\Qm_{\pv},\Qm_{\dot{\pv}},\allowbreak\Qm_{\etav},\Qm_{\omegav},\Qm_{\ddot{\pv}},\Qm_{\dot{\omegav}}), \ \forall k \in \{ 0,\dots,N \}$.
The values of the trajectory parameters and the diagonal sub-blocks $\Qm_{\bullet}$ chosen for the quadrotor experiments are displayed in Tab.~\ref{tab:qr_exp}.
\textbf{•}The latter ones are the result of a trial-and-error procedure that we performed, in compliance with some heuristic guidelines, in order to obtain satisfactory tracking performance.
In particular, the weights associated with the orientation error have been selected much smaller than the ones related to the position error, given the impossibility for the particular platform to track the roll and pitch references. On the other hand, the yaw error has a larger impact w.r.t. the other two angular components, as the authority around this axis is still present despite the under-actuation. Finally, the feed-forward terms related to $\ddot{\pv}_d$ and $\dot{\omegav}_d$ turned out to be not very relevant in these experiments. This explains why their entries are weighted with null gains.

With reference to the desired trajectory, the platform is required (if possible) to keep a flat orientation while moving laterally along the x-axis. This motion is unfeasible for a UDT aerial vehicle, since the only way it has to steer the thrust force is by re-orienting its body chassis, with no possibility to keep it horizontal. We provided an unfeasible rotational profile on purpose with the intent of showing that the proposed NMPC scheme can manage the re-generation and tracking of a generic trajectory, subject to the limitations imposed by the particular MRAV under analysis, without the need to resort, e.g., to differential flatness. In this way, the user does not need to explicitly compute the particular platform-dependent feasible trajectory, but can delegate this task to the predictive controller, which automatically adapts the reference profile according to the robot constraints. Of course, position errors are made even smaller if a feasible reference trajectory is available and is provided to the controller. However, this was not the major  point to be shown in the experiments.

\begin{figure}[t!]
	\begin{center}
		\includegraphics[width=\figWidth\columnwidth]{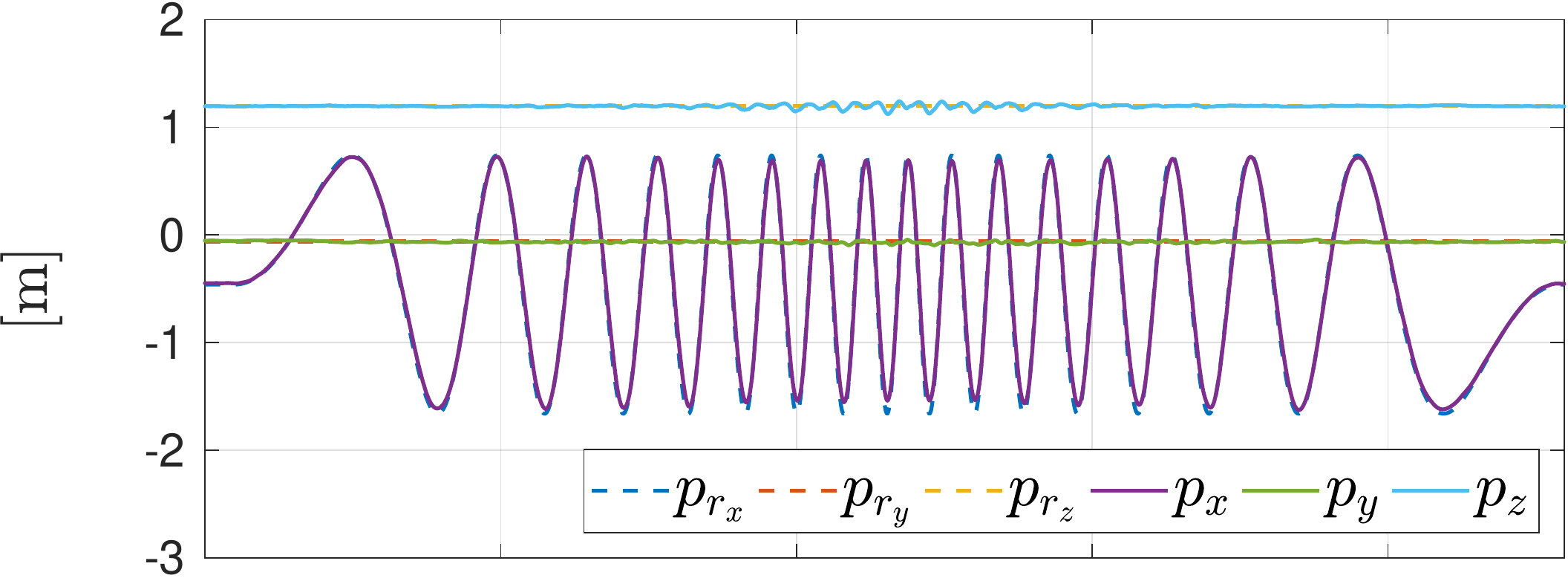}\\
		\includegraphics[width=\figWidth\columnwidth]{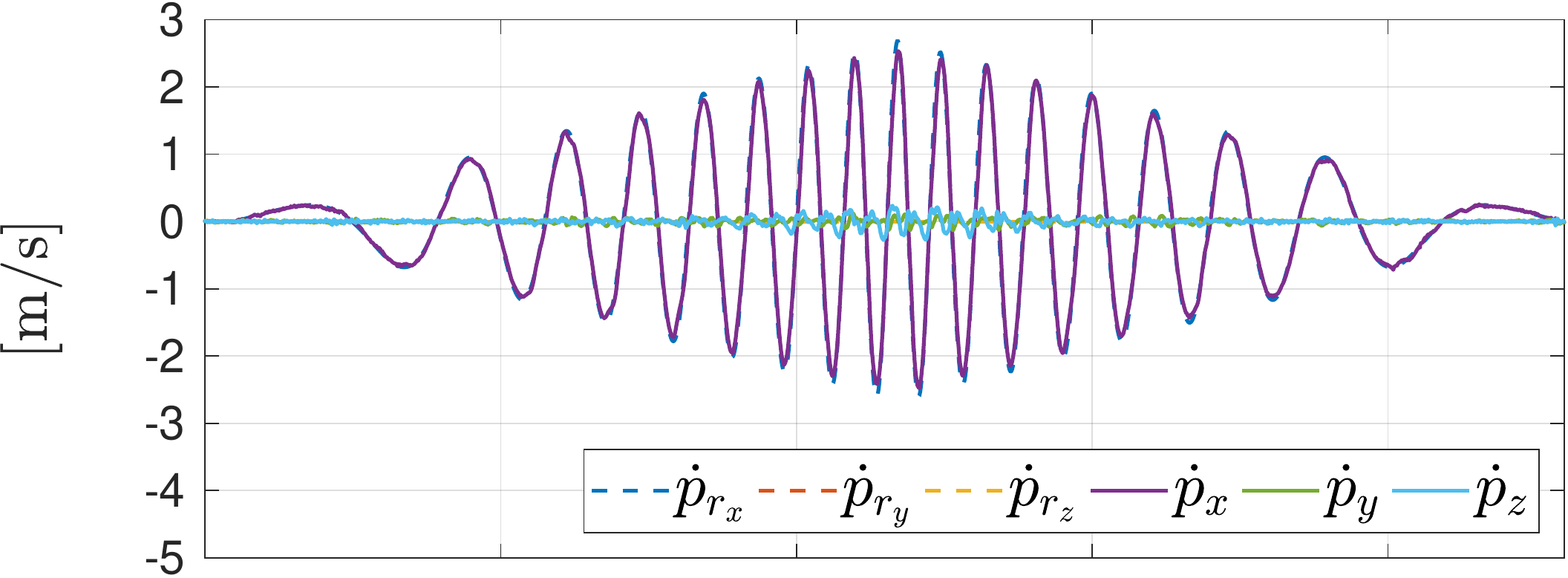}\\
		\includegraphics[width=\figWidth\columnwidth]{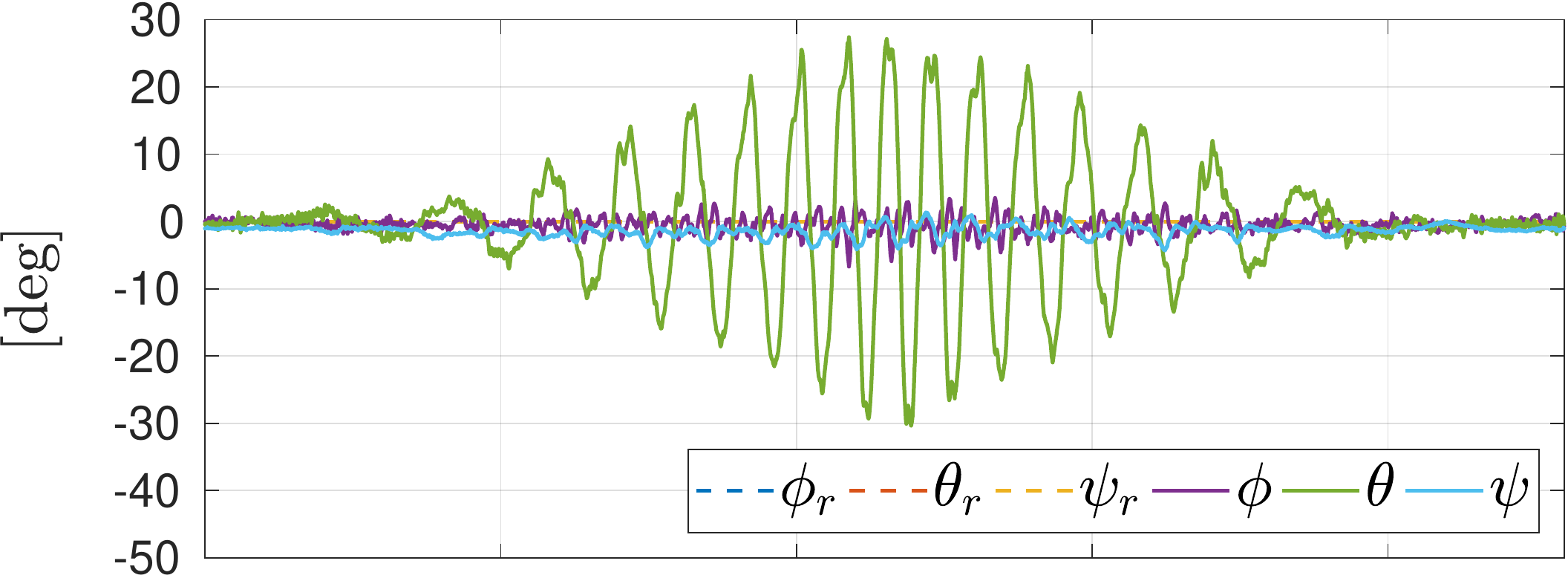}\\
		\includegraphics[width=\figWidth\columnwidth]{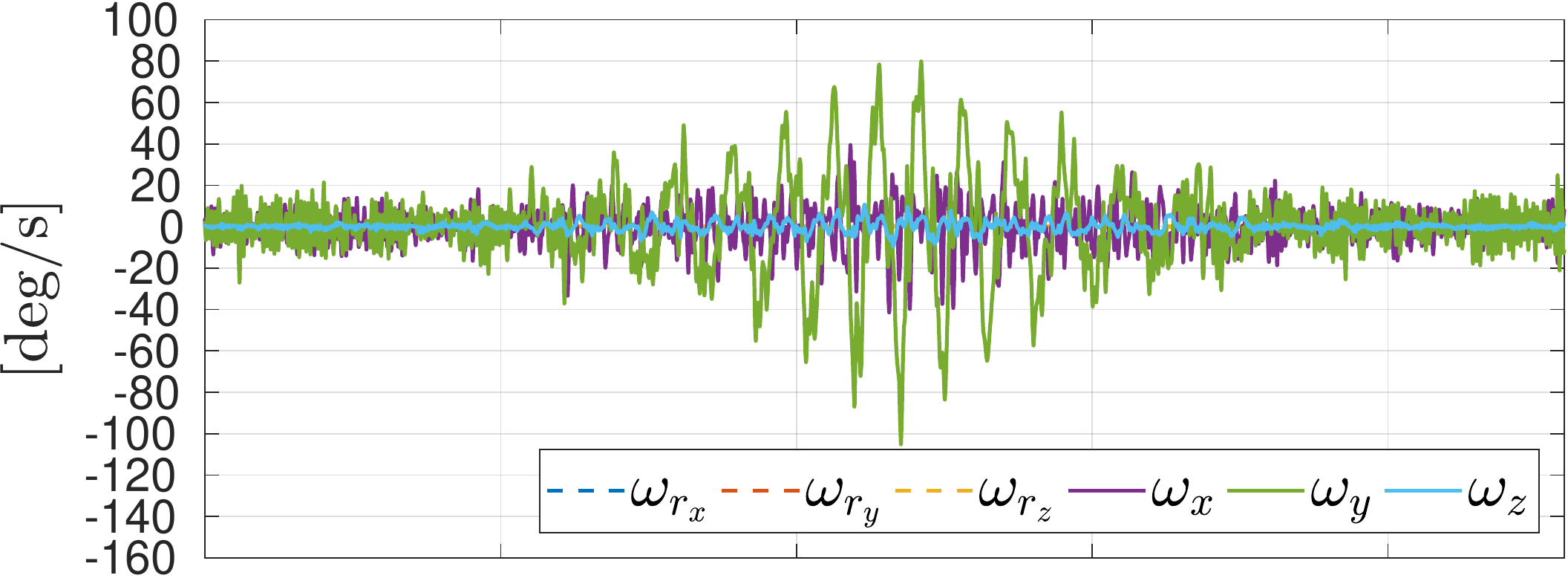}\\
		\includegraphics[width=\figWidth\columnwidth]{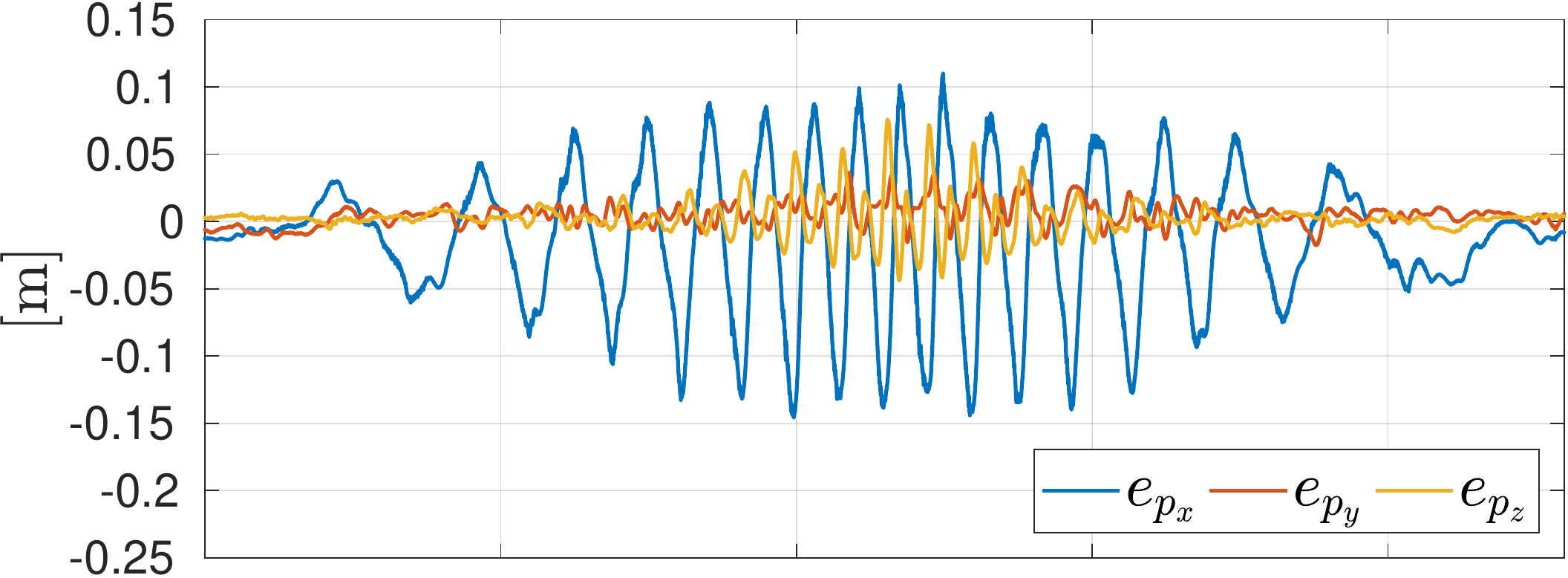}\\
		\includegraphics[width=\figWidth\columnwidth]{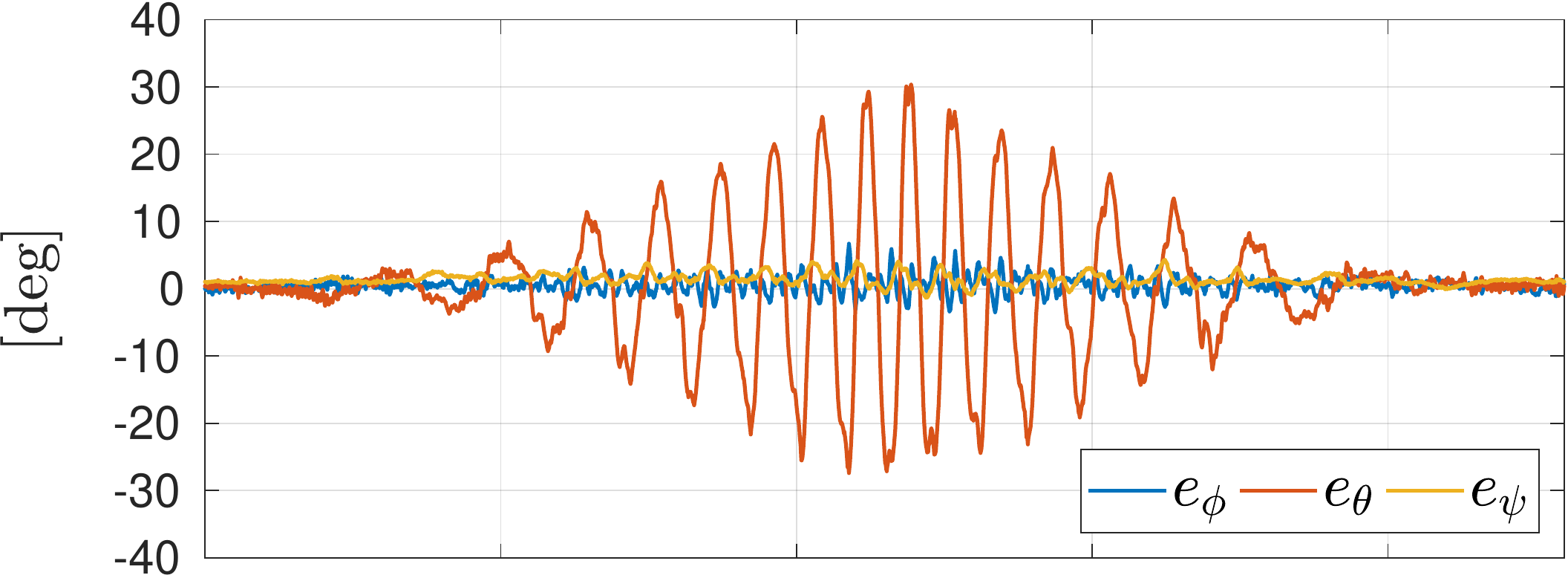}\\
		\includegraphics[width=\figWidth\columnwidth]{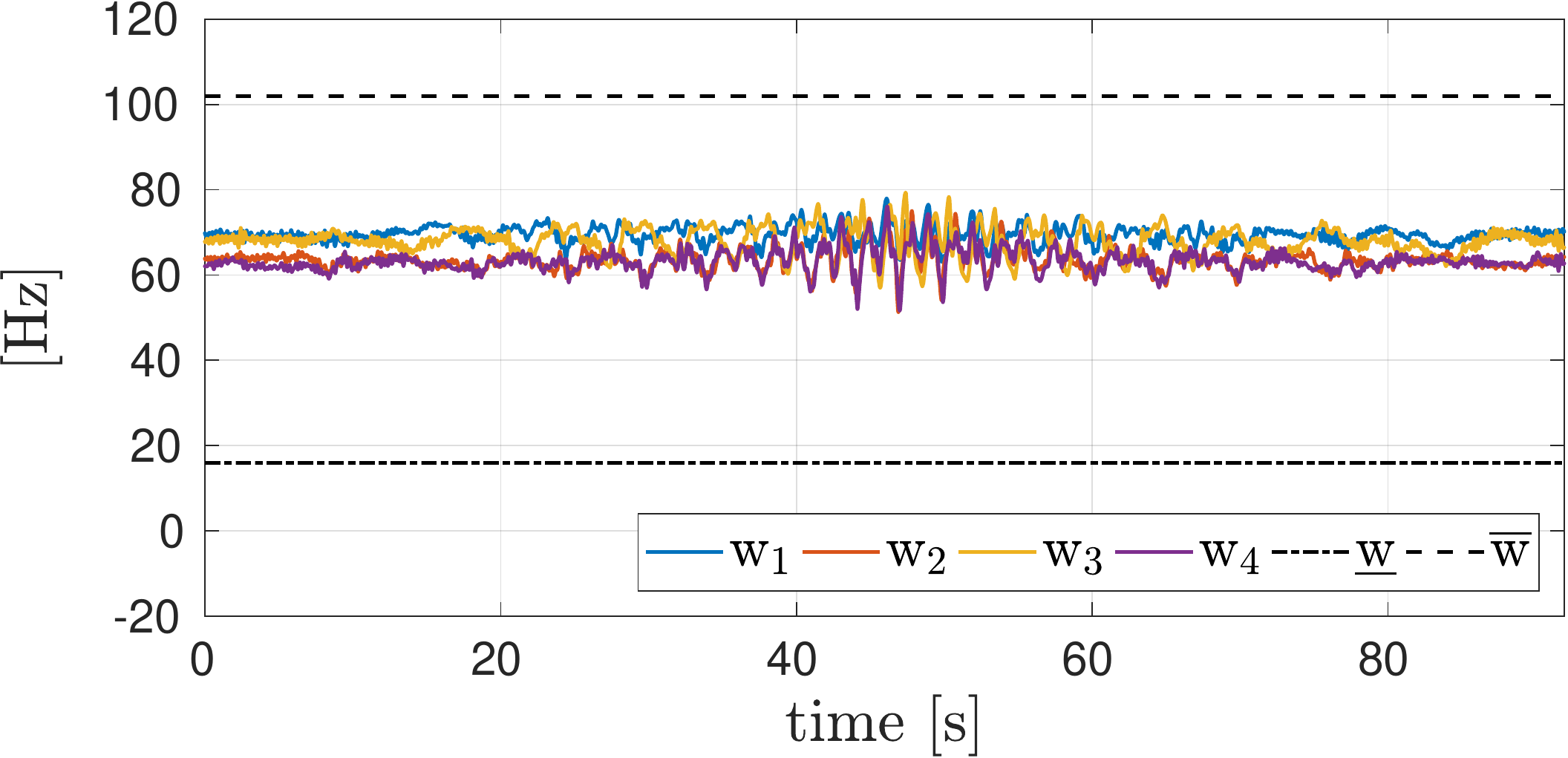}
	\end{center}
	\caption{Plots of the quadrotor performing a chirp trajectory on the x-axis. From top to bottom, the position, linear velocity, orientation and angular velocity tracking, the position and orientation errors, and the actuator spinning velocities.}
	\label{fig:plots_QR_chirp_traj}
\end{figure}

\begin{figure}[t!]
	\begin{center}
		\includegraphics[width=\figWidth\columnwidth]{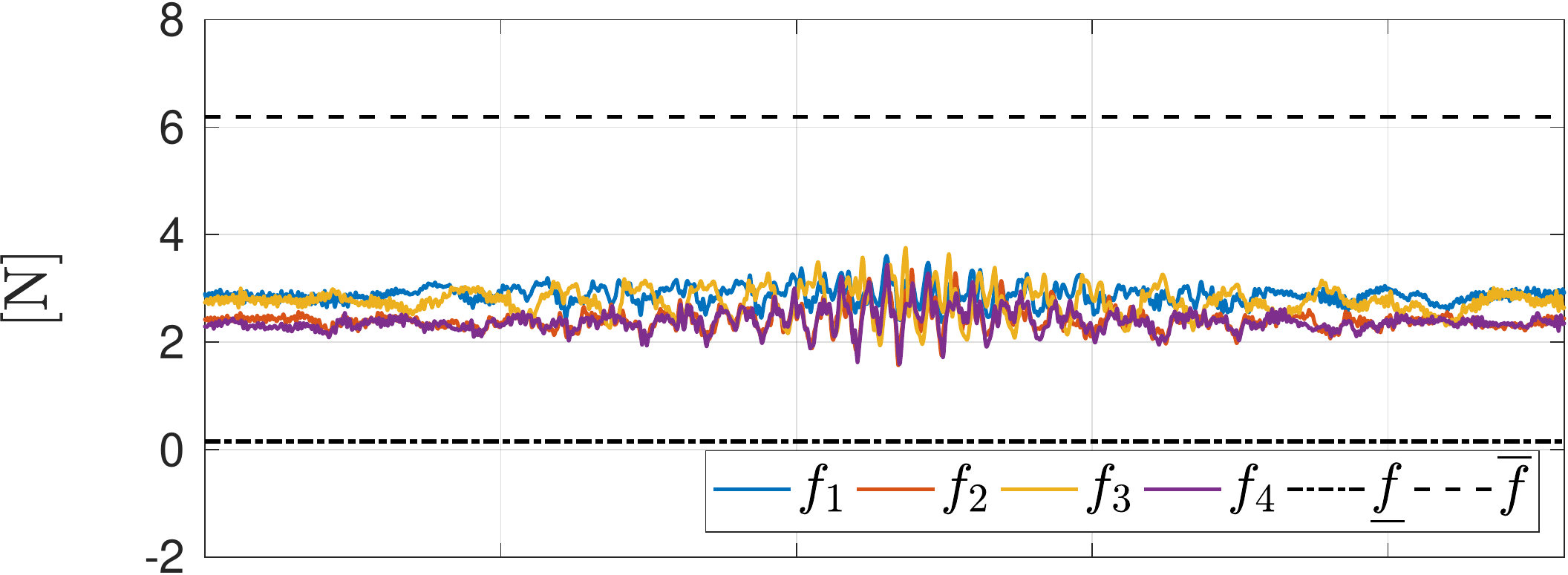}\\
		\includegraphics[width=\figWidth\columnwidth]{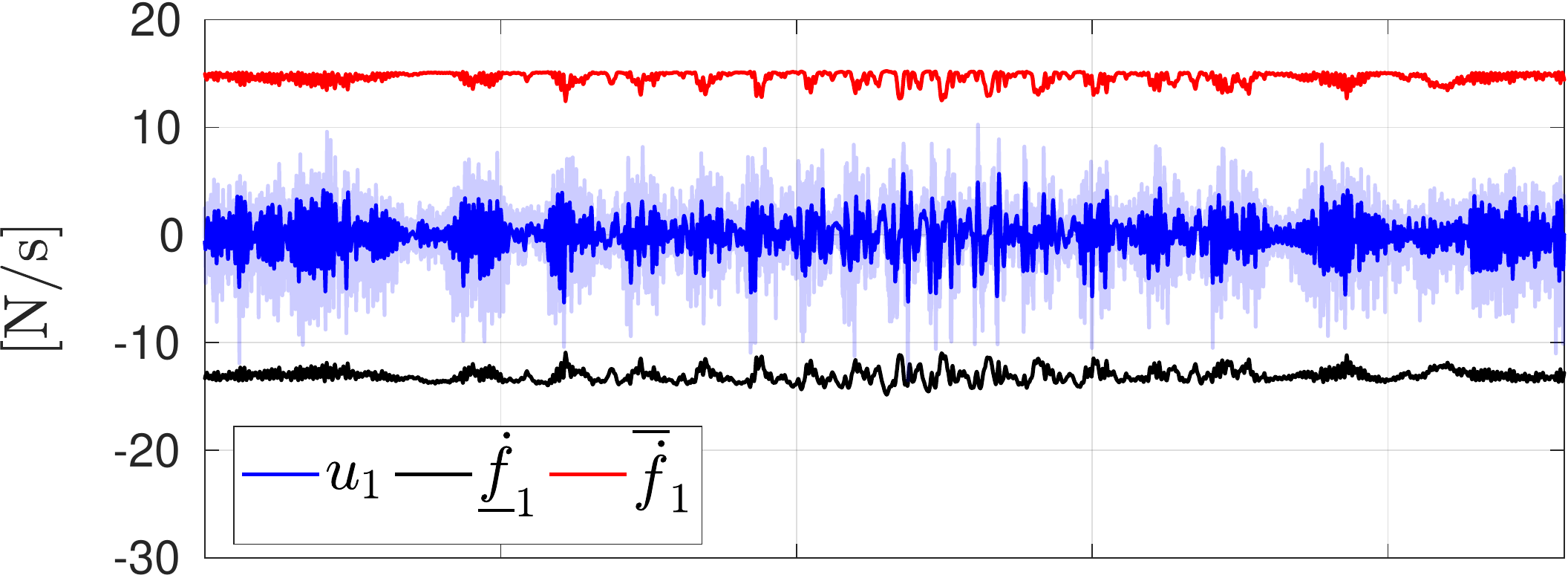}\\
		\includegraphics[width=\figWidth\columnwidth]{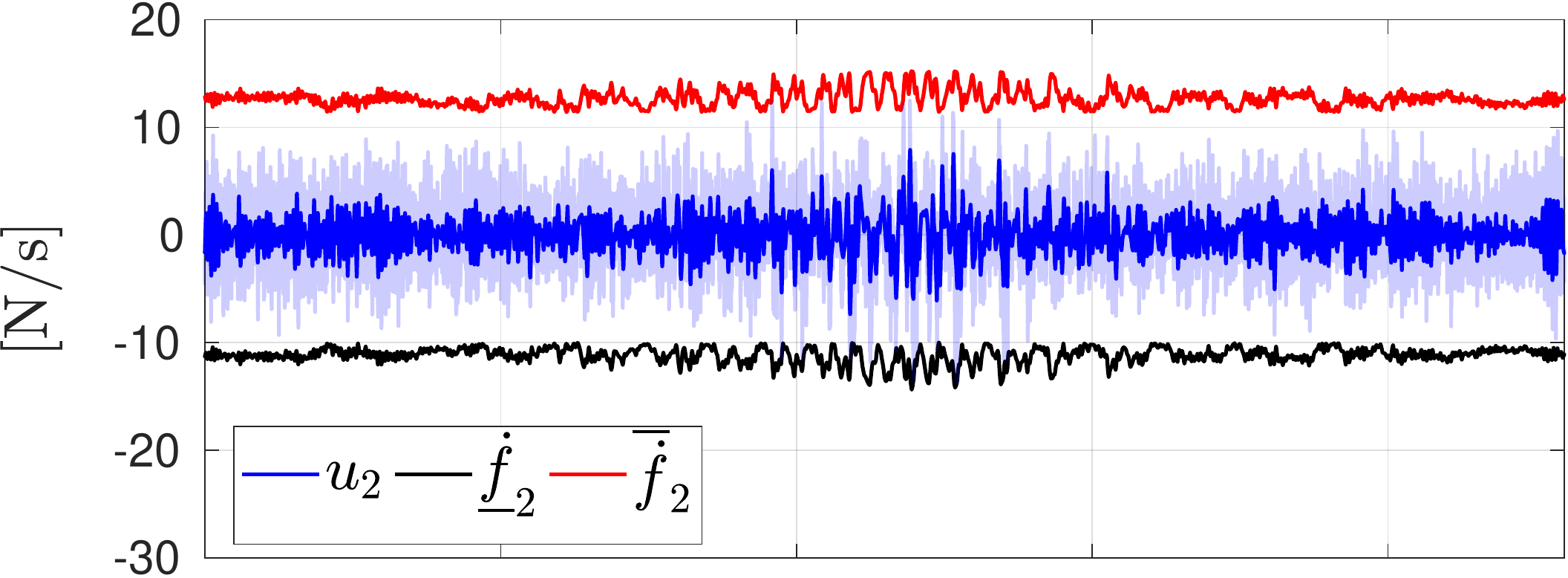}\\
		\includegraphics[width=\figWidth\columnwidth]{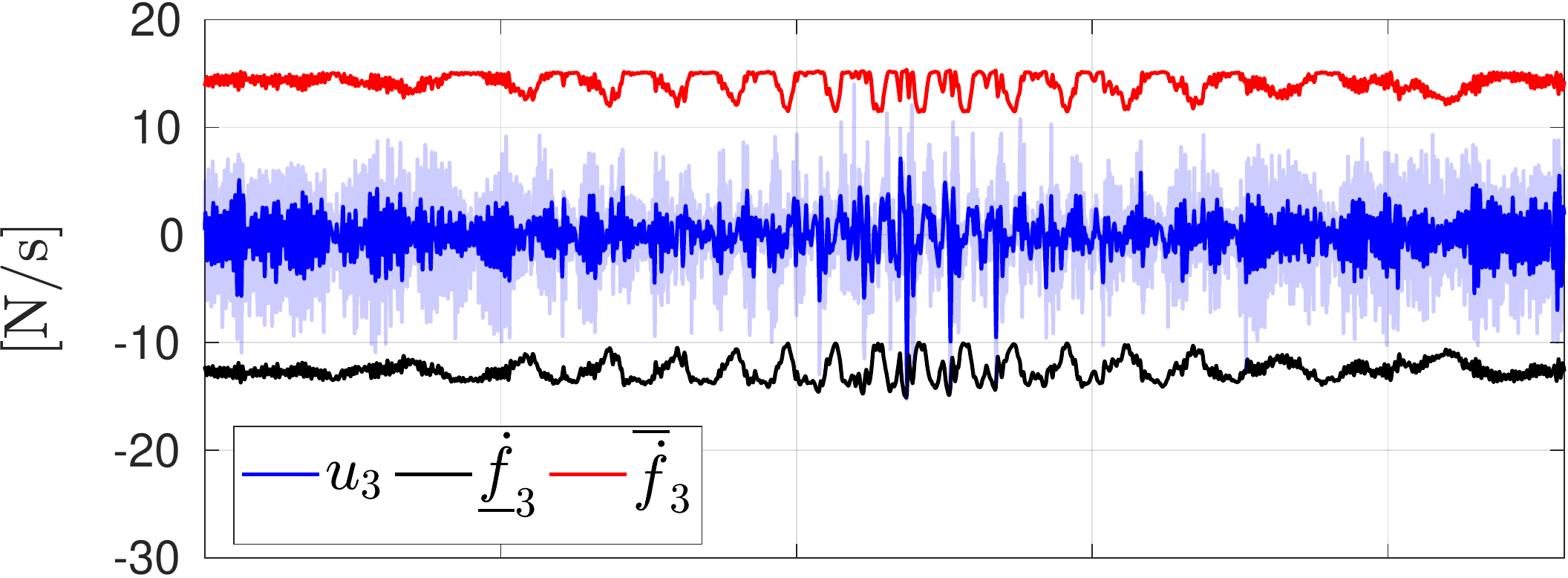}\\
		\includegraphics[width=\figWidth\columnwidth]{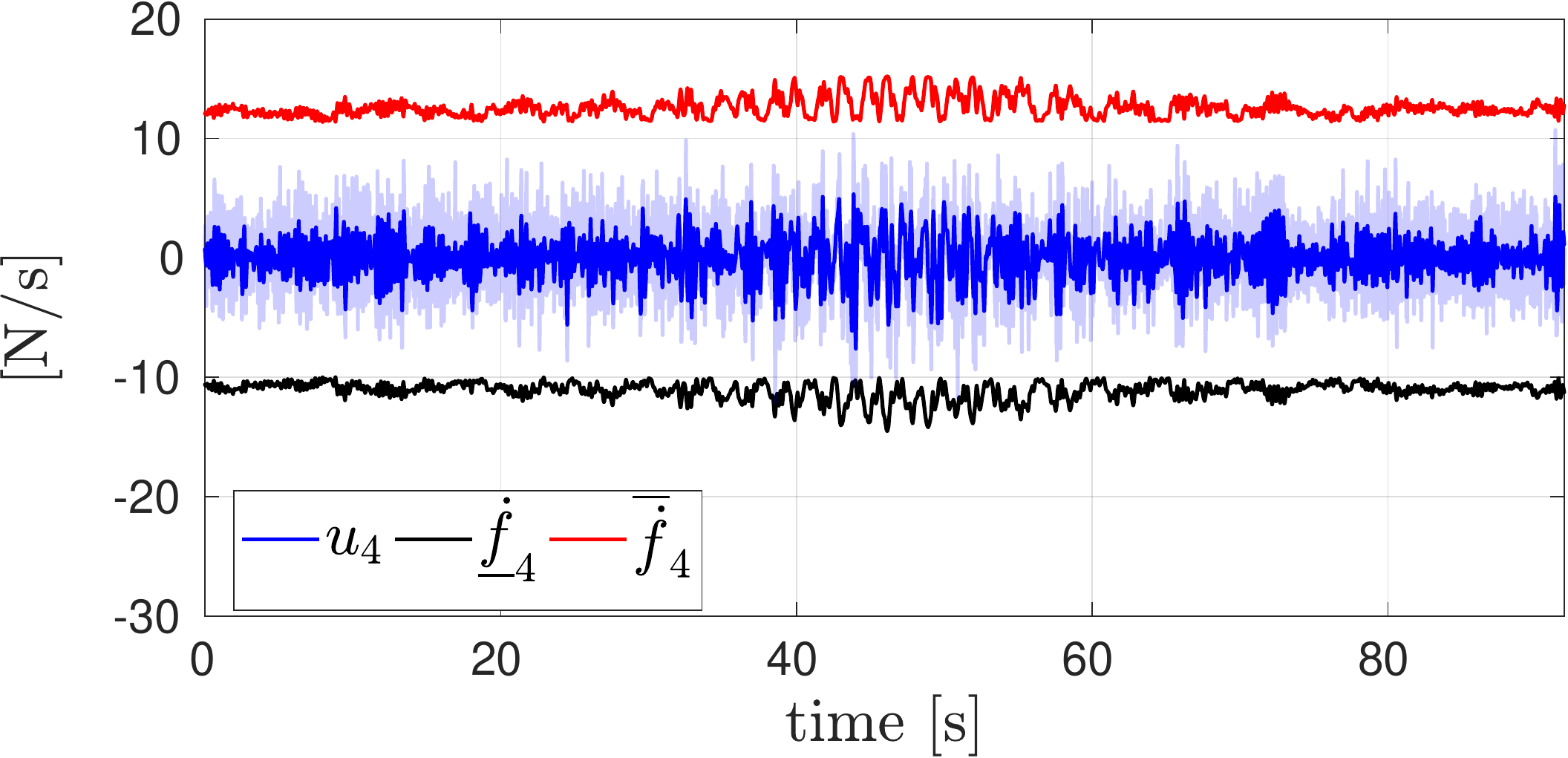}
	\end{center}
	\caption{Plots of the quadrotor performing a chirp trajectory on the x-axis. From top to bottom, the actuator forces and their derivatives. In particular, all the signals remain inside the feasible region delimited by the identified constraints. Notably, the noisy references $u_i$ are overlapped by their filtered profiles.}
	\label{fig:plots_QR_chirp_limits}
\end{figure}

The plots related to the trajectory tracking are depicted in Fig.~\ref{fig:plots_QR_chirp_traj}. As it is visible from the first one, related to the position tracking, the trajectory is symmetric w.r.t. the time instant $t=\bar{t}$. While the position and the linear velocity are globally well tracked, the second components of the orientation and the angular velocity deviate consistently from their reference signals.
This is a natural consequence of the platform inability to produce any lateral force in body frame, which causes its under-actuation. More in detail, the peaks in the measured robot pitch $\theta$ in the third plot are synchronized with the ones of the position $p_x$ in the first plot. Indeed, the edge points on the sine corresponds to the moments of maximum lateral acceleration, which can be attained only by a re-orientation of the platform frame.
With regard to the position error, illustrated in the fifth plot from the top, it is possible to observe that the negative peaks are more pronounced w.r.t. the positive ones. This asymmetry is caused by the lateral force disturbance acting on the platform due to the presence of the serial data cable, which pulls the robot in a more severe way towards the positive direction of the x-axis. The very same outcome can be consistently recognized also in the corresponding plot of Fig.~\ref{fig:plots_TiltHex_chirp_traj}, since the cable configuration remains unchanged throughout the experiments.
Apart from the contribution of the external disturbance, the inexact position tracking is also a side effect of the unfeasible flat orientation given as reference to the predictive controller.

The velocities of the MRAV rotors, whose plot is illustrated in the bottom of Fig.~\ref{fig:plots_QR_chirp_traj}, are centered on the mean value needed to compensate the gravity force while the aerial vehicle is hovering. The small offset between the velocity of rotors 1-3 and 2-4 suggests that the serial cable also generates a small clockwise torque around the z-axis, which is balanced in order to keep the platform aligned with the yaw reference.
In particular remark the fact that, even if the trajectory is rapidly-varying (with a linear acceleration peak of $5.85\si{m/s^2}$), the rotor velocities (equivalently their produced forces, presented in the first of plot of Fig.~\ref{fig:plots_QR_chirp_limits}) take values close to the hovering set-point, without the need to span a large set of values. 
This happens because the body torque needed to re-orient the aerial vehicle requires just small differences between the rotor spinning rates.
As a consequence, in this experiment the actuator force derivatives do not need to assume large values.
This intuition is confirmed by the plots 2-5 of Fig.~\ref{fig:plots_QR_chirp_limits}, which show that the input components $u_i$, represented in blue, remain distinctively far from their lower and upper bounds, drawn in black and red, respectively. This evidence suggests that in the case of UDT-MRAVs, the limits on the input and on the state components related to the actuator forces can be reached only with rapidly-varying trajectories, designed in order to produce sudden changes in the rotor commands. This motivated the next experiment.

\subsubsection{Discontinuous trajectory}\label{subsubsec:quad_steps}

Since in the chirp experiment the input limits were far from being approached, we designed a discontinuous trajectory to test the controller stability and its compliance to the actuator constraints in a critical case. For this purpose, we generated as position reference signal a sequence of steps from an initial position $\pv_1$ to a final one $\pv_2 = \pv_1 + \boldsymbol{\epsilon}_{\pv}$, with $\boldsymbol{\epsilon}_{\pv}=[-0.5\, 0.2\, 0.2]^{\top}$\,\si{m}.  On the other hand, all the other reference profiles were set to zero. In this way, the vehicle was always required to reach the next hovering configuration, with an horizontal attitude and zero translational and rotational velocities, in a short time. Moreover, in order to make the experiment even more challenging, we limited on purpose the predictive capability of the controller, i.e., the NMPC algorithm was made aware about the transitions in the position reference only at the time in which such changes effectively occurred. This strategy emulated an unforeseen event against which the algorithm had to promptly and safely react. In this way, the instantaneous appearance of a consistent error in the controller easily pushed the actuator commands towards their limitations.
Throughout this experiment, the identified input constraints on the actuator force derivative were re-scaled with gains $\rho$, taking values in $[0.125,1.75]$, spanning from very conservative - obtained with $\rho=0.125$ - to larger than the identified ones - obtained with $\rho=1.75$. The input limits in the controller were manually increased, after each discontinuous motions of the robot, by the operator by means of a joystick connected to the ground station. This allowed us to empirically assess the validity of the bounds resulting from our identification.

\begin{figure}[t]
	\begin{center}
		\includegraphics[width=\figWidth\columnwidth]{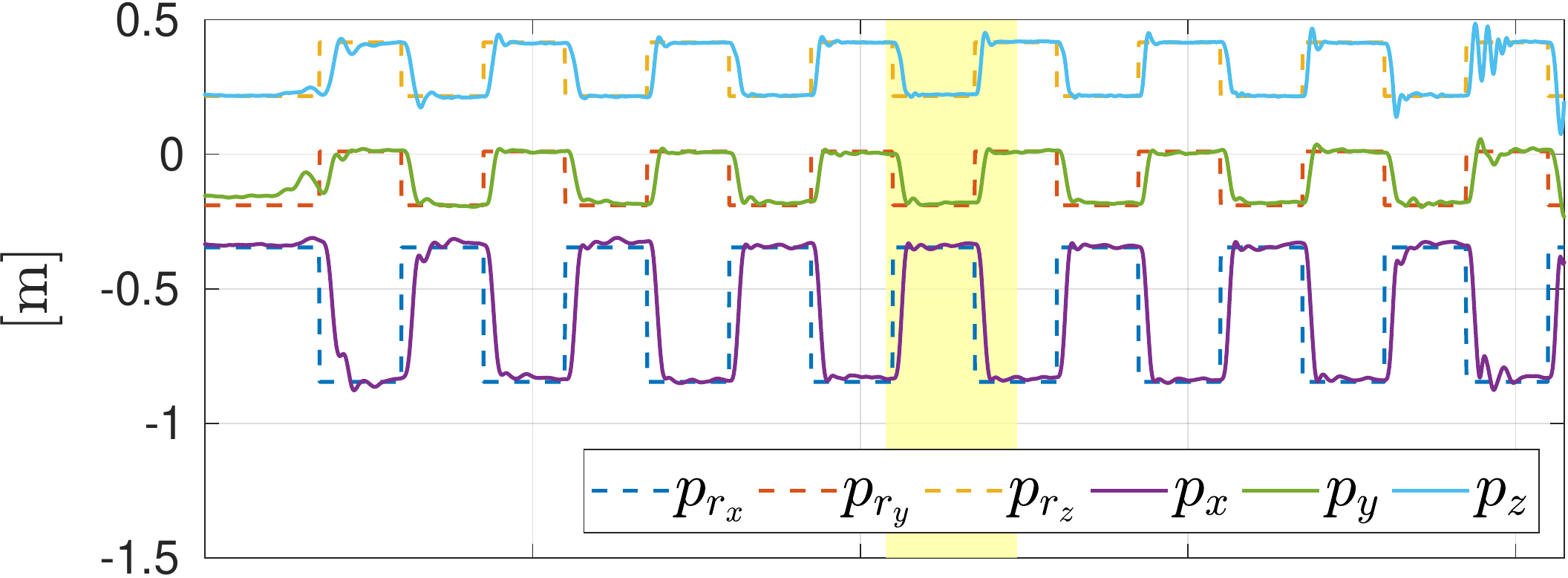}\\
		\includegraphics[width=\figWidth\columnwidth]{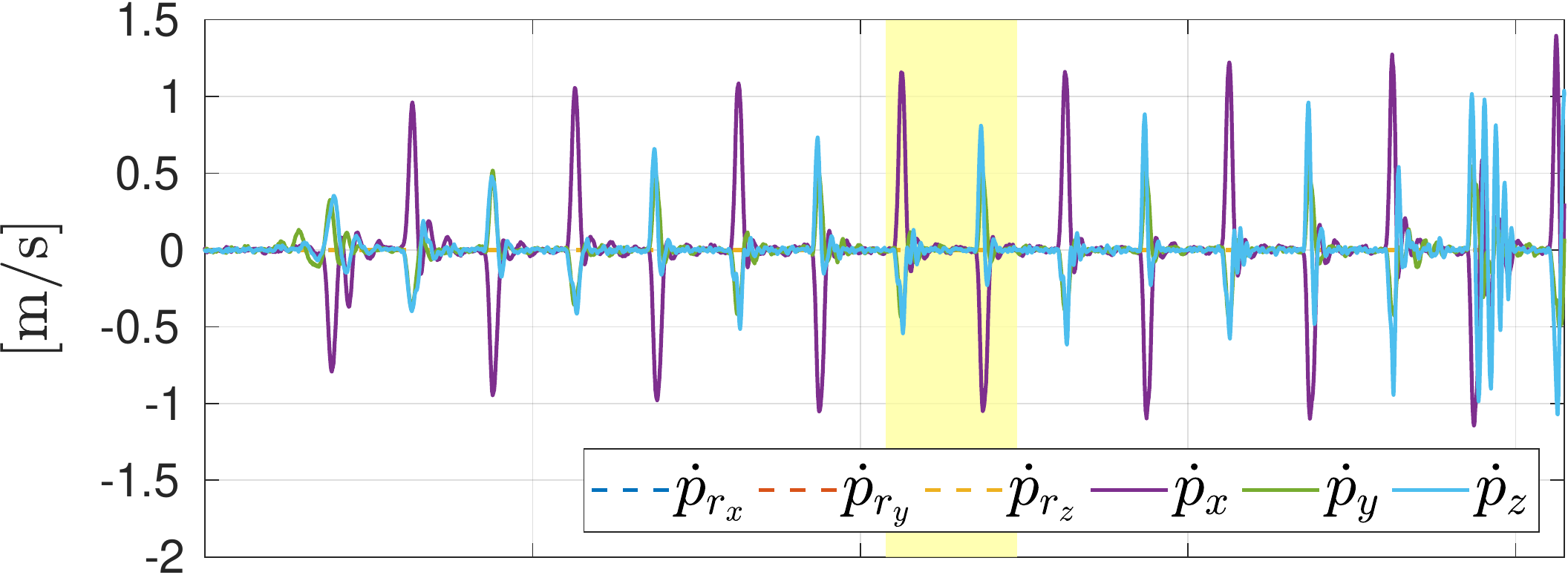}\\
		\includegraphics[width=\figWidth\columnwidth]{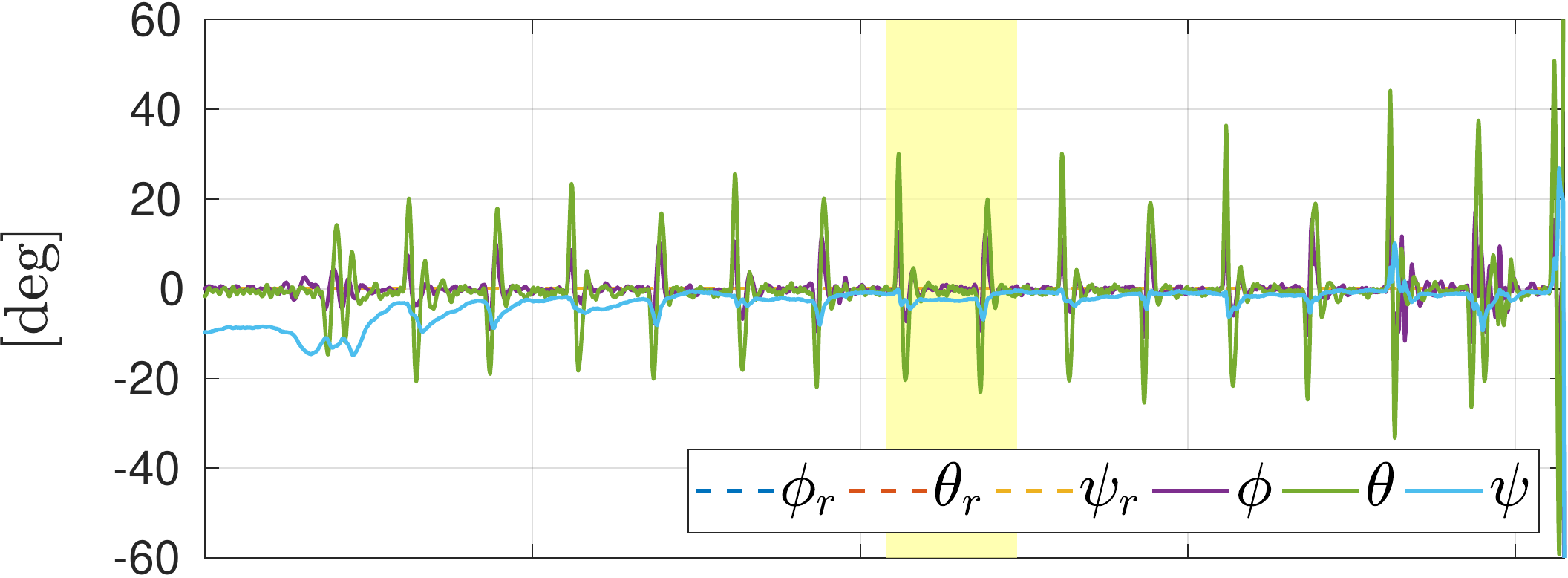}\\
		\includegraphics[width=\figWidth\columnwidth]{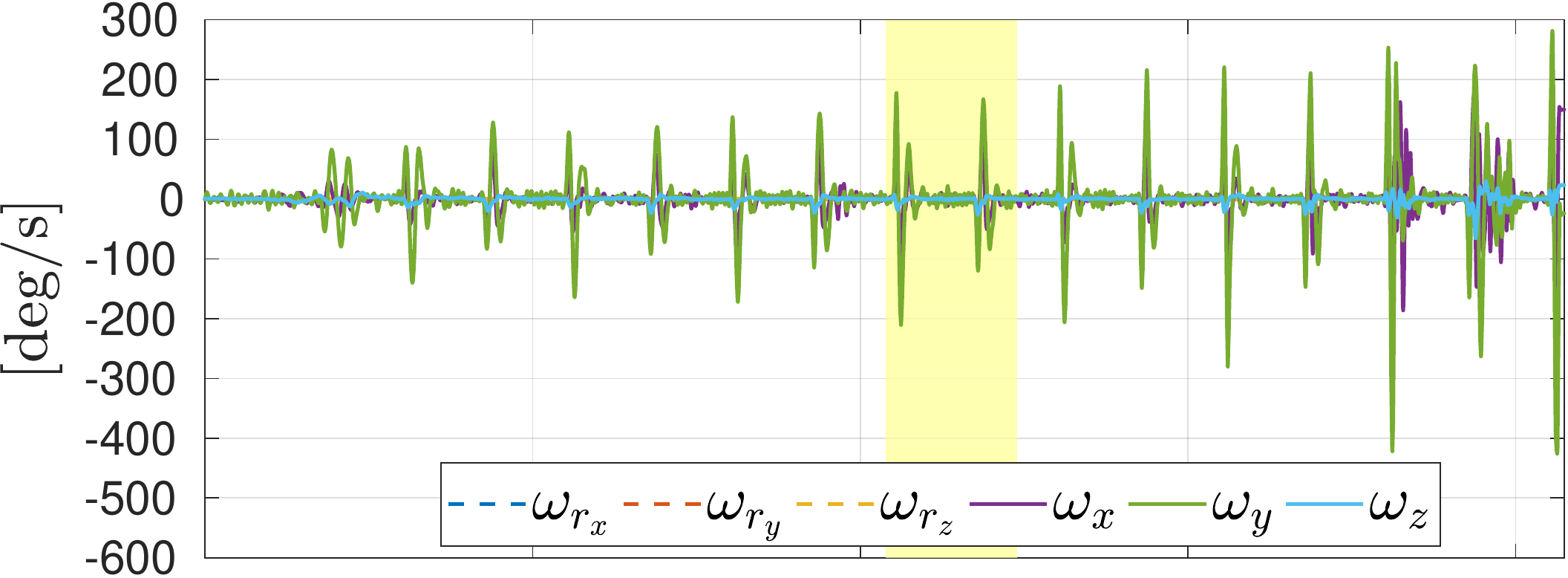}\\
		\includegraphics[width=\figWidth\columnwidth]{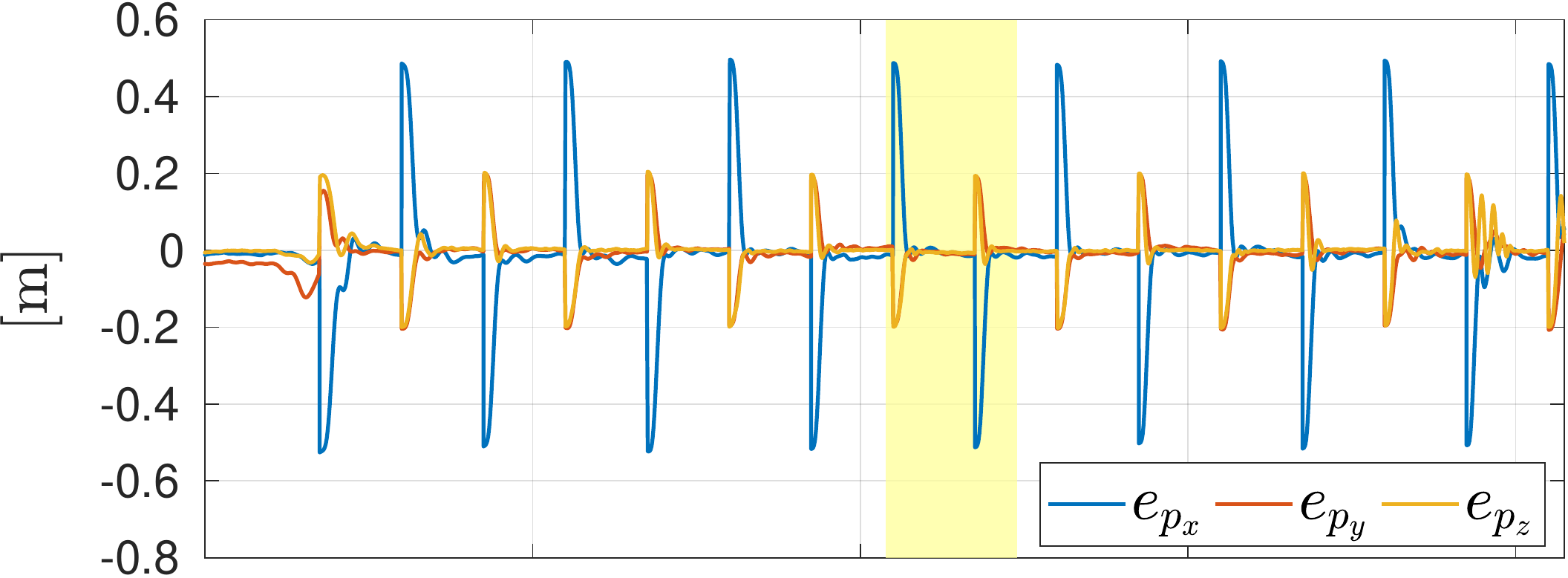}\\
		\includegraphics[width=\figWidth\columnwidth]{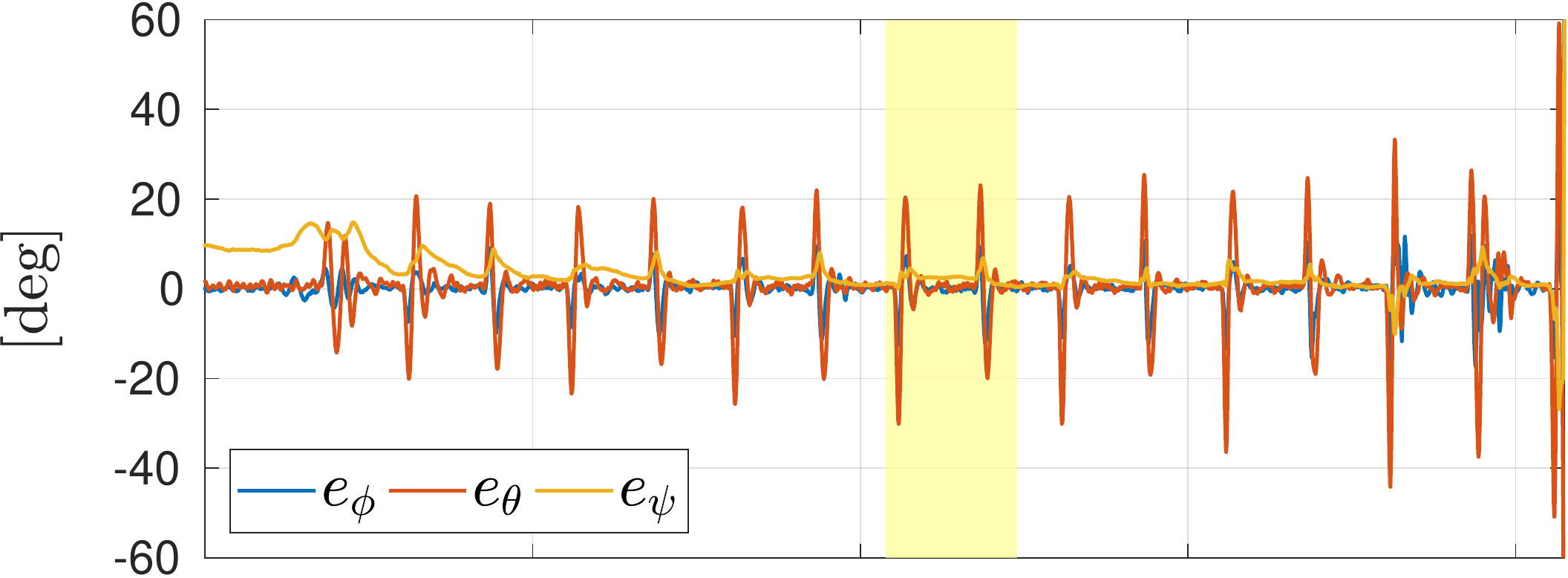}\\
		\includegraphics[width=\figWidth\columnwidth]{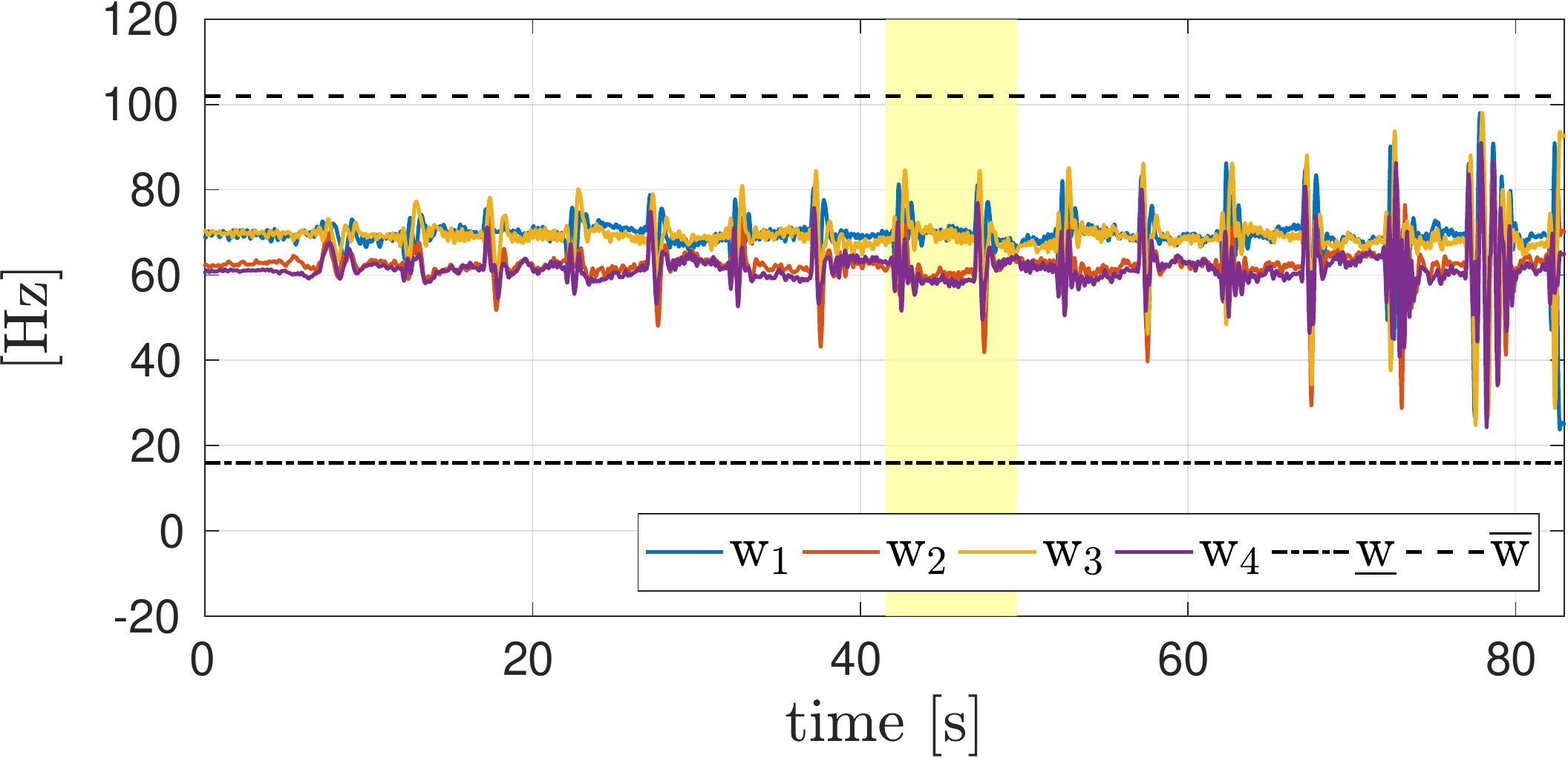}
	\end{center}
	\caption{Plots of the quadrotor tracking a discontinuous trajectory with steps in the position, while the controller limits are increased (the yellow region highlights the use of the identified ones). From top to bottom, the position, linear velocity, orientation and angular velocity tracking, the position and orientation errors, and the actuator spinning velocities.}
	\label{fig:plots_QR_steps_traj}
\end{figure}

\begin{figure}[t]
	\begin{center}
		\includegraphics[width=\figWidth\columnwidth]{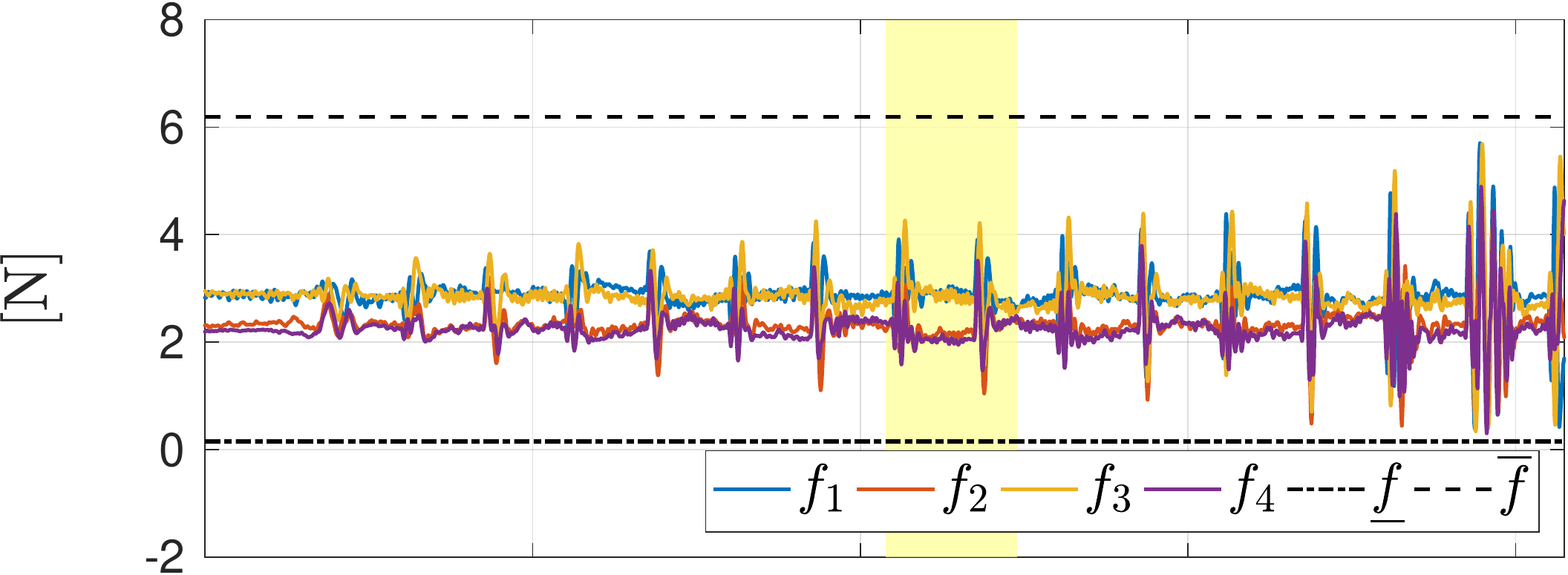}\\
		\includegraphics[width=\figWidth\columnwidth]{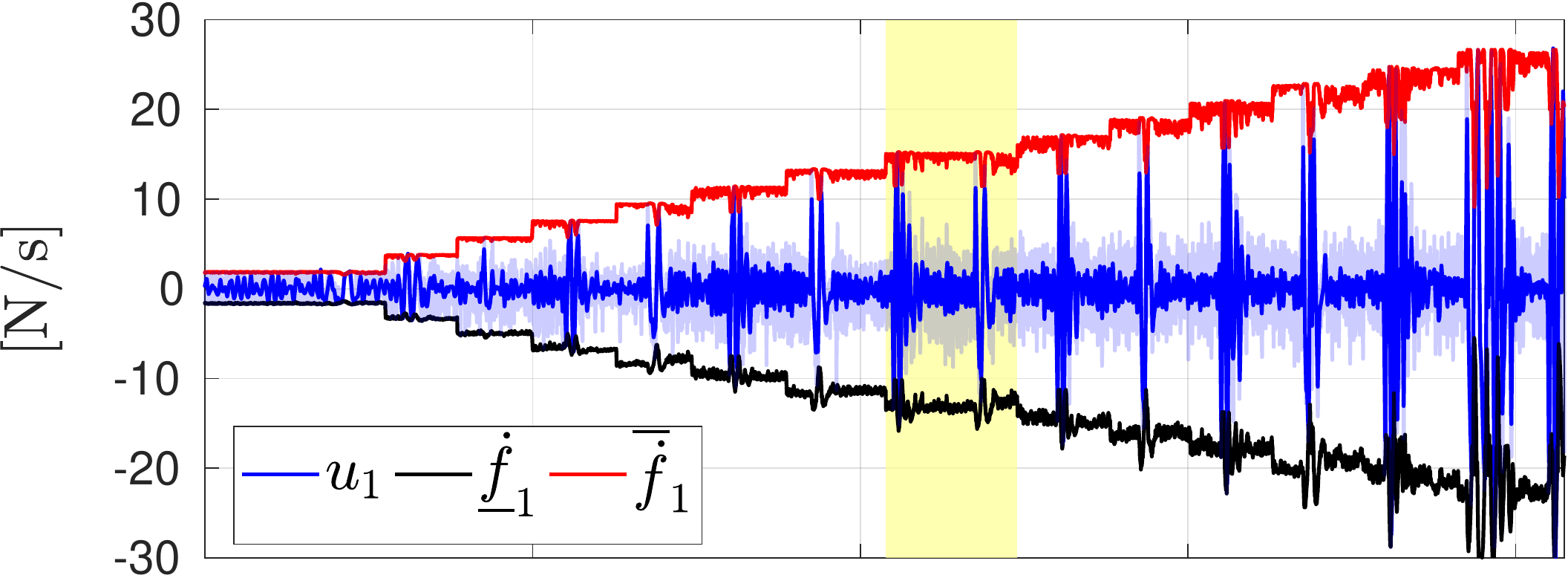}\\
		\includegraphics[width=\figWidth\columnwidth]{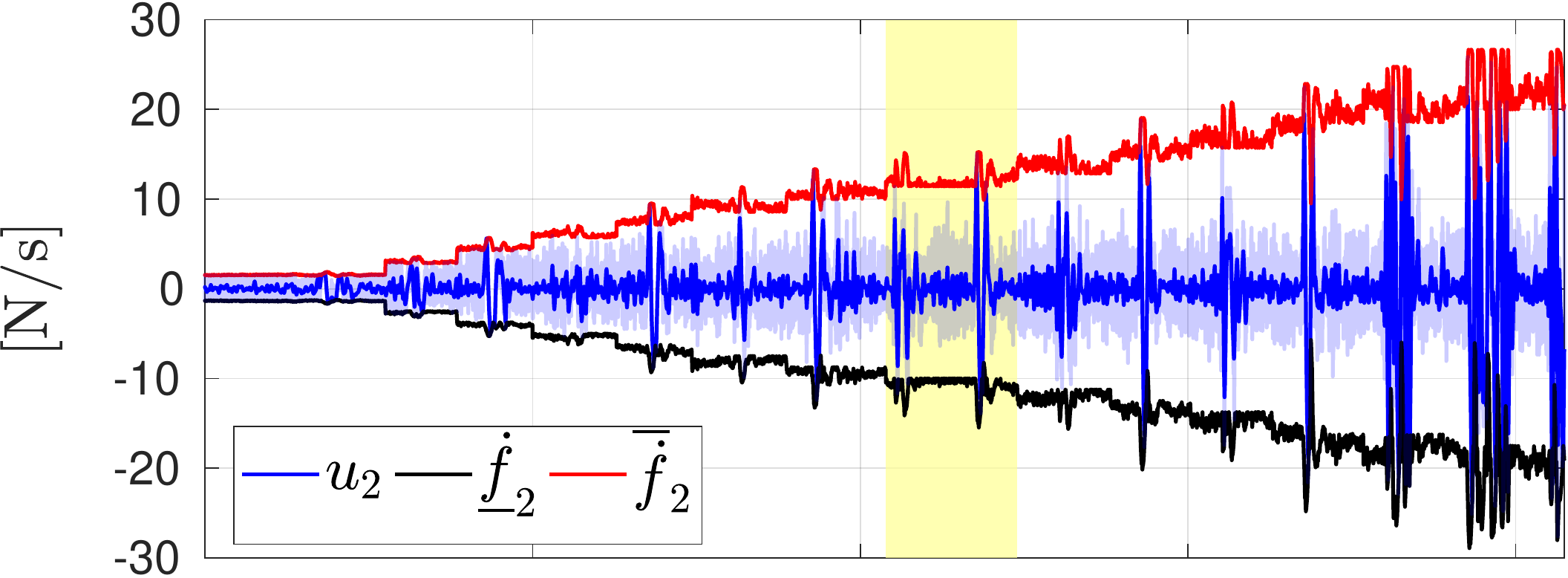}\\
		\includegraphics[width=\figWidth\columnwidth]{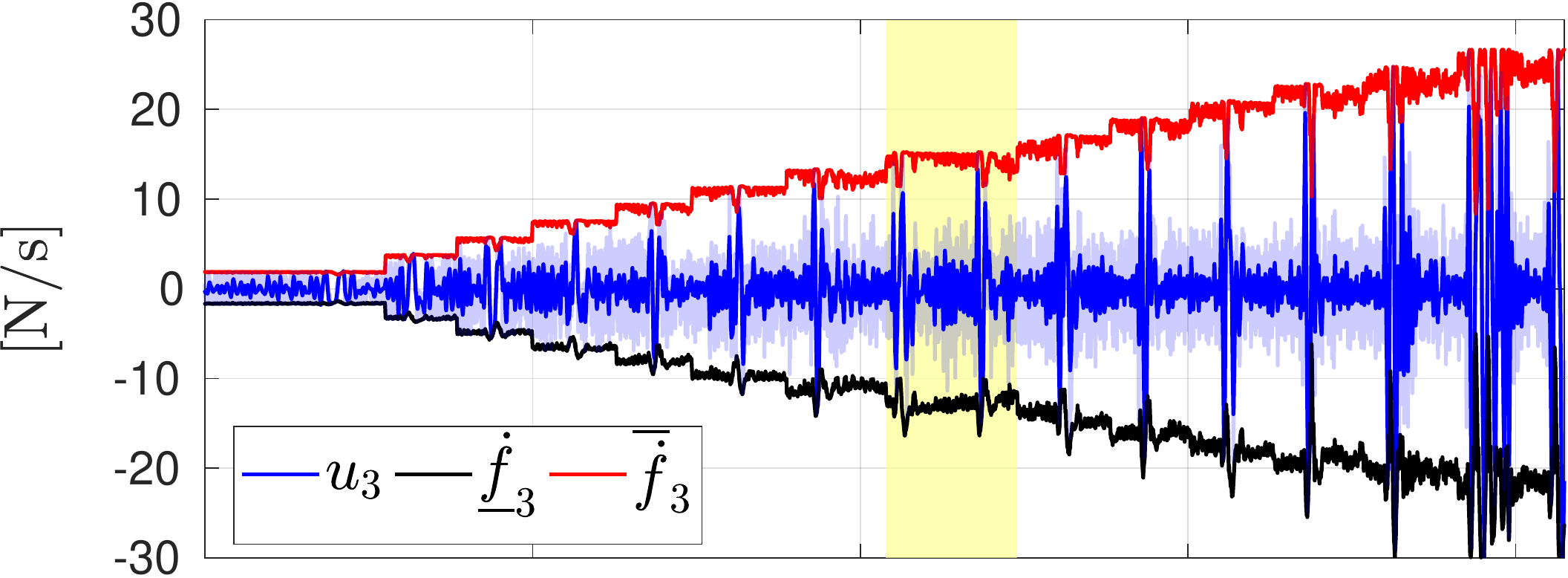}\\
		\includegraphics[width=\figWidth\columnwidth]{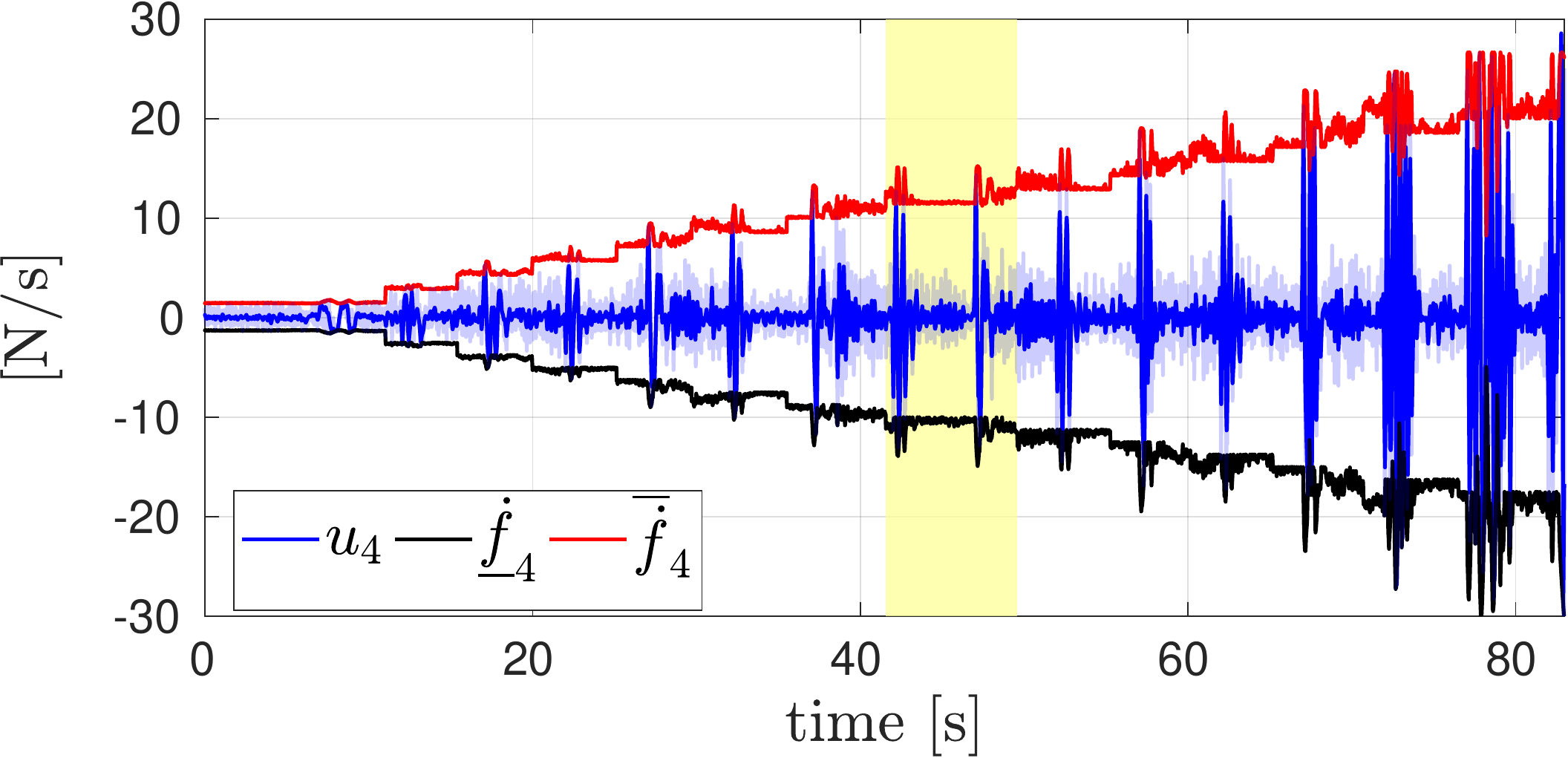}
	\end{center}
	\caption{Plots of the quadrotor tracking a discontinuous trajectory with steps in the position, while the controller limits are increased (the yellow region highlights the use of the identified ones). From top to bottom, the actuator forces and their derivatives. In particular, all the signals remain inside the feasible region delimited by the constraints.}
	\label{fig:plots_QR_steps_limits}
\end{figure}

The tracking results related to the trajectory of this specific experiment are shown in Fig.~\ref{fig:plots_QR_steps_traj}, where the yellow region highlights the time interval in which the enforced limits correspond exactly to the ones previously identified.
The position tracking, depicted in the first plot from the top, shows that very conservative bounds for the actuators, i.e., $\rho\in [\frac{1}{8} \, \frac{1}{4}]$, cause step responses with a remarkable settling time and extended oscillations. Furthermore, the reduced capability to produce a change in the actuator forces seems to affect the tracking of the yaw, that has a non-negligible error for low values of $\rho$. As already ascertained in the previous experiment, this disturbance is induced by the communication cable.
On the other hand, the oscillations in the step responses result much more restrained as soon as the control saturations approach the identified ones. Nevertheless, an additional increase in the control bounds imply growing overshoots, especially on the z-axis.
Ultimately, the instability is reached at $t \approx 84\si{s}$, when $\rho=\frac{7}{4}$. In this moment, the associated limits become almost the double of the identified ones and they induce the platform to reach a configuration from which it was not able to recover. This is confirmed by the plot of the orientation error, where the pitch error reaches almost $e_{\theta}=60$\,\si{deg}.
The MRAV instability, which causes the experiment to abort, can be particularly well appreciated from the multimedia attachment. Regarding this point, it is worthwhile to make some considerations. First, the tracking results suggest the identified limits to be suitable to ensure the platform stability, also in such a critical experiment. Moreover, this is true within some robustness margin, which was sought in order to avoid an excessive stress for the motor currents. Finally, the plots of Fig.~\ref{fig:plots_QR_steps_limits} deserve a particular attention. With reference to the first one, we can observe that the aerial vehicle becomes unstable even if the actuator forces never reach their limitations, even when instability finally happens. On the other hand, we see from the other plots that their derivatives closely approach the lower and upper bounds. This fact suggests that, neglecting the constraints on the force derivatives, as done in other works, may jeopardize, not only the system performances, but also its stability properties.

\subsection{Experiments with the Tilt-Hex}

Compared to the quadrotor model, the one of the Tilt-Hex is characterized by two more state and input components to describe the dynamics related to the presence of the additional actuators. As a matter of fact, $\xv \in \mathbb{R}^{18}$ and $\uv \in \mathbb{R}^{6}$. With this configuration, the average NMPC solver time is $t_{\text{solv}}=4.1\si{ms}$. In the validation campaign, we made the Tilt-Hex track both the trajectories presented in the previous experiments.
The values for the cost function diagonal matrices used in this experiment are reported in Tab.~\ref{tab:hr_exp}.

\begin{table}[t]
	\renewcommand\arraystretch{1.3}
	\caption{Parameters used in the Tilt-Hex experiment.}
	\label{tab:hr_exp}
	\centering
	\resizebox{\figWidth\columnwidth}{!}{%
	\begin{tabular}{CCC}
		\hline
		\text{Parameter}  & \text{Value} & \text{Unit} \\
		\hline
		\nu & 1.2 & \si{m} \\
		\xi & 0.025 & \frac{\si{rad}}{\si{s^2}} \\
		\bar{t} & 44.84 & \si{s} \\
		\boldsymbol{\epsilon}_{\pv} & [-0.4\, 0.3\, 0.2]^{\top} & \si{m} \\
		\rho & 0.125:0.125:1.75 & [\ ] \\
		\hline
		\Qm_{\pv}(j,j)|_{j=1,2,3} & 500,200,200 & [\ ] \\
		\Qm_{\dot{\pv}}(j,j)|_{j=1,2,3} & 25,20,20 & [\ ] \\
		\Qm_{\etav}(j,j)|_{j=1,2,3} & 10,6,10 & [\ ] \\
		\Qm_{\omegav}(j,j)|_{j=1,2,3} & 0.5,0.5,0.5 & [\ ] \\
		\Qm_{\ddot{\pv}}(j,j)|_{j=1,2,3} & 0.01,0.01,0.01 & [\ ] \\
		\Qm_{\dot{\omegav}}(j,j)|_{j=1,2,3} & 0,0,0 & [\ ] \\
		\Rm_h(j,j)|_{j=1,\dots{},6} & 0,0,0,0,0,0 & [\ ] \\
		\hline
	\end{tabular}  } 
\end{table}

\subsubsection{Position chirp trajectory}

Thanks to the tilting of its actuators, the Tilt-Hex can exert a 3D set of forces which is not anymore restrained to the body-frame z-axis.
In particular, the polytope of forces with zero moment, computed in compliance with all the admissible actuator forces, can be appreciated from the left side of Fig.~3 in our previous work~\cite{2018d-FraCarBicRyl}.
Thanks to this feature, the vehicle can track decoupled references in position and orientation.
However, despite this additional capability, the Tilt-Hex cannot track any decoupled trajectory, due to the unavoidable limitations still present in the actuators.

\begin{figure}[b]
	\centering
		\includegraphics[width=\figWidth\columnwidth]{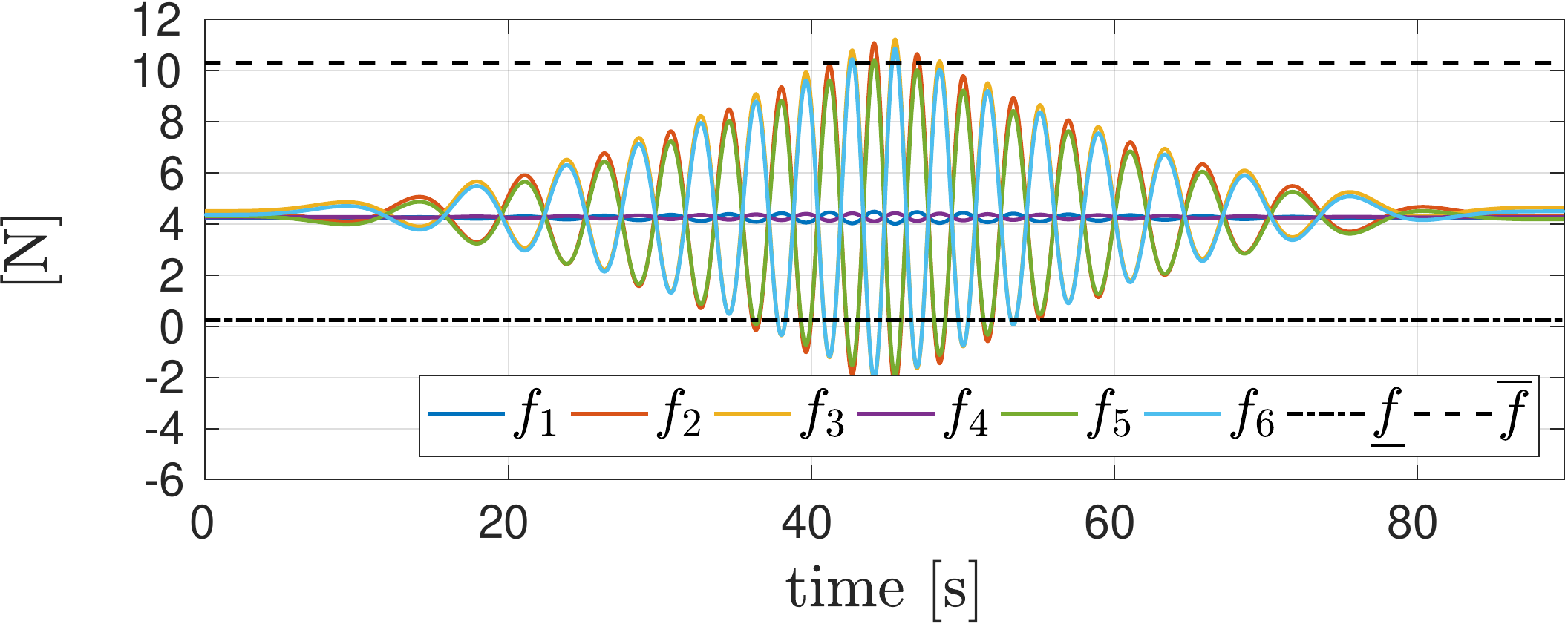}
	\caption{Desired profile for the actuators forces obtained by inverting the model dynamics. This chirp trajectory results unfeasible also with respect to the Tilt-Hex limitations.}
	\label{fig:TiltHex_chirp_thrust_inv}
\end{figure}

It should be appreciated that the previously defined chirp trajectory was generated with the goal to be unfeasible also w.r.t. the Tilt-Hex actuation capabilities. In fact, by plugging the desired trajectory and the physical parameters in~\eqref{eq:Newton_Euler_Law} and isolating the vector $[\fv_B^{\top} \, \tauv_B^{\top}]^{\top}$, we obtain the analytic expression of the wrench needed to ideally follow the 6D reference  profile. At this point, inverting~\eqref{eq:alloc_mat} -- which is possible in this case since $\Gm$ is square and full-rank -- provides the ideal (no noise or disturbance were involved in this computation) evolution of the actuator forces. As shown in Fig.~\ref{fig:TiltHex_chirp_thrust_inv}, the desired actuator force trajectories, obtained via such dynamic inversion, are not compliant with the lower and upper bounds. This means that, also in this case, a new feasible trajectory has to be re-computed by the NMPC strategy. Nevertheless, we expect to obtain improved tracking performances compared to the quadrotor experiment.

\begin{figure}[t!]
	\begin{center}
		\includegraphics[width=\figWidth\columnwidth]{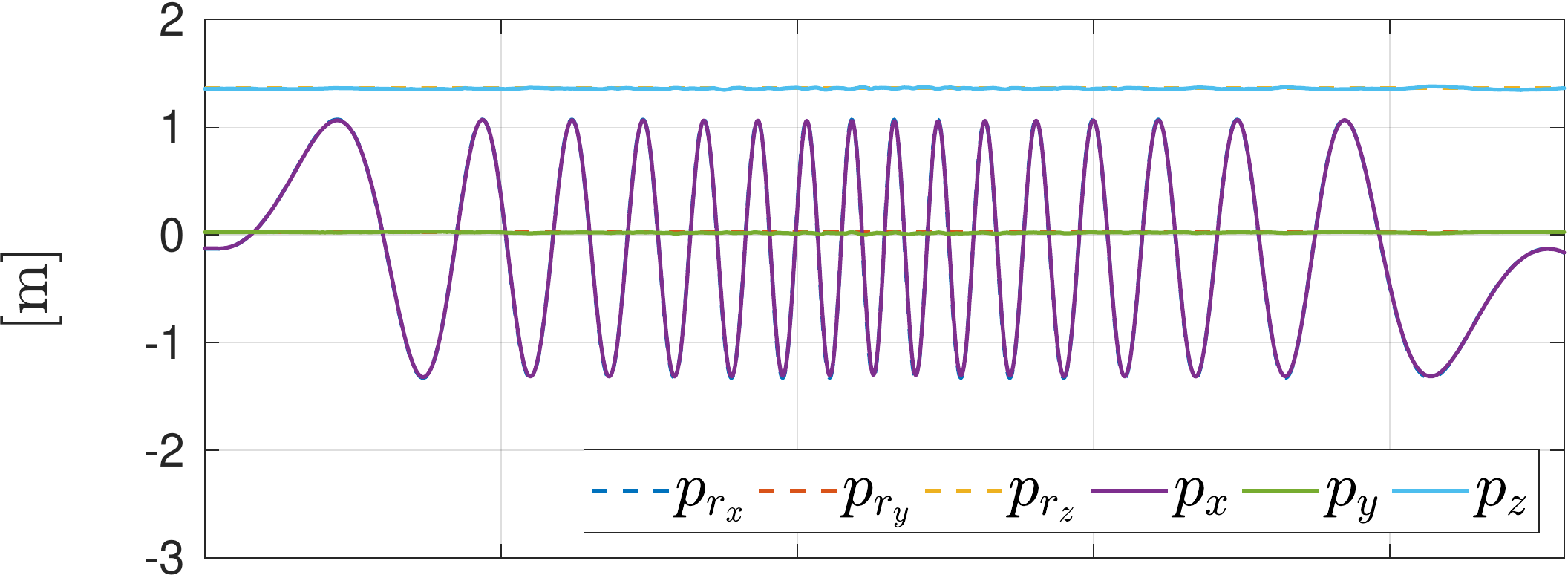}\\
		\includegraphics[width=\figWidth\columnwidth]{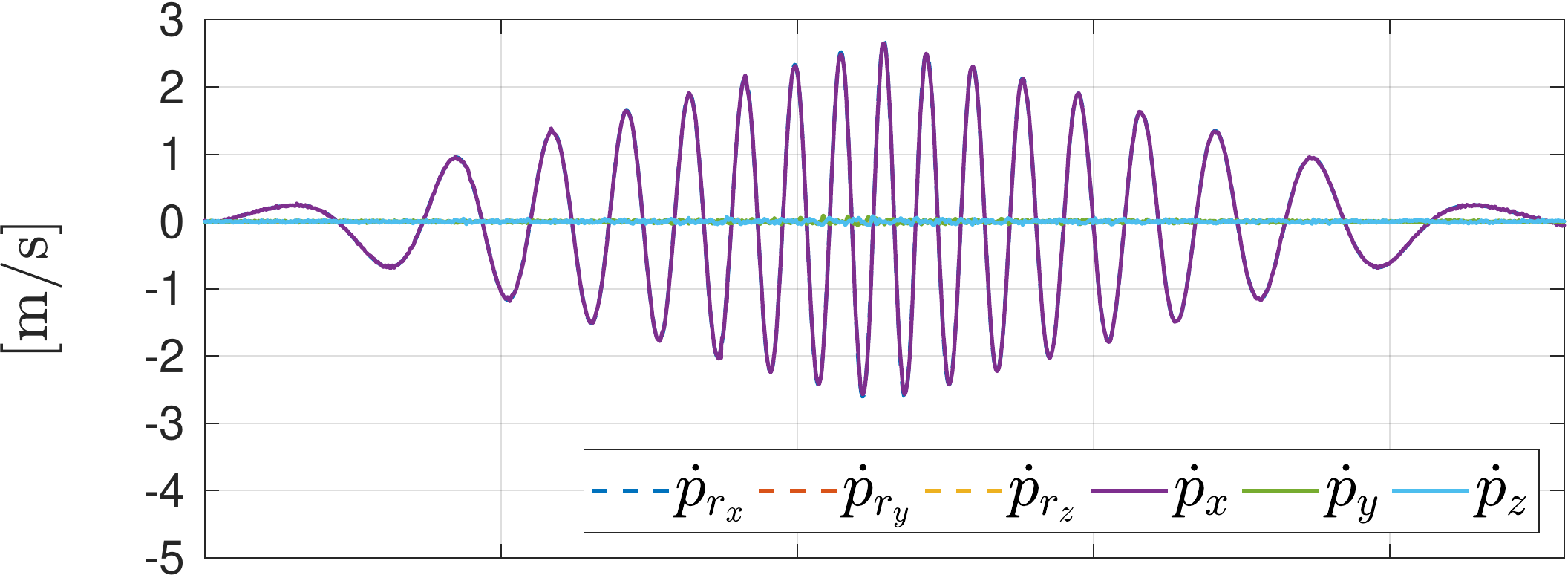}\\
		\includegraphics[width=\figWidth\columnwidth]{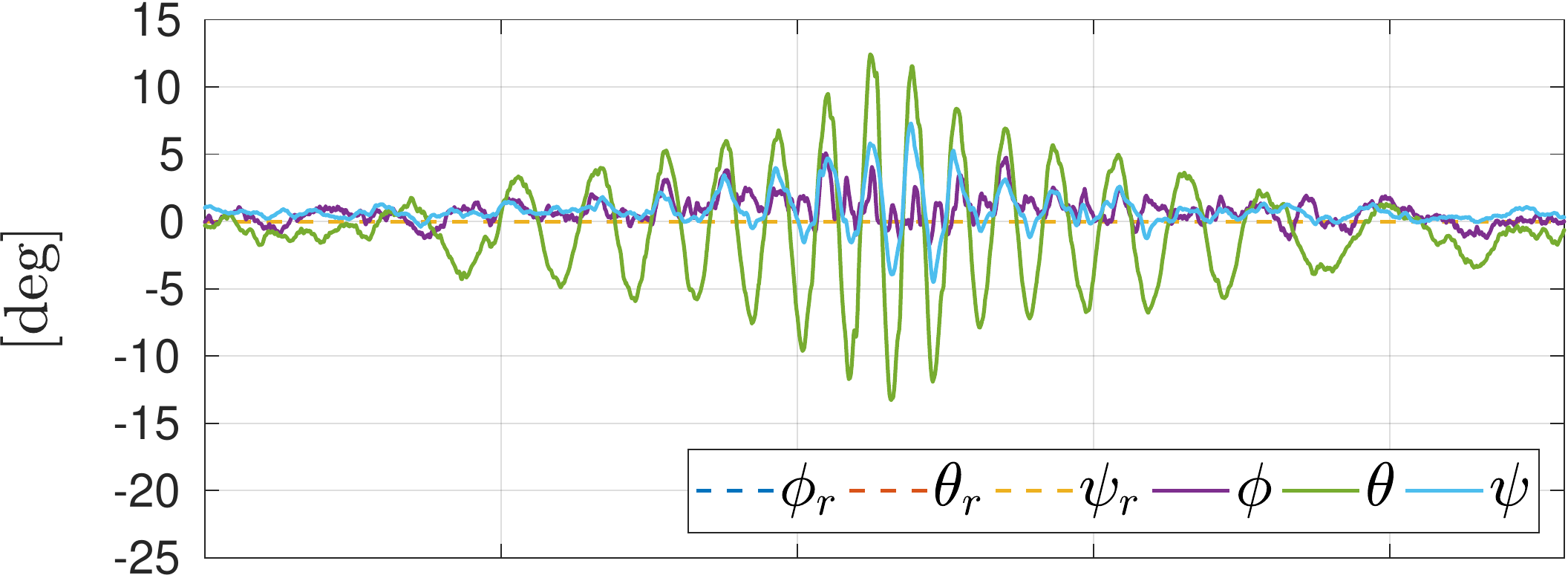}\\
		\includegraphics[width=\figWidth\columnwidth]{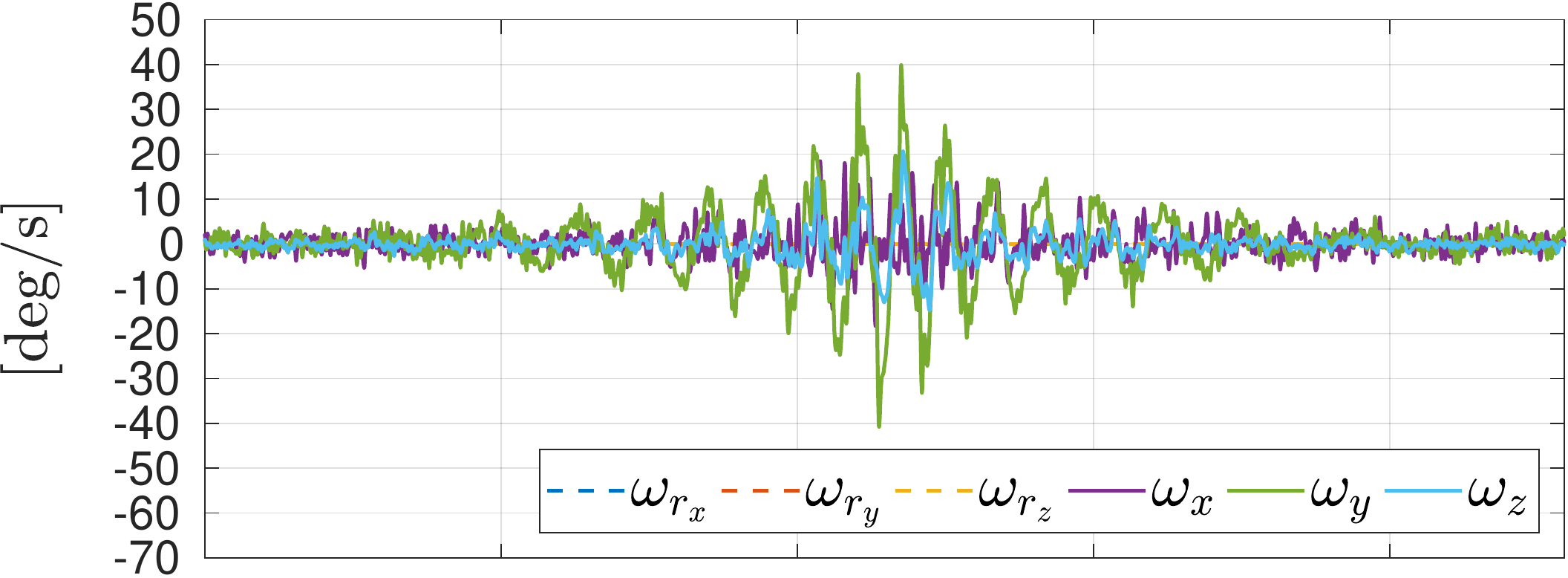}\\
		\includegraphics[width=\figWidth\columnwidth]{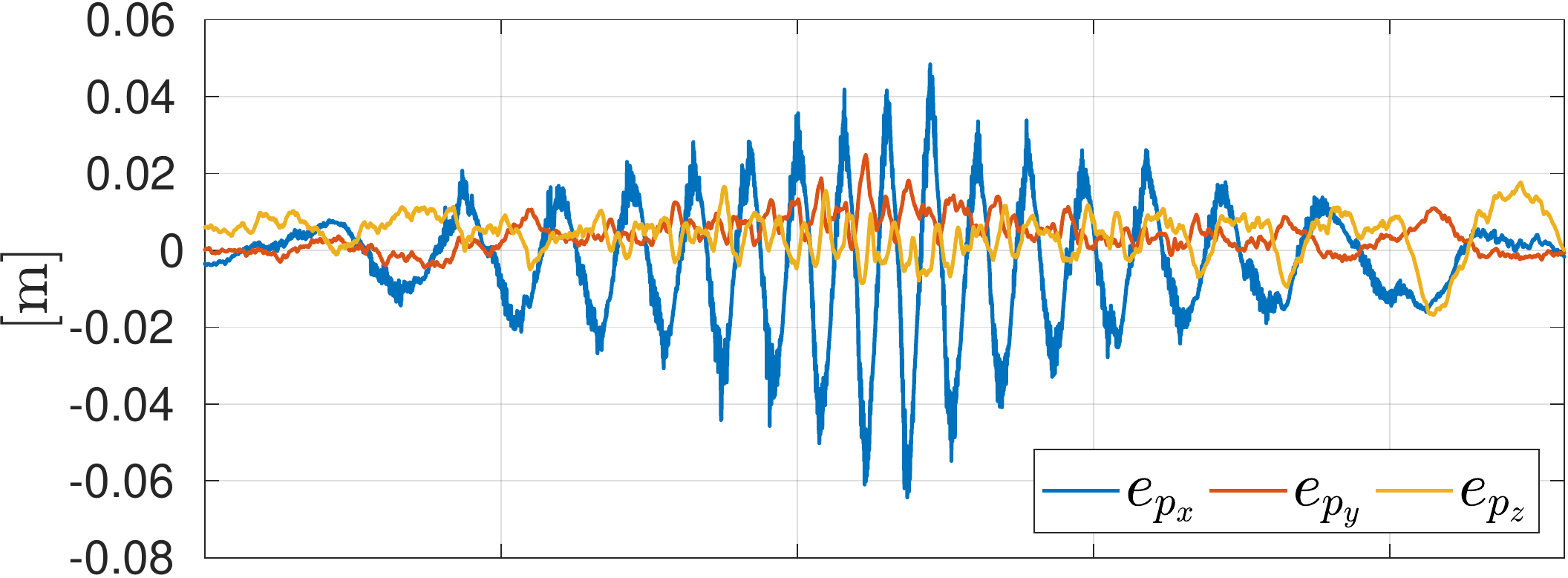}\\
		\includegraphics[width=\figWidth\columnwidth]{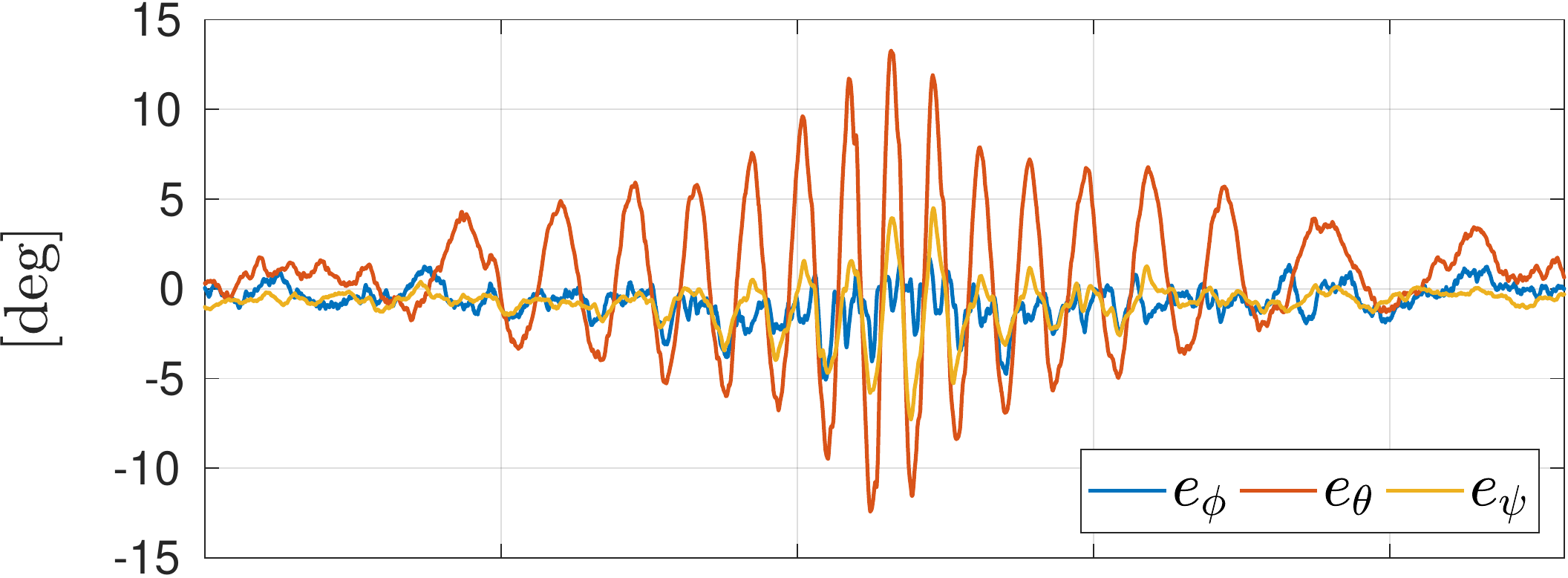}\\
		\includegraphics[width=\figWidth\columnwidth]{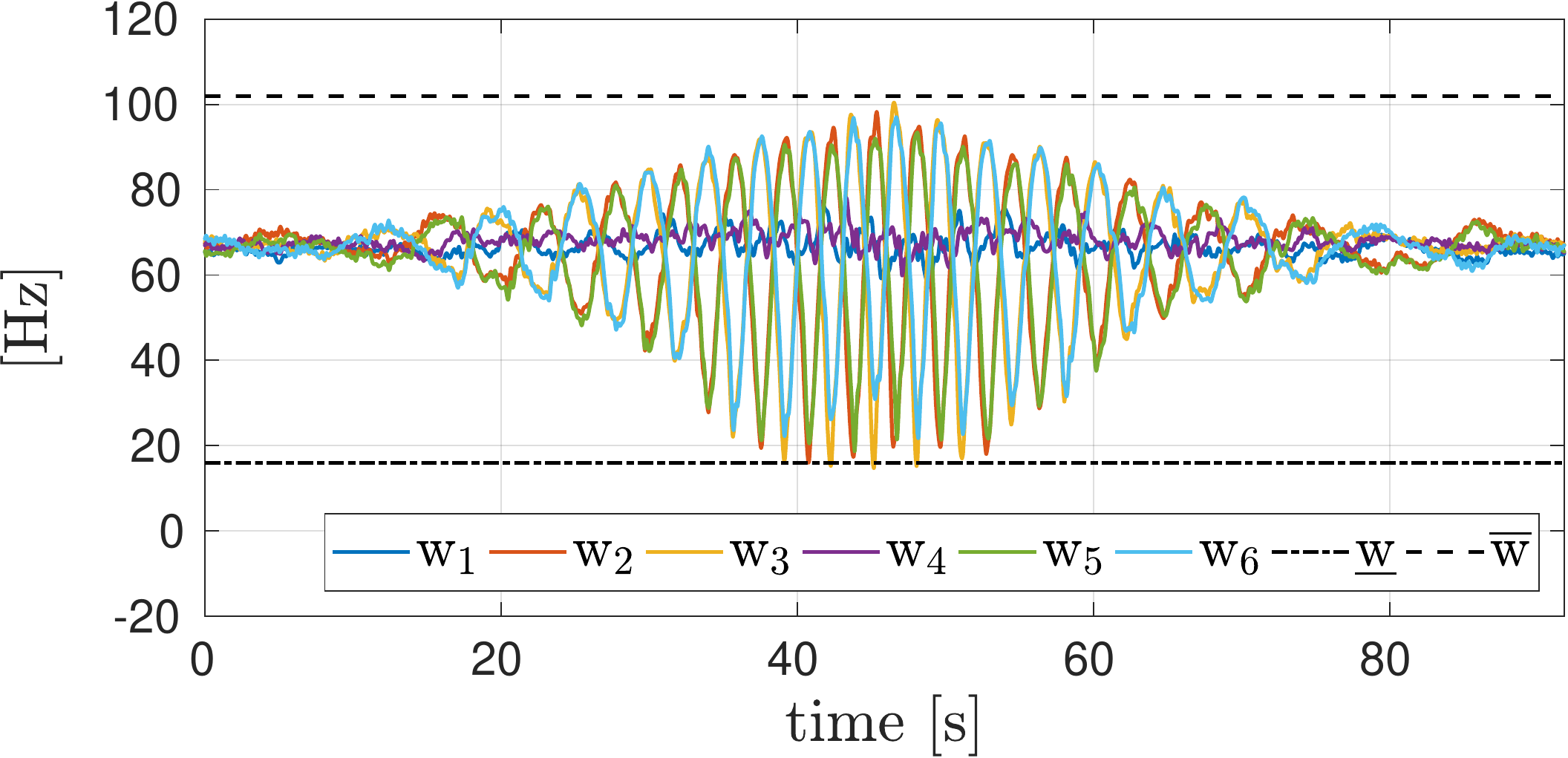}
	\end{center}
	\caption{Plots of the Tilt-Hex performing a chirp trajectory on the x-axis. From top to bottom, the position, linear velocity, orientation and angular velocity tracking, the position and orientation errors, and the actuator velocities.}
	\label{fig:plots_TiltHex_chirp_traj}
\end{figure}

The plots related to the trajectory tracking for this experiment are shown in Fig.~\ref{fig:plots_TiltHex_chirp_traj}. As shown on the two top sub-figures, the translational references are followed in a more precise way compared to Fig.~\ref{fig:plots_QR_chirp_traj}. In particular, this is true also around the central peaks, which correspond to the most rapidly-varying part of the trajectory, i.e., where the lateral acceleration takes the largest values. From the third and the fourth plots it can be observed that the deviations from the orientation and the angular velocity references are significantly reduced w.r.t. the ones produced by the quadrotor with the very same trajectory. Such remarkable improvement is a direct consequence of the benefits induced by the multi-directionality of the thrust. On the other hand, also the position error is consistently reduced, with a maximum peak of $6.4\si{cm}$ (in absolute value) against the $14.5\si{cm}$ of the quadrotor experiment. This suggests that the full actuation also helps improving the position tracking, as already observed in our previous work. With reference to the fifth plot, the systematic small asymmetry in the position error is caused again by the cable disturbance.
To ease the reader's analysis of the experimental results, we translated the graphical comparison offered by the plots into a quantitative analysis of the data, which we present in Tab.~\ref{tab:exp_comp}.

\begin{table}[t]
	\caption{Numerical comparison of the NMPC performance achieved in the experimental validation with different MRAVs.}
	\label{tab:exp_comp}
	\centering
	\renewcommand\arraystretch{1.4}
	\resizebox{\columnwidth}{!}{
	\begin{tabular}{CCC}
		\hline
		\multicolumn{3}{c}{\textbf{Chirp trajectory}} \\
		\hline
		\text{Parameter [$\bullet$]} & \text{Quadrotor} & \text{Tilt-Hex} \\
		\hline
		e_{p,\text{MAX}}\ [\si{m}]  &  [0.146,\ 0.036,\ 0.076]^\top & [\mathbf{0.064,\ 0.025,\ 0.018}]^\top \\
		e_{p,\text{RMS}}\ [\si{m}]  &  [0.054,\ 0.010,\ 0.014]^\top & [\mathbf{0.015,\ 0.006,\ 0.007}]^\top \\
		e_{\eta,\text{MAX}}\ [\si{deg}]  &  [6.7,\ 30.4,\ \mathbf{4.3}]^\top & [\mathbf{5.1,\ 13.3},\ 7.3]^\top \\
		e_{\eta,\text{RMS}}\ [\si{deg}]  &  [1.2,\ 10.0,\ 1.6]^\top & [\mathbf{1.2,\ 3.9,\ 1.4}]^\top \\
		f_{i,\text{MIN}},\ f_{i,\text{MAX}}\ [\si{N}] & 1.495,\ 3.876 & \mathbf{0.231,\ 10.251} \\
		\dot{f}_{i,\text{MIN}},\ \dot{f}_{i,\text{MAX}}\ [\si{N/s}] & -15.264,\ 13.998 & \mathbf{-25.320,\ 28.321} \\
		\hline
		\multicolumn{3}{c}{\textbf{Step trajectory with identified input limits ($\rho=1$)}} \\
		\hline
		\text{Parameter [$\bullet$]} & \text{Quadrotor} & \text{Tilt-Hex} \\
		\hline
		\renewcommand\arraystretch{1.2}
		e_{\eta,\text{MAX}}\ [\si{deg}]  &  [12.6,\ 30.1,\  6.9]^\top & [\mathbf{4.0,\ 5.1,\ 2.0}]^\top \\
		e_{\eta,\text{RMS}}\ [\si{deg}]  &  [3.4,\ 7.9,\ 2.6]^\top & [\mathbf{1.1,\ 1.1,\ 1.2}]^\top \\
		f_{i,\text{MIN}},\ f_{i,\text{MAX}}\ [\si{N}] & 1.031,\ 4.396 & \mathbf{0.972,\ 8.054} \\
		\dot{f}_{i,\text{MIN}},\ \dot{f}_{i,\text{MAX}}\ [\si{N/s}] & -15.162,\ 15.204 & \mathbf{-25.348,\ 28.103} \\
		\hline
	\end{tabular}
	}
\end{table}

\begin{figure}[t]
	\begin{center}
		\includegraphics[width=\figWidth\columnwidth]{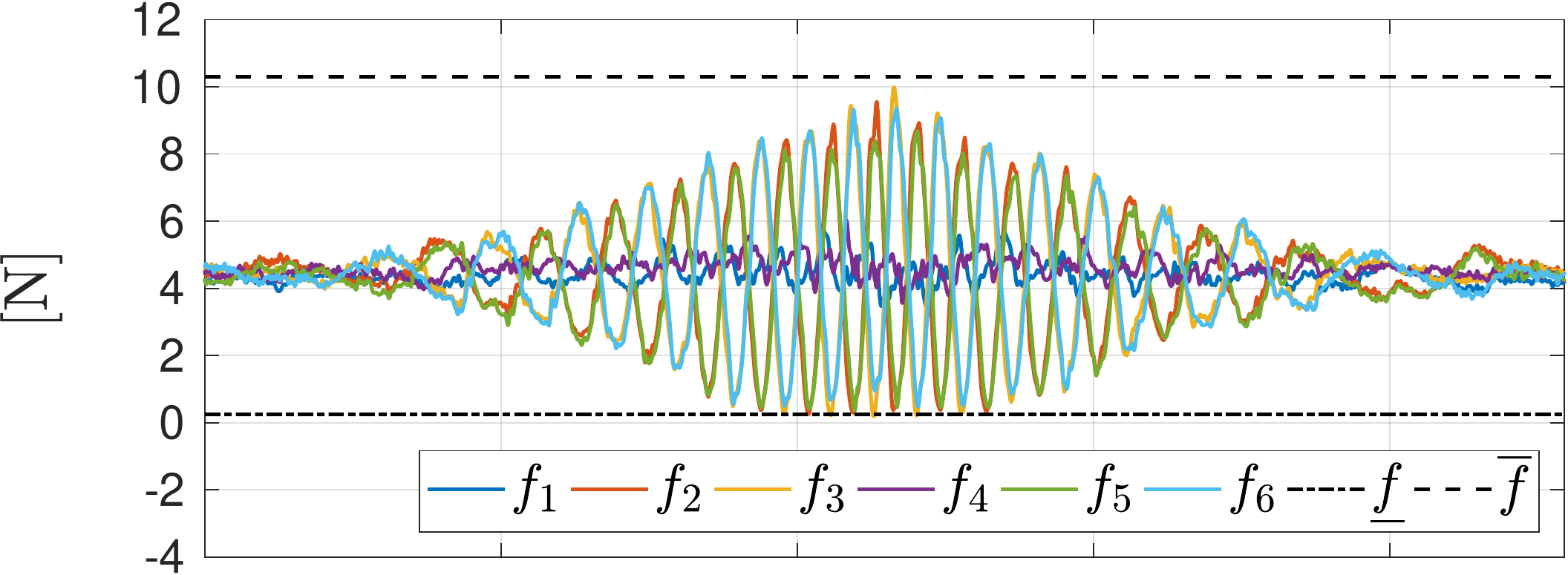}\\
		\includegraphics[width=\figWidth\columnwidth]{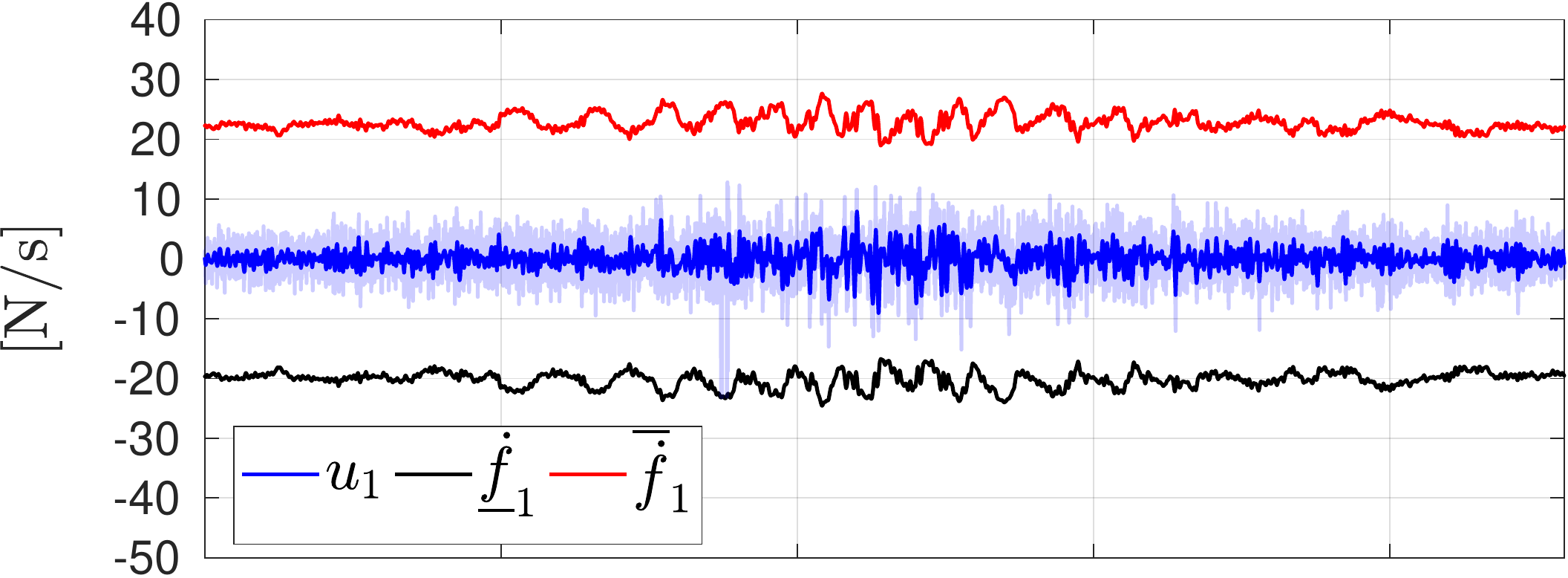}\\
		\includegraphics[width=\figWidth\columnwidth]{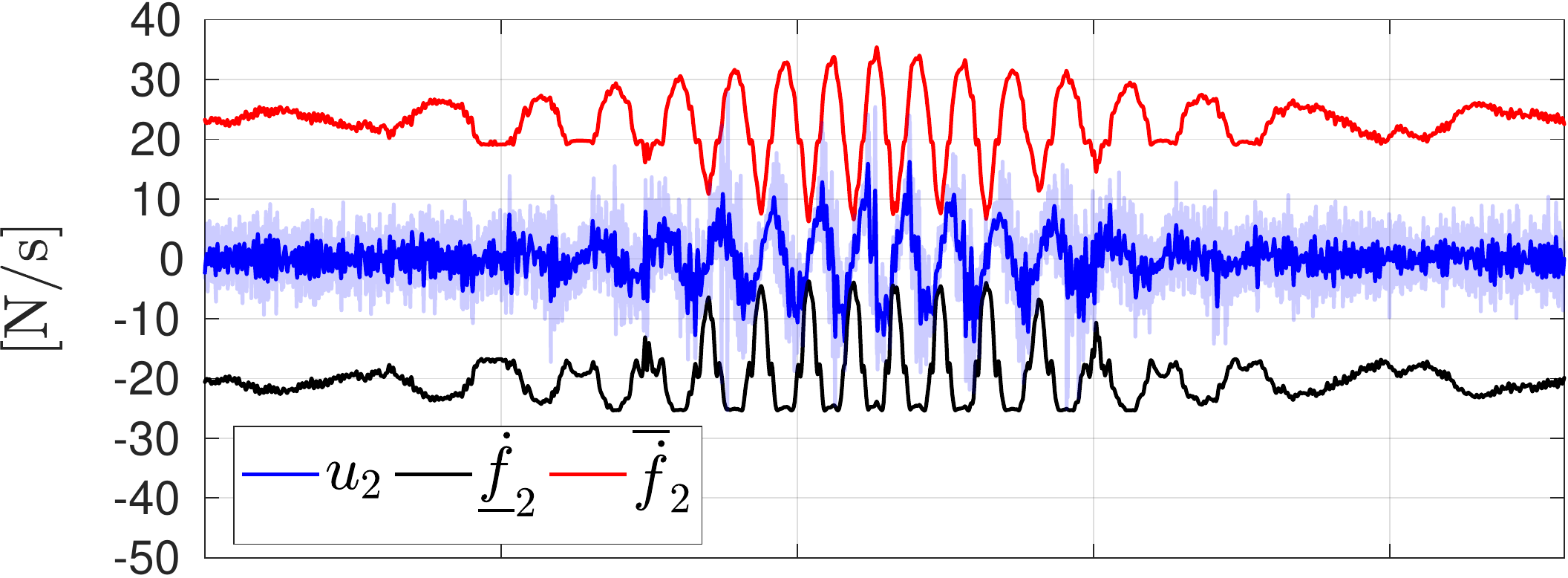}\\
		\includegraphics[width=\figWidth\columnwidth]{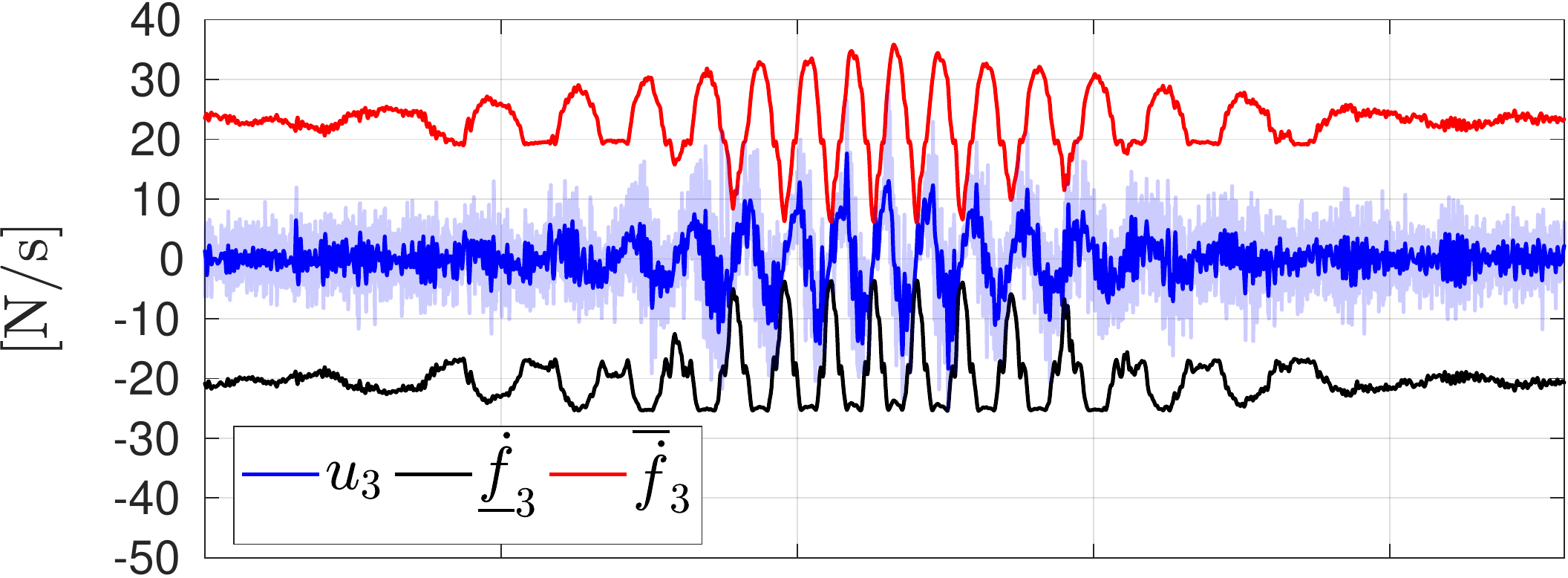}\\
		\includegraphics[width=\figWidth\columnwidth]{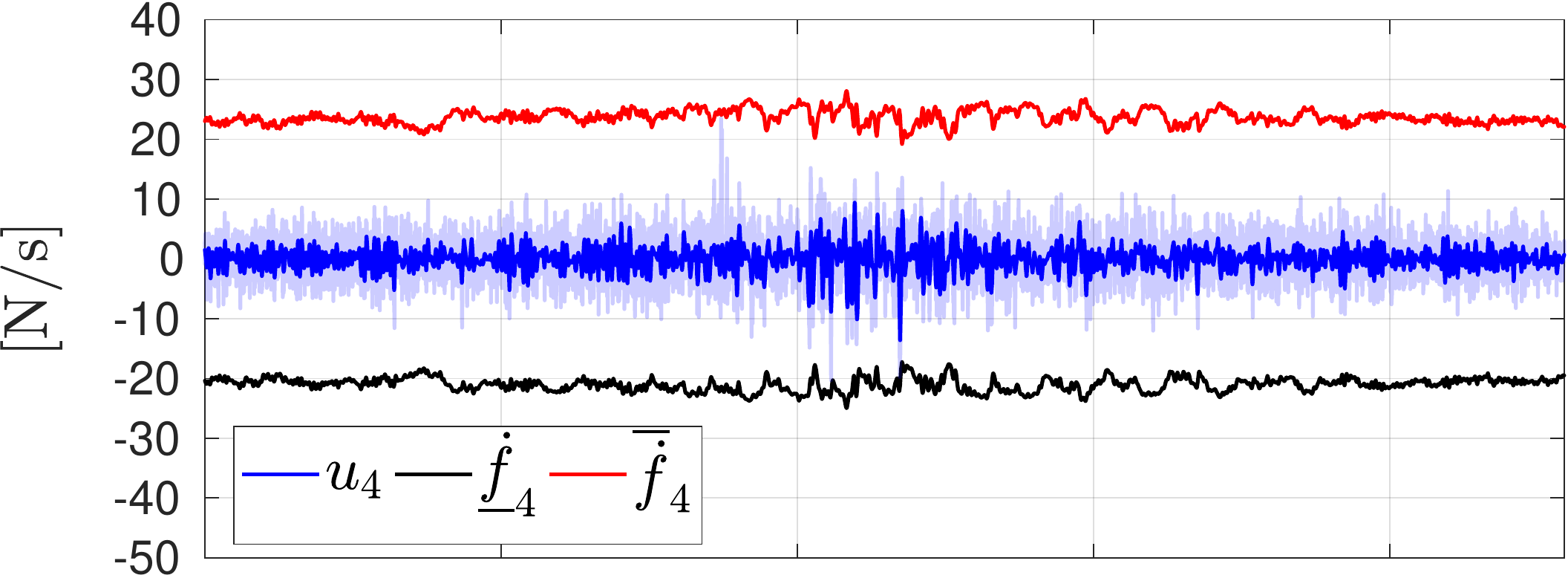}\\
		\includegraphics[width=\figWidth\columnwidth]{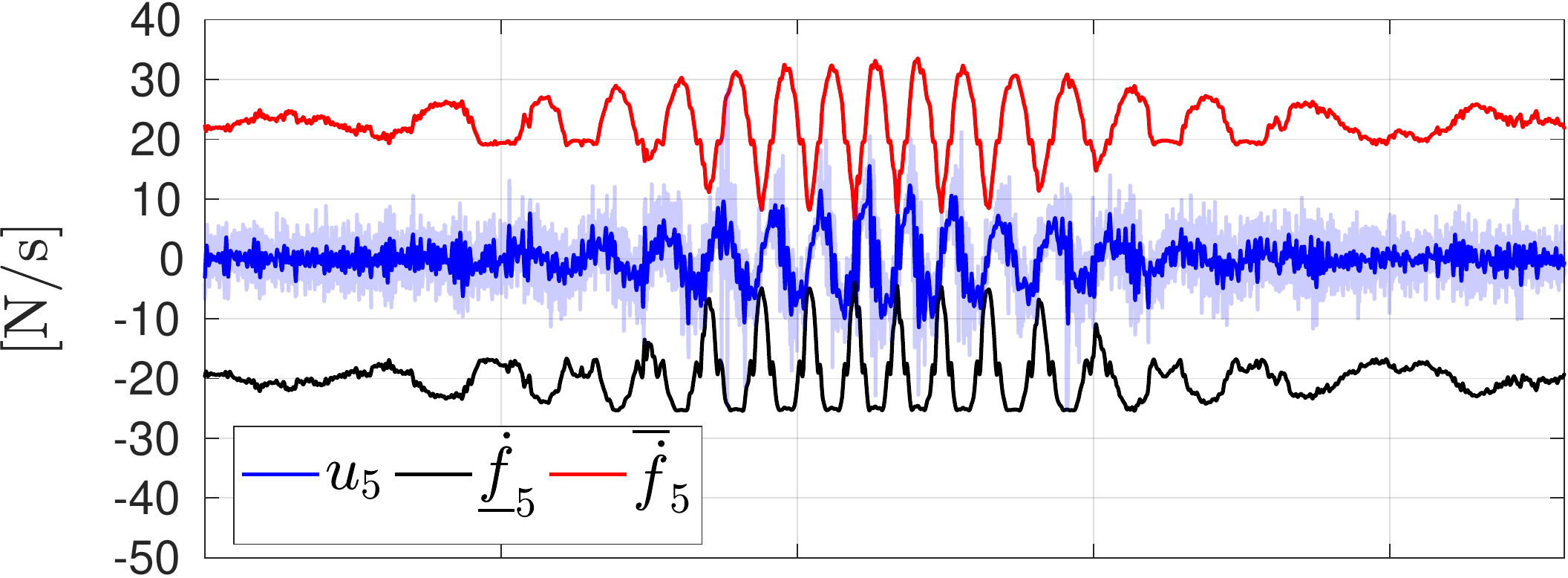}\\
		\includegraphics[width=\figWidth\columnwidth]{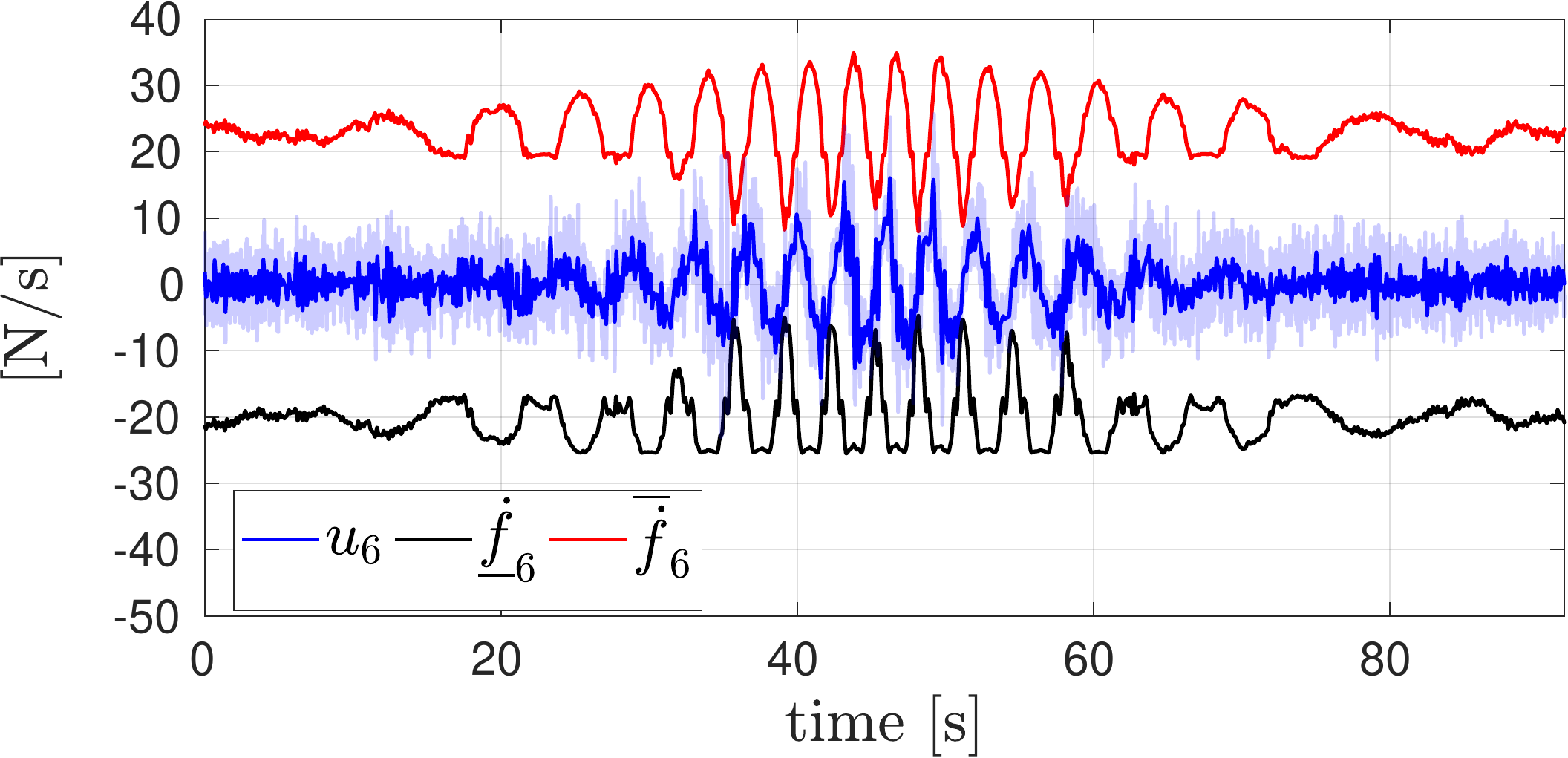}
	\end{center}
	\caption{Plots of the Tilt-Hex performing a chirp trajectory on the x-axis. From top to bottom, the actuator forces and their derivatives. In particular, all the signals remain inside the feasible region delimited by the constraints.}
	\label{fig:plots_TiltHex_chirp_limits}
\end{figure}

As far as the actuator data are concerned, consider the last plot of Fig.~\ref{fig:plots_QR_chirp_traj} and the bottom one in Fig.~\ref{fig:plots_TiltHex_chirp_traj}. The rotor velocities in the second case span the feasible set in a wider way. While in the quadrotor experiment the rotor speed constraints are not even approached, in the Tilt-Hex case they become frequently active. The same applies for the generated thrust forces, as shown in the first plots of 
Fig.~\ref{fig:plots_QR_chirp_limits} and Fig.~\ref{fig:plots_TiltHex_chirp_limits}. 
In the specific case, the fact that the lower bounds are reached more often than the upper ones is simply due to the platform mass. Indeed, from the first plot of Fig.~\ref{fig:plots_TiltHex_chirp_limits} we can see that the mean hovering value per actuator is approximately $4\si{N}$, which is closer to the lower saturation level. On the other hand, the velocities of a more massive vehicle would have approached more easily the upper part of the plot.
Finally, from the other plots of Fig.~\ref{fig:plots_TiltHex_chirp_limits} it should be appreciated how the Tilt-Hex force derivatives related to actuators $\{2,3,5,6\}$ and their state-dependent limitations oscillate in a much more dynamic way compared to the ones of the quadrotor, depicted in Fig.~\ref{fig:plots_QR_chirp_limits}. 
This is due to the larger span of the feasible force set required by the controller to these actuators, as a consequence of their geometric arrangement, for the tracking of the same trajectory. 

We now shortly compare the results achieved by this NMPC algorithm with the ones obtained by the reactive static-feedback controller designed in our previous work~\cite{2018d-FraCarBicRyl}. In particular, Fig.~5 of~\cite{2018d-FraCarBicRyl} presents the tracking results related to the same trajectory and the same MRAV.
To provide a better mean for the reader to appreciate the experimental results, we report a quantitative comparison of the data obtained with the two controllers in Tab.~\ref{tab:ctrl_comp}. 
Regarding the position error, we achieved a reduced root mean square position (RMS) error with the NMPC regulator, in particular in the two lateral tails of the trajectory, where the error is always bounded within $4\si{cm}$). Furthermore, while the error profile obtained with the reactive controller was more or less uniformly distributed along the trajectory, in the present case its trend seems to be proportional to the chirp frequency, which also has a triangular envelope.
This effect could be explained by the predictive nature of the algorithm discussed in this paper. Indeed, while the reactive regulator always acts in relation to the instantaneous value of the desired trajectory, the NMPC response is affected by the future evolution of the former, which depends on the chirp frequency.
As far as the orientation tracking is concerned, a relevant improvement is achieved. As a matter of fact, the maximum pitch error is reduced from $23$\,\si{deg} to $13$\,\si{deg}, i.e., a decrease of more than $43\%$.
Furthermore, analyzing the plot of the rotor velocities we see that now they evolve in a larger range, meaning that the NMPC regulator is exploiting the actuator capabilities in a more efficient way. This is a consequence of the fact that the previous controller deals with a less precise -- and more conservative -- model of the platform.

\begin{table}[t]
	\caption{Numerical comparison of the performance achieved in the experimental validation with the NMPC presented in this paper and the static-feedback reactive controller in~\cite{2018d-FraCarBicRyl}.}
	\label{tab:ctrl_comp}
	\centering
	\renewcommand\arraystretch{1.4}
	\resizebox{\columnwidth}{!}{
	\begin{tabular}{CCC}
		\hline
		\multicolumn{3}{c}{\textbf{Chirp trajectory}} \\
		\hline
		\text{Parameter [$\bullet$]} & \text{Tiltex (CTRL~\cite{2018d-FraCarBicRyl})} & \text{Tilt-Hex (NMPC)} \\
		\hline
		e_{p,\text{MAX}}\ [\si{m}]  &  [0.076,\ 0.032,\ 0.041]^\top & [\mathbf{0.064,\ 0.025,\ 0.018}]^\top \\
		e_{p,\text{RMS}}\ [\si{m}]  &  [0.020,\ 0.009,\ 0.013]^\top & [\mathbf{0.015,\ 0.006,\ 0.007}]^\top \\
		e_{\eta,\text{MAX}}\ [\si{deg}]  &  [\mathbf{2.2},\ 23.2,\ \mathbf{1.7}]^\top & [5.1,\ \mathbf{13.3},\ 7.3]^\top \\
		e_{\eta,\text{RMS}}\ [\si{deg}]  &  [\mathbf{0.8},\ 6.0,\ \mathbf{0.6}]^\top & [1.2,\ \mathbf{3.9},\ 1.4]^\top \\
		f_{i,\text{MIN}},\ f_{i,\text{MAX}}\ [\si{N}] & 1.803,\ 6.836 & \mathbf{0.231,\ 10.251} \\
			\end{tabular}
	}
\end{table}


\subsubsection{Discontinuous trajectory}

\begin{figure}[ht]
	\centering
		\includegraphics[width=\figWidth\columnwidth]{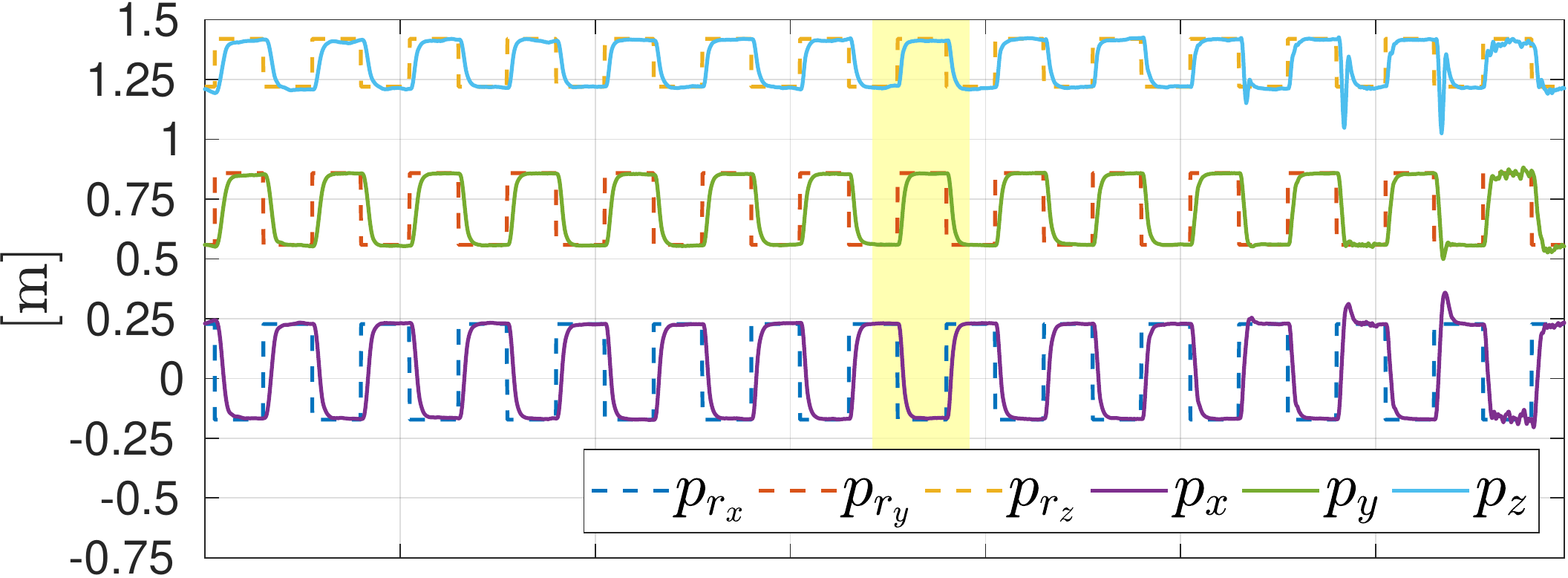}\\
		\includegraphics[width=\figWidth\columnwidth]{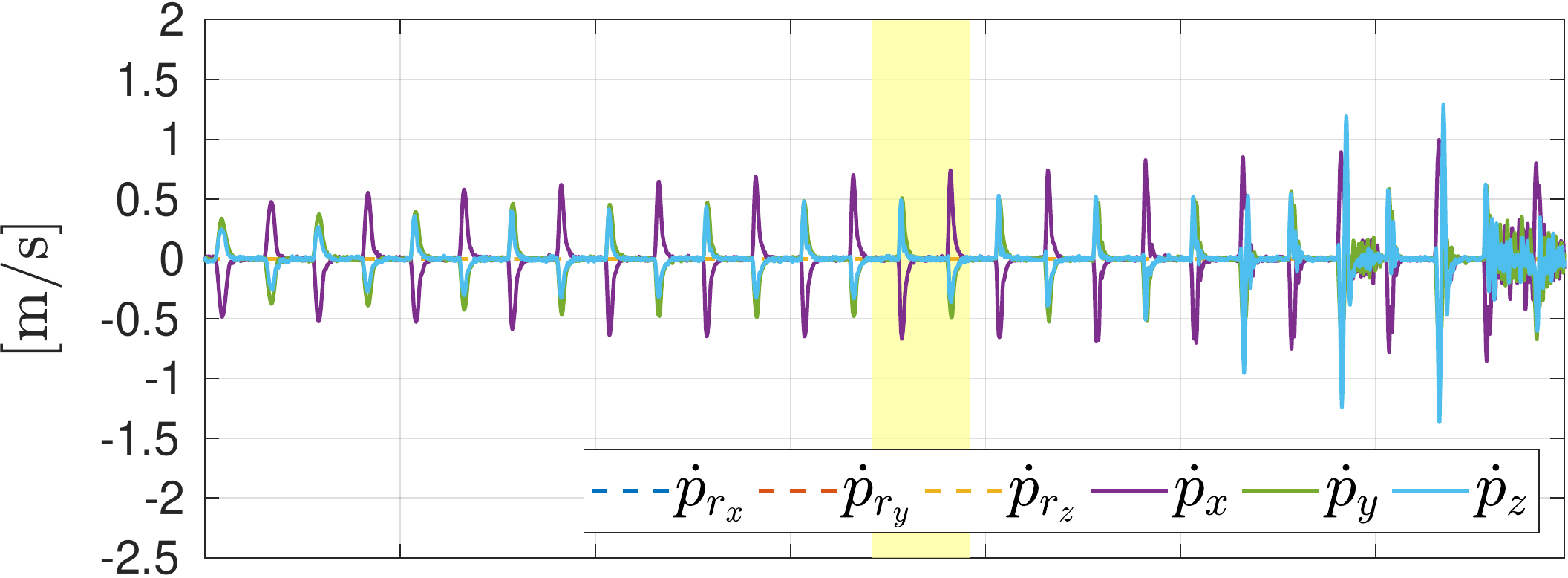}\\
		\includegraphics[width=\figWidth\columnwidth]{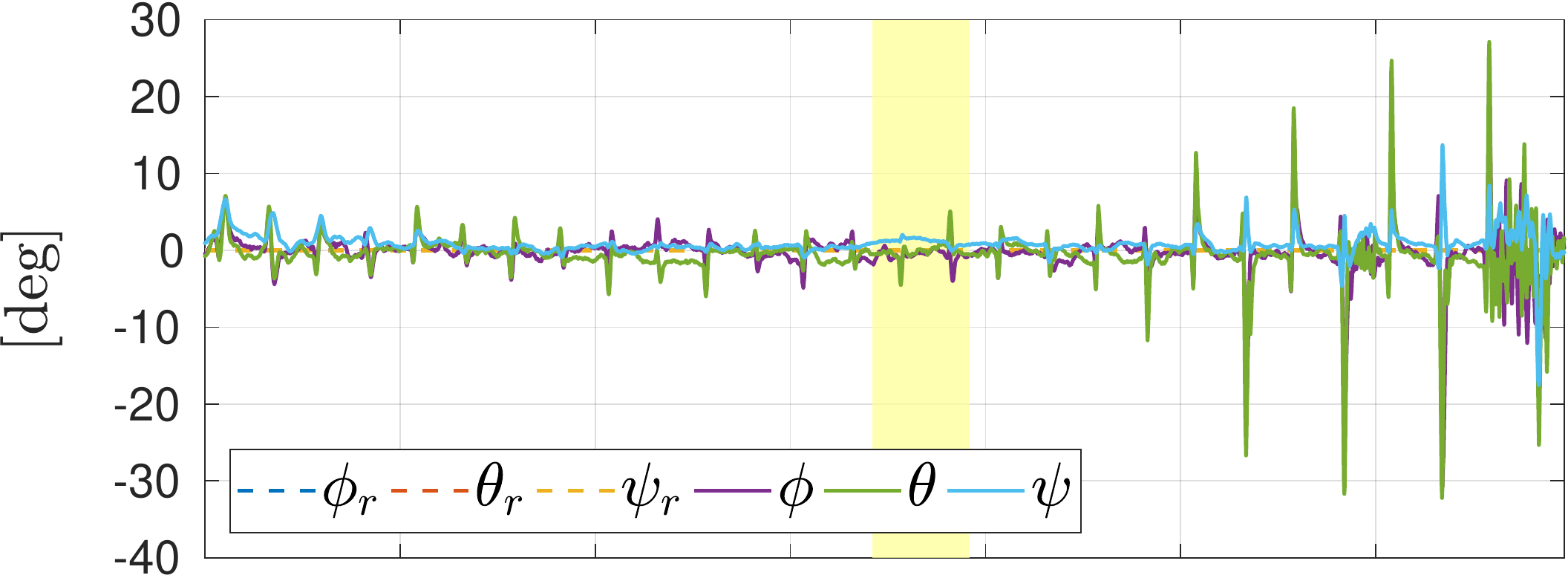}\\
		\includegraphics[width=\figWidth\columnwidth]{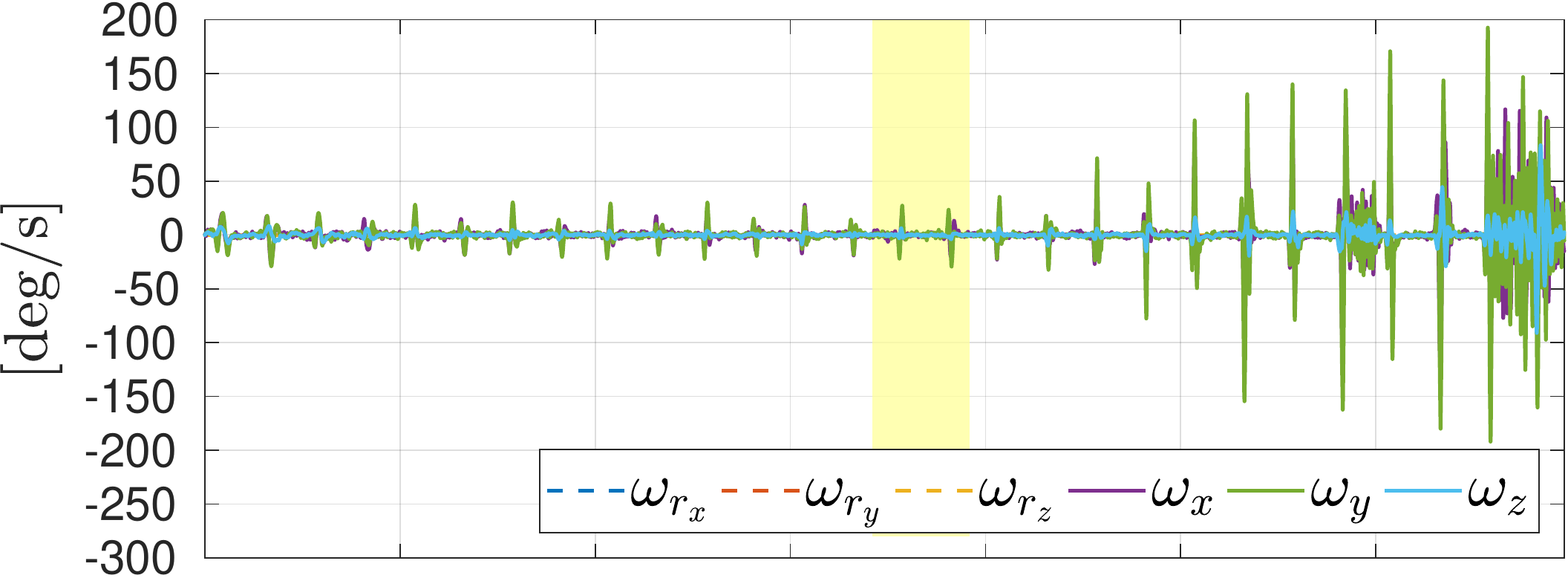}\\
		\includegraphics[width=\figWidth\columnwidth]{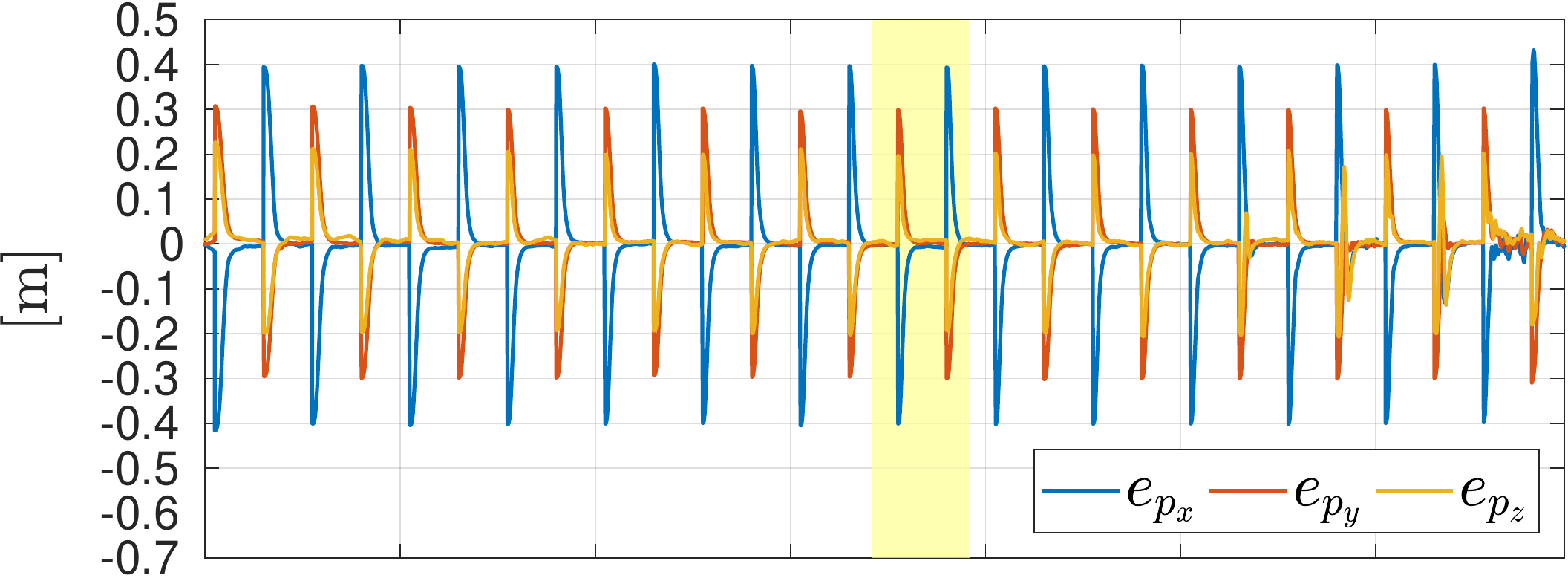}\\
		\includegraphics[width=\figWidth\columnwidth]{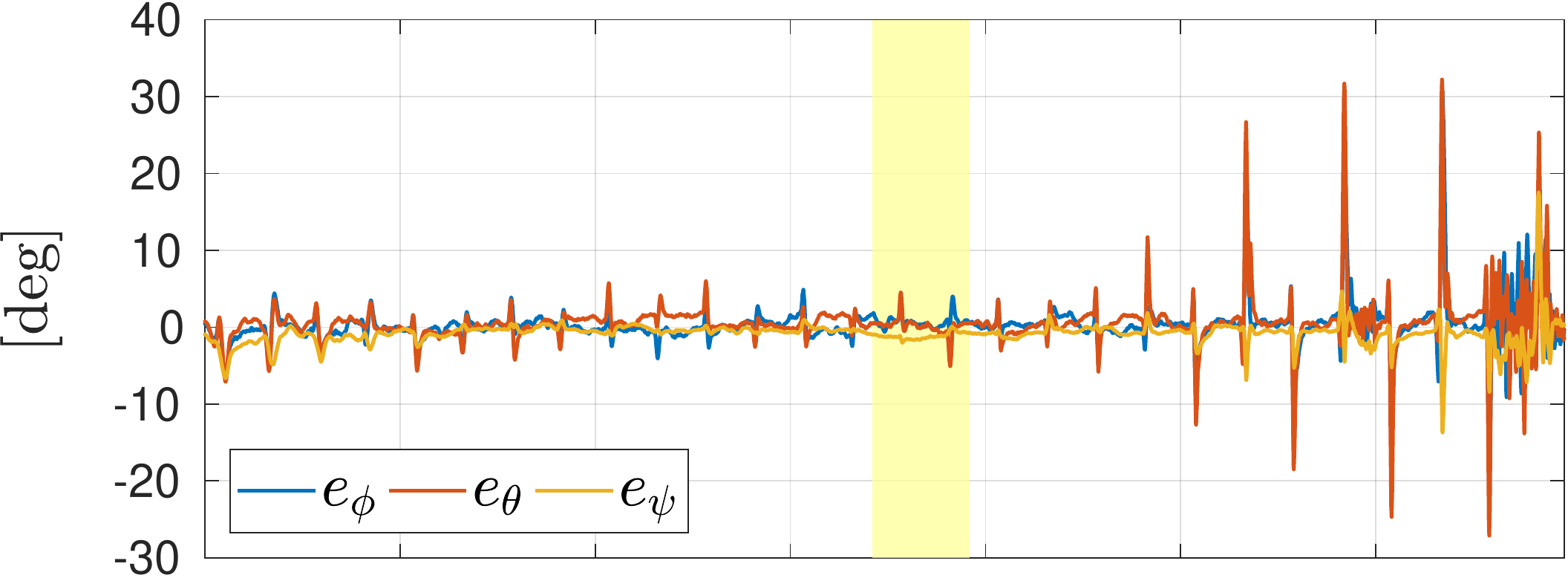}\\
		\includegraphics[width=\figWidth\columnwidth]{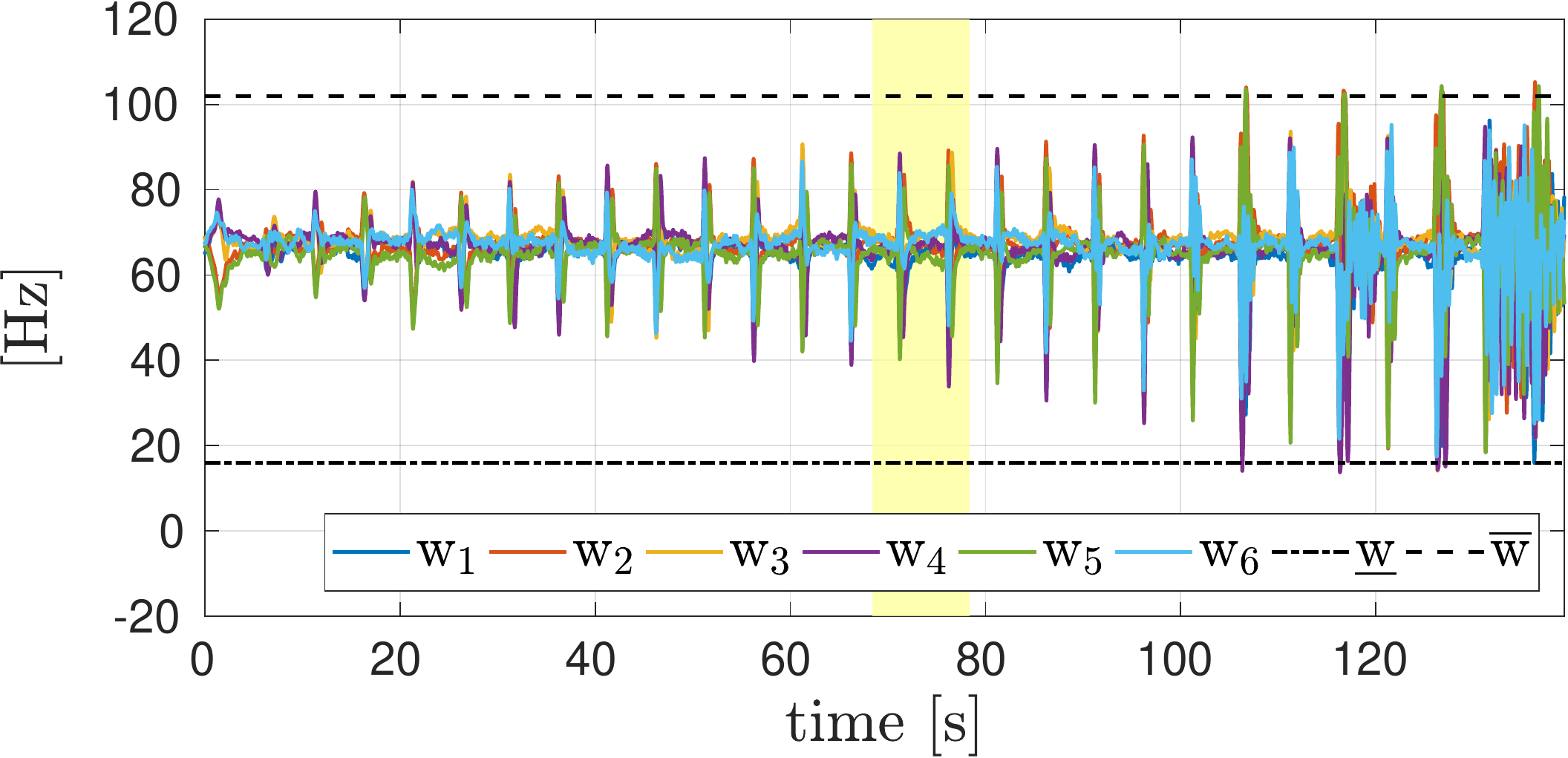}
	\caption{Plots of the Tilt-Hex tracking a discontinuous trajectory with steps in the position, while the controller limits are increased (the yellow region highlights the use of the identified ones). From top to bottom, the position, linear velocity, orientation and angular velocity tracking, the position and orientation errors, and the actuator spinning velocities.}
	\label{fig:plots_TiltHex_steps_traj}
\end{figure}

\begin{figure}[ht]
	\centering
		\includegraphics[width=\figWidth\columnwidth]{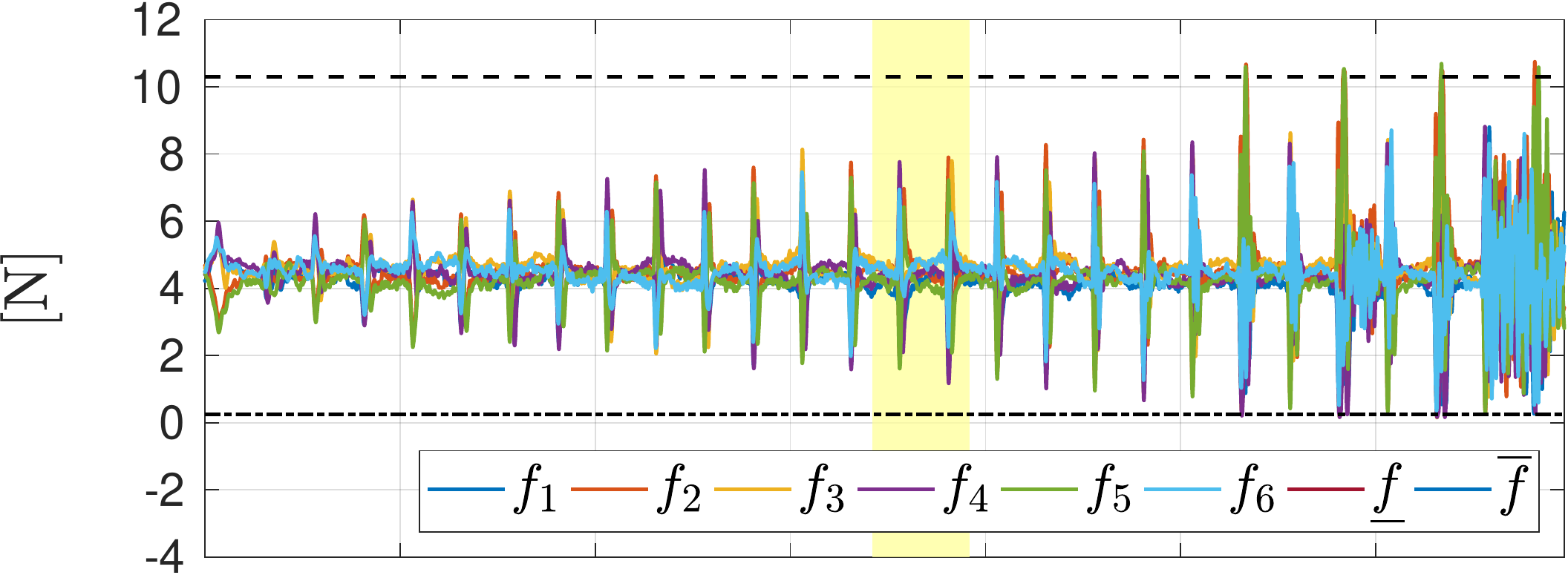}\\
		\includegraphics[width=\figWidth\columnwidth]{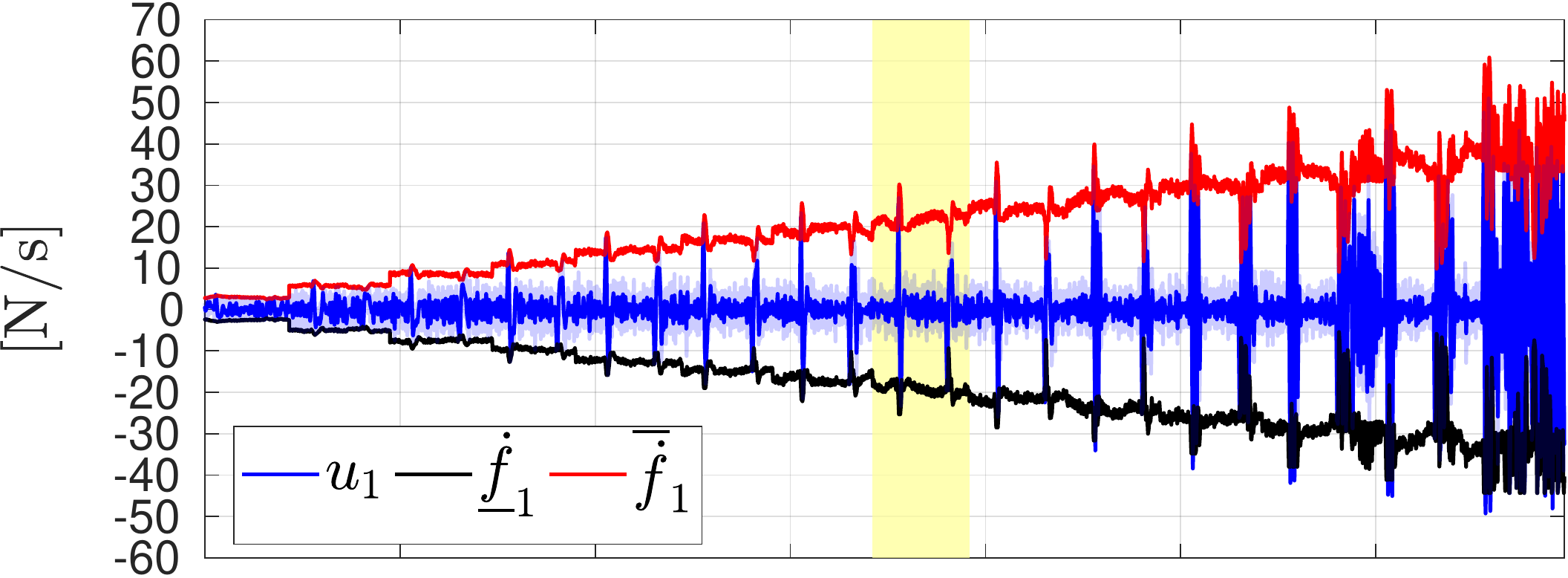}\\
		\includegraphics[width=\figWidth\columnwidth]{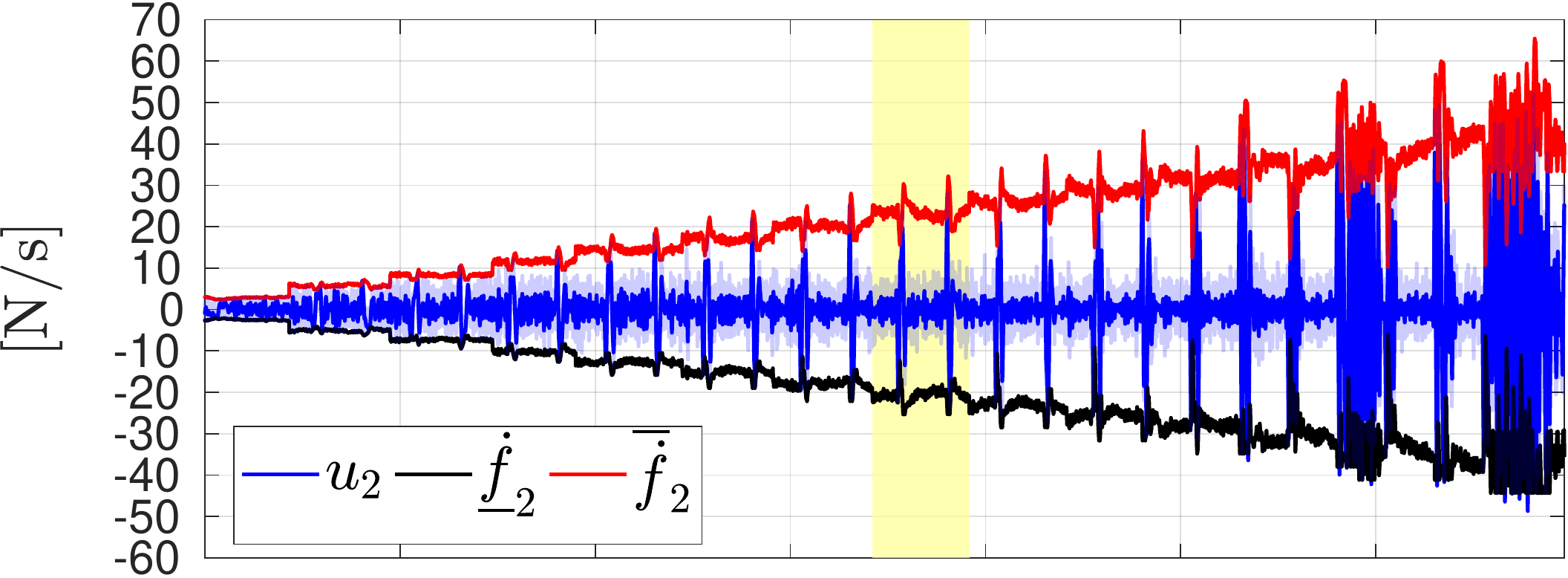}\\
		\includegraphics[width=\figWidth\columnwidth]{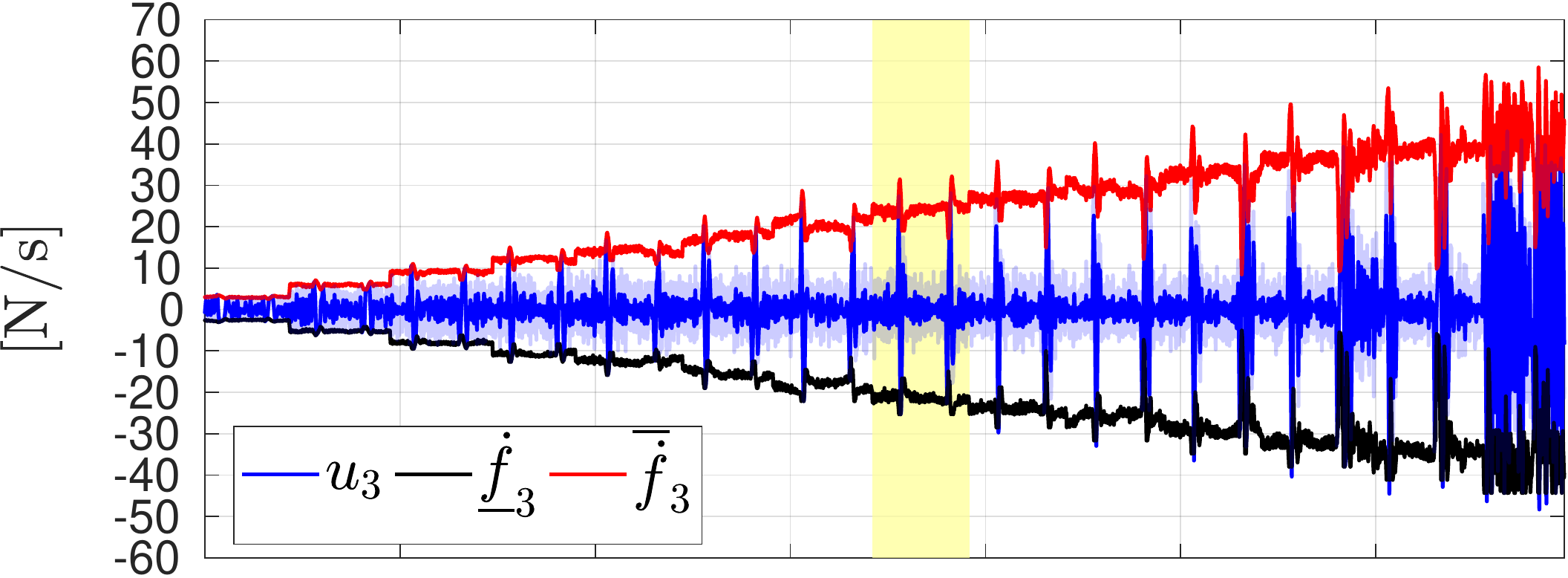}\\
		\includegraphics[width=\figWidth\columnwidth]{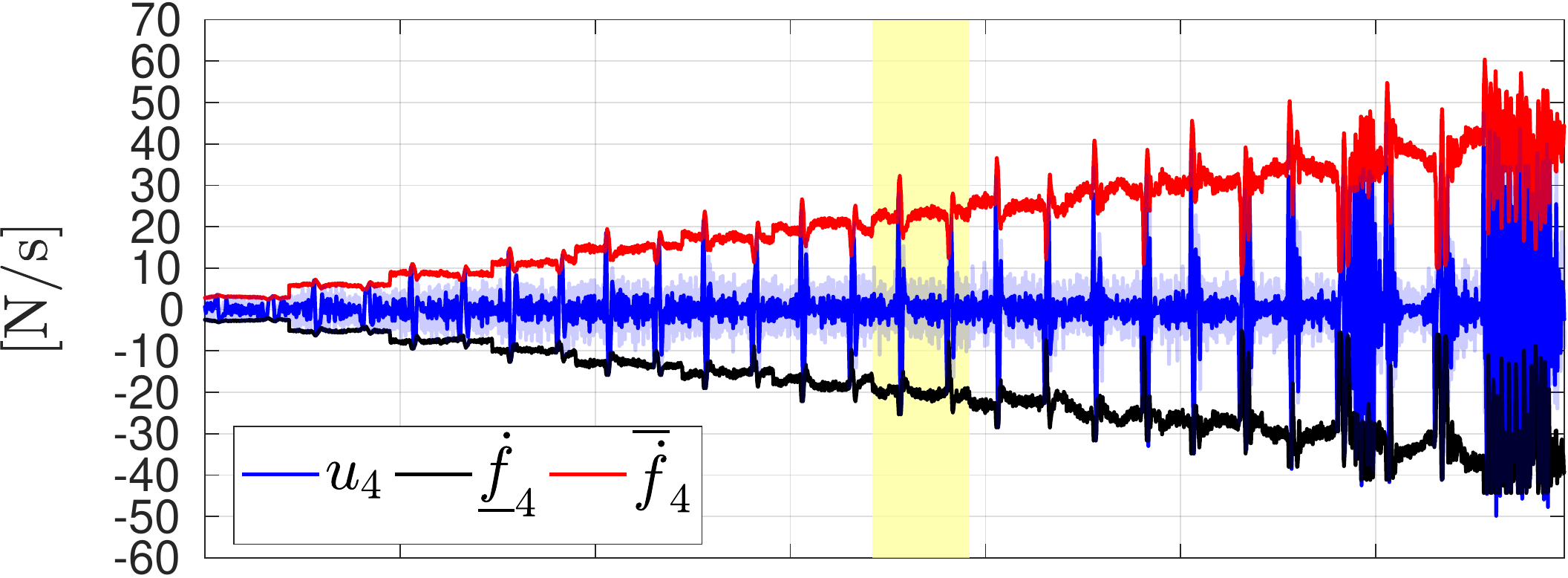}\\
		\includegraphics[width=\figWidth\columnwidth]{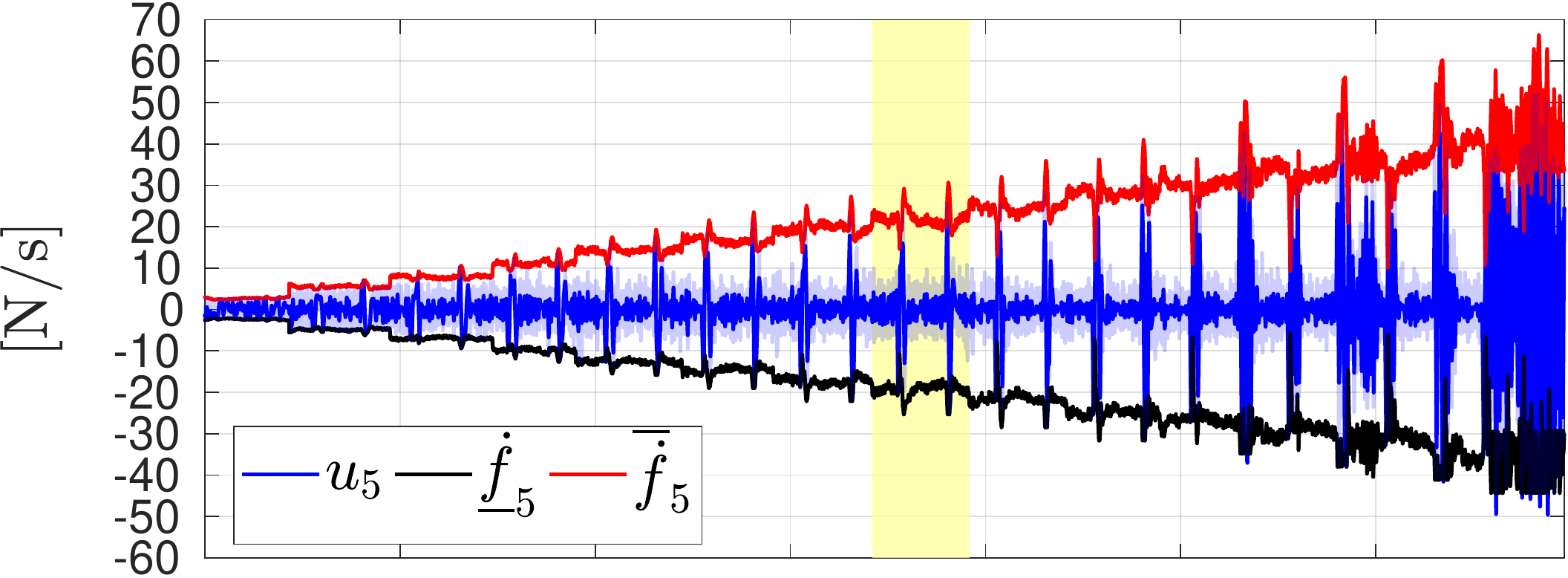}\\
		\includegraphics[width=\figWidth\columnwidth]{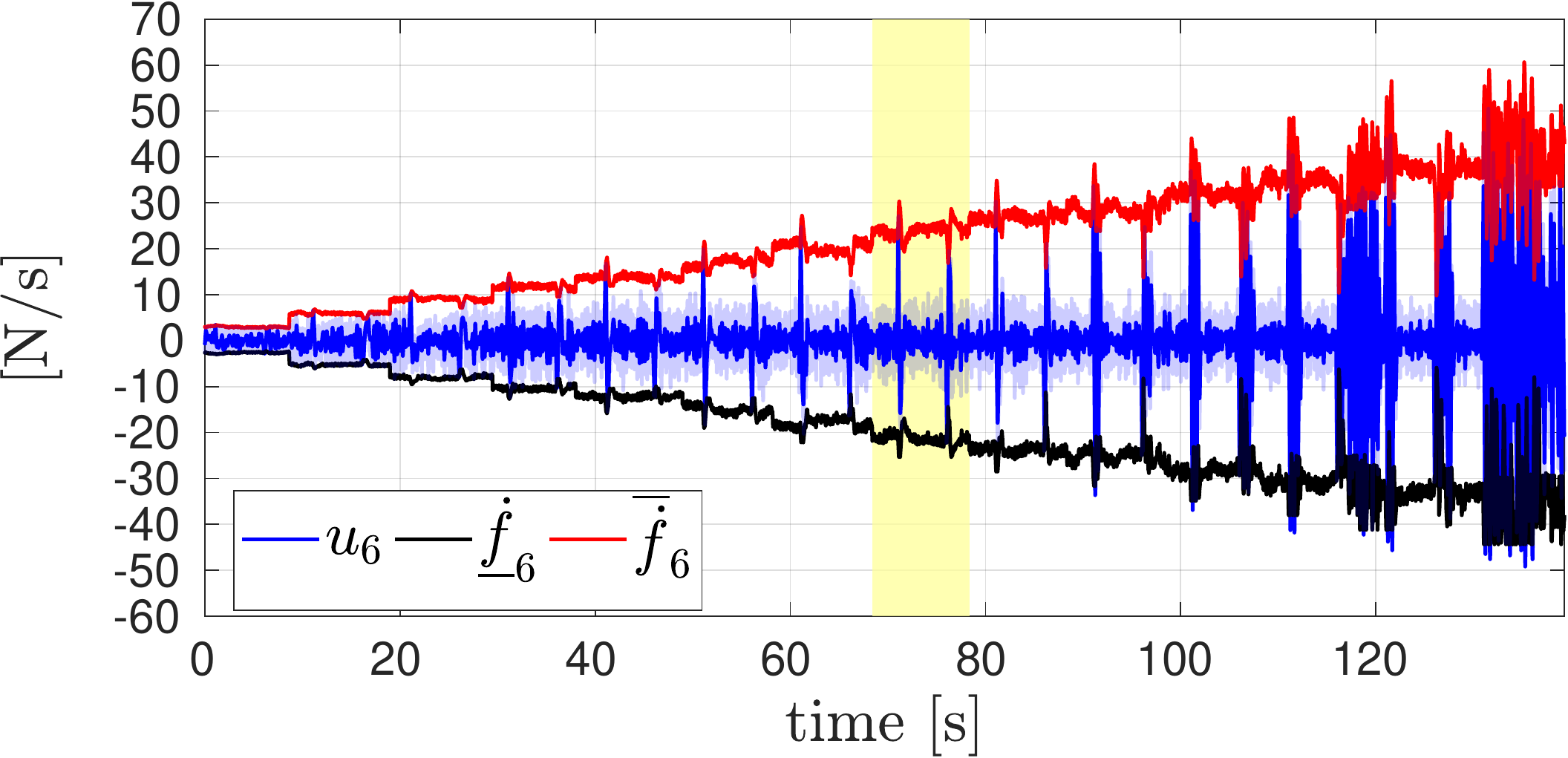}
	\caption{Plots of the Tilt-Hex tracking a discontinuous trajectory with steps in the position, while the controller limits are increased (the yellow region highlights the use of the identified ones). From top to bottom, the actuator forces and their derivatives. In particular, all the signals remain inside the feasible region delimited by the constraints.}
	\label{fig:plots_TiltHex_steps_limits}
\end{figure}

To assess the effectiveness of our procedure in identifying meaningful actuator limitations for a non-specific hardware setup, we replicated the experiment described in~\ref{subsubsec:quad_steps} using the Tilt-Hex robot.
The plots related to this test are depicted in Fig.~\ref{fig:plots_TiltHex_steps_traj} and in Fig.~\ref{fig:plots_TiltHex_steps_limits}. For this experiment, the limits to the NMPC were scaled by the user after two consecutive jumps of the MRAV, while $\boldsymbol{\epsilon}_{\pv}=[-0.5\, 0.3\, 0.2]^{\top}$\,\si{m}.
The experiment outcomes show that the best step responses are achieved when the actuator limits are closer to the identified ones. This confirms again the validity of our approach. 
Furthermore, also in this case, the instability is reached when $\rho=\frac{7}{4}$, i.e., when the force derivative bounds are almost the double of the identified ones, which shows an adequate margin of conservativeness for the chosen limits.
Also for this experiment, a numerical comparison of the most interesting results obtained with the two aforementioned aerial vehicles driven with the identified limits ($\rho=1$), is provided in Tab.~\ref{tab:exp_comp}.
Given the discontinuous nature of the reference position trajectory, it is not very meaningful to analyze the maximum or the RMS position error, in this case. On the other hand, it is much more interesting to compare the orientation tracking in the two cases. While the UDT platform has to consistently deviate from the reference flat-hovering orientation in order to generate the needed lateral force to track the position step, the MDT robot almost does not need any re-orientation of its chassis. Such remarkable effect can be visually appreciated in the attached multimedia content, where the motion of the two MRAVs have been juxtaposed. To conclude, we highlight that also in this case a bigger span of the feasible rotor velocities is obtained with the MDT MRAV.

\section{Validation with real-time simulations}

\begin{figure}[t]
	\centering
		\includegraphics[width=0.49\columnwidth]{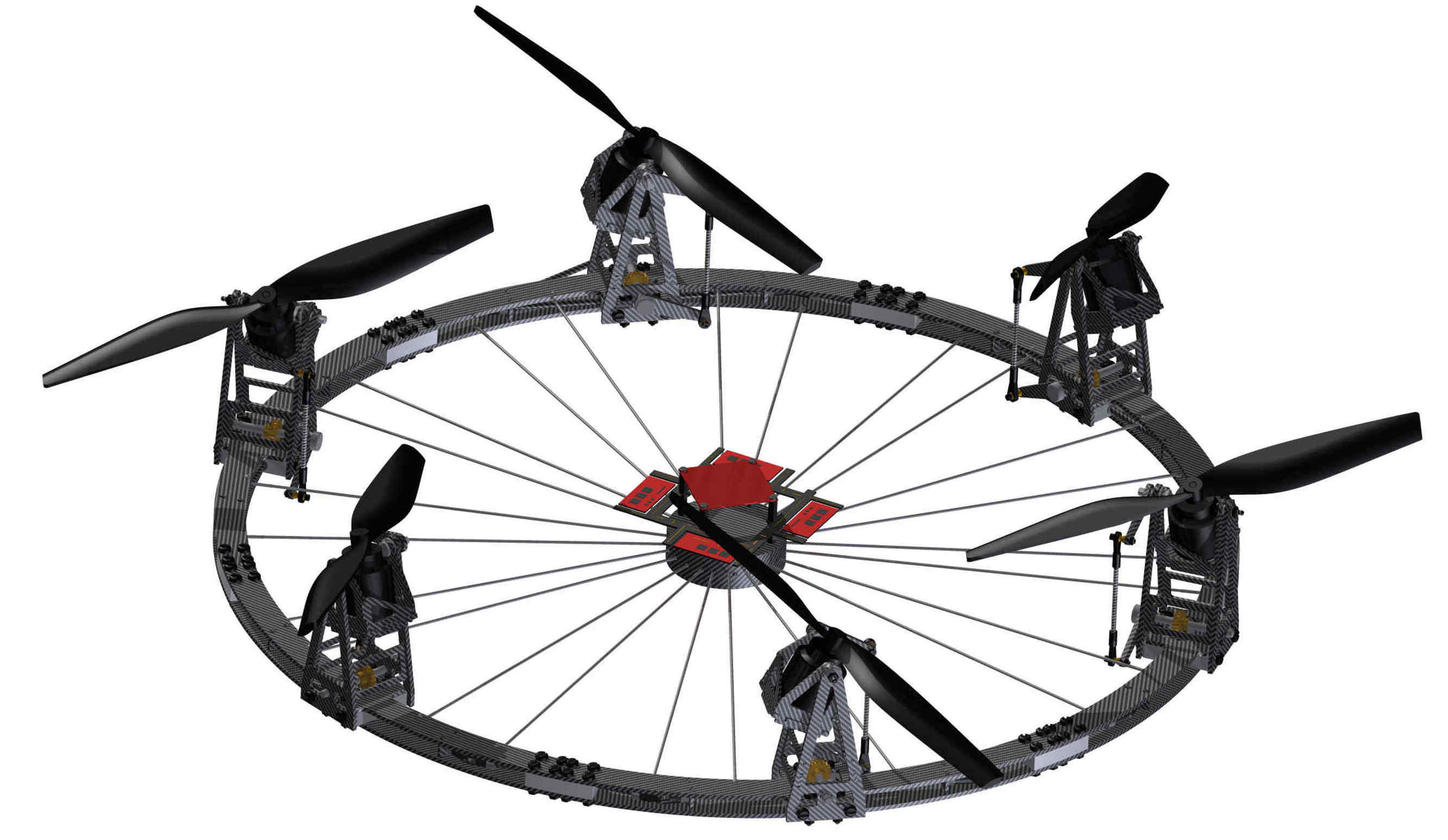}
		\includegraphics[width=0.49\columnwidth]{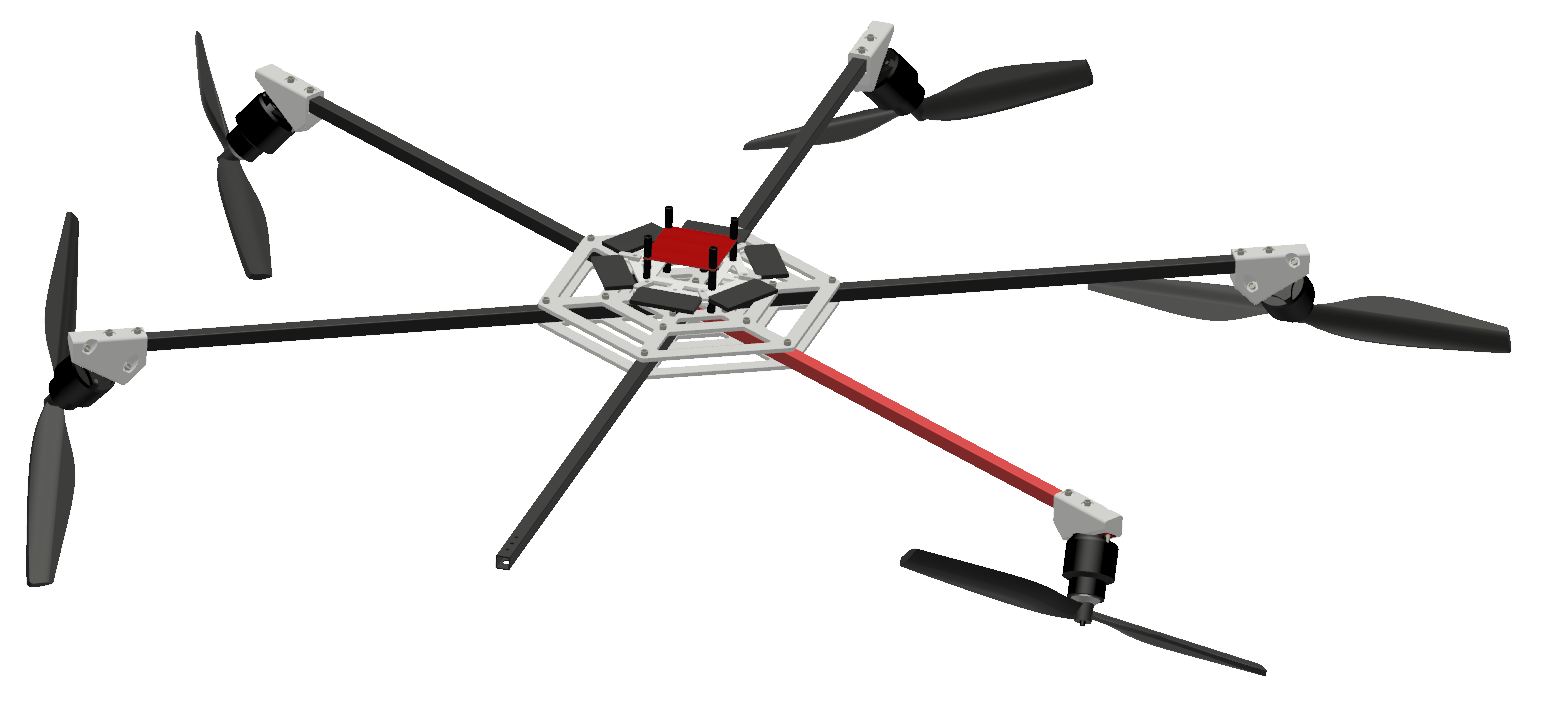}
	\caption{Photos of the FAST-Hex (left) and the rotor failed Tilt-Hex (right).}
	\label{fig:robots2}
\end{figure}

In order to support the claim that our framework can deal with a generic MRAV design, we provide additional numerical validations with two other different vehicle models, shown in Fig.~\ref{fig:robots2}.
The first one, depicted on the left, is called FAST-Hex, i.e., \emph{Fully-Actuated by Synchronized Tilting propellers hexarotor}. This original MRAV, introduced in our previous work~\cite{2016j-RylBicFra}, has the capability of synchronously modifying the orientation of its actuators, thanks to a single additional servo-motor. Exploiting this further degree of freedom, the FAST-Hex can actively regulate the angle $\alpha$, c.f. Fig.~\ref{fig:modeling}, allowing the robot to pass from an UDT configuration to an MDT one, and conversely. Moreover, the value of $\alpha$ can be automatically regulated, i.e., without the need of an external planner.
The physical parameters of the FAST-Hex are condensed in Tab.~\ref{tab:robots_parameters_fasthex}.

The second vehicle, shown on the right of Fig.~\ref{fig:robots2}, is a pentarotor (a multi-rotor with five propellers) obtained as a failed Tilt-Hex MRAV, i.e., the platform already described in the experimental validation, but after a rotor failure. In particular the $6-th$ rotor is not allowed to spin, due to, e.g., a technical problem, and cannot exert a thrust force and generate a drag torque. For this reason, from a control point of view, we consider that such actuator is not present.
The rotor failure essentially modifies the available set of body forces and torques. As already pointed out in previous contributions, in case $\alpha=\beta=0$ it is not possible with five uni-directional actuators to generate torques in pitch and roll without generating a residual disturbing torque in the yaw axis, cf.~\cite{achtelik2012design,giribet2016analysis}. In the more general case in which $(\alpha,\beta)\ne(0,0)$, however, the platform maintains the ability to hover, see~\cite{giribet2016analysis}. Nevertheless, the hovering orientation can not be flat any more, and depends of the actuator tilting angles, cf.~\cite{2018a-MicRylFra}. We show that the NMPC controller can satisfactorily deal with the problem of static hovering, without the need to a-priori compute the steady-state orientation.

\subsection{Simulations with the FAST-Hex}

In order to take into account the evolution of the angle $\alpha$ and to let the NMPC algorithm manage its automatic regulation, we expanded the state and the input vector, defined in~\eqref{eq:state_vector} and in~\eqref{eq:input_vector}, in the following way
 \begin{align}
	\label{eq:state_vector_fast}
	\mathbf{x} &:=
	\begin{bmatrix}
		\pv^\top\,\dot{\pv}^{\top}\,\etav^{\top}\,\omegav^{\top}\,\gammav^{\top},\alpha
	\end{bmatrix}
	^{\top}
	\\
	\label{eq:input_vector_fast}
	\mathbf{u} &:=
	\begin{bmatrix}
		\dot{\gammav},\dot{\alpha}
	\end{bmatrix}
\end{align}
The angle $\alpha$ is now a component of the state vector, while $\dot{\alpha}$ is regarded as an additional control input to be optimized at each control iteration. This allows to constrain the synchronous tilting angle and its derivative within their feasible sets, computed accordingly to the data of the real MRAV prototype designed in~\cite{2016j-RylBicFra}.
According to this choice, for the FAST-Hex model we have $\xv \in \mathbb{R}^{19}$ and $\uv \in \mathbb{R}^{7}$.

\begin{table}[t]
	\renewcommand\arraystretch{1.3}
	\caption{Physical parameters of the FAST-Hex.}
	\label{tab:robots_parameters_fasthex}
	\centering	
	\resizebox{\figWidth\columnwidth}{!}{%
	\begin{tabular}{CCC}
		\hline
		\multicolumn{3}{c}{\textbf{FAST-Hex}} \\
		\hline
		\text{Parameter}  & \text{Value} & \text{Unit} \\
		\hline
		m & 2.4 & \si{Kg} \\
		\Jm(:,1) & [0.042\ 0 \ 0]^{\top} & \si{Kg\,m^2} \\
		\Jm(:,2) & [0 \ 0.042 \ 0]^{\top} & \si{Kg\,m^2} \\
		\Jm(:,3) & [0 \ 0 \ 0.083]^{\top} & \si{Kg\,m^2} \\
		c_i & (-1)^i & [\ ] \\
		c_f^{\tau} & \num{1.9e-2} & \si{m} \\
		c_f & \num{9.9e-4} & \si{N/Hz^2} \\
		\hline
		\Rm_{A_i}^B & \Rm_z \big ((i-1)\frac{\pi}{3})\big)\Rm_x(\alpha_i)\Rm_y(\beta) & [\ ] \\
		\pv_{A_i}^B & \Rm_z \big((i-1)\frac{\pi}{3})\big) [\ell\ 0\ 0]^{\top} & [\ ] \\
		\alpha_i & (-1)^{i-1}\ |\alpha| & \si{deg} \\
		\beta & 0 & \si{deg} \\
		\ell & 0.315 & \si{m} \\
		\hline
	\end{tabular}	} 
\end{table}

As it can be appreciated from Fig.~3 in~\cite{2016j-RylBicFra}, the larger the angle $\alpha$, the larger the set of body-frame lateral forces. This translates also into the possibility of decoupling the control of the body force and moment in a larger extent, which becomes particularly useful in many realistic scenarios, ranging from 6D trajectory tracking, see~\cite{2018d-FraCarBicRyl}, to aerial physical interaction tasks, see~\cite{2018g-StaBicSabAreMisFra,2019h-RylMusPieCatAntCacFra}, and disturbance rejection in general.
On the other hand, the increase of the tilting angle implies also an increment in the energy consumption. In fact, the progressive decrease in the projection of the thrust vector along $\zW$ must be compensated by an increase in the thrust intensity.
In view of these considerations, it might be beneficial to regulate the angle $\alpha$ w.r.t. the particular task to be accomplished, while trying to minimize the energy consumption. In order to fulfill this requirement, we expanded also the output vector as follows
\begin{align}
{\bf y}(t) = {\bf h} \left(\xv(t), \uv(t) \right) = \left[
\begin{array}{c}
\pv(t)\\
\dot{\pv}(t)\\
\ddot{\pv}\left(\xv(t), \uv(t) \right) \\
\etav(t) \\
{\omegav}(t)\\
\dot{\omegav}\left(\xv(t), \uv(t) \right) \\
c_e\left(\xv(t), \uv(t) \right)
\end{array}
\right]
\end{align}
where the cost related to power consumption is taken into account using the following additional cost
\begin{align}
c_e\left(\xv(t), \uv(t) \right) = \sum_{i=1}^n f_i^2
\end{align}
which is integrated along the prediction horizon.
Such model has been chosen mainly due to its simple dependency on the state components $f_i$, but other models can be employed.

In this context, we target the classical problem of trajectory regulation to a certain 6D configuration, i.e., the flat hovering, adding the effect of an external unknown disturbance from the environment, which emulates, in a simplified but meaningful way, the scenario of a physical interaction task or an external wind. In the first simulation, we exploit the possibility of regulating $\alpha$. In this way we show how our NMPC algorithm can automatically and actively manipulate the additional control input $\dot{\alpha}$, thus improving our previous work~\cite{2016j-RylBicFra}. Furthermore, in order to demonstrate the usefulness of this supplementary degree of freedom, we present the results of the same simulation, where the tilting angle $\alpha$ is forced to assume different fixed values and cannot be regulated.

\begin{table}[t]
	\renewcommand\arraystretch{1.3}
	\caption{Parameters used in the FAST-Hex simulation.}
	\label{tab:FAST_sim}
	\centering
	\resizebox{\figWidth\columnwidth}{!}{%
	\begin{tabular}{CCC}
		\hline
		\text{Parameter}  & \text{Value} & \text{Unit} \\
		\hline
		\sigmav_{\pv} & [\sqrt{0.005}\ \sqrt{0.005}\ \sqrt{0.005}]^{\top} & \si{m} \\
		\sigmav_{\dot{\pv}} & [\sqrt{0.02}\ \sqrt{0.02}\ \sqrt{0.02}]^{\top} & \si{m/s} \\
		\sigmav_{\etav} & [\sqrt{1}\ \sqrt{1}\ \sqrt{1}]^{\top} & \si{deg} \\
		\sigmav_{\omegav} & [\sqrt{0.15}\ \sqrt{0.15}\ \sqrt{0.05}]^{\top} & \si{deg/s} \\
		\Omega_{\text{filt}} & {25} & \si{rad/s} \\
		\hline
		t_1 & 10 & \si{s} \\
		t_2 & 20 & \si{s} \\
		\fv_{\text{dist}}(t_2) & 3\ [cos(\frac{\pi}{3})\ sin(\frac{\pi}{3})\ 0]^{\top} & \si{N} \\
		\hline
		\underline{{\alpha}} \ ,\ \overline{{\alpha}} & -35 \ ,\ 35 & \si{deg} \\
		\underline{\dot{\alpha}} \ ,\ \overline{\dot{\alpha}} & -8.75 \ ,\ 8.75 & \si{deg/s} \\
		\hline
		\Qm_{\pv}(j,j)|_{j=1,2,3} & 50,50,50 & [\ ] \\
		\Qm_{\dot{\pv}}(j,j)|_{j=1,2,3} & 0.5,0.5,0.5 & [\ ] \\
		\Qm_{\etav}(j,j)|_{j=1,2,3} & 15,15,15 & [\ ] \\
		\Qm_{\omegav}(j,j)|_{j=1,2,3} & 0.01,0.01,0.01 & [\ ] \\
		\Qm_{\ddot{\pv}}(j,j)|_{j=1,2,3} & 0.0001,0.0001,0.0001 & [\ ] \\
		\Qm_{\dot{\omegav}}(j,j)|_{j=1,2,3} & 0,0,0 & [\ ] \\
		\Rm_h(j,j)|_{j=1,\dots{},7} & 0,0,0,0,0,0,0 & [\ ] \\
		Q_{e_c} & 0.0005 & [\ ] \\
		\hline
	\end{tabular}  } 
\end{table}

In order to make the simulations more realistic, we added to the measured state a noise, obtained by filtering a zero-mean white Gaussian noise with a first-order causal low-pass filter having a cut-off frequency $\Omega_{\text{filt}}$, whose value has been estimated analyzing real experimental data.
The noise standard deviation values $\sigmav_{\bullet}$ are collected in Tab.~\ref{tab:FAST_sim}, together with the other trajectory parameters, state/input bounds and cost function weights. In particular, the values of $\sigmav_{\bullet}$ are related to very unfavorable conditions compared to the use of typical sensors such as MoCap and gyros.

\subsubsection{Hovering trajectory with unknown lateral force disturbance}

Alongside the presented simulations, the FAST-Hex is required to hover maintaining a flat orientation, i.e.,
\begin{align}
\pv_r &= [0.6\ 0.6\ 0.75]^{\top} ,\
\dot{\pv}_r = \ddot{\pv}_r = [0\ 0\ 0]^{\top} \nonumber \\
\etav_r &= \omegav_r = \dot{\omegav}_r = [0\ 0\ 0]^{\top}
\end{align}
under the effect of a lateral force disturbance $\fv_{\text{dist}}$ with a triangular profile. Such force, unknown to the controller, has a triangular shape from $t_1$ to $t_1+t_2$, with a peak in module of $3$\,\si{N} at $t_2$, while it is $\fv_{\text{dist}}=\mathbf{0}$\,\si{N} elsewhere. 
As the steady-state orientation is not known a priori, the reference of the energetic term $c_{e,r}$ is constantly equal to the one needed for hovering horizontally with $\alpha=0$, i.e., $c_{e,r}=\sum_{i=1}^n (\frac{mg}{6})^2$.

\begin{figure}[t]
	\centering
		\includegraphics[width=\figWidth\columnwidth]{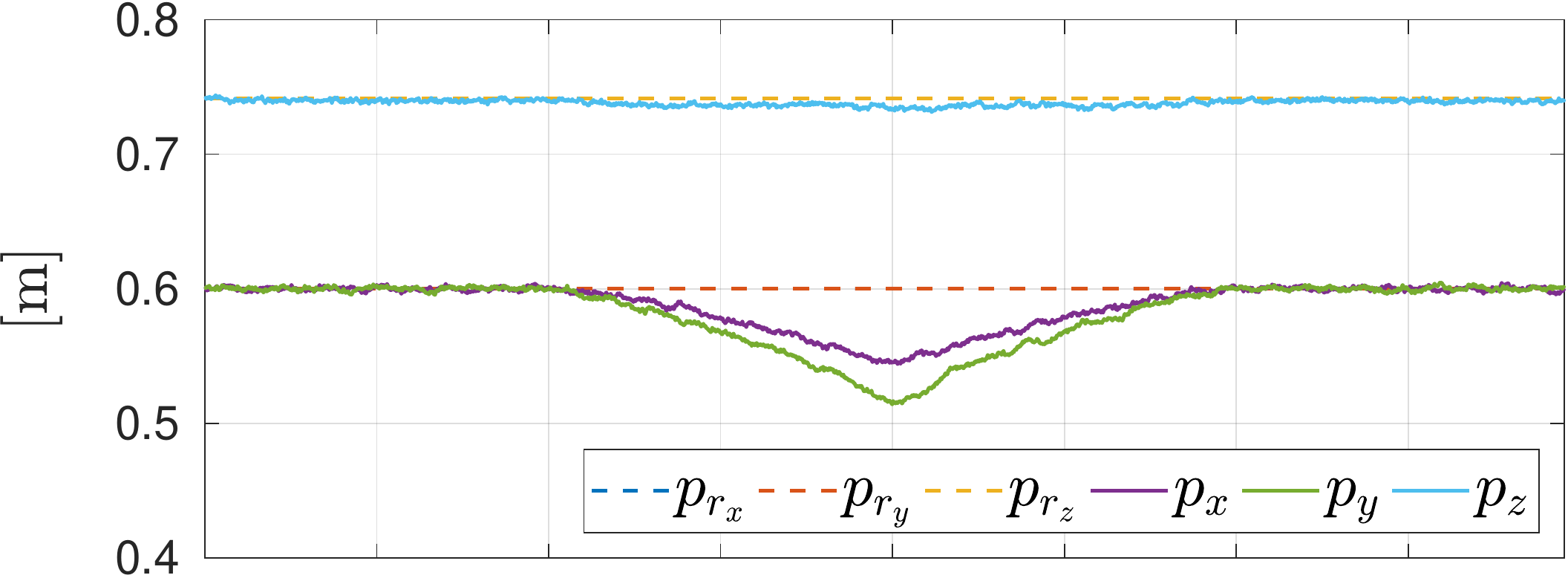}\\
		\includegraphics[width=\figWidth\columnwidth]{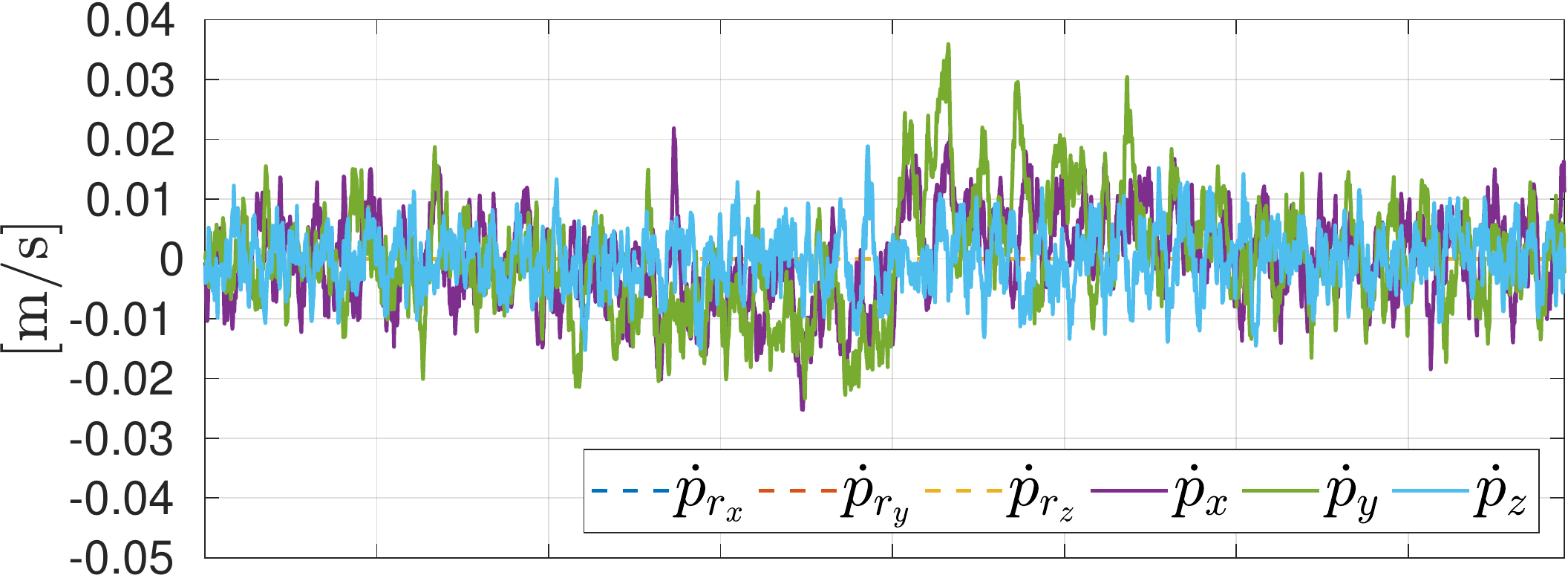}\\
		\includegraphics[width=\figWidth\columnwidth]{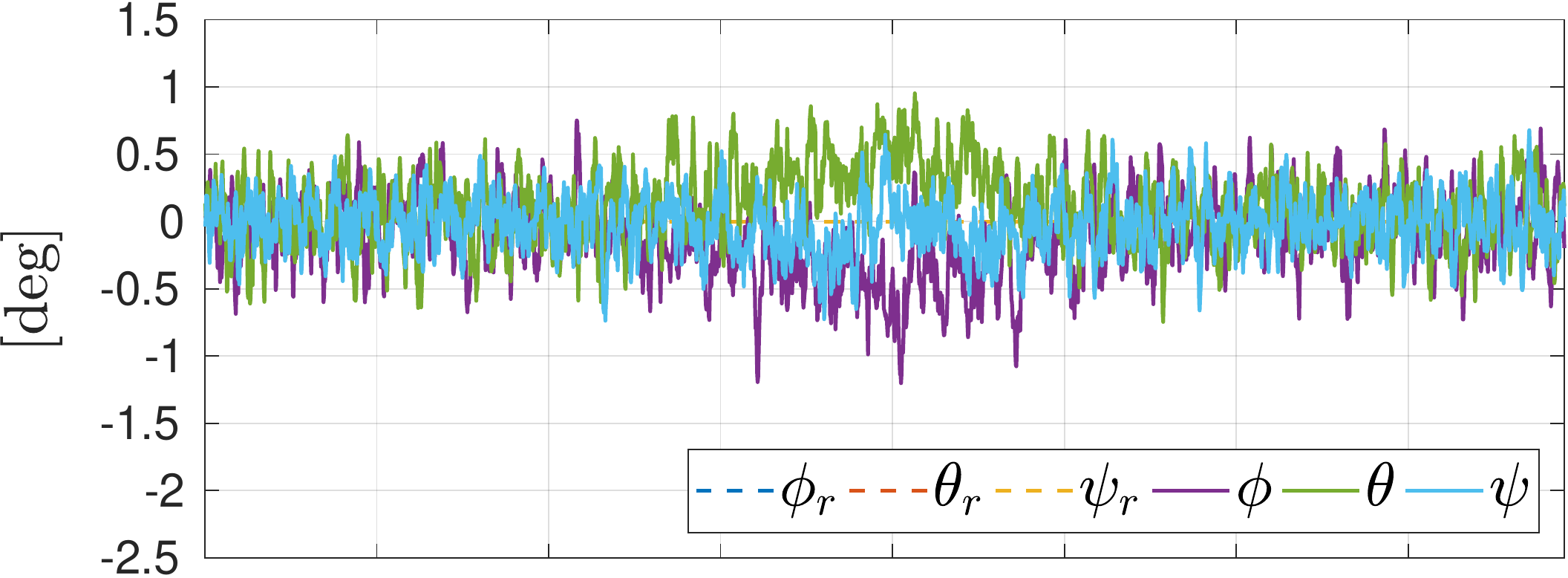}\\
		\includegraphics[width=\figWidth\columnwidth]{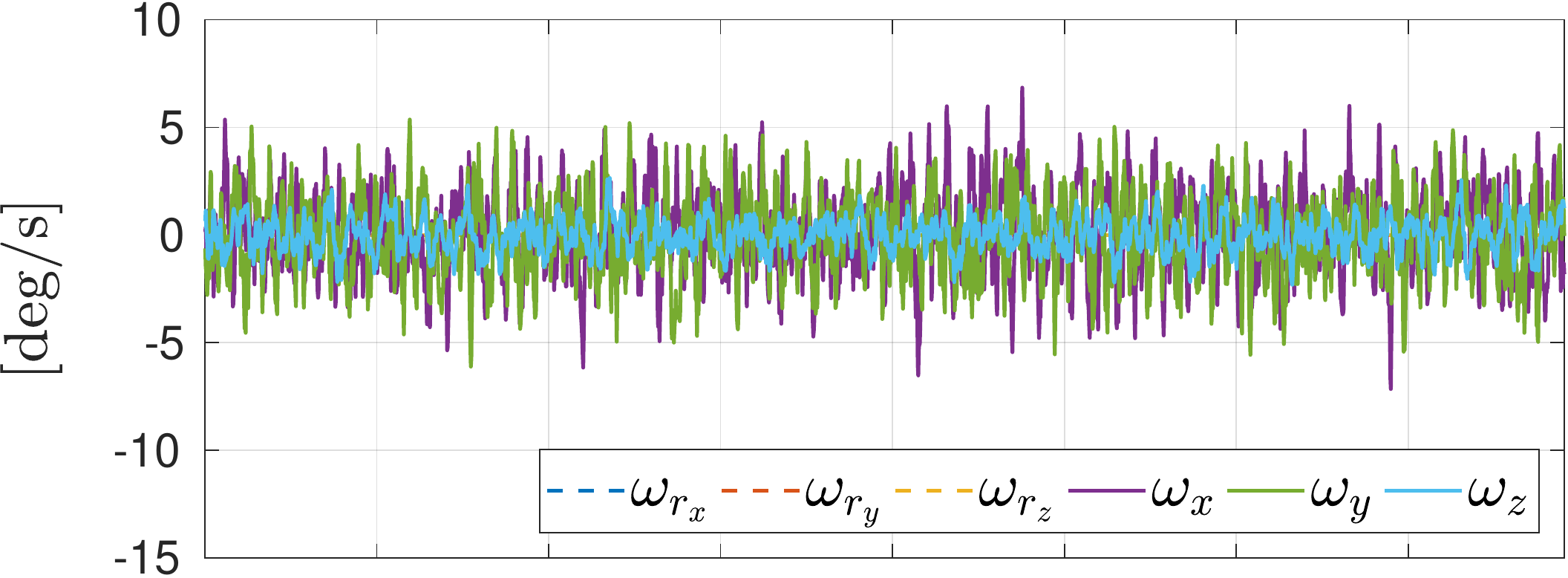}\\
		\includegraphics[width=\figWidth\columnwidth]{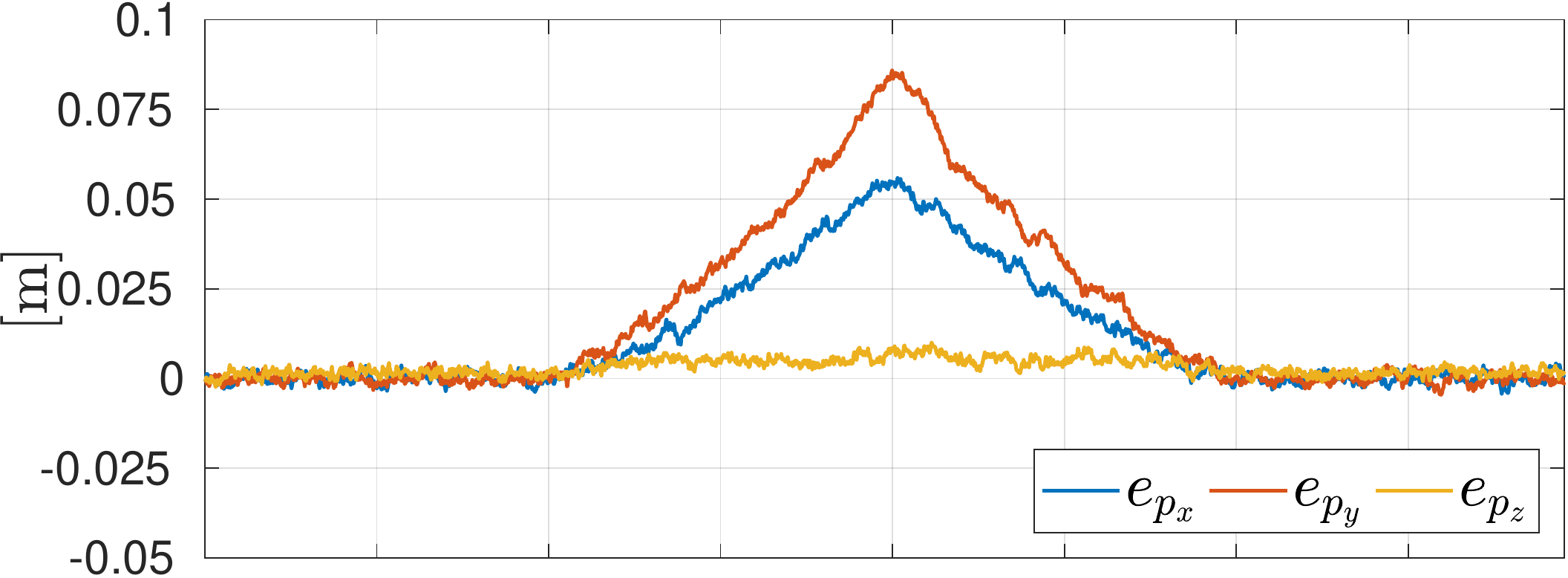}\\
		\includegraphics[width=\figWidth\columnwidth]{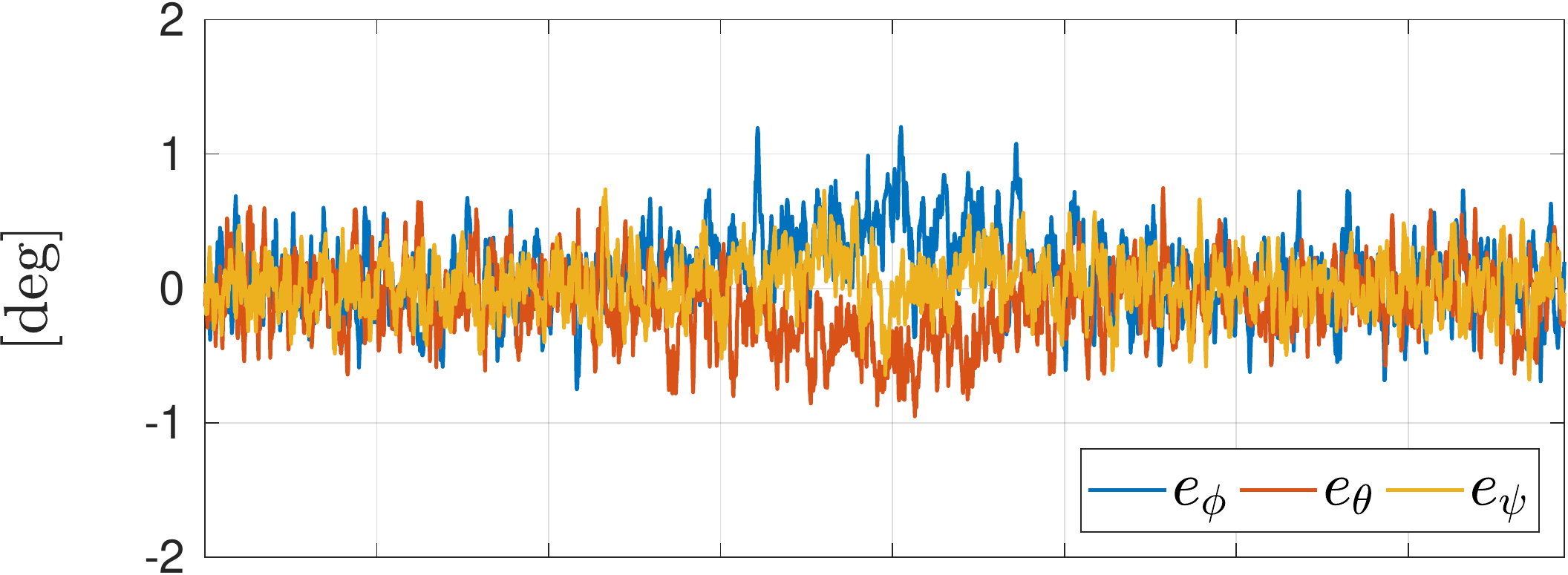}\\
		\includegraphics[width=\figWidth\columnwidth]{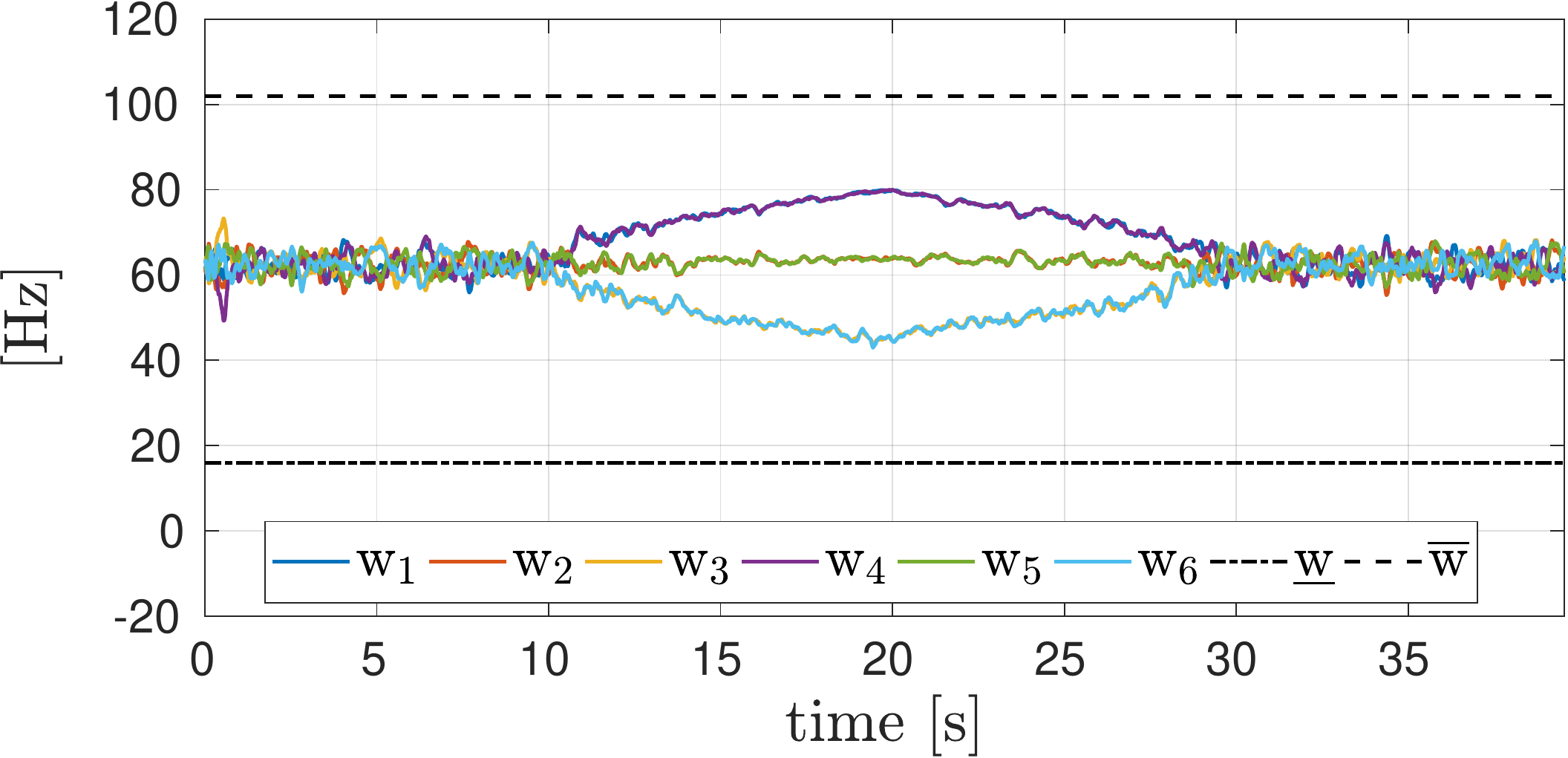}
	\caption{Plots of the FAST-Hex (with variable $\alpha$ regulated from the MPC algorithm) while hovering. The robot is disturbed with an external lateral force with a triangular profile. From top to bottom, the position, linear velocity, orientation and angular velocity tracking, the position and orientation errors, and the actuator spinning velocities.}
	\label{fig:plots_FASTHex_hov_traj}
\end{figure}

\begin{figure}[t]
	\centering
		\includegraphics[width=\figWidth\columnwidth]{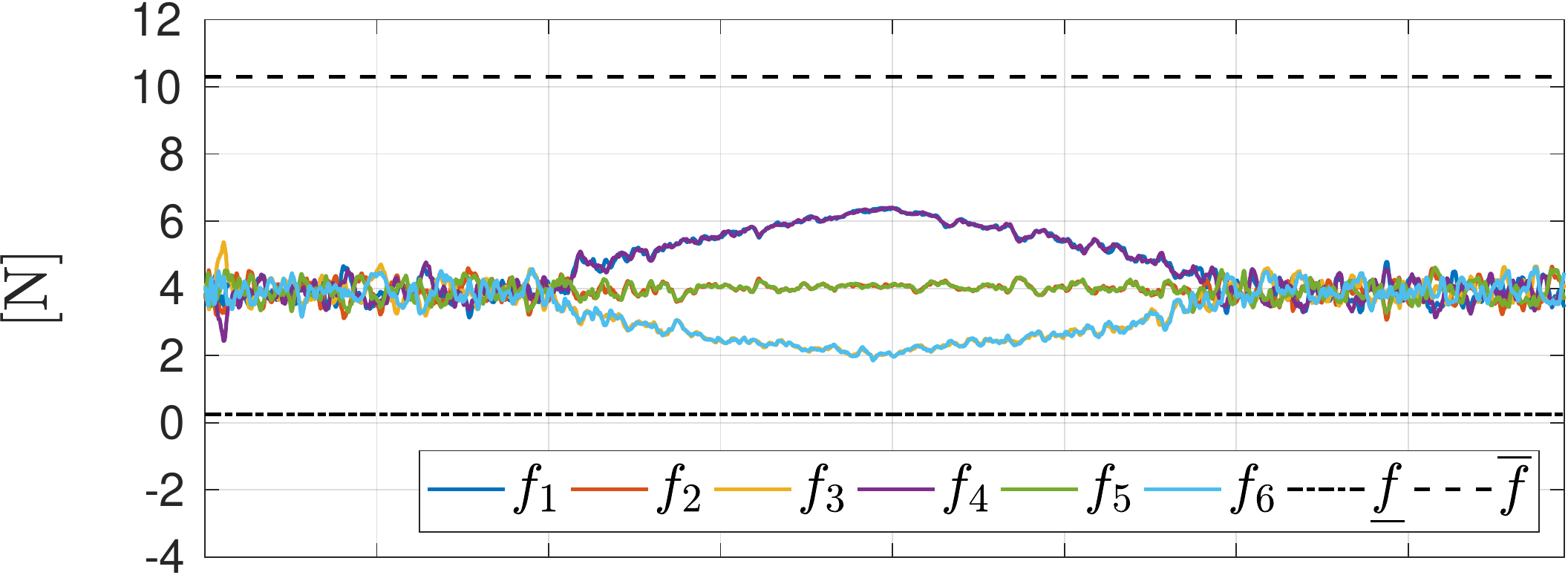}\\
		\includegraphics[width=\figWidth\columnwidth]{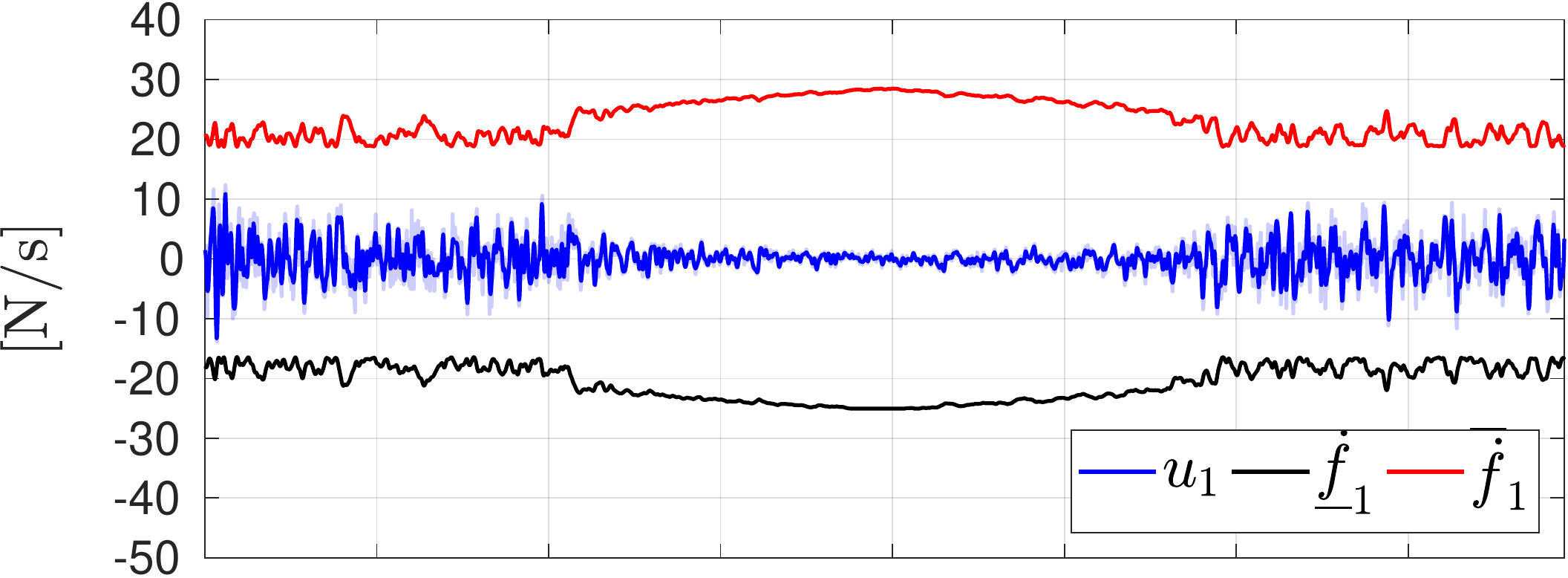}\\
		\includegraphics[width=\figWidth\columnwidth]{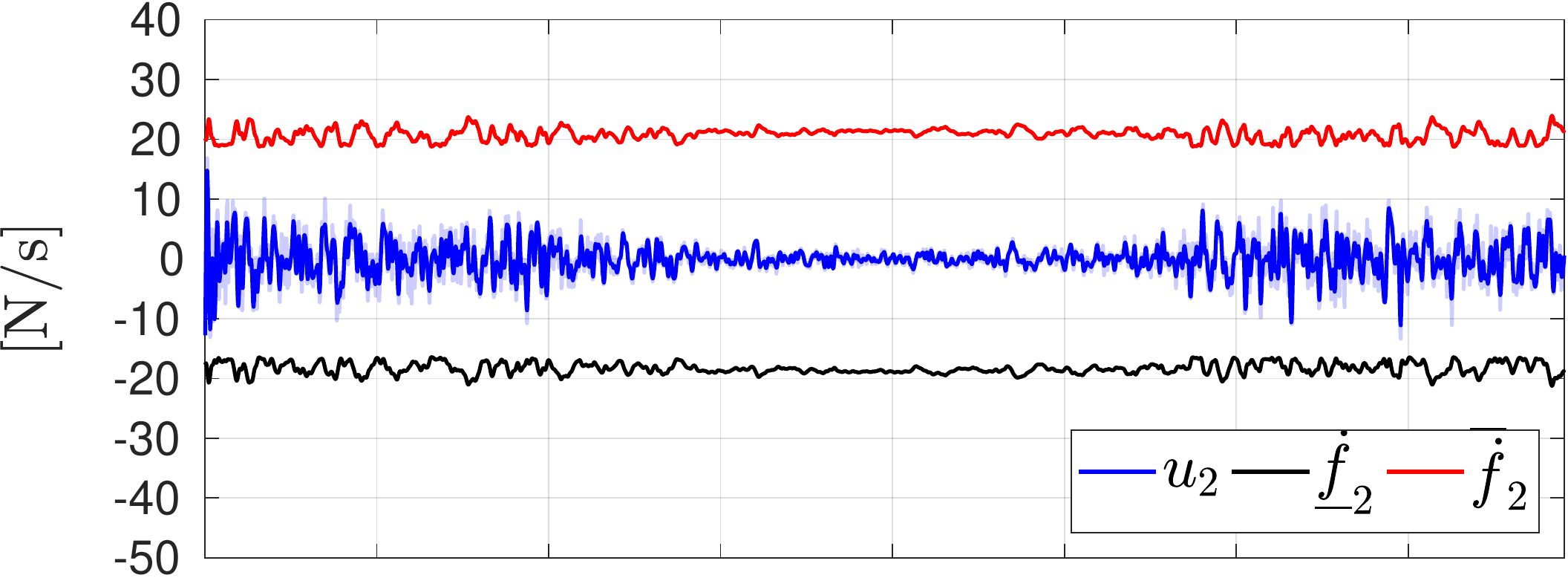}\\
		\includegraphics[width=\figWidth\columnwidth]{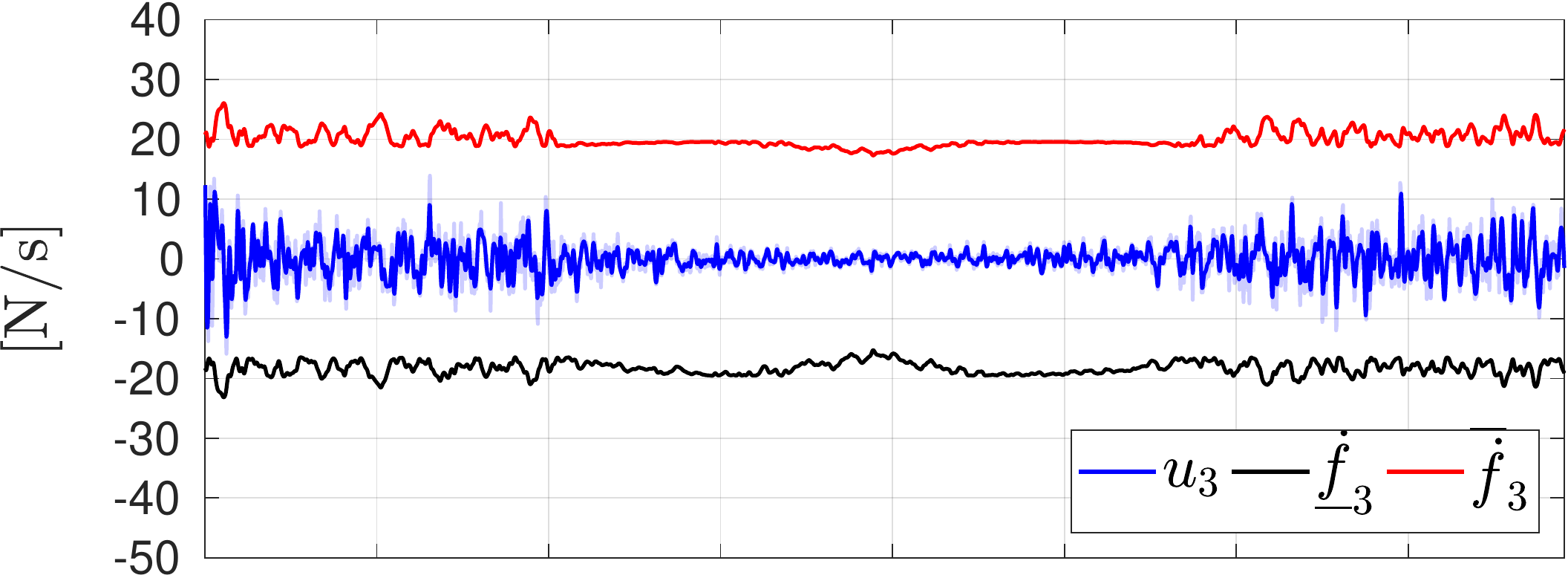}\\
		\includegraphics[width=\figWidth\columnwidth]{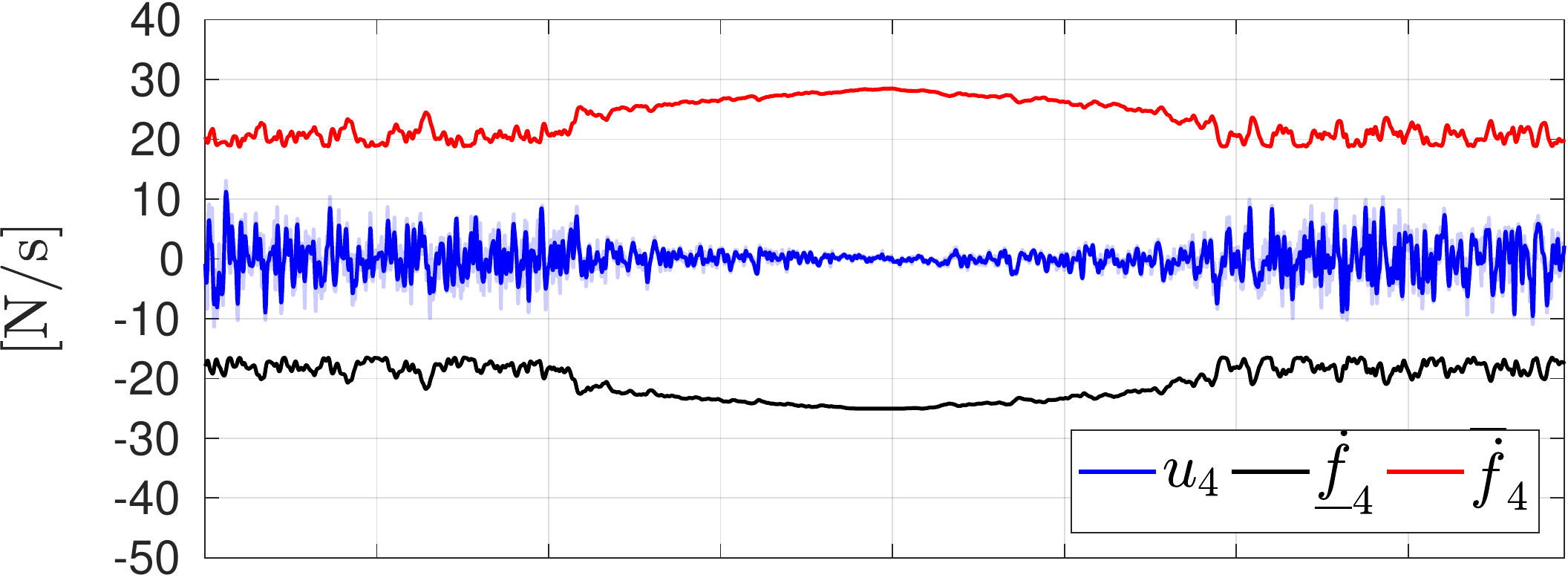}\\
		\includegraphics[width=\figWidth\columnwidth]{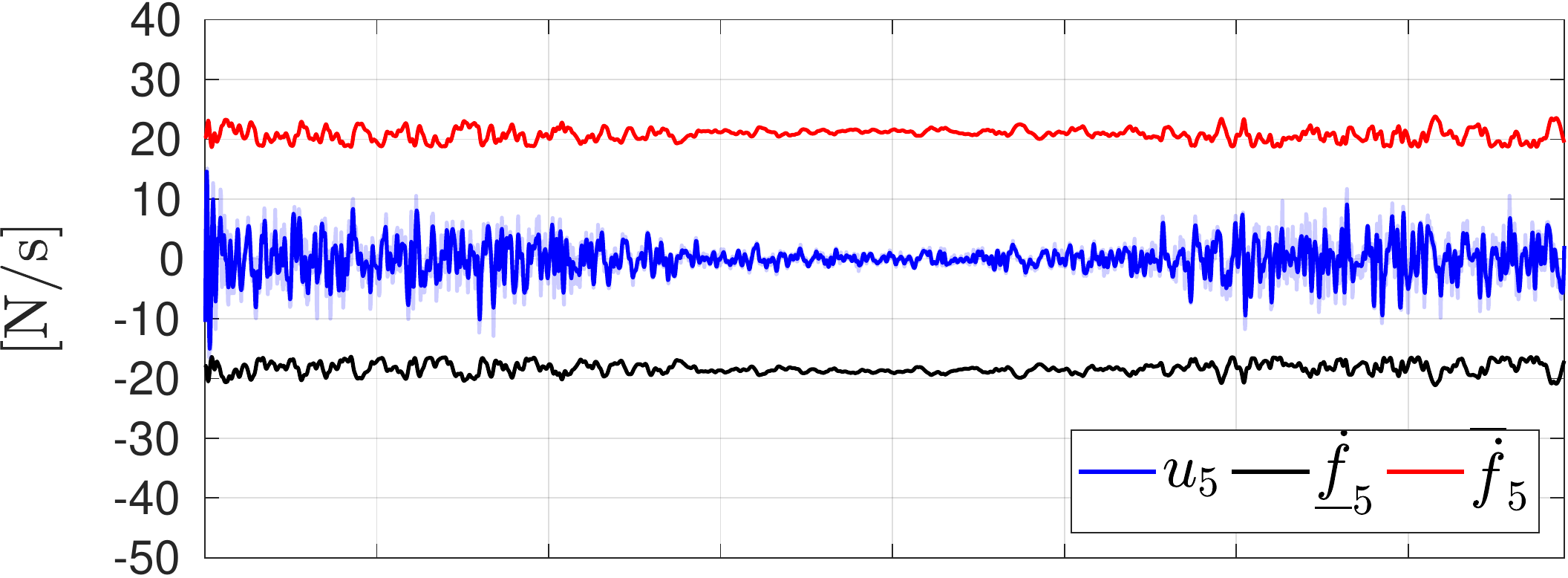}\\
		\includegraphics[width=\figWidth\columnwidth]{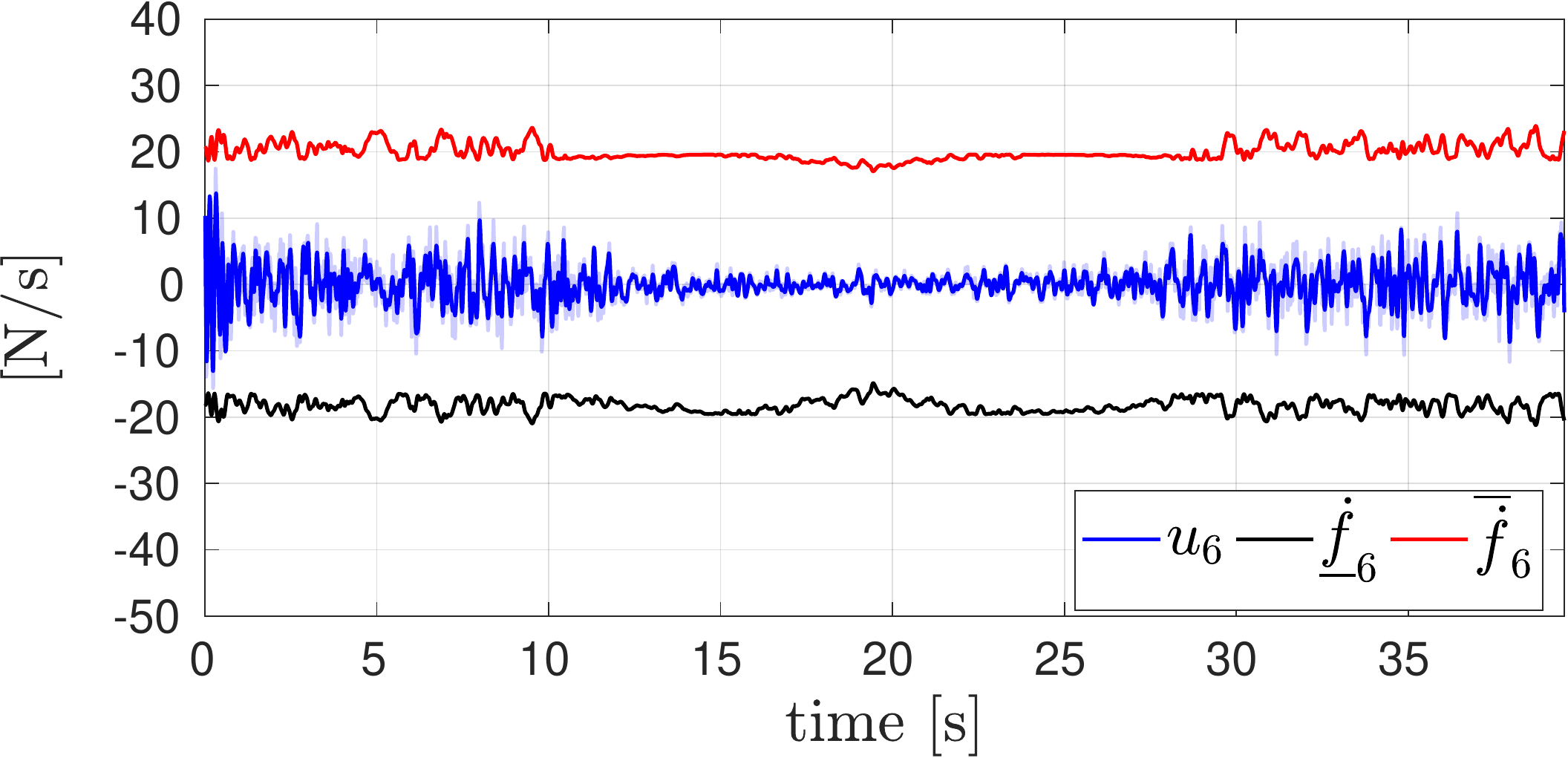}
	\caption{Plots of the FAST-Hex (with variable $\alpha$ regulated from the MPC algorithm) while hovering. The robot is disturbed with an external lateral force with a triangular profile. From top to bottom, the actuator forces and their derivatives. In particular, all the signals remain inside the feasible region delimited by the constraints.}
	\label{fig:plots_FASTHex_hov_limits}
\end{figure}

As long as the disturbance is not active, the NMPC algorithm should try to maintain $\alpha$ small, ideally equal to zero. This claim is motivated by the fact that this trajectory does not need the MDT capability in order to be tracked. On the other hand, as soon as the lateral force is activated, the platform can react to it either tilting its actuators or re-orienting its chassis. In this choice, the relative values of the cost function weights play a fundamental role. Intuitively, if the energy cost is weighted consistently (w.r.t. the tracking error terms on the states), the control algorithm should try to produce an input with low energy consumption, giving less priority to the trajectory tracking. In particular, the task of maintaining a flat orientation should be somehow discouraged by the controller, since the generation of a lateral force in this configuration would require a consistent increase of some of the actuator forces, thus raising up the energy consumption. Conversely, if the weight related to the energy cost is small, the controller would always privilege the trajectory tracking, acting on input $\alpha$.

In the first simulation, related to the case in which $\alpha$ is actively regulated, we try to achieve a good \emph{trade-off} between the two tasks. 
In the other simulations, corresponding to different fixed configurations for $\alpha$, all the parameters are left untouched, in order to fairly compare the resulting performance, in terms of the overall cost function, w.r.t. the variable case.

\begin{figure}[t]
	\centering
		\includegraphics[width=\figWidth\columnwidth]{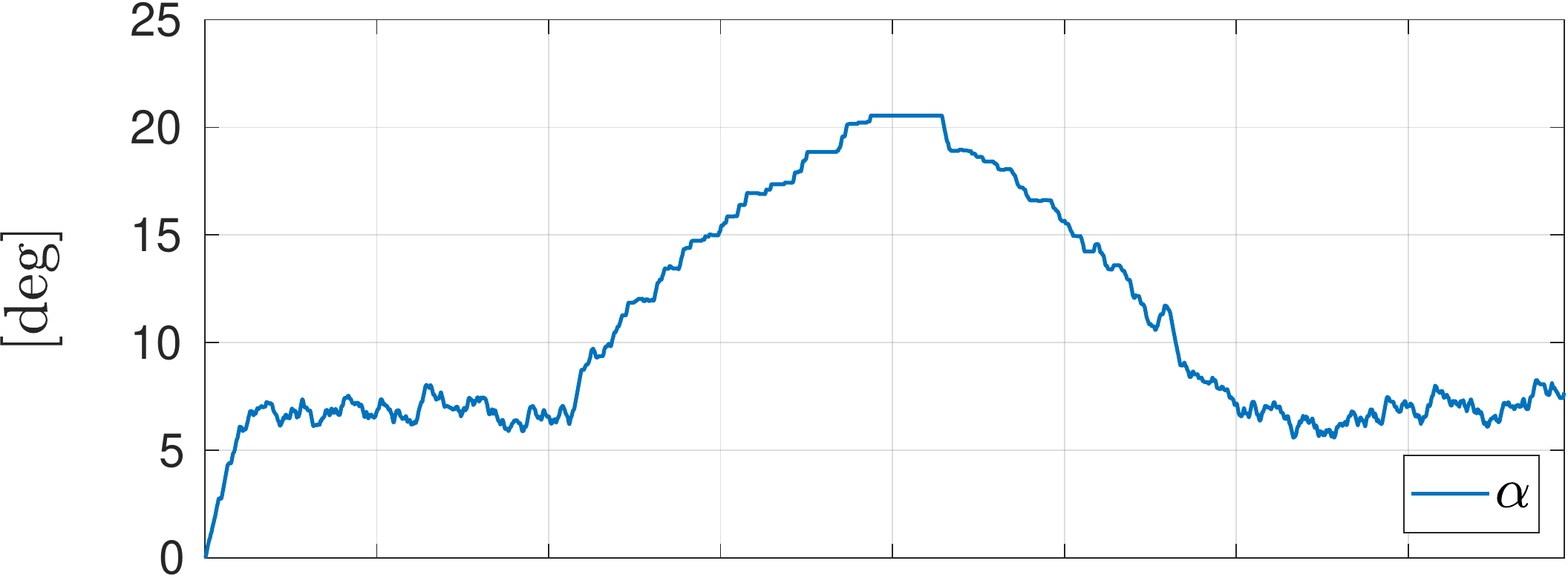} \\
		\includegraphics[width=\figWidth\columnwidth]{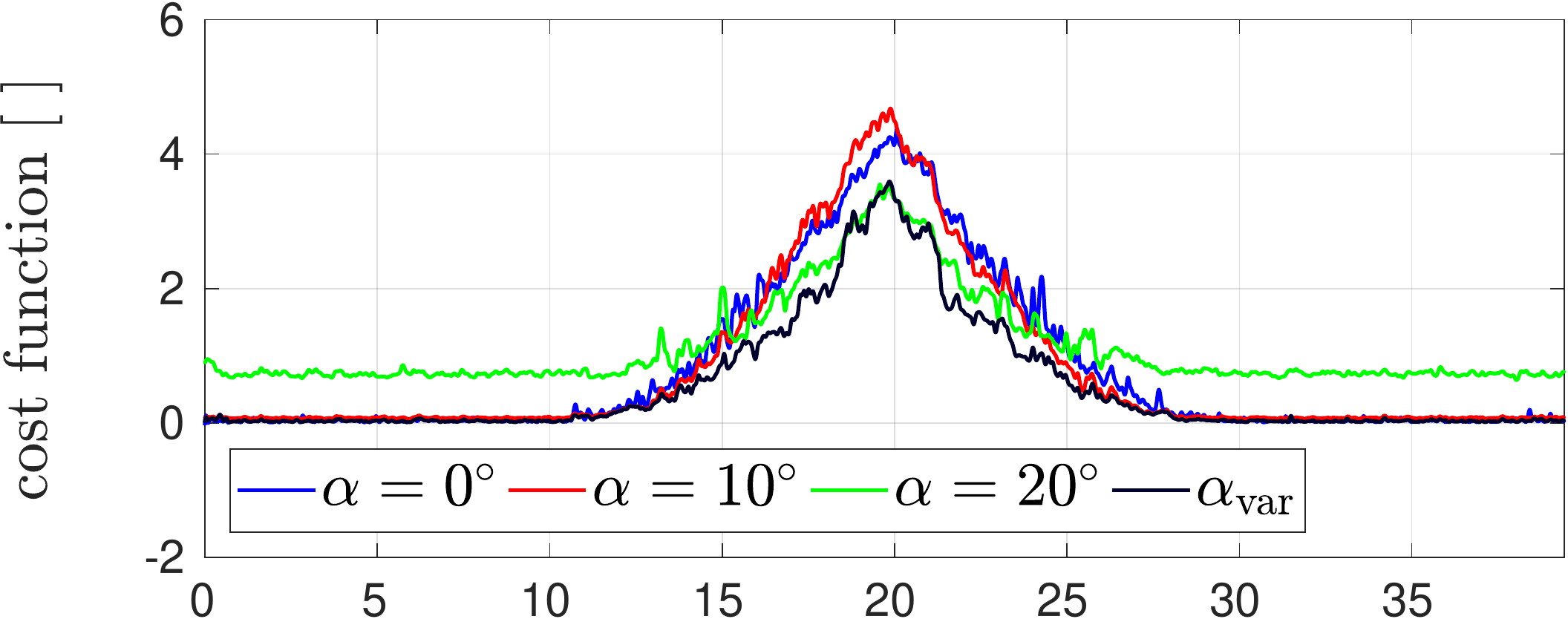} \\
		\includegraphics[width=\figWidth\columnwidth]{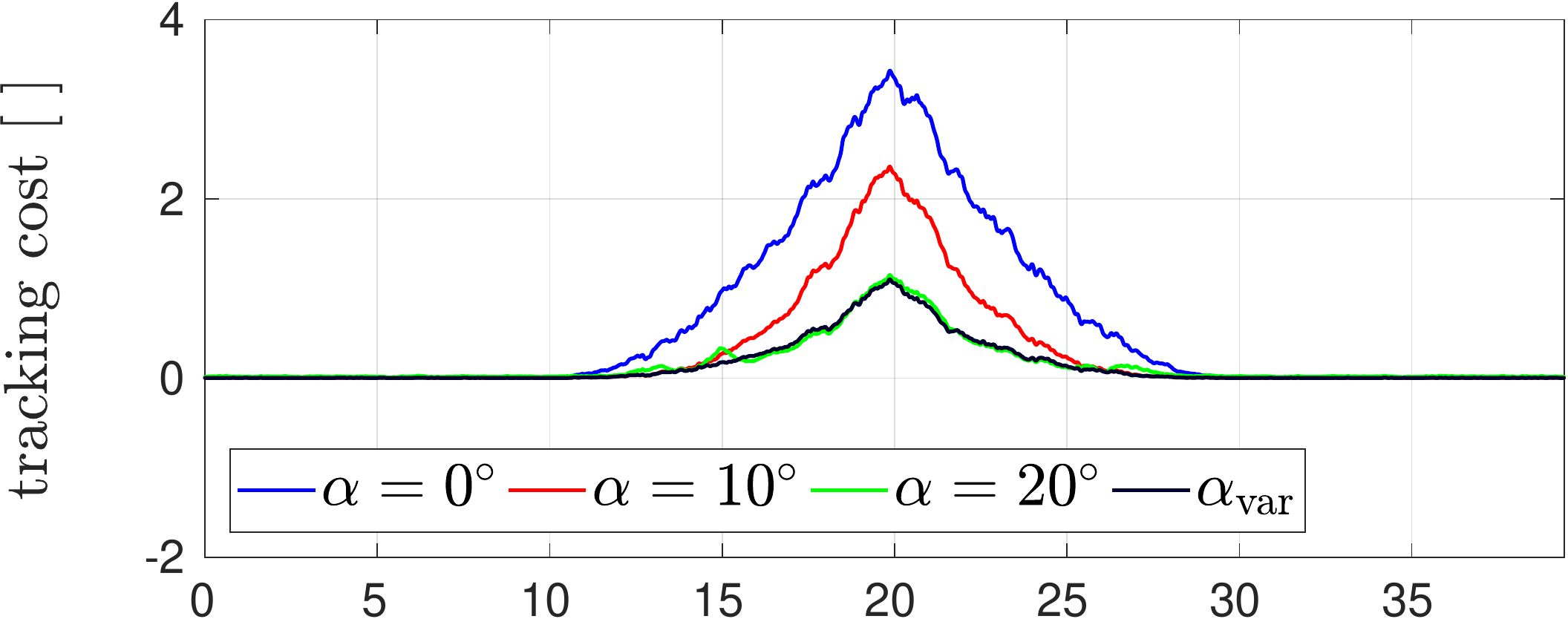} \\
		\includegraphics[width=\figWidth\columnwidth]{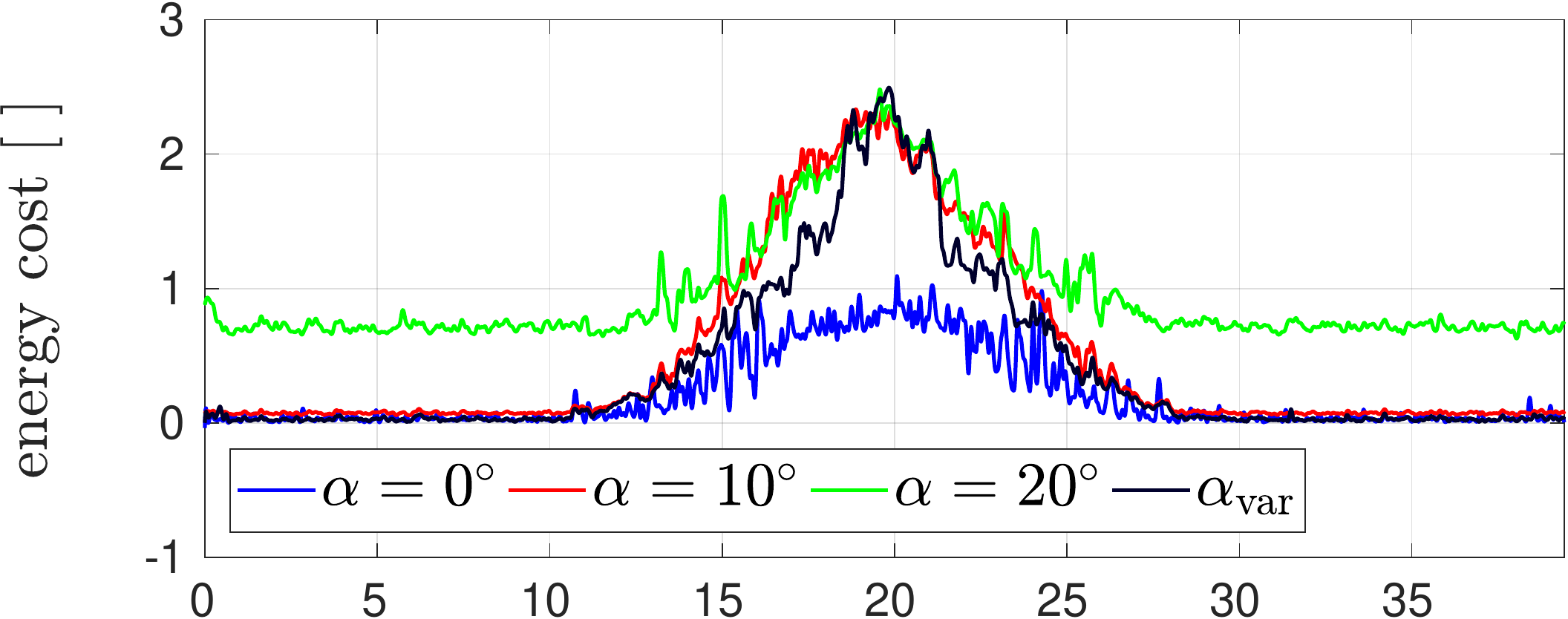} \\
		\includegraphics[width=\figWidth\columnwidth]{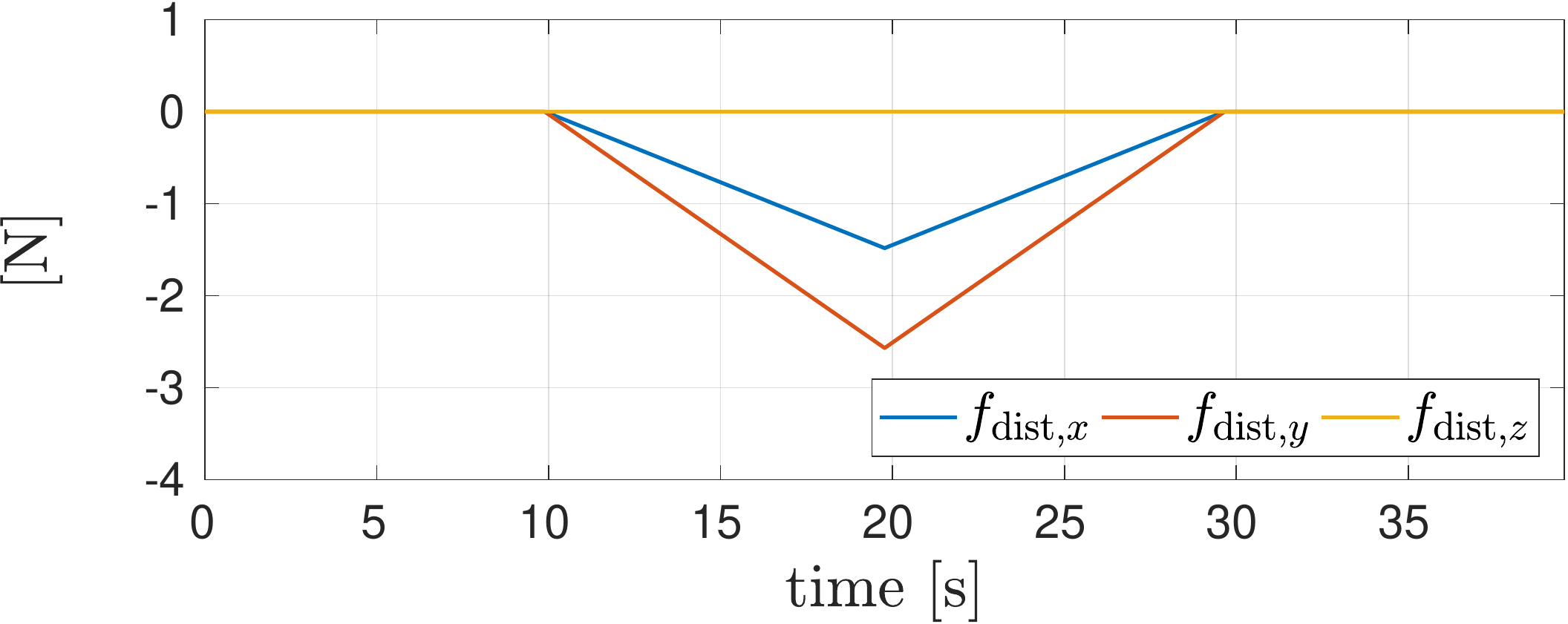}
	\caption{Plots of the FAST-Hex while hovering. In the first two plots, the evolution of $\alpha$ in the variable case and the comparison of the total cost function for different cases of constant $\alpha$ and the variable $\alpha$ (regulated from the MPC algorithm). Then, the comparison of the partial costs related to the tracking and the energy terms. Finally, the profile of the external force disturbance (in World Frame).}
	\label{fig:plots_FASTHex_costs}
\end{figure}

The plots related to the trajectory tracking in the variable case are depicted in
Fig.~\ref{fig:plots_FASTHex_hov_traj}. The first four plots, exhibiting the trajectory tracking of the state components, outline the good performance of the controller.
Indeed, the measured linear velocity, orientation, and angular velocity tracking keep very close to their reference profile, which are constantly equal to zero on all components. On the other hand, the measured position visibly
deviates from the reference one when the disturbance is acting on the robot.
Nevertheless, the position error keeps bounded, with a peak of less than $9$\,\si{cm} on its second component, which corresponds to the direction mostly affected by the lateral force, as shown in the last plot of Fig~\ref{fig:plots_FASTHex_costs}. This error could be considerably reduced by increasing the relative weights inside the NMPC algorithm cost function: this is confirmed by the sixth plot of Fig.~\ref{fig:plots_FASTHex_hov_traj}, where the MRAV maintains the orientation error below $1$\,\si{\deg}. We were able to achieve such result by properly weighting the attitude term in relation to the others, in particular in relation to the energy cost.
Moreover, the last plot of the same figure shows that the external disturbance can be counteracted without an excessive effort of the rotors, since their spinning velocities (and so the generated forces) safely remain with the bounds.
The plots related to the force derivatives, which are presented in Fig.~\ref{fig:plots_FASTHex_hov_limits}, confirm that a static trajectory, combined with a slowly-varying disturbance, does not produce large values for the inputs.

Consider the first plot of Fig.~\ref{fig:plots_FASTHex_costs}, which depicts the trajectory of $\alpha$. During the middle phase, the tilting angle is increased up to $\approx 21$\,\si{deg} in order to counteract the lateral force and to keep the platform flat at the same time. On the other hand, the reason why $\alpha$ is regulated to a constant value of $\approx 7$\,\si{deg} and not exactly to zero, is due to the noise introduced in the simulation, in particular to the one related to the translational part of the state $[p_x\ p_y\ \dot{p}_x\ \dot{p}_y]^{\top}$. 
Indeed, the control algorithm is informed about a non-zero error in these components, and continuously tries to annihilate it by selecting a small tilting angle, in order to be able to exert a lateral force and stay horizontal at the same time. 

In order to demonstrate the benefit of the active regulation of the tilting angle, we additionally performed other three simulations (with the same parameters) imposing $\alpha = 0,10,20\ $\,\si{deg}, respectively.
The comparison of the overall NMPC cost functions for the different fixed cases and the variable one is displayed in the second plot of Fig.~\ref{fig:plots_FASTHex_costs}. As it can be appreciated, the regulated case, denoted with $\alpha_{\text{var}}$, gives the best trade-off between tracking performance and consumed energy.
In the unperturbed hovering phases (lateral parts of the plots), $\alpha$ is regulated to a small value in order to avoid unnecessary energy waste, while in the middle phase, when the disturbance force is activated, $\alpha$ is increased in order to improve the trajectory tracking, in particular the one related to the orientation.
The third and the fourth plots of the same figure outline the partial costs related to the tracking errors and the energy cost. Among the fixed configurations, the one with the largest tilting angle, i.e. $\alpha=20$\,\si{deg}, generates the smallest tracking cost along all the simulation. This confirms that the MDT capability drastically improves the MRAV tracking performance. However, it unavoidably causes a larger energy cost as the angle takes larger values.
This is why the additional degree of freedom on $\alpha$ might be very convenient in many applications.

\subsection{Simulations of the Tilt-Hex with rotor failure}

The problem of the robustness of a MRAV in case of a rotor failure is not new in the literature. Indeed, the analysis and the design of a tilted-rotor hexarotor for fault tolerance has been considered in~\cite{giribet2016analysis}, while formal definitions as well as the design of an analytic controller based on the identification of a direction in the force space, along which the intensity of the control force can be assigned independently from the torque, can be found in~\cite{2017f-MicRylFra,2018a-MicRylFra}.
Given the importance of such topic in the aerial robotics panorama, we decided to target this problem, showing that our NMPC algorithm can deal with this problem in a very efficient and general way.

\begin{table}[t]
	\renewcommand\arraystretch{1.3}
	\caption{Parameters used in the rotor-failed Tilt-Hex simulation.}
	\label{tab:Tilt_fail_sim}
	\centering
	\resizebox{\figWidth\columnwidth}{!}{%
	\begin{tabular}{CCC}
		\hline
		\text{Parameter}  & \text{Value} & \text{Unit} \\
		\hline
		\sigmav_{\pv} & [\sqrt{0.005}\ \sqrt{0.005}\ \sqrt{0.005}]^{\top} & \si{m} \\
		\sigmav_{\dot{\pv}} & [\sqrt{0.02}\ \sqrt{0.02}\ \sqrt{0.02}]^{\top} & \si{m/s} \\
		\sigmav_{\etav} & [\sqrt{1}\ \sqrt{1}\ \sqrt{1}]^{\top} & \si{deg} \\
		\sigmav_{\omegav} & [\sqrt{0.15}\ \sqrt{0.15}\ \sqrt{0.05}]^{\top} & \si{deg/s} \\
		\Omega_{\text{filt}} & {25} & \si{rad/s} \\
		\hline
		t_1 & 5 & \si{s} \\
		\tauv_{\text{dist}} & \frac{1}{250}\ [0.68\ 0.39\ 0.62]^{\top} & \si{Nm} \\
		\hline
		\Qm_{\pv}(j,j)|_{j=1,2,3} & 10,10,10 & [\ ] \\
		\Qm_{\dot{\pv}}(j,j)|_{j=1,2,3} & 0.5,0.5,0.5 & [\ ] \\
		\Qm_{\etav}(j,j)|_{j=1,2,3} & 1.5,1.5,1.5 & [\ ] \\
		\Qm_{\omegav}(j,j)|_{j=1,2,3} & 0.0005,0.0005,0.0005 & [\ ] \\
		\Qm_{\ddot{\pv}}(j,j)|_{j=1,2,3} & 0,0,0 & [\ ] \\
		\Qm_{\dot{\omegav}}(j,j)|_{j=1,2,3} & 0,0,0 & [\ ] \\
		\Rm_h(j,j)|_{j=1,\dots{},5} & 0,0,0,0,0 & [\ ] \\
		\hline
	\end{tabular}  } 
\end{table}

The failure of one rotor in the Tilt-Hex is modeled removing one state and one input, i.e., those related to the $6-th$ actuator. As a matter of fact, $\xv \in \mathbb{R}^{17}$ and $\uv \in \mathbb{R}^{5}$ in this case.
In the following, we present the hovering performance in two different configurations of the angle $\beta$, c.f. Fig.~\ref{fig:modeling}, in order to highlight the importance of such angle in relation to the fault tolerance capabilities, c.f.~\cite{2018a-MicRylFra}.
The parameters related to these simulations are reported in Tab.~\ref{tab:Tilt_fail_sim}.

As already pointed out in~\cite{2017f-MicRylFra}, given the particular arrangement of the Tilt-Hex actuators, which are symmetrically disposed in a star-configuration with alternated $\alpha$ and equal $\beta$ angles, it is convenient to switch off the actuator located in the mirrored position w.r.t. the broken one, when the failure is detected. In this case, this corresponds to the $3-rd$ one. This choice represents the best solution in order to balance the control effort, c.f.~\cite{2017f-MicRylFra}, Fig.~3.
In the following simulations, this behavior is emulated by setting $\underline{f}=0\si{N}$, i.e., letting the controller the possibility to completely switch off the actuators. 

\subsubsection{Hovering trajectory with unknown torque disturbance}

In this case, the reference trajectory is again a static hovering,
\begin{align*}
\pv_r &= [0\ 0\ 0.75]^{\top} ,\
\dot{\pv}_r = \ddot{\pv}_r = [0\ 0\ 0]^{\top} \\
\etav_r &= \omegav_r = \dot{\omegav}_r = [0\ 0\ 0]^{\top}
\end{align*}

\begin{figure}[t]
	\centering
		\includegraphics[width=\figWidth\columnwidth]{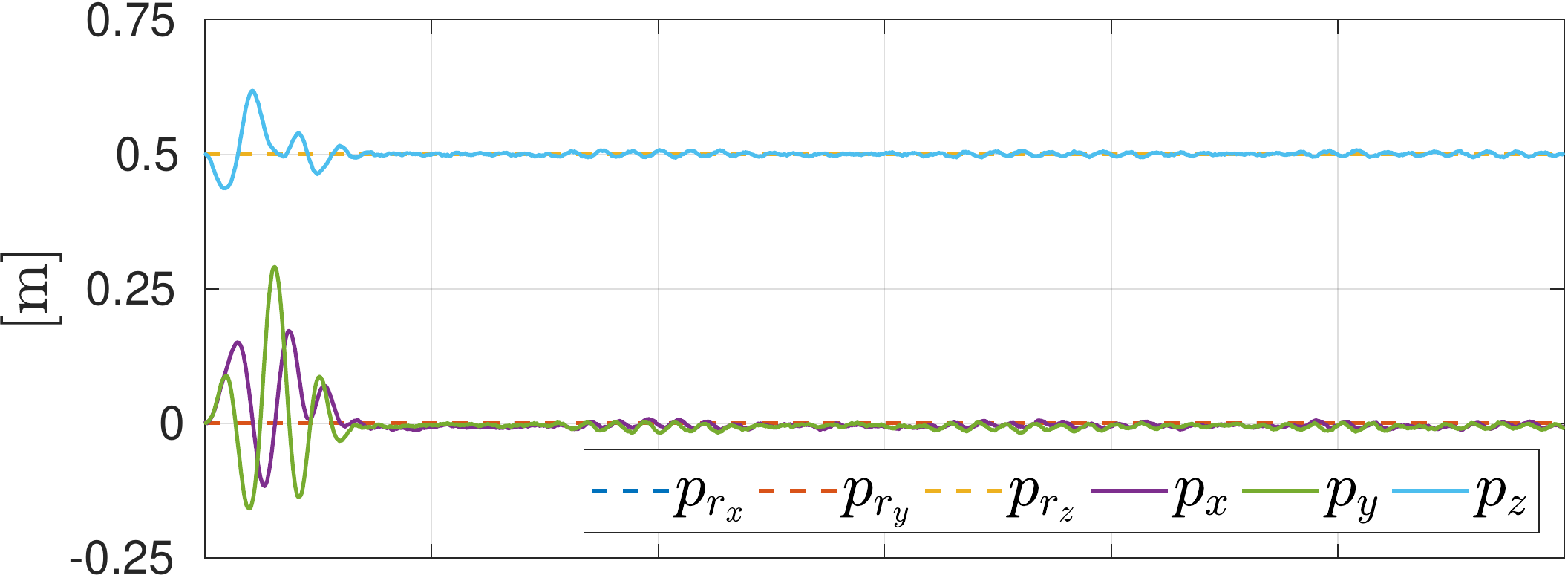}\\
		\includegraphics[width=\figWidth\columnwidth]{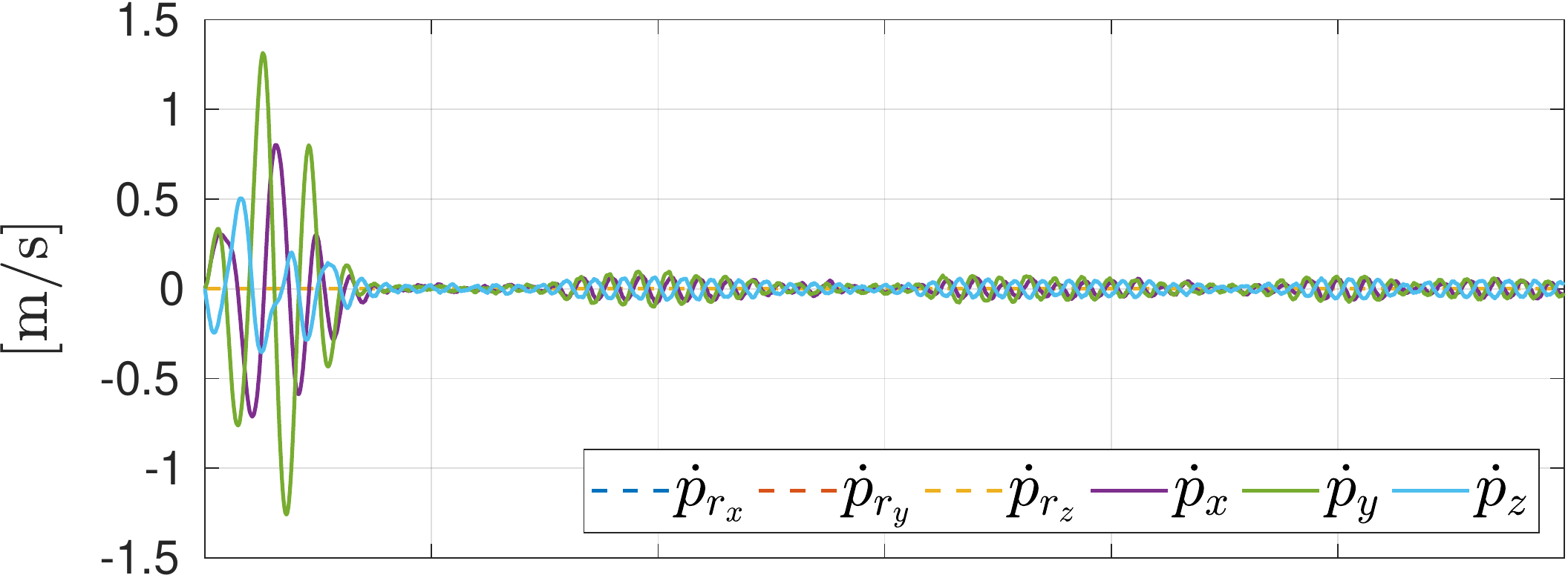}\\
		\includegraphics[width=\figWidth\columnwidth]{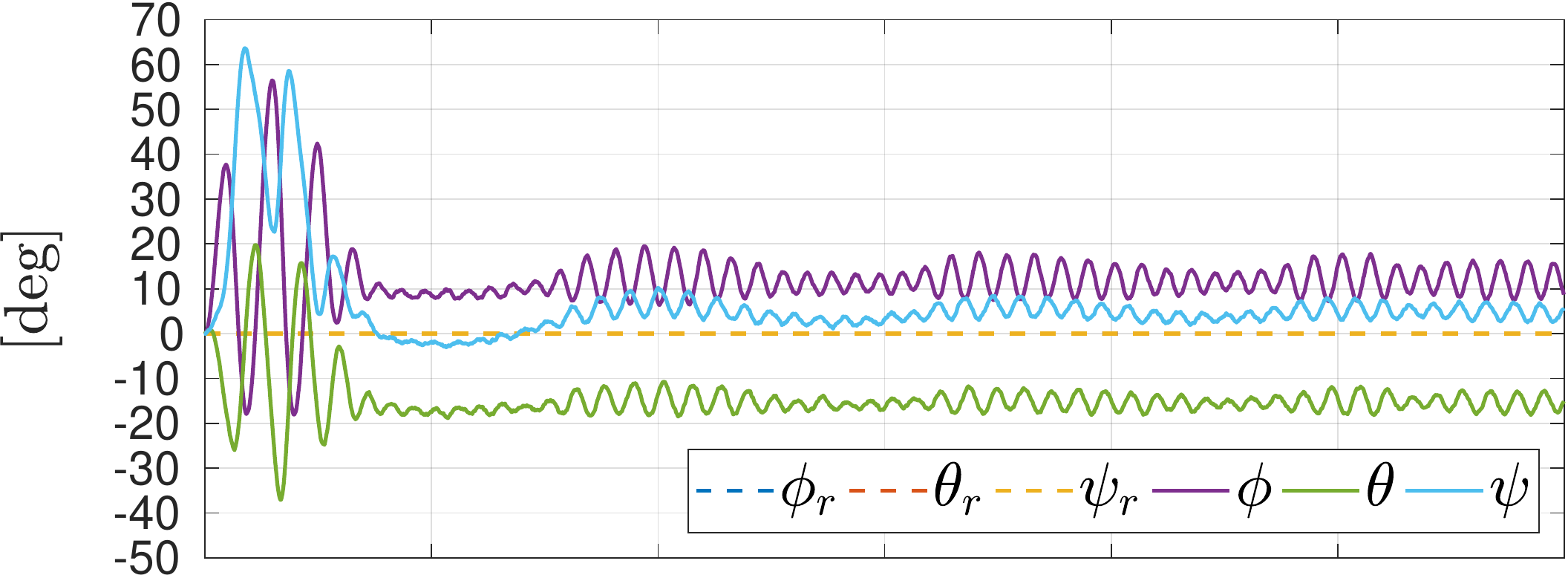}\\
		\includegraphics[width=\figWidth\columnwidth]{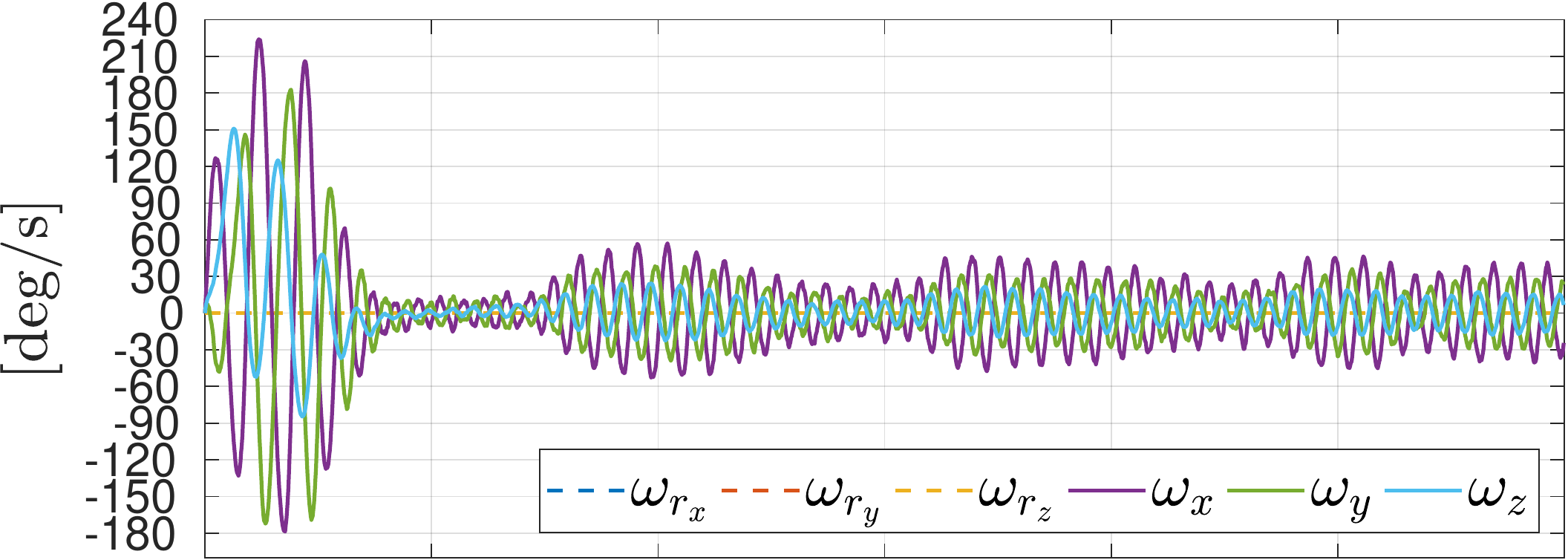}\\
		\includegraphics[width=\figWidth\columnwidth]{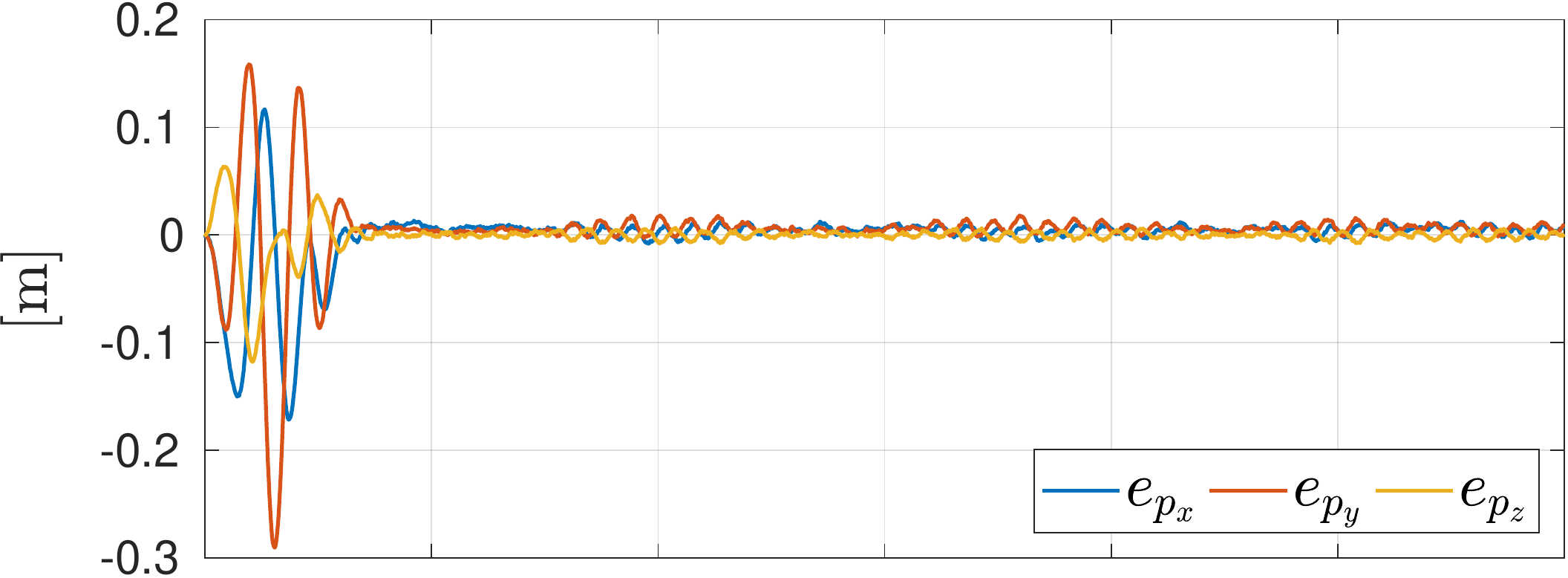}\\
		\includegraphics[width=\figWidth\columnwidth]{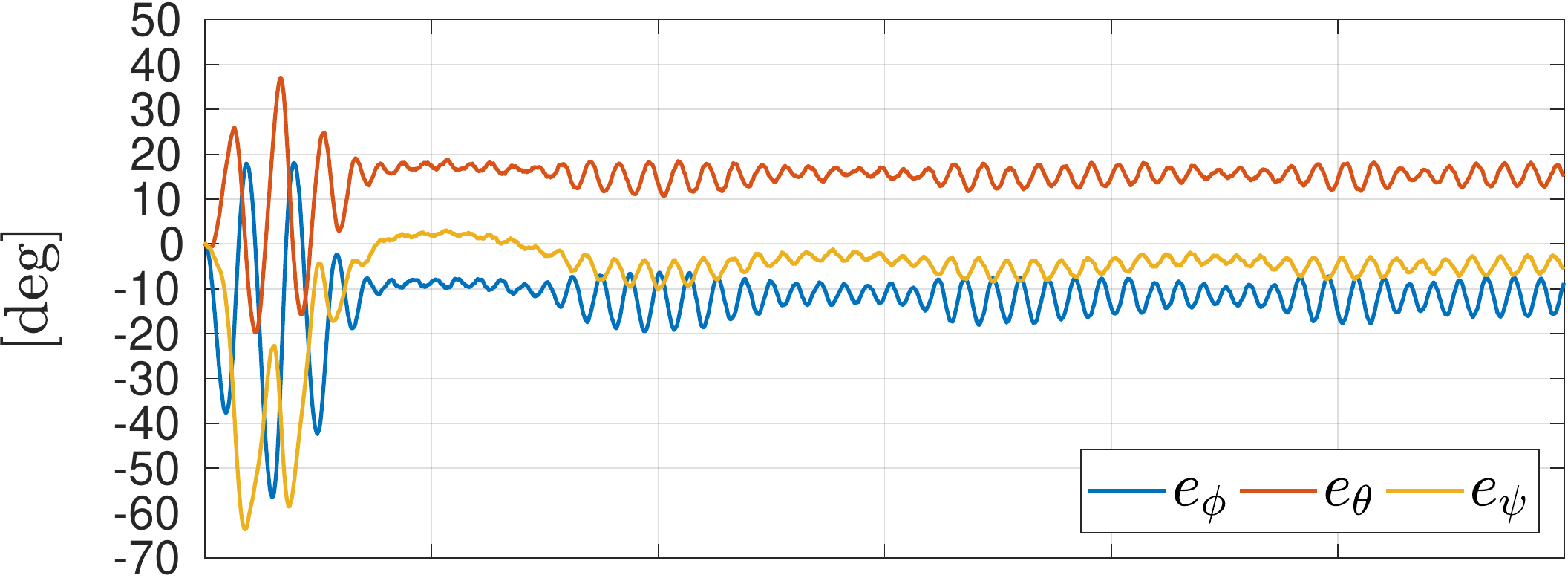}\\
		\includegraphics[width=\figWidth\columnwidth]{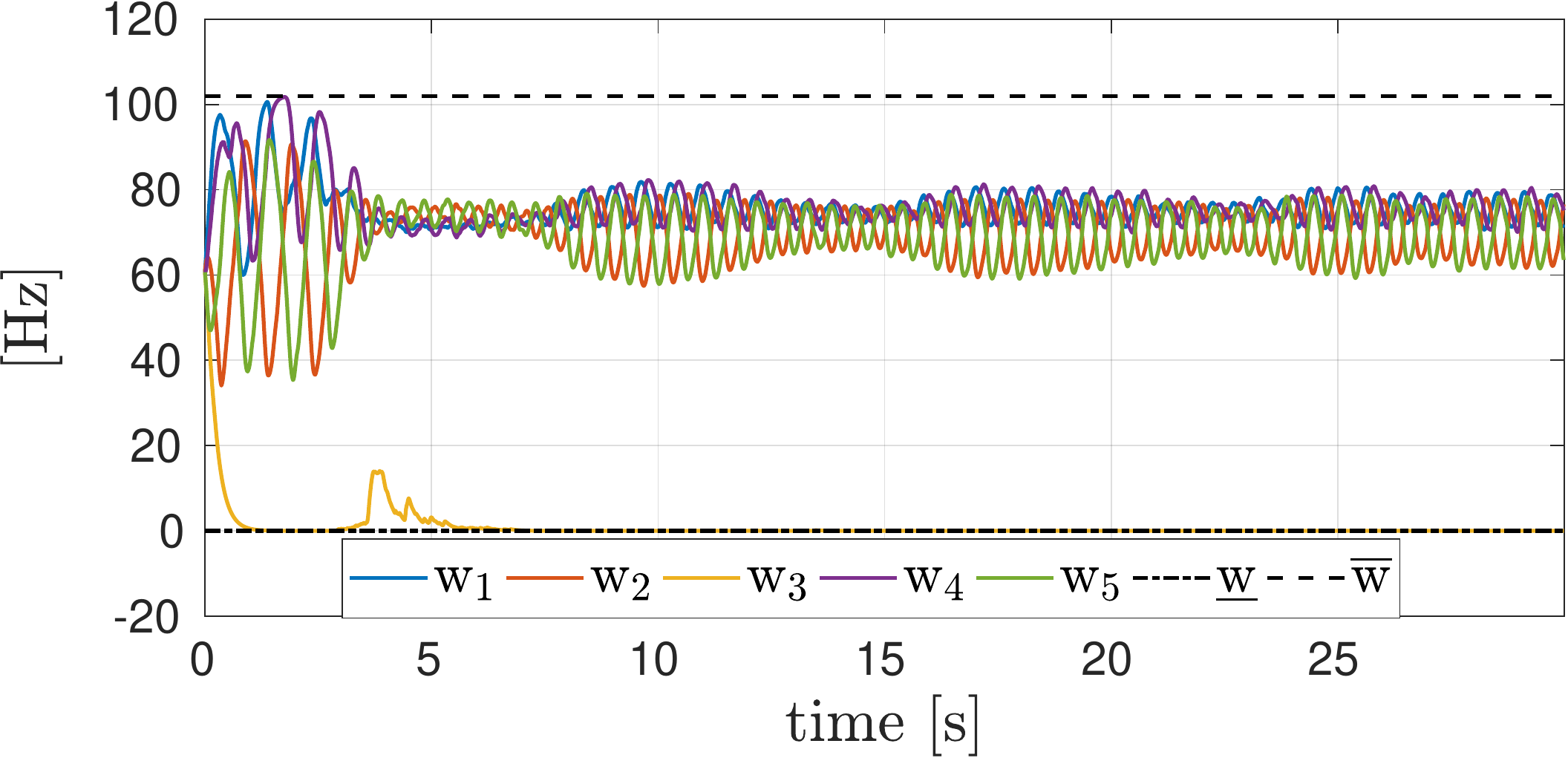}
	\caption{Plots of the Tilt-Hex with rotor failure and $\beta=0$\,\si{deg} while hovering. The robot is disturbed with a constant external torque. From top to bottom, the position, linear velocity, orientation and angular velocity tracking, the position and orientation errors, and the actuator spinning velocities.}
	\label{fig:plots_TiltHexFail1_hov_traj}
\end{figure}

\begin{figure}[t]
	\centering
		\includegraphics[width=\figWidth\columnwidth]{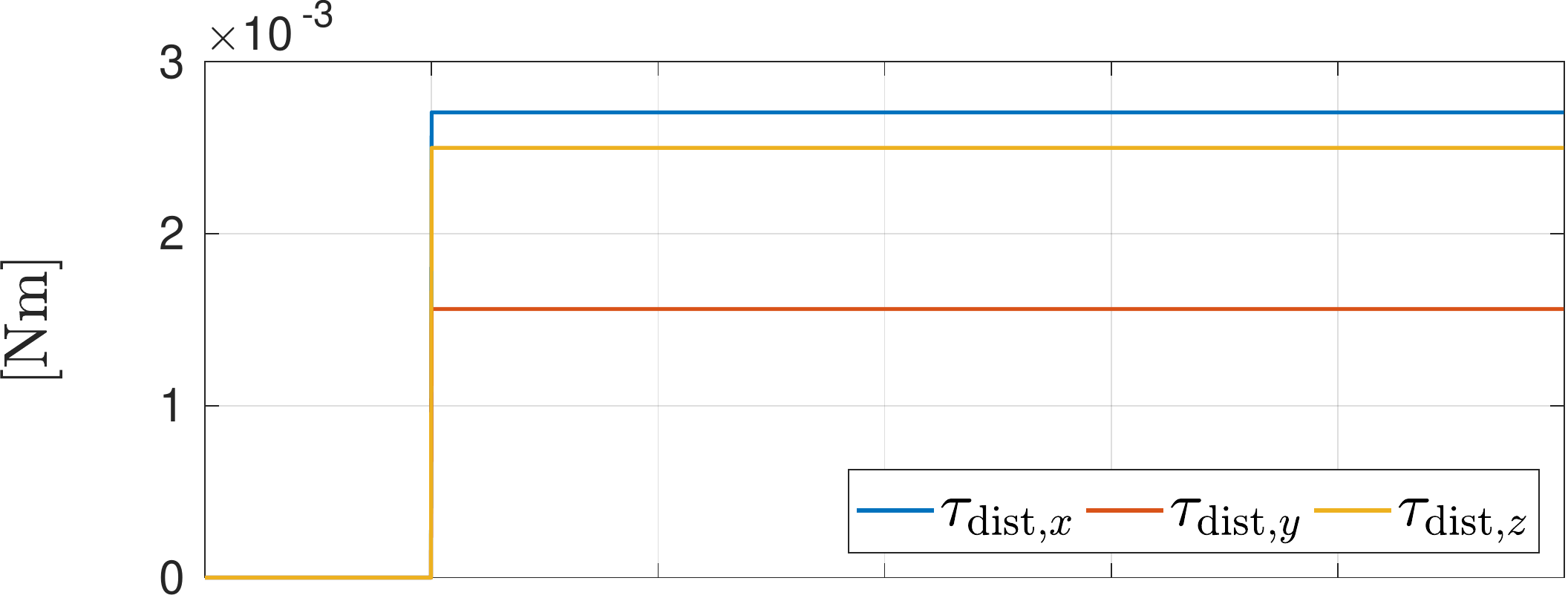}\\
		\includegraphics[width=\figWidth\columnwidth]{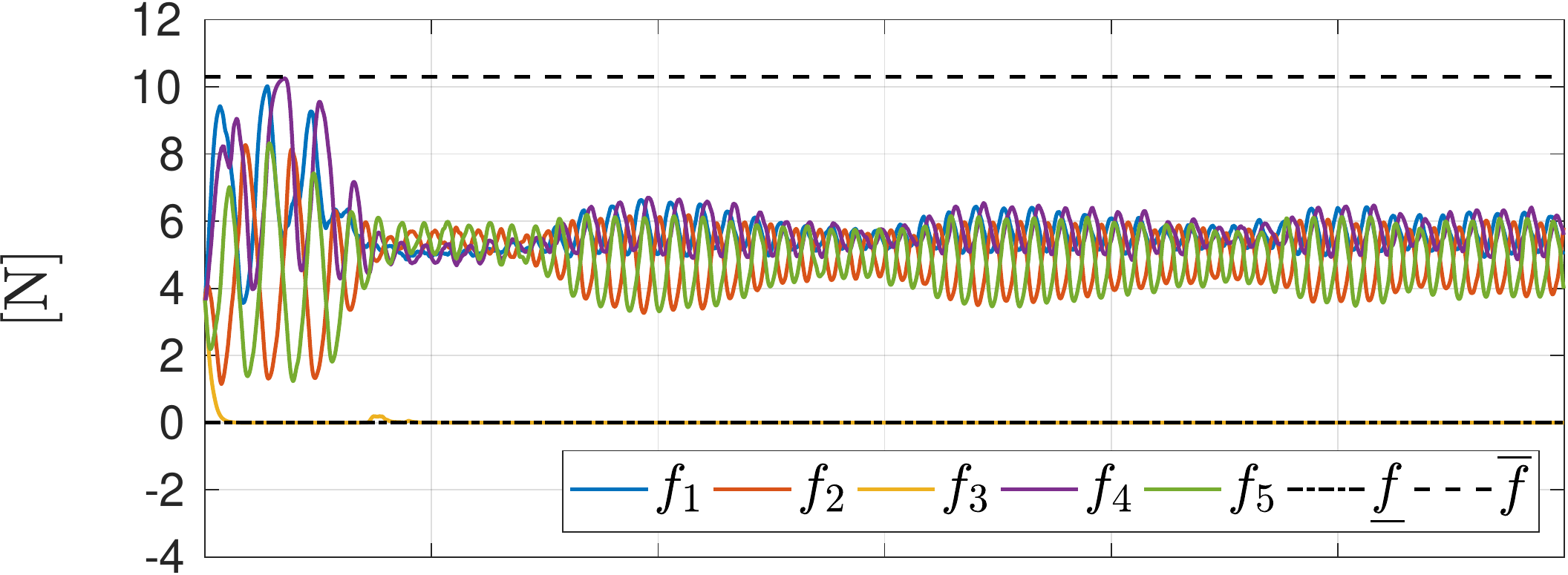}\\
		\includegraphics[width=\figWidth\columnwidth]{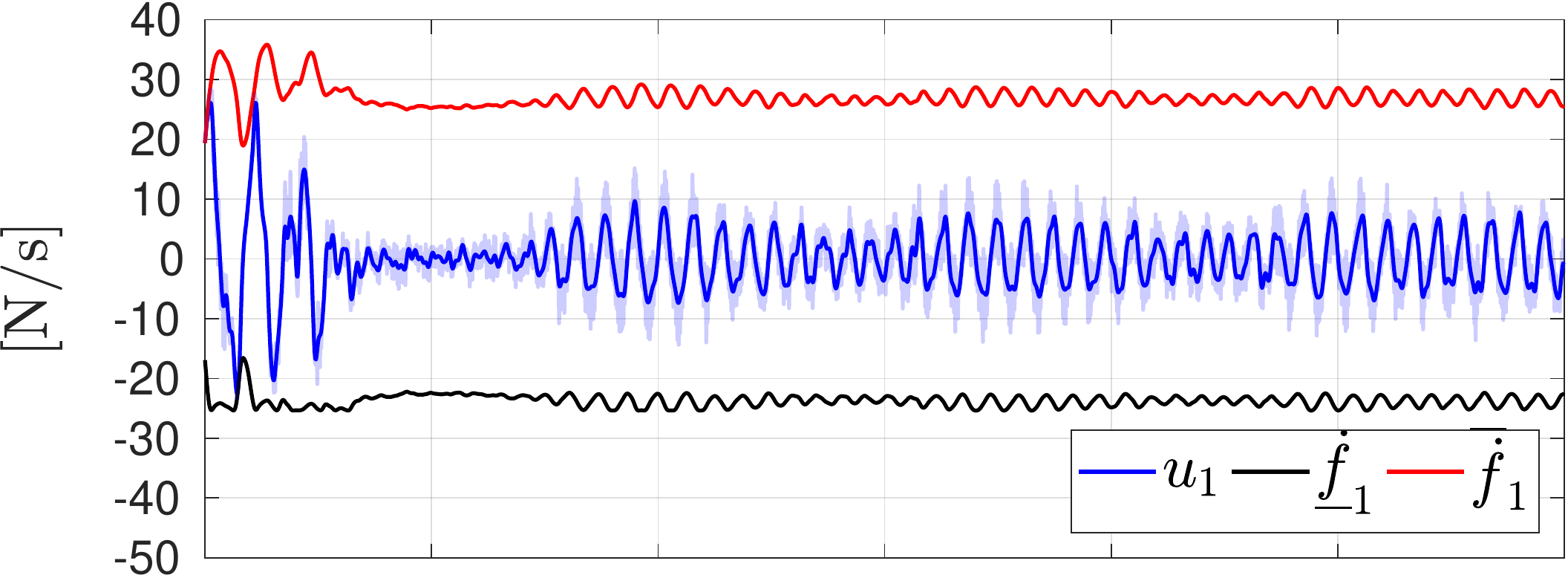}\\
		\includegraphics[width=\figWidth\columnwidth]{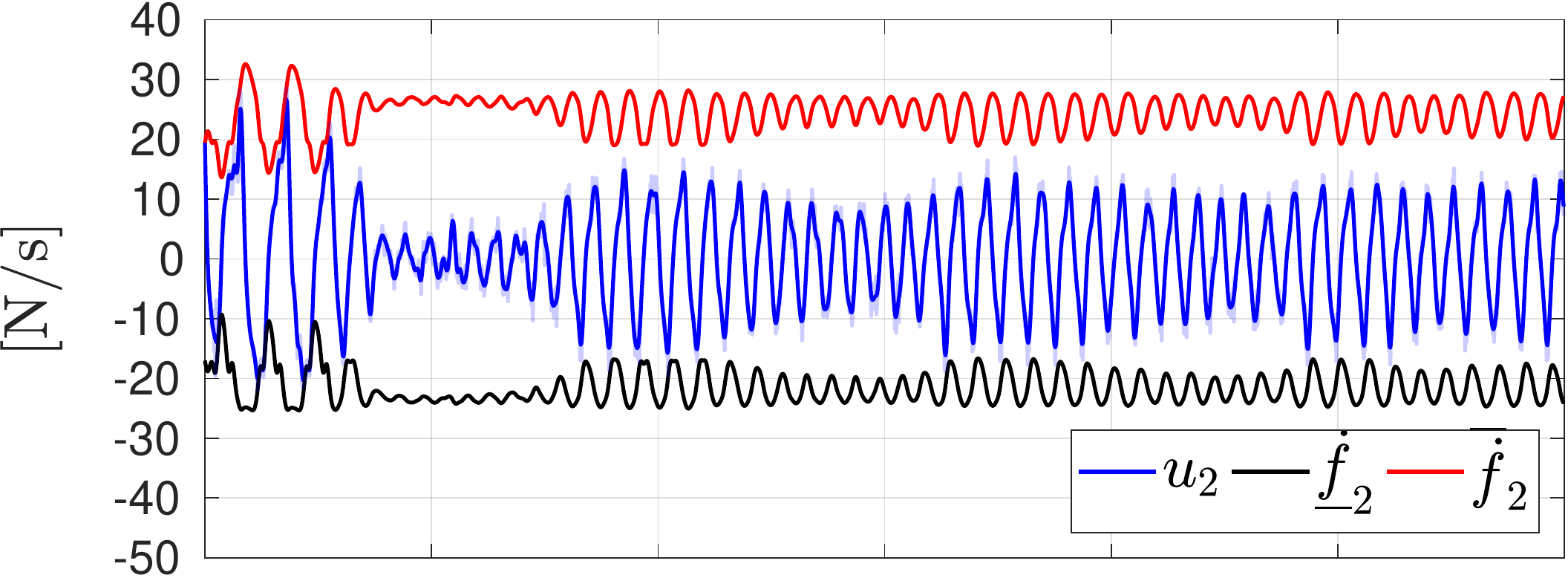}\\
		\includegraphics[width=\figWidth\columnwidth]{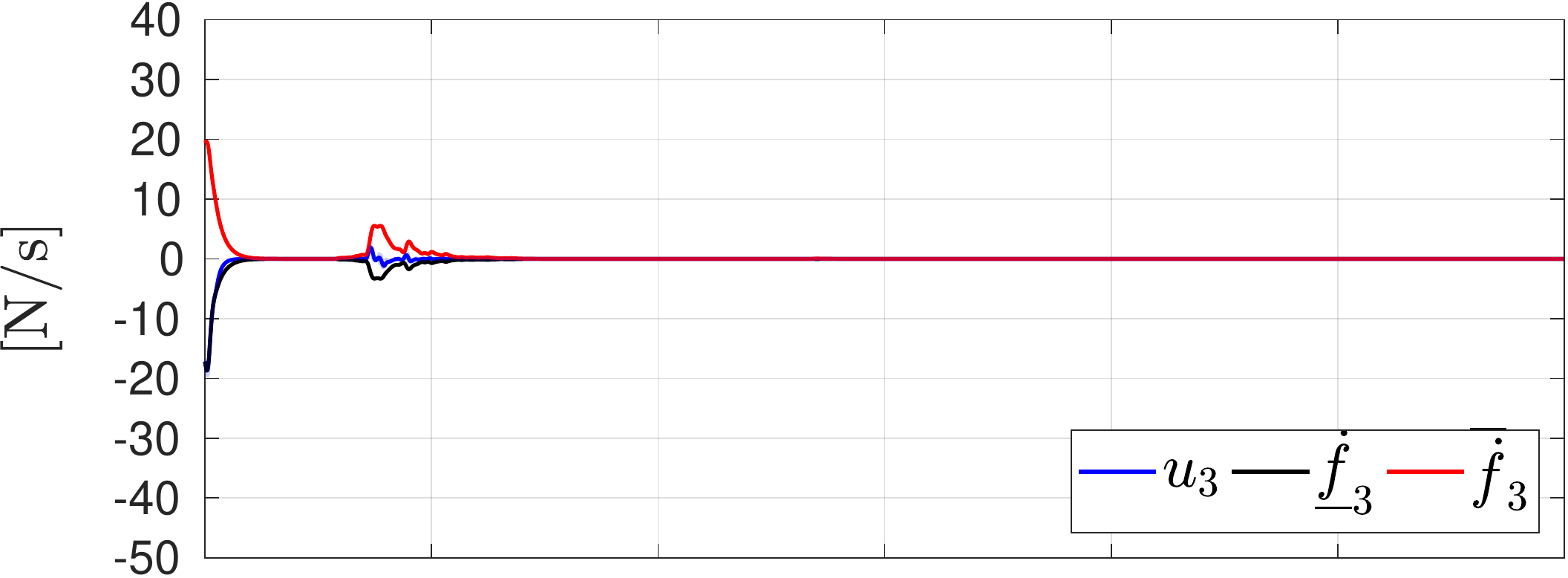}\\
		\includegraphics[width=\figWidth\columnwidth]{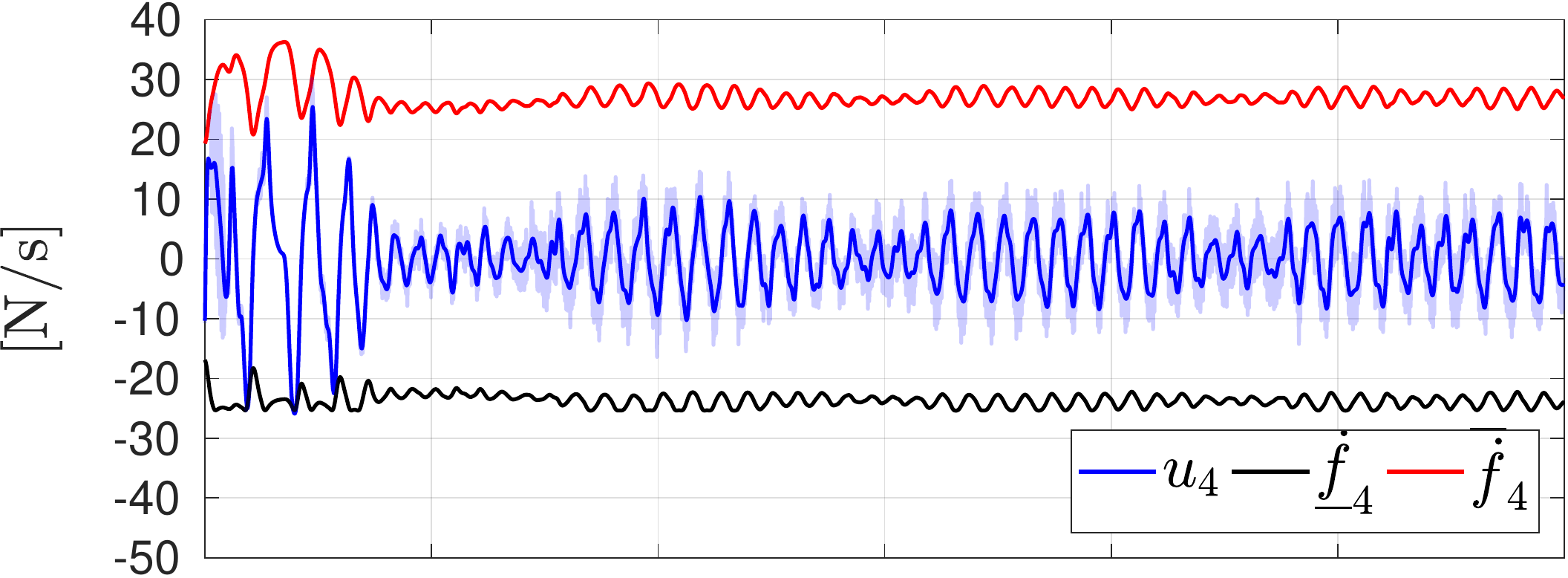}\\
		\includegraphics[width=\figWidth\columnwidth]{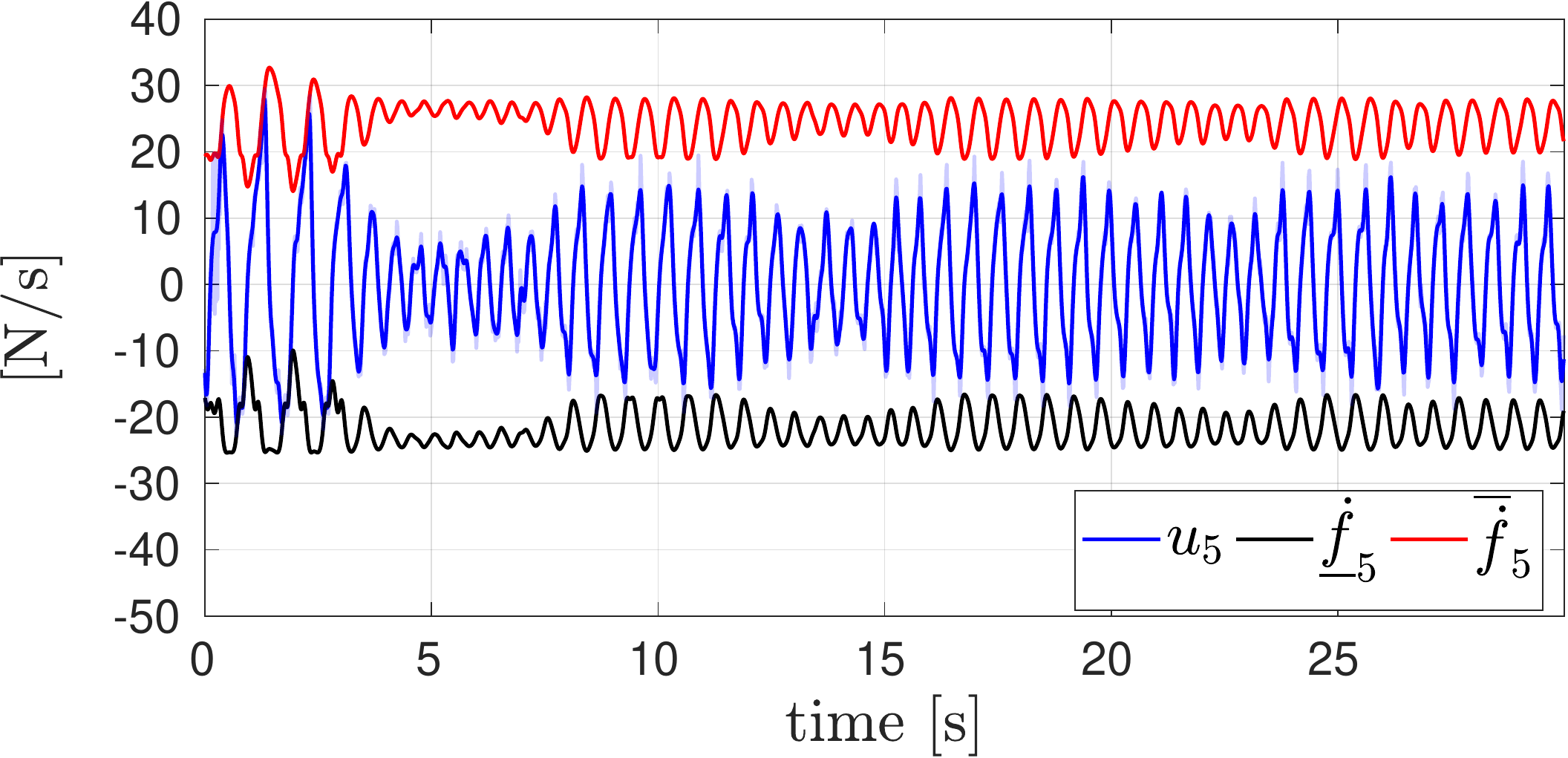}
	\caption{Plots of the Tilt-Hex with rotor failure and $\beta=0$\,\si{deg} while hovering. The robot is disturbed with a constant external torque $\tauv_{\text{dist}}$, activated at $t=t_1$. From top to bottom, the disturbance torque, the actuator forces and their derivatives. In particular, all the signals remain inside the feasible region delimited by the constraints.}
	\label{fig:plots_TiltHexFail1_hov_limits}
\end{figure}

In order to make the simulations even more realistic, in addition to the already introduced measurement noise, we add a torque disturbance to the platform, whose magnitude can be compared to typical values that one could experience in a real experiment due to parameter mismatches and/or external perturbations.
The way this torque $\tauv_{\text{dist}}$ is computed deserves some explanations.
In the case $\beta=0$, when both the $6-th$ and the $3-rd$ actuators are switched off, the moments generated by the
other four propellers lie all on a 2-dimensional plane, c.f.~\cite{2018a-MicRylFra}, Fig.~3. This can be verified by analyzing the rank of the allocation sub-matrix $^3\Gm_2^6=\Gm_2(:,1,2,4,5)$, i.e., the sub-part related to the torque actuation deprived of the columns related to the actuators which are broken (the $6-th$) and off (the $3-rd$), respectively.
At this point, we select the normal to such plane by finding an orthonormal base $\{\vv_1\, \vv_2\}$ for the column span of $^3\Gm_2^6$ and operate the cross product $\vv_3 = \vv_1 \times \vv_2$.
This unit vector indicates the direction of the most unfavorable torque disturbance for the platform when $\beta=0$ and only actuators $\{1,2,4,5\}$ are effectively working. In order ensure that such perturbation cannot be compensated by a MRAV with this tilting configuration, even if the $3-rd$ actuator is actively used, we verify that $\vv_3$ has a positive projection along the direction of the total torque $\tauv_3^B$ that can be generated by such actuator.
In mathematical terms, we select $\vv_3' = \text{sgn}(\vv_3^{\top} \tauv_3^B) \vv_3$.
Finally, we scale down the vector norm in order to obtain a meaningful order of magnitude for the disturbance, i.e., $\tauv_{\text{dist}} = \frac{1}{250} \vv_3'$. In the presented simulations, it is activated at $t=t_1$. The evolution of such perturbation, constant in body frame, is depicted in the first plots of Fig.~\ref{fig:plots_TiltHexFail1_hov_limits} and Fig.~\ref{fig:plots_TiltHexFail2_hov_limits}.

\begin{figure}[t]
	\centering
		\includegraphics[width=\figWidth\columnwidth]{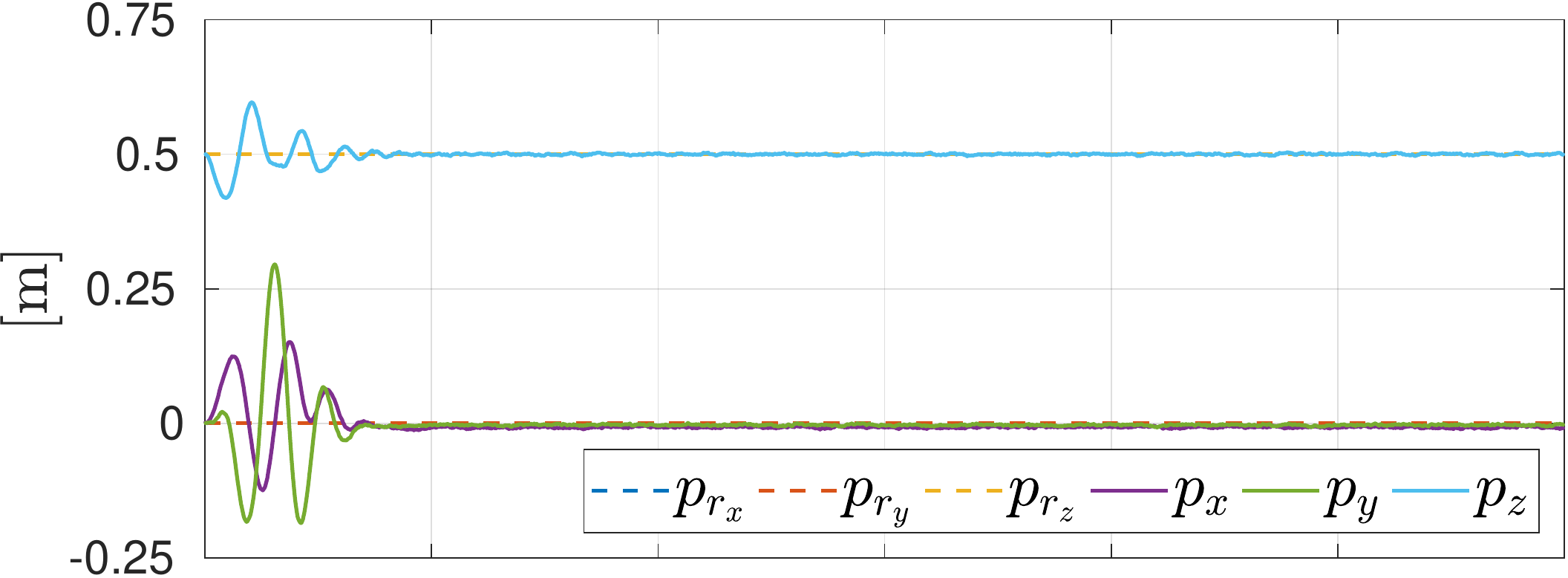}\\
		\includegraphics[width=\figWidth\columnwidth]{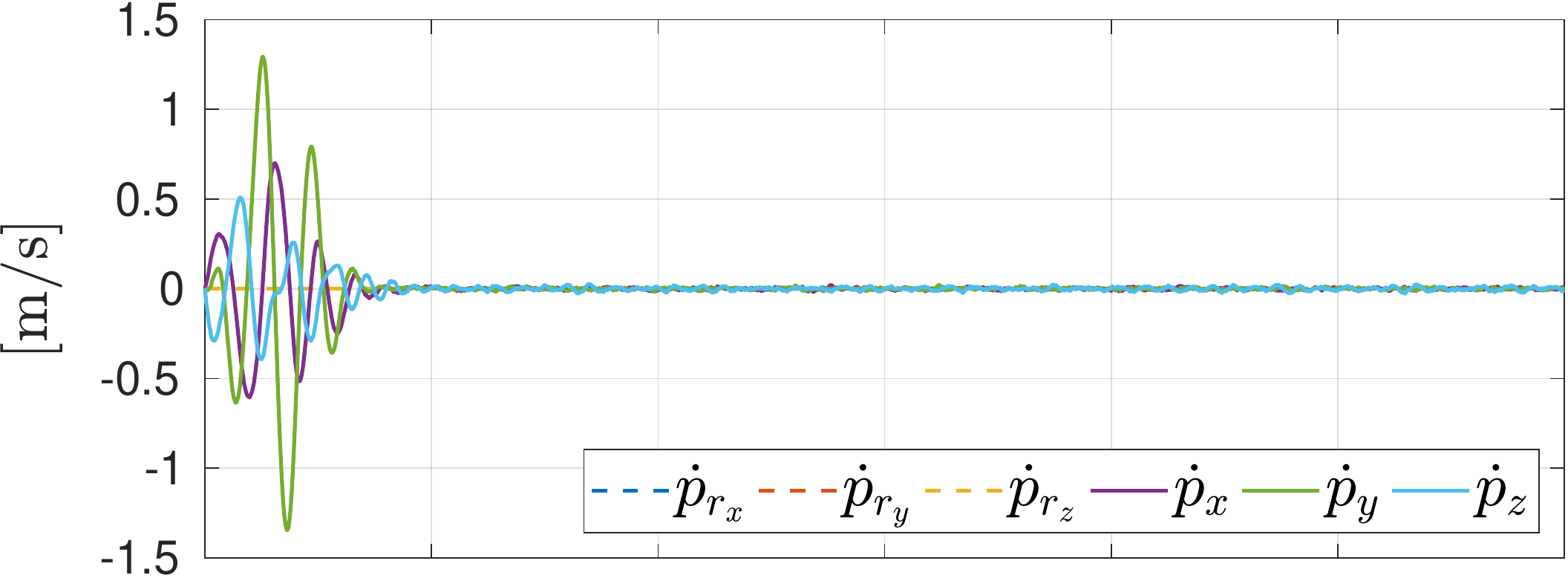}\\
		\includegraphics[width=\figWidth\columnwidth]{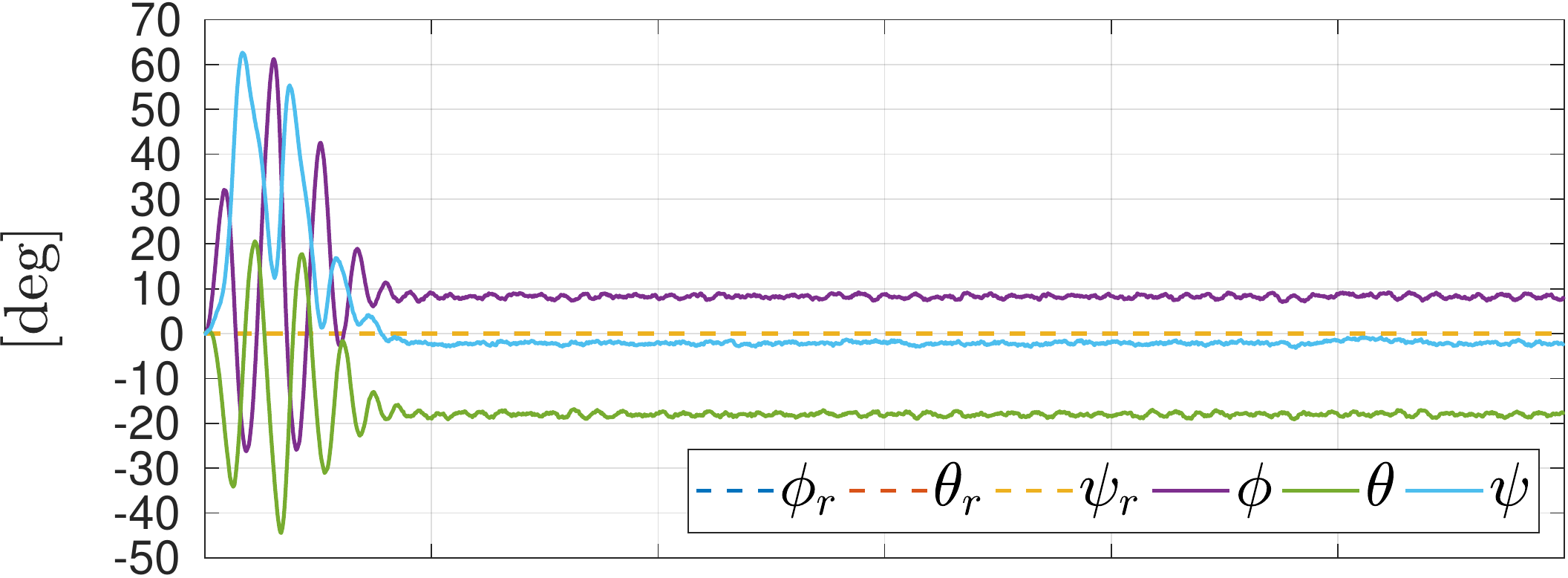}\\
		\includegraphics[width=\figWidth\columnwidth]{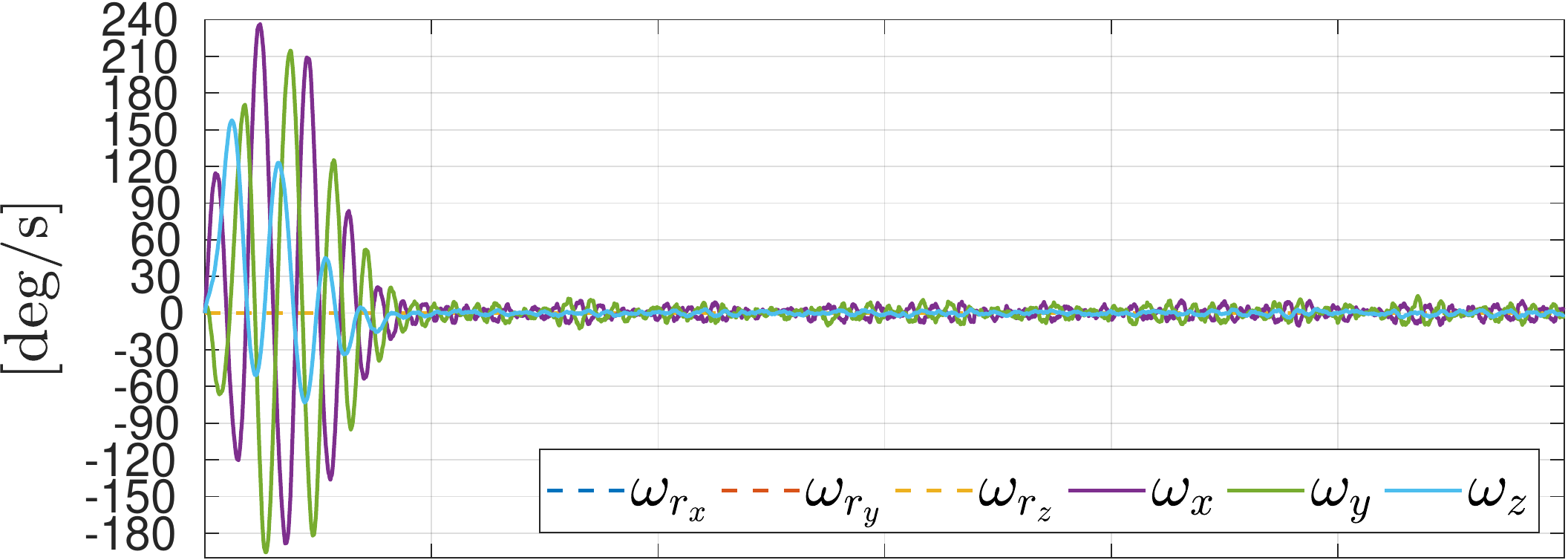}\\
		\includegraphics[width=\figWidth\columnwidth]{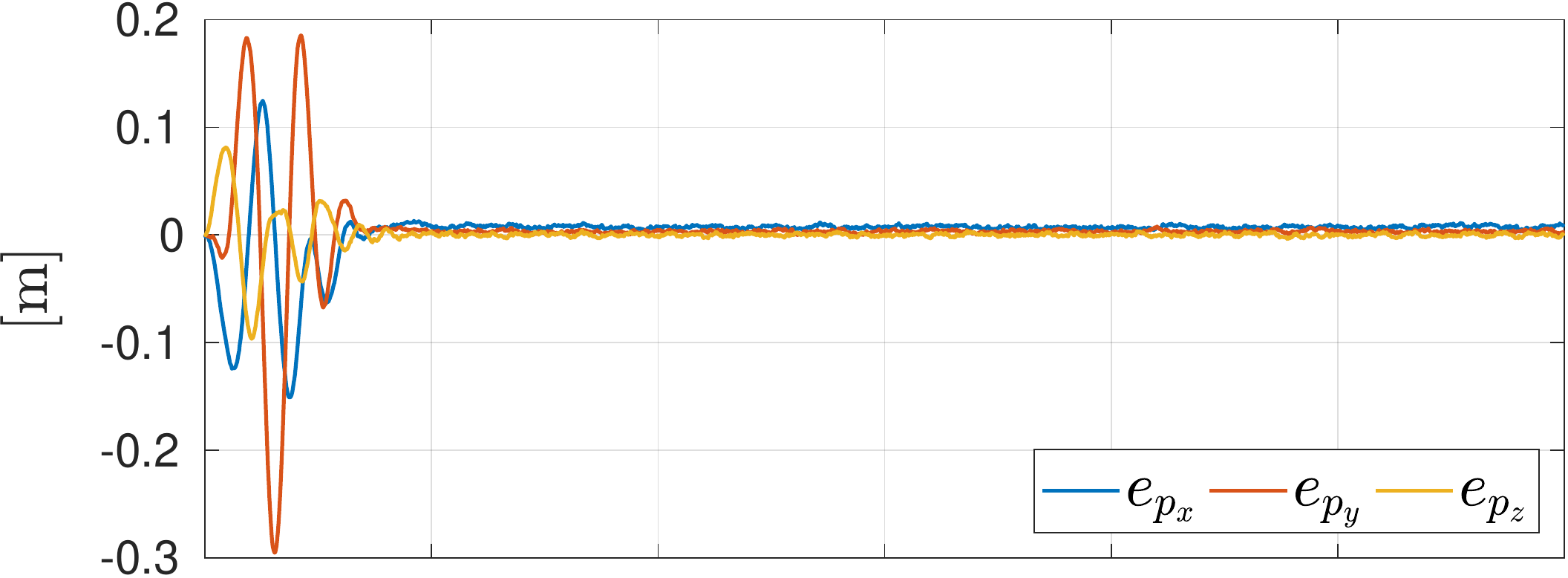}\\
		\includegraphics[width=\figWidth\columnwidth]{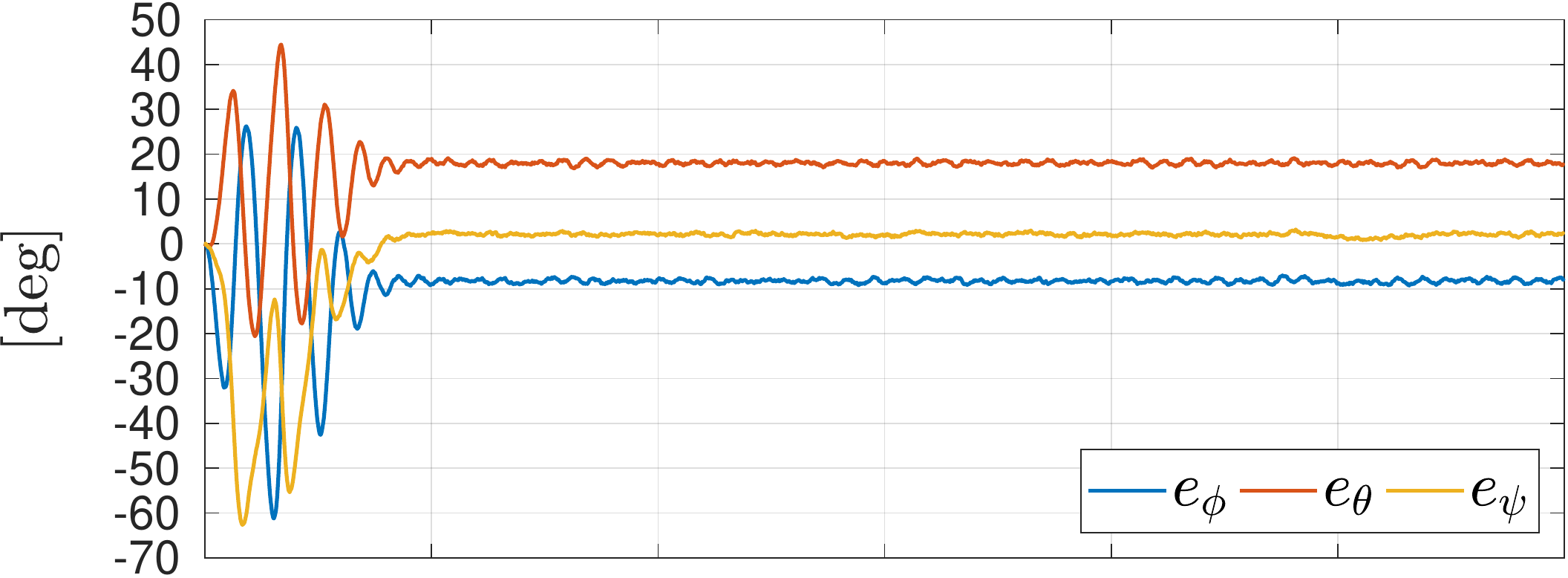}\\
		\includegraphics[width=\figWidth\columnwidth]{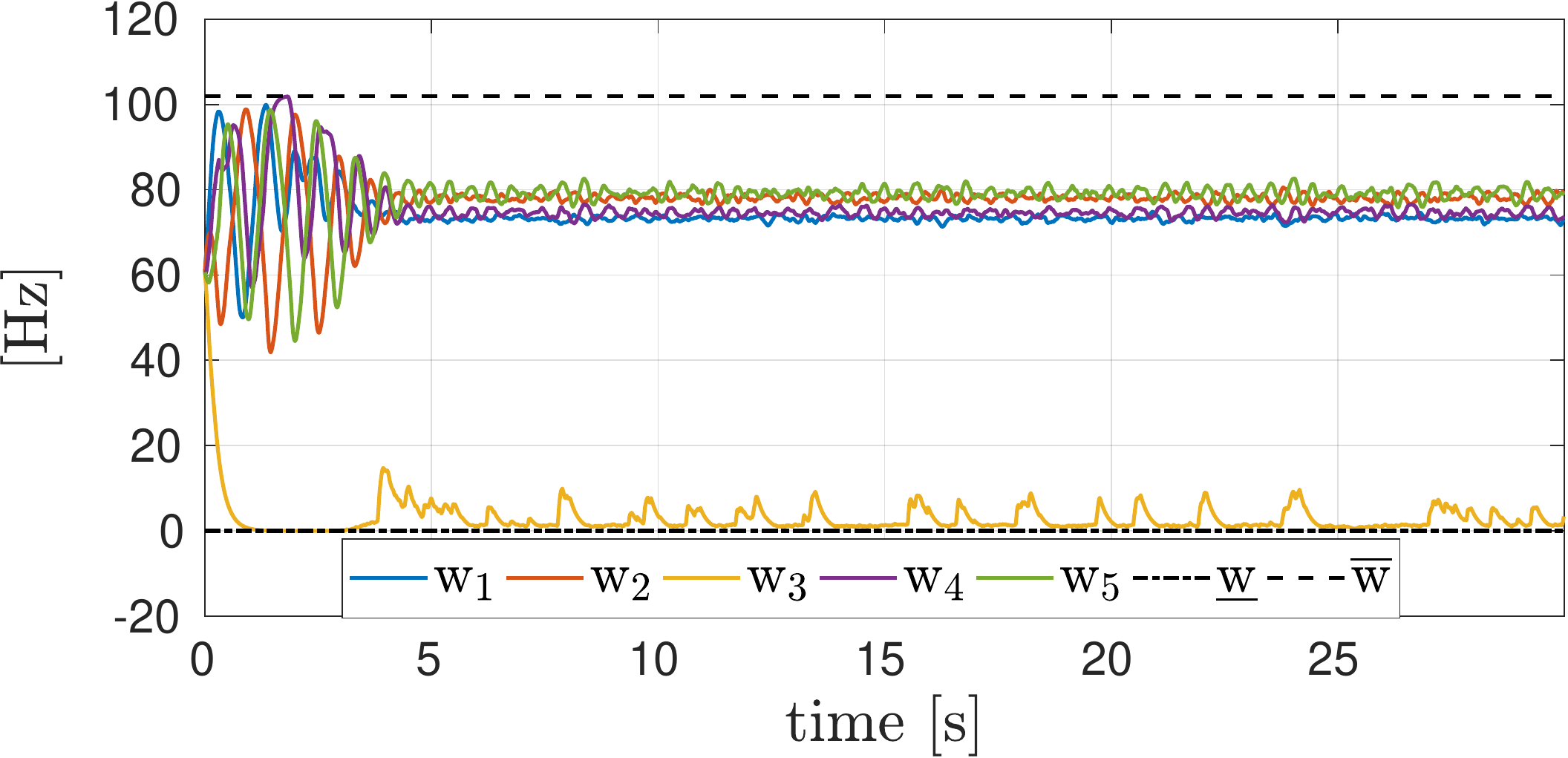}
	\caption{Plots of the Tilt-Hex with rotor failure and $\beta=-25$\,\si{deg} while hovering. The robot is disturbed with a constant external torque. From top to bottom, the position, linear velocity, orientation and angular velocity tracking, the position and orientation errors, and the actuator spinning velocities.}
	\label{fig:plots_TiltHexFail2_hov_traj}
\end{figure}

\begin{figure}[t]
	\centering
		\includegraphics[width=\figWidth\columnwidth]{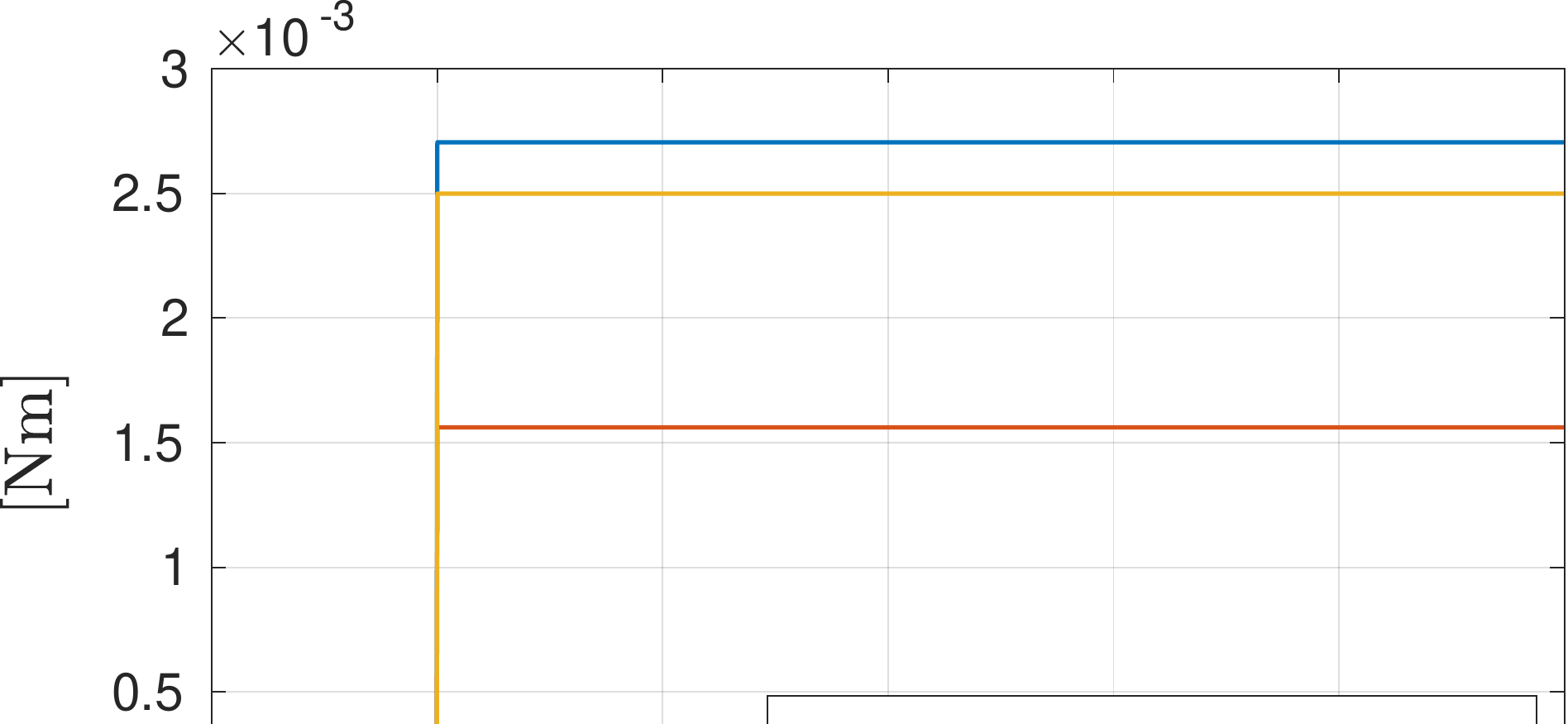}\\
		\includegraphics[width=\figWidth\columnwidth]{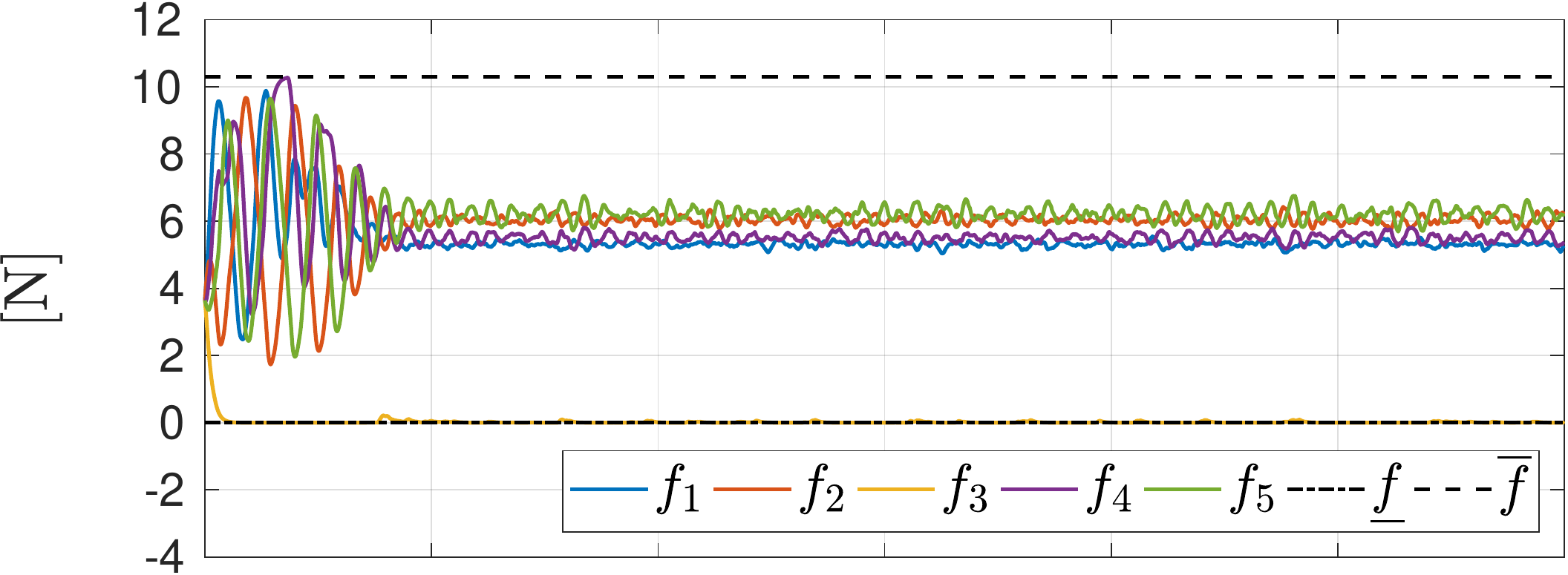}\\
		\includegraphics[width=\figWidth\columnwidth]{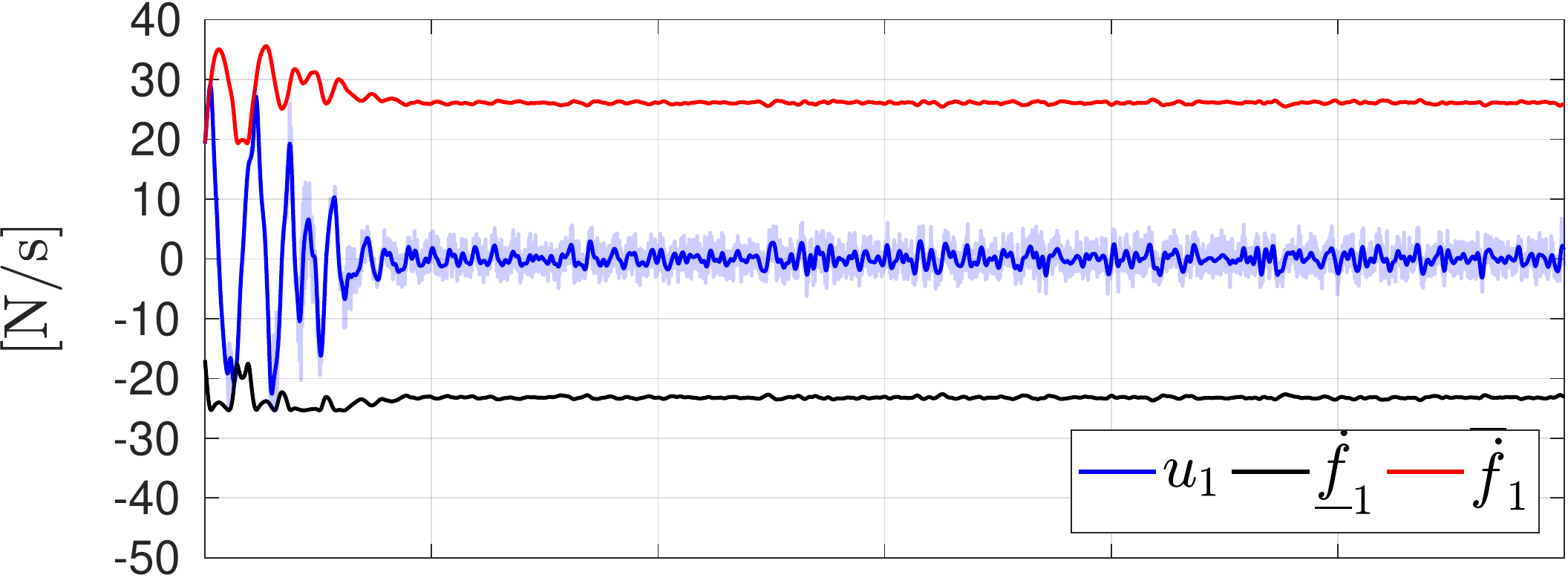}\\
		\includegraphics[width=\figWidth\columnwidth]{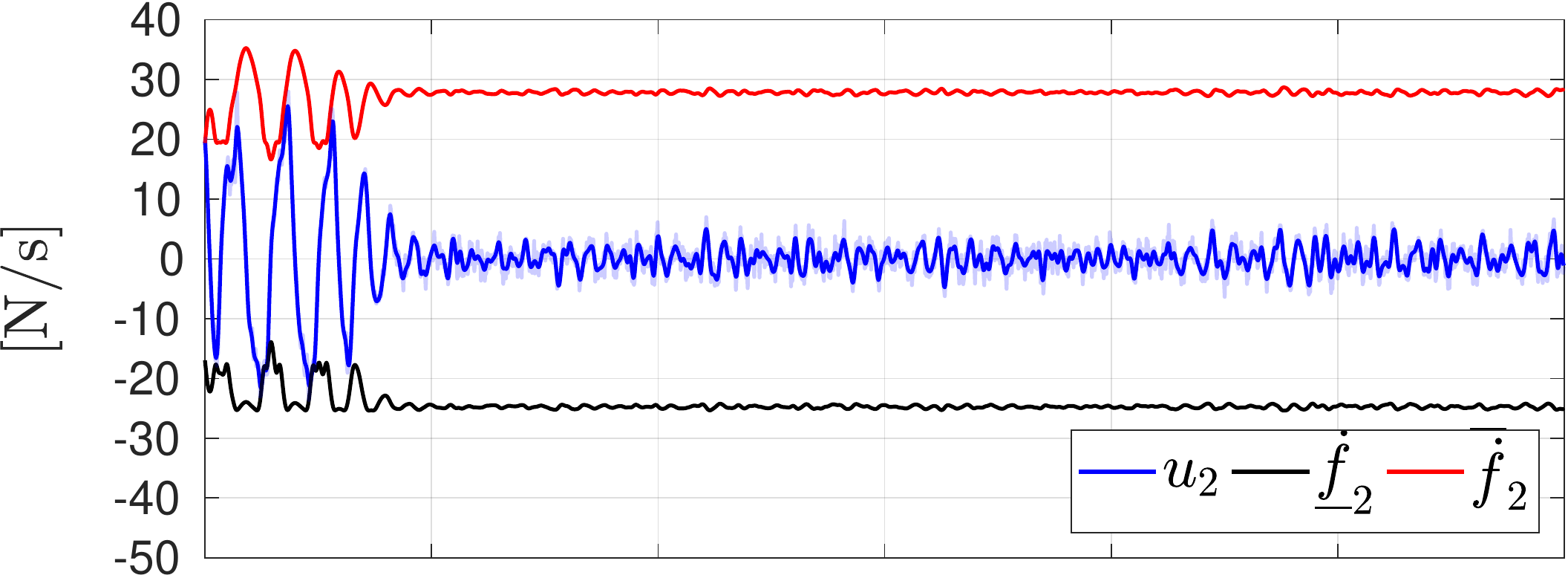}\\
		\includegraphics[width=\figWidth\columnwidth]{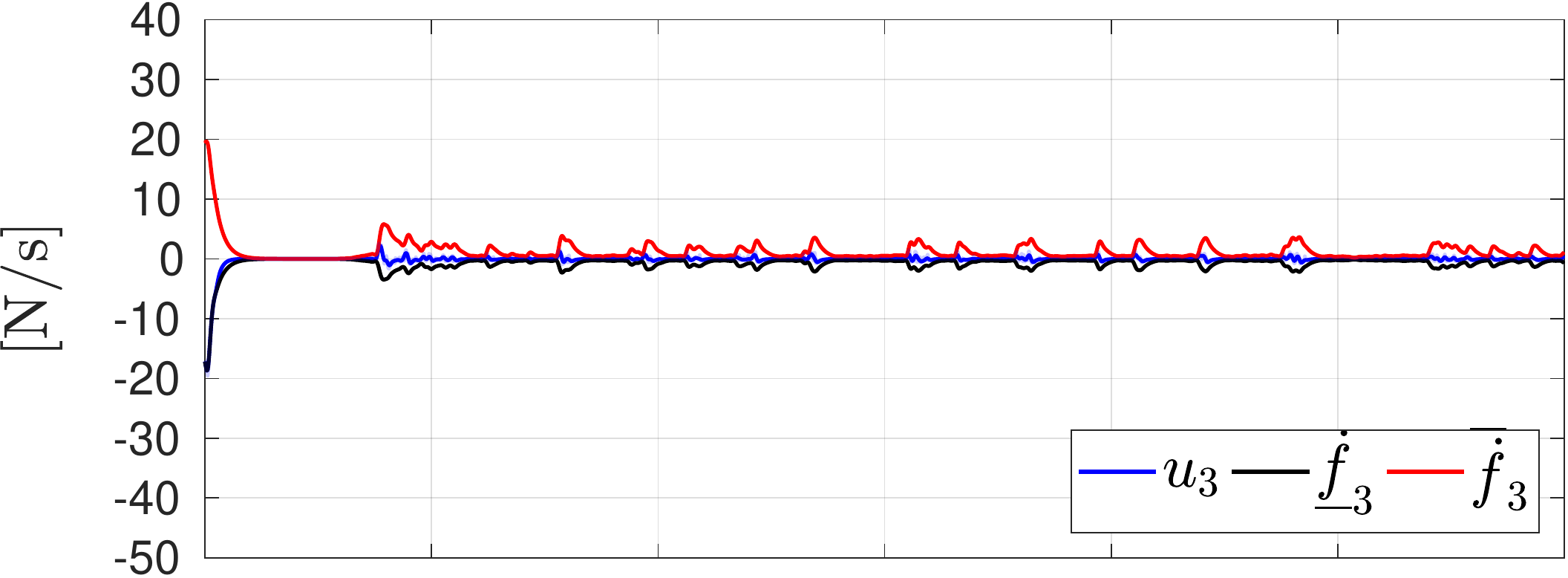}\\
		\includegraphics[width=\figWidth\columnwidth]{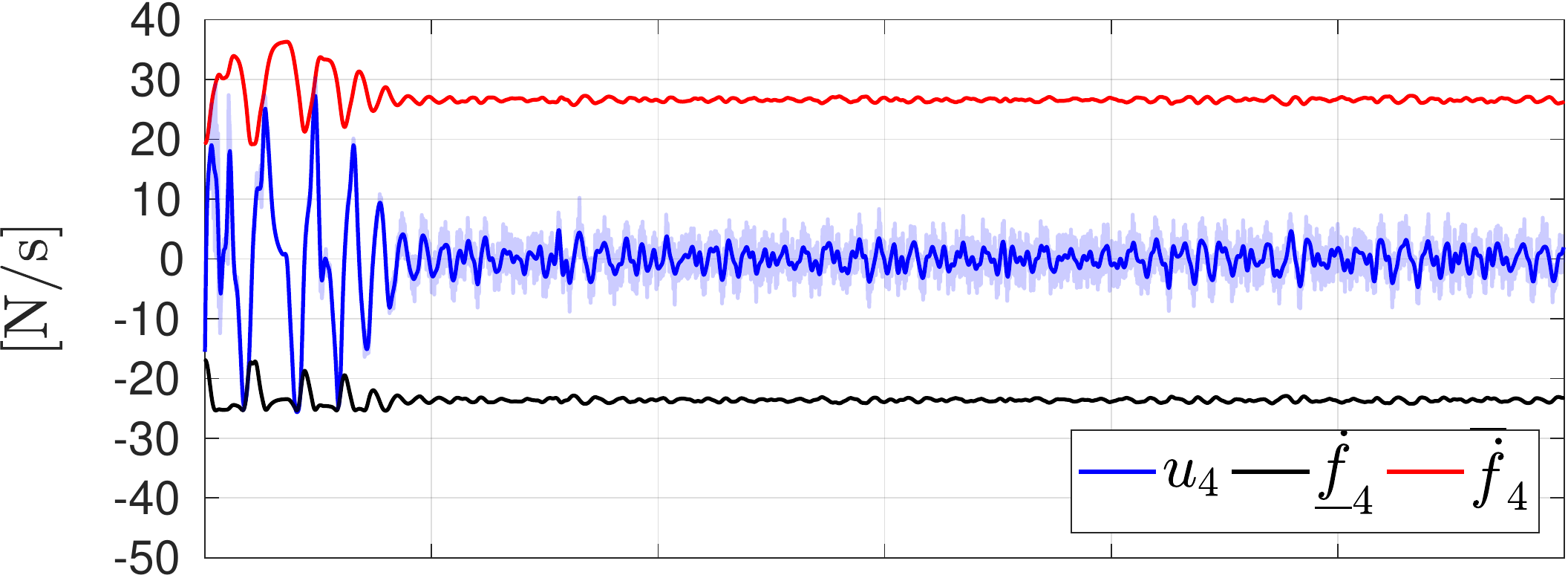}\\
		\includegraphics[width=\figWidth\columnwidth]{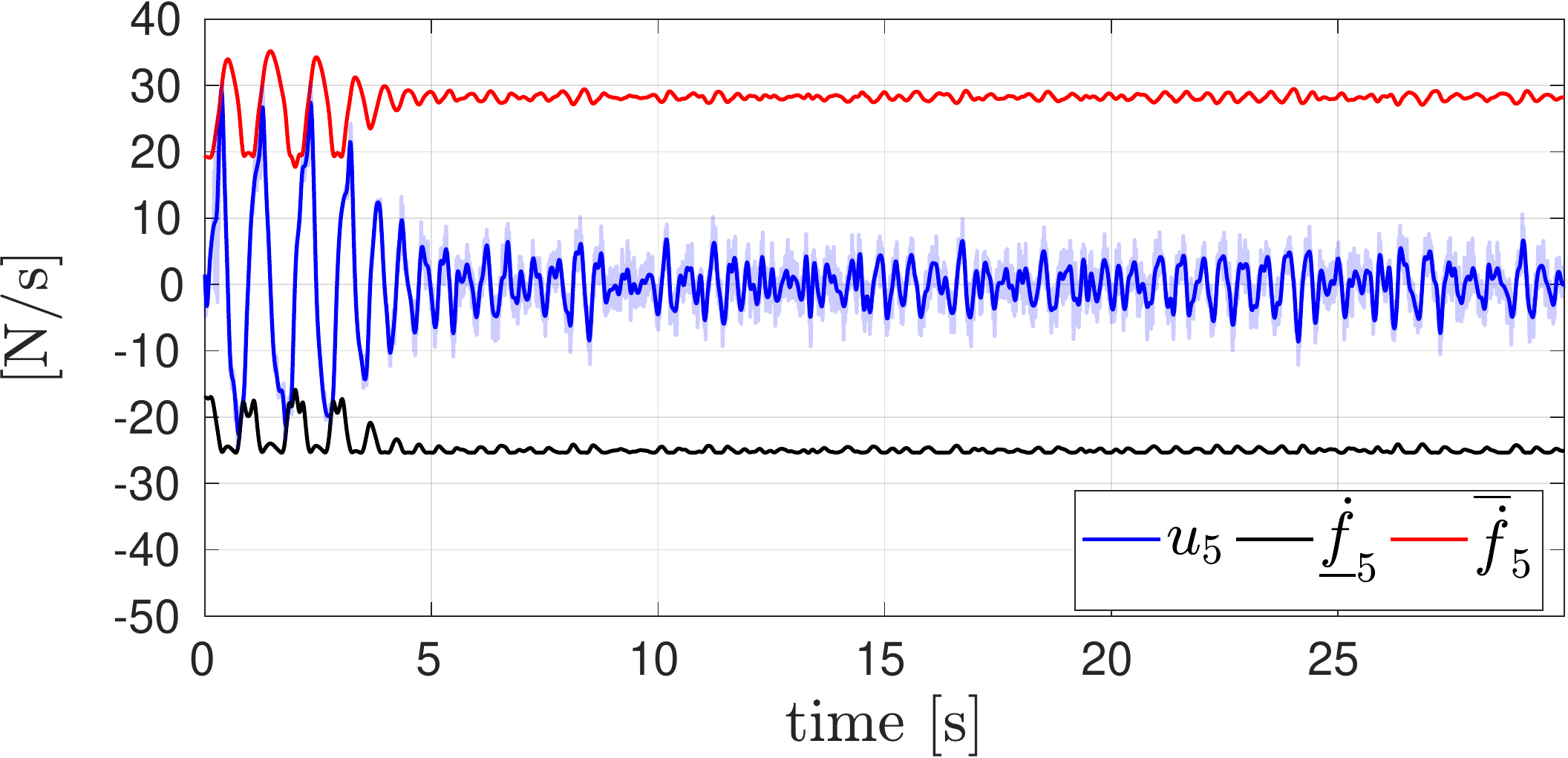}
	\caption{Plots of the Tilt-Hex with rotor failure and $\beta=-25$\,\si{deg} while hovering. The robot is disturbed with a constant external torque $\tauv_{\text{dist}}$, activated at $t=t_1$. From top to bottom, the disturbance torque, the actuator forces and their derivatives. In particular, all the signals remain inside the feasible region delimited by the constraints.}
	\label{fig:plots_TiltHexFail2_hov_limits}
	\vspace{-1em}
\end{figure}

The plots of this simulation related to the case $\beta=0$\,\si{deg} are depicted in
Fig.~\ref{fig:plots_TiltHexFail1_hov_traj} and in Fig.~\ref{fig:plots_TiltHexFail1_hov_limits}, while the ones obtained with $\beta=-25$\,\si{deg} are portrayed in Fig.~\ref{fig:plots_TiltHexFail2_hov_traj} and in
Fig.~\ref{fig:plots_TiltHexFail2_hov_limits}.
Comparing both the position and the orientation errors in the two cases, we can see that for $\beta=0$\,\si{deg} the platform cannot hover statically, since it periodically oscillates, with peaks of almost $\pm2$\,\si{cm} and $\pm7.5$\,\si{deg}, around the steady-state configurations. On the other hand, for $\beta=-25$\,\si{deg} the MRAV can fulfill the challenging goal of remaining still. This is a consequence of the fact that, for $\beta\ne0$ the span of $^3\Gm_2^6$ is already 3-dimensional and so the perturbation can be annihilated while being in static hovering. In both cases, the first part of the simulation is characterized by consistent oscillations of the state components, as it is clear from the plots 1-4 of the two figures. In particular, these transients are caused by the fact that the initial robot orientation is $\etav_0=[0\, 0\, 0]^{\top}\, \si{deg}$, which is not attainable in steady-state for the MRAV in both configurations.

Some final remarks are detailed in order. First of all, the aforementioned claim that, in this case, the $3-rd$ actuator is almost never used is confirmed by the last plots of Fig.~\ref{fig:plots_TiltHexFail1_hov_traj} and Fig.~\ref{fig:plots_TiltHexFail2_hov_traj}. Indeed, the control algorithm regulates to zero the related force component almost everywhere. In particular, during the initial transient phase, we see how the rotor velocities (and so the generated thrust forces) approach their upper bounds. Regulating the spinning rate of the $3-rd$ rotor to a value grater than zero, would cause the other components to saturate, with large chances to destabilize the platform.
Secondly, the platform orientation converges (for $\beta=-25$\,\si{deg}) to a certain value, as depicted in the third and in the sixth plots of Fig.~\ref{fig:plots_TiltHexFail2_hov_traj}. Note that such steady-state orientation value, which depends on $\alpha,\, \beta$ and on $\tauv_{\text{dist}}$, is automatically computed by the NMPC algorithm, in relation to the state and input limitations, and it is not a-priori given.
This feature guarantees the optimality of the trajectory w.r.t. the robot dynamic capabilities and relieves the user from performing any explicit computations.
Finally, remark that the proposed controller can achieve better results compared to the one designed in~\cite{2017f-MicRylFra,2018a-MicRylFra}, since the errors on the state keep bounded without diverging also in the case $\beta=0$, despite the addition of a constant challenging disturbance which remain unknown to the NMPC algorithm. This fact highlights the potentiality of predictive controllers compared to reactive static feedback ones.

\section{Conclusions}

In this paper, we have presented an NMPC framework tailored to generic multi-directional thrust MRAVs with arbitrarily positioned and oriented rotors, which considers a novel and more representative model for the actuators of such systems compared to the ones often employed by other works. 
More in detail, the time derivatives of the propeller thrust forces are considered as the control inputs to be optimized by the predictive controller, as they are directly related to the torques applied to the motors, which constitute the lowest-level control inputs for multi-rotor systems.
Thanks to the simple but effective model for the actuator dynamics that we designed by leveraging available experimental data, it is possible to indirectly take into account multiple low-level physical effects such as the ones induced by the rotor inertia, the aerodynamic drag, and other highly nonlinear hidden electrical phenomena, just by modeling the maximum force derivatives (equivalently, the maximum rotor accelerations) as a function, identifiable thanks to the proposed methodology, of the instantaneous propeller forces and the user-defined accuracy w.r.t. the set-point thrust values.
During the resolution of the optimal control problem, we constrain only the inputs and the part of the model state related to the actuators to lie within the identified feasible set, avoiding the imposition of any fictitious limits in the robot orientation, angular velocity, body thrust and moment, or any other non-physical limitations. To improve its computational efficiency, the control algorithm is implemented using a state-of-the-art real time iteration scheme with partial sensitivity update method. To demonstrate its real-time capabilities, the controller has been validated with four different multi-rotor platforms, both in experiments and realistic simulations, showing its versatility and applicability to different challenging scenarios. At the best of our knowledge, this is the first time that an NMPC framework with all such features is presented and extensively validated both with under-actuated and fully-actuated aerial robots, and both with fixed and orientable propellers. Ultimately, we have provided a unified framework for the predictive control of generic multi-rotor aerial vehicles which can be particularized for the specific platform at hand just by applying the proposed identification procedure for the actuators limits, which constitutes an additional contribution of our work.

Future work includes the automatic regulation of the cost function weights, which are a fundamental part of predictive controllers, in order to reduce the tuning time and effort for the user. The use of neural networks could be envisioned. Moreover, we plan to conduct additional validation tests, e.g., including perception objectives inside the cost function, and controlling other different MDT-MRAVs.
Ultimately, we aim to transfer the technology presented in the experimental validation to an outdoor MoCap-denied scenario where only on-board computational resources are used. The final goal is to be able to extend and use the presented framework to fulfill aerial physical interaction tasks.

\section*{Appendix}

\subsection*{Allocation matrix identification}

The nominal values of the entries of the allocation matrix $\Gv$ can be calculated from the system's geometrical properties, consistently with~\eqref{eq:alloc_mat_detail}. However, the real physical parameters of the robot could be quite different from the ideal ones, due to mechanical inaccuracies unavoidably associated with the manufacturing and the assembly of the robot parts. This may dramatically affect the control system performances. For this reason, in this work the entries of the allocation matrix are identified from experimental data.
In the following we briefly outline the used identification method, which is extensively used in the literature and very well-known from the community, so it is not considered as a contribution.

First of all, we used the nominal allocation matrix to design a simple but robust controller, applied on the platform. Accordingly, the so-obtained control system is used to track
suitable \emph{persistently exciting} 6D trajectories. To  this purpose chirp signals are used, i.e., sinusoidal trajectories with increasing frequencies.
While doing this, we collected the measured data $\pv$ and $\Rm$ thanks to the MoCap system, used as ground truth. In particular, we made the assumption to be able to measure the CoM location. Then, thanks to a properly tuned post-processing of the data which mainly consisted in a constant frame-rate signal re-sampling, an anti-causal low-pass filtering and the computation of numerical derivatives, we were able to retrieve a precise-enough estimation of $\ddpv, \omegav, \dot{\omegav}$ defined in~\eqref{eq:Newton_Euler_Law}. On the other hand, $\gammav$ was reconstructed by collecting the measured spinning rates of the motors $\doubleu_i$ and using the thrust model~\eqref{eq:f_i}.
Finally, $m$ was directly measured and $\vect J$ estimated by a precise CAD model of the robot.
At this point, we re-wrote~\eqref{eq:Newton_Euler_Law} as
\begin{align}
	\label{eq:ident}
	\underbrace{
	\begin{bmatrix}
		m \mat R ^{\top} (\ddpv+g\ev_3) \\
		\Jm \dot{\omegav} + \omegav\times\Jm\omegav
	\end{bmatrix}
	}_{:=\yv}
	=
	\underbrace{
	\begin{bmatrix}
		f_1 \mat I _3\ \dots\ f_n\mat I _3   &   \mat 0 _{3 \times 3n} \\
				\mat 0 _{3 \times 3n}		    &   f_1 \mat I _3\ \dots\ f_n\mat I _3
	\end{bmatrix}
	}_{:=\mathbf{A}}
	\boldsymbol{\beta}
\end{align}
with $\mathbf{A} \in \mathbb{R}^{6 \times 6n}$. In such form, the equation allows to express the vector of measurable quantities $\yv \in \mathbb{R}^{6 \times 1}$ as a linear function of a vector of parameters $\boldsymbol{\beta} \in \mathbb{R}^{6n \times 1}$, obtained re-arranging the entries of $\Gm$
\begin{align}
	\label{eq:beta_params}
	\boldsymbol{\beta}
	:=
	\begin{bmatrix}
		{\Gm_1(:,1)}^{\top}\, \dots\, {\Gm_1(:,n)}^{\top}\, {\Gm_2(:,1)}^{\top}\, \dots {\Gm_2(:,n)}^{\top}
	\end{bmatrix}
	^{\top}
\end{align}
Collecting a large number of measurements $p>>6n$ and stacking them in vectorial form, we obtained
\begin{align}
	\label{eq:ident_vect}
	(	
	\boldsymbol{\xi}
	=
	\boldsymbol{\Lambda}
	\boldsymbol{\beta}
	)
	:=
	(
	\begin{bmatrix}
		\yv_1 \\
		\vdots \\
		\yv_p
	\end{bmatrix}
	=
	\begin{bmatrix}
		A_1 \\
		\vdots \\
		A_p
	\end{bmatrix}
	\boldsymbol{\beta}
	)
\end{align}
At this point, applying the standard least-squares identification method, the vector of parameters which minimizes the 2-norm of the error $|| \boldsymbol{\Lambda}\boldsymbol{\beta} - \boldsymbol{\xi} ||^2$ is obtained as
\begin{align}
	\label{eq:pinv}
	\hat{\boldsymbol{\beta}}
	=
	\boldsymbol{\Lambda}^{\dagger}
	\boldsymbol{\xi}
\end{align}
Finally, re-arranging the element of the vector $\hat{\boldsymbol{\beta}}$ using the convention of~\eqref{eq:beta_params}, we obtained the identified allocation matrix $\hat{\Gv}$ that we used in the presented experiments.

\begin{figure}[t]
	\centering
	\includegraphics[width=0.75\columnwidth]{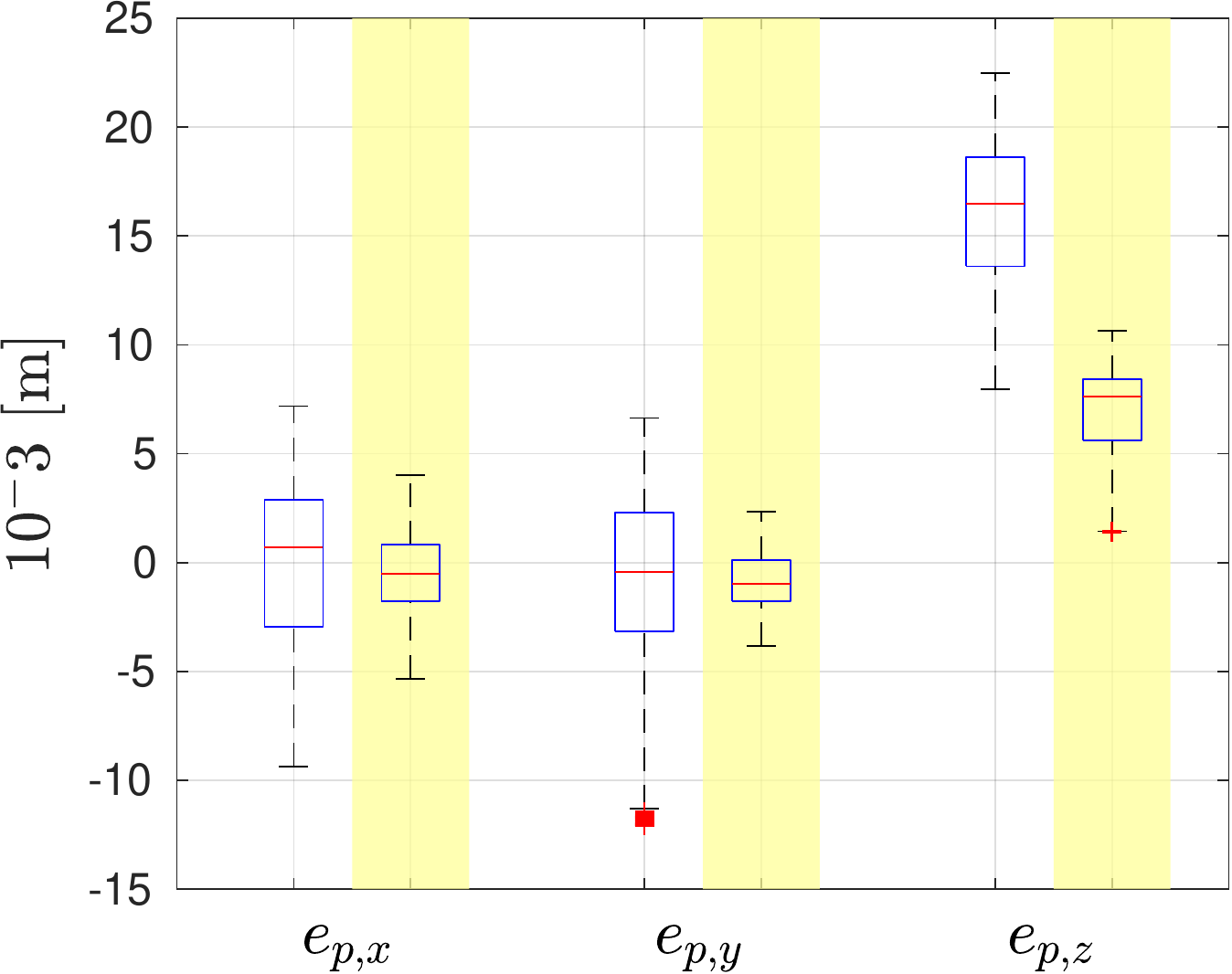} \\ \vspace{1em}
	\includegraphics[width=0.75\columnwidth]{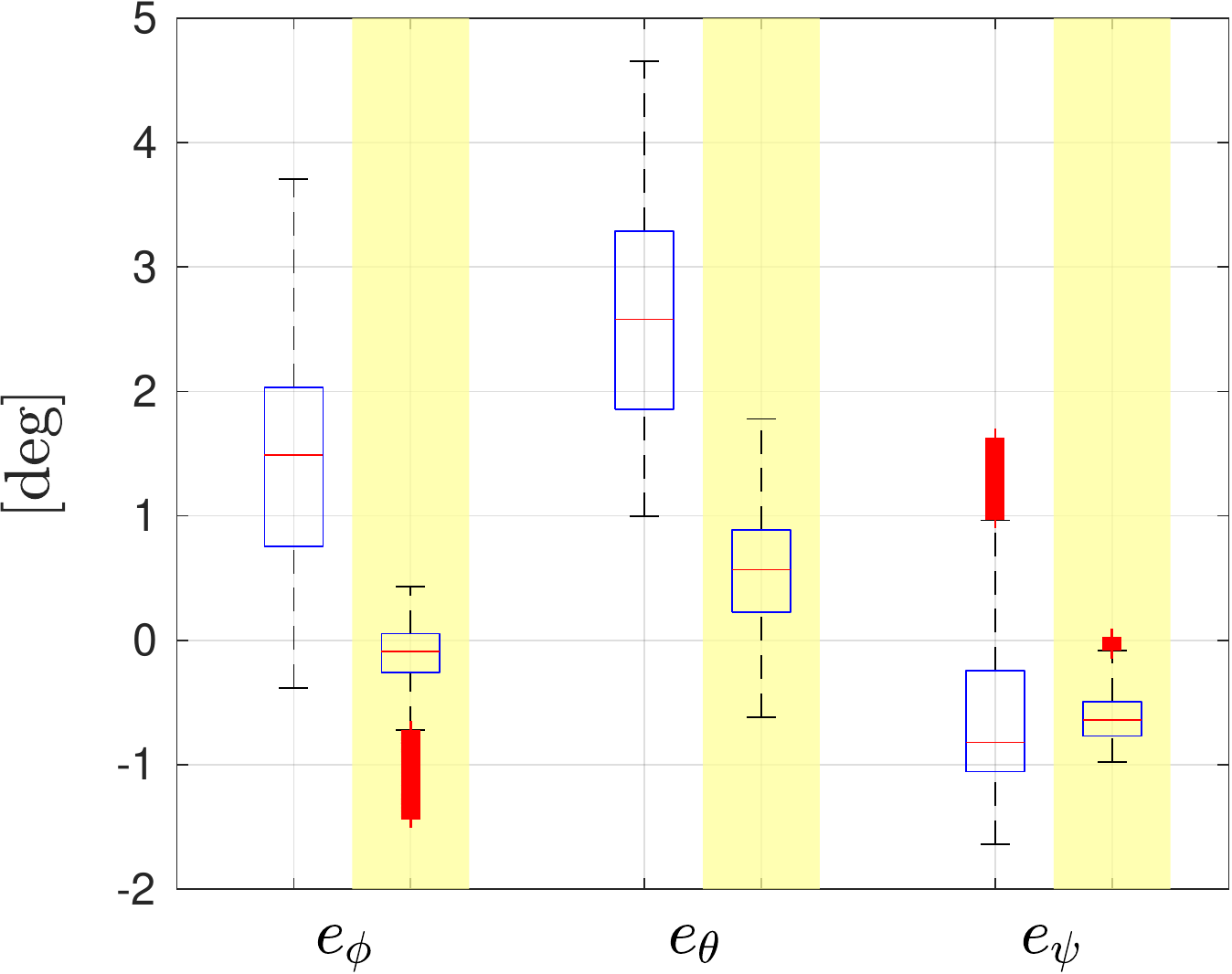}
	\caption{Box-plots for the position error (above) and the orientation error (below) of the Tilt-Hex when hovering using the nominal and the identified allocation matrices. The results for the latter case have been highlighted with yellow bands. We can appreciate how the error mean and variance is reduced.}
	\label{fig:boxplot_ident_alloc}
\end{figure}

Comparing the entries of the nominal and the identified allocation matrices in the hexarotor (Tilt-Hex) case, notice that the difference between some elements is pretty consistent. This confirms that the physical parameters of the real robot can be very dissimilar from the nominal ones.
\begin{align}
	\label{eq:e_G}
	e_{G,\%}
	=
	100\ \bigg[\frac{g_{i,j}-\hat{g}_{i,j}}{g_{i,j}}\bigg]
	=
\left[\begin{smallmatrix}
	   42 & -12 & -18 & 41 & 9 & 16 \\
	    4 & 21 & 25 & 4 & 80 & 104 \\
	    4 & 6 & 11 & 10 &  5 & 2 \\
	   72 & 31 & 30 & 58 & 25 & 28 \\
	   26 & 24 & 31 & 27 & 29 & 28 \\
	    9 & 15 & 16 & 12 & 14 & 13
\end{smallmatrix}\right]
\end{align}

To conclude, we would like to point out that using the identified matrix in the controller instead of the nominal one allowed to consistently reduce both the position and the orientation errors in all the experiments that we performed. This happens already in hovering condition, as it is shown in the box-plots of Fig.~\ref{fig:boxplot_ident_alloc}.

\begin{acknowledgements}
We thank Anthony Mallet (LAAS-CNRS) for his contribution to the development of the software architecture exploited for the experiments and Yutao Chen (University of Padova -- Eindhoven University) for the development of the MATMPC framework, integrated in our setup.
\end{acknowledgements}

%
%

\bibliographystyle{IEEEtran} 			 
\bibliography{bibAlias,bibMain,bibNew,bibAF,./bibCustom}   


\end{document}